\documentclass[12pt,oneside]{report}
\usepackage[T1]{fontenc}
\usepackage[utf8x]{inputenc}
\usepackage[american]{babel}

\usepackage{parskip}

\usepackage[a4paper,
	    left=2.5cm,
	    right=2.5cm,
	    top=1.5cm,
	    bottom=1.5cm,
	    marginparsep=5mm,
	    marginparwidth=10mm,
	    headheight=7mm,
	    headsep=1.2cm,
	    footskip=1.5cm,
	    includeheadfoot]{geometry}

\usepackage{fancyhdr}
\renewcommand{\chaptermark}[1]{
  \markboth{Chapter \thechapter. #1}{}}

\fancypagestyle{plain}{
  \fancyhf{}
  \fancyfoot[C]{\thepage}
  \fancyhead[R]{\IhreArbeitEN}
  \fancyhead[L]{\IhrVorname\ \IhrNachname}
}

\usepackage{amsmath}
\usepackage{amssymb}
\usepackage[intlimits]{empheq}

\usepackage[sc]{mathpazo}
\usepackage{pifont}

\newcommand{\cmark}{\ding{51}}
\newcommand{\xmark}{\ding{55}}

\usepackage[format=hang,
            font={footnotesize},
            labelfont={bf},
            margin=1cm,
            aboveskip=5pt,
            position=bottom]{caption}

\usepackage{graphicx}
\usepackage[svgnames,cmyk,table,hyperref]{xcolor}
\usepackage{tikz}
\usetikzlibrary{positioning,arrows,plotmarks}
\usetikzlibrary{calc,math,shapes.geometric,intersections,arrows.meta,decorations.pathreplacing,angles}
\usetikzlibrary{babel}
\usetikzlibrary{quotes}
\usepackage{listings}
\usepackage{microtype,relsize}
\newcommand*{\Sperren}[1]{\textls*[100]{#1}}

\usepackage[backend=biber,
            style=numeric,
            alldates=iso,
            seconds=true]{biblatex}

\usepackage{csquotes}
\usepackage{acronym}

\newcommand*{\IhrVorname}{Tobias J.}
\newcommand*{\IhrNachname}{Bauer}
\newcommand*{\IhrStudiengang}{Künstliche Intelligenz}
\newcommand*{\IhreArbeit}{Masterarbeit}
\newcommand*{\IhreArbeitEN}{Master's Thesis}
\newcommand*{\IhrTitelDE}{Blickverfolgung anhand RGBD-Bilder unter Verwendung von Transformern zur Feature-Zusammenführung}
\newcommand*{\IhrTitelEN}{RGBD Gaze Tracking Using Transformer for Feature Fusion}
\newcommand*{\IhrBearbeitungszeitraumVON}{03. Mai 2023}
\newcommand*{\IhrBearbeitungszeitraumBIS}{02. November 2023}
\newcommand*{\IhrErstpruefer}{Prof. Dr.-Ing. Gerald Pirkl}
\newcommand*{\IhrZweitpruefer}{Prof. Dr. rer. nat. Daniel Loebenberger}
\newcommand*{\IhreFirma}{Hochschule}
\newcommand*{\IhrFirmenbetreuer}{---}
\newcommand*{\IhreZusammenfassung}{%
Gegenstand dieser Arbeit ist die Implementierung einer KI-gestützten Blickverfolgung anhand von RGBD-Bildern, welche neben Farb- (RGB) auch Tiefeninformationen (Depth) liefern. Zur Zusammenführung der aus den Bilddaten gewonnen Features soll ein Modul nach der Transformer-Architektur eingesetzt werden. Gewählt wurde die Kombination aus RGBD-Eingabebildern und Transformern, da sie bisher noch nicht untersucht wurde. Weiterhin soll für das Training der KI-Modelle ein eigener Datensatz erstellt werden, da bisherige Datensätze entweder keine Tiefeninformationen oder nur Labels zur Blickpunktbestimmung enthalten, welche sich nicht für die Blickwinkelbestimmung eignen. Verschiedene Modellkonfigurationen werden auf insgesamt drei unterschiedlichen Datensätzen trainiert, validiert und evaluiert. Die trainierten Modelle sollen dann im Rahmen einer echtzeitfähigen Anwendung eingesetzt werden können, um die Blickrichtung und damit auch den Blickpunkt einer sich vor einem Computer-Monitor befindlichen Person zu bestimmen.

Die in dieser Arbeit verwendete KI-Modellarchitektur basiert auf einer früheren Arbeit von \citeauthor*{Lian2019}, welche ein Generatives Adversiales Netzwerk (GAN) sowohl zur Artefaktentfernung im Tiefenbild, als auch als Feature-Extraktionsnetzwerk verwendet. Auf ihrem eigenen Datensatz, \stg, erreicht das Modell von \citeauthor*{Lian2019} einen mittleren Abstandsfehler von \SI{38.7}{\milli\meter}. Eine Modellarchitektur mit Transformer-Modul zur Feature-Zusammenführung erreicht in dieser Arbeit auf demselben Datensatz einen mittleren Abstandsfehler von \SI{55.3}{\milli\meter}, wobei gezeigt wird, dass ein Verzicht auf ein vortrainiertes GAN-Modul zu einem mittleren Abstandsfehler von \SI{30.1}{\milli\meter} führt. Ein Ersetzen des Transformer-Moduls durch ein mehrlagiges Perzeptron (MLP) führt zu einer Verbesserung auf \SI{26.9}{\milli\meter}.

Die Ergebnisse stimmen auch auf den anderen untersuchten Datensätzen überein. Auf dem \xgaze-Datensatz erreicht das Modell mit Transformer-Modul einen mittleren Winkelfehler von \ang{3.59} und ohne Transformer-Modul \ang{3.26}, wohingegen die grundlegend andere Modellarchitektur der Datensatzautoren \citeauthor*{Zhang2020} einen mittleren Winkelfehler von \ang{2.04} erreicht. Auf dem für diese Arbeit erstellten Datensatz \oge\ erreicht das Modell mit Transformer-Modul einen mittleren Winkelfehler von \ang{4.84}, das Modell ohne Transformer-Modul \ang{4.71} und ein Modell, welches ausschließlich auf RGB-Bildern operiert, \ang{6.29}.
}

\newcommand*{\IhreSchluesselwoerter}{Appearance-based Gaze Estimation, Vision-based Gaze Estimation, RGBD, Feature Fusion Transformer, Machine Learning, Deep Learning, Computer Vision, Eye Tracking}
\newcommand*{\IhreSchluesselwoerterDE}{Erscheinungsbildbasierte Blickverfolgung, Bildgestützte Blickverfolgung, RGBD, Feature-Zusammenführung mit Transformern, Maschinelles Lernen, Deep Learning, Computer-Vision, Eye-Tracking}
\newcommand*{\IhreZusammenfassungEN}{%
Subject of this thesis is the implementation of an AI-based Gaze Tracking system using RGBD images that contain both color (RGB) and depth (D) information. To fuse the features extracted from the images, a module based on the Transformer architecture is used. The combination of RGBD input images and Transformers was chosen because it has not yet been investigated. Furthermore, a new dataset is created for training the AI models as existing datasets either do not contain depth information or only contain labels for Gaze Point Estimation that are not suitable for the task of Gaze Angle Estimation. Various model configurations are trained, validated and evaluated on a total of three different datasets. The trained models are then to be used in a real-time pipeline to estimate the gaze direction and thus the gaze point of a person in front of a computer screen.

The AI model architecture used in this thesis is based on an earlier work by \citeauthor*{Lian2019}. It uses a Generative Adversarial Network (GAN) to simultaneously remove depth map artifacts and extract head pose features. \Citeauthor*{Lian2019} achieve a mean Euclidean error of \SI{38.7}{\milli\meter} on their own dataset \stg. In this thesis, a model architecture with a Transformer module for feature fusion achieves a mean Euclidean error of \SI{55.3}{\milli\meter} on the same dataset, but we show that using no pre-trained GAN module leads to a mean Euclidean error of \SI{30.1}{\milli\meter}. Replacing the Transformer module with a Multilayer Perceptron (MLP) improves the error to \SI{26.9}{\milli\meter}.

These results are coherent with the ones on the other two datasets. On the \xgaze\ dataset, the model with Transformer module achieves a mean angular error of \ang{3.59} and without Transformer module \ang{3.26}, whereas the fundamentally different model architecture used by the dataset authors \citeauthor*{Zhang2020} achieves a mean angular error of \ang{2.04}. On the \oge\ dataset created for this thesis, the model with Transformer module achieves a mean angular error of \ang{4.84}, the model without Transformer module \ang{4.71}, and an RGB-only model \ang{6.29}.
}

\usepackage[bookmarks, raiselinks, pageanchor,
            hyperindex, colorlinks,
            citecolor=black, linkcolor=black,
            urlcolor=black, filecolor=black,
            menucolor=black]{hyperref}
\hypersetup{pdftitle={\IhrTitelEN},
            pdfauthor={\IhrVorname\ \IhrNachname},
            pdfsubject={\IhreArbeitEN},
            pdfkeywords={\IhreSchluesselwoerter}}

\usepackage{cleveref}
\usepackage{enumitem}
\setlist[itemize]{labelsep=0.25cm, leftmargin=!, itemindent=0pt, labelindent=0.5em}
\usepackage{tabto}
\usepackage{siunitx}
\DeclareSIUnit{\inch}{''}
\usepackage{subcaption}
\usepackage{multirow}
\usepackage{xfrac}
\usepackage[l3]{csvsimple}
\usepackage[defaultlines=2,all]{nowidow}
\usepackage{nicematrix}
\AtBeginDocument{\renewcommand{\today}{10. Oktober 2023}}

\newcommand{\xgaze}{\textit{ETH-XGaze}}
\newcommand{\oge}{\textit{OTH-Gaze-Estimation}}
\newcommand{\mpii}{\textit{MPIIGaze}}
\newcommand{\ed}{\textit{Eyediap}}
\newcommand{\stg}{\textit{ShanghaiTechGaze+}}
\newcommand{\et}{Eye Tracking}
\newcommand{\gt}{Gaze Tracking}
\newcommand{\ges}{Gaze Estimation}
\newcommand{\dl}{Deep Learning}
\newcommand{\ml}{Machine Learning}
\newcommand{\bn}{Batch Normalization}
\newcommand{\mpool}{Max Pooling}
\newcommand{\lrelu}{Leaky \ac{ReLU}}
\newcommand{\sigm}{Sigmoid}
\newcommand{\package}[1]{\texttt{#1}}

\newcommand{\tableHeaderHeight}{0cm}
\newcommand{\tableHeaderR}[1]{\rotatebox{90}{\parbox{\tableHeaderHeight}{#1}}}

\begin{document}
  \pagenumbering{roman}
  \begin{titlepage}
    \thispagestyle{empty}
    \begin{center}
      \Large
      Ostbayerische Technische Hochschule Amberg-Weiden\\
      Fakultät Elektrotechnik, Medien und Informatik\\[1cm]
      Studiengang \IhrStudiengang\\[1cm]
      \textbf{\IhreArbeit}\\[1cm]
      von\\[1cm]
      \IhrVorname\ \Sperren{\textbf{\IhrNachname}}\\[1cm]
      \textbf{\IhrTitelDE}\\[1cm]
      \IhrTitelEN
      \vfill
      \textcolor{red}{\normalsize Note: some images have been redacted due to unclear copyright situation}
    \end{center}
  \end{titlepage}
  \clearpage
  \thispagestyle{empty}
  \mbox{}
  \clearpage
  \thispagestyle{empty}
  \begin{center}
    \Large
    Ostbayerische Technische Hochschule Amberg-Weiden\\
    Fakultät Elektrotechnik, Medien und Informatik\\[1cm]
    Studiengang \IhrStudiengang\\[1cm]
    \textbf{\IhreArbeit}\\[1cm]
    von\\[1cm]
    \IhrVorname\ \Sperren{\textbf{\IhrNachname}}\\[1cm]
    \textbf{\IhrTitelDE}\\[1cm]
    \IhrTitelEN
  \end{center}
  \vspace*{5cm}
  \begin{tabbing}
    \underbar{Bearbeitungszeitraum:}\qquad\= von\qquad\=\IhrBearbeitungszeitraumVON\\
                                          \> bis      \>\IhrBearbeitungszeitraumBIS
  \end{tabbing}
  \vspace*{1cm}
  \underbar{1. Prüfer:}\qquad\IhrErstpruefer\par
  \underbar{2. Prüfer:}\qquad\IhrZweitpruefer
  \clearpage
  \include{formblatt_selbststaendigkeitserklaerung}
  \include{formblatt_summary}
  \tableofcontents
  \newpage
  \chapter*{Abbreviations}
  \begin{acronym}[DCGAN]
    \acro{API}{Application Programming Interface}
    \acro{CNN}{Convolutional Neural Network}
    \acro{CV}{Computer Vision}
    \acro{DCGAN}{Deep Convolutional Generative Adversarial Network}
    \acro{GAN}{Generative Adversarial Network}
    \acro{LSTM}{Long Short-Term Memory}
    \acro{MLP}{Multilayer Perceptron}
    \acro{MSE}{Mean Squared Error}
    \acro{NAS}{Neural Architecture Search}
    \acro{NLP}{Natural Language Processing}
    \acro{NN}{Neural Network}
    \acro{ReLU}{Rectified Linear Unit}
    \acro{ViT}{Vision Transformer}
  \end{acronym}
  \newpage
  \pagenumbering{arabic}
\colorlet{blu20}{blue!20}
\colorlet{blu40}{blue!40}
\colorlet{blu60}{blue!60}
\colorlet{blu80}{blue!80}

\colorlet{grn20}{green!20}
\colorlet{grn40}{green!40}
\colorlet{grn60}{green!60}
\colorlet{grn80}{green!80}

\colorlet{red20}{red!20}
\colorlet{red40}{red!40}
\colorlet{red60}{red!60}
\colorlet{red80}{red!80}

\colorlet{yel20}{yellow!20}
\colorlet{yel40}{yellow!40}
\colorlet{yel60}{yellow!60}
\colorlet{yel80}{yellow!80}

\definecolor{purple}{RGB}{106,13,173}
\colorlet{pur20}{purple!20}
\colorlet{pur40}{purple!40}
\colorlet{pur60}{purple!60}
\colorlet{pur80}{purple!80}

\colorlet{ora20}{orange!20}
\colorlet{ora40}{orange!40}
\colorlet{ora60}{orange!60}
\colorlet{ora80}{orange!80}

\colorlet{cya20}{cyan!20}
\colorlet{cya40}{cyan!40}
\colorlet{cya60}{cyan!60}
\colorlet{cya80}{cyan!80}

\colorlet{mag20}{magenta!20}
\colorlet{mag40}{magenta!40}
\colorlet{mag60}{magenta!60}
\colorlet{mag80}{magenta!80}

\colorlet{vio20}{violet!20}
\colorlet{vio40}{violet!40}
\colorlet{vio60}{violet!60}
\colorlet{vio80}{violet!80}

\colorlet{tea20}{teal!20}
\colorlet{tea40}{teal!40}
\colorlet{tea60}{teal!60}
\colorlet{tea80}{teal!80}

\colorlet{gry20}{gray!20}
\colorlet{gry40}{gray!40}
\colorlet{gry60}{gray!60}
\colorlet{gry80}{gray!80}

\colorlet{blk20}{black!20}
\colorlet{blk40}{black!40}
\colorlet{blk60}{black!60}
\colorlet{blk80}{black!80}

\colorlet{good}{green!10!white}
\colorlet{okay}{yellow!10!white}
\colorlet{bad}{red!10!white}

\tikzset{
    arrow/.style = {-latex, thick},
    arrowr/.style = {latex-, thick},
    arrowd/.style = {latex-latex, thick},
}

\tikzset{
    vertical centered/.style = {midway, rotate=90, align=center, anchor=center},
}

\tikzset{
    box/.style 2 args = {rectangle, draw, thick, minimum width=#1, minimum height=#2, text width=#1, align=center},
    boxv/.style 2 args = {rectangle, draw, thick, minimum width=#1, minimum height=#2, text width=#1, align=center, rotate=90, anchor=center},
    img/.style = {inner sep=0pt},
}

\tikzset{
    minheight/.store in = \minheight, minheight = 1cm,
    minwidth/.store in = \minwidth, minwidth = 1cm,
    padding/.store in = \padding, padding = 5pt,
}

\tikzset{
    cuboid/.pic = {
        \tikzset{%
            every edge quotes/.append style={midway, auto},
            every edge/.append style={pic actions, densely dashed, opacity=0.25},
            /cuboid/.cd,
            #1
        }
        \draw [pic actions, line join=bevel]
        (\cubescale*\cubex*0.5, \cubescale*\cubey*0.5, \cubescale*\cubez*0.5) coordinate (o) -- ++(-\cubescale*\cubex,0,0) coordinate (a) -- ++(0,-\cubescale*\cubey,0) coordinate (b) edge coordinate [pos=1] (g) ++(0,0,-\cubescale*\cubez)  -- ++(\cubescale*\cubex,0,0) coordinate (c) -- cycle
        (o) -- ++(0,0,-\cubescale*\cubez) coordinate (d) -- ++(0,-\cubescale*\cubey,0) coordinate (e) edge (g) -- (c) -- cycle
        (o) -- (a) -- ++(0,0,-\cubescale*\cubez) coordinate (f) edge (g) -- (d) -- cycle;
        \ifthenelse{\equal{\cubedraft}{true}}{%
        \path node at (o) {o} node at (a) {a} node at (b) {b} node at (c) {c} node at (d) {d} node at (e) {e} node at (f) {f} node at (g) {g};%
        }{}
    },
    /cuboid/.search also={/tikz},
    /cuboid/.cd,
    width/.store in=\cubex,
    height/.store in=\cubey,
    depth/.store in=\cubez,
    scale/.store in=\cubescale,
    draft/.store in=\cubedraft,
    width=10,
    height=10,
    depth=10,
    scale=.1,
    draft=false,
}

\tikzset{
    pic/.style = {
        local bounding box=#1,
    },
}

\tikzset{
    cnn basic/.style = {trapezium, draw, inner sep=\padding, text width=#1, text centered, minimum width=\minheight, minimum height=\minwidth, trapezium angle=75, trapezium stretches=true, shape border rotate=270},
    cnn basic/.default = {1.5cm},
}

\makeatletter
\newcommand{\ifempty}[1]{\ifnum\pdf@strcmp{#1}{}=\z@}
\makeatother


\def\getlen#1{%
\pgfmathsetmacro{\lenarray}{0}%
\foreach \i in #1{%
\pgfmathtruncatemacro{\lenarray}{\lenarray+1}%
\global\let\lenarray\lenarray}%
}

\tikzset{
    fc/.pic = {
        \tikzset{
            every node/.append style={scale=\fcscale},
            /fc/.cd,
            #1
        }

        \def\dotsdefs{}
        \pgfmathsetmacro{\lastnnodes}{0}
        \pgfmathsetmacro{\xoffset}{0}
        \pgfmathsetmacro{\layer}{0}
        \getlen{\nodedef}\pgfmathsetmacro{\nodedeflen}{\lenarray}

        \ifthenelse{\equal{\fcvertical}{true}}{
            \pgfmathsetmacro{\verticaldots}{"\hdots"}
            \pgfmathsetmacro{\horizontaldots}{"\vdots"}
        }{
            \pgfmathsetmacro{\verticaldots}{"\vdots"}
            \pgfmathsetmacro{\horizontaldots}{"\hdots"}
        }

        \pgfmathsetmacro{\maxnodes}{0}
        \def\layerlens{} 
        \foreach \j in {1,...,\nodedeflen}
        {
            \pgfmathsetmacro{\n}{{\nodedef}[\j-1][0]}
            \pgfmathsetmacro{\labeldef}{{\nodedef}[\j-1][1]}
            \ifthenelse{\equal{\labeldef}{.}}{
            }{
                \ifnum\n=0
                    \pgfmathsetmacro{\thislayernodes}{3.5}
                \else
                    \pgfmathsetmacro{\thislayernodes}{\n}
                \fi
                \xappto{\layerlens}{\thislayernodes,}
                \pgfmathsetmacro{\maxnodes}{max(\maxnodes, \thislayernodes)}
                \global\let\maxnodes\maxnodes
            }
        }

        \foreach \j in {1,...,\nodedeflen}
        {
            \pgfmathsetmacro{\n}{{\nodedef}[\j-1][0]}
            \pgfmathsetmacro{\labeldef}{{\nodedef}[\j-1][1]}

            \pgfmathsetmacro{\ll}{\layer}
            \ifthenelse{\equal{\labeldef}{.}}{
            }{
                \pgfmathsetmacro{\layer}{\intcalcAdd{\layer}{1}}
            }
            \global\let\layer\layer

            \pgfmathsetmacro{\layeryoffset}{-0.5 * ({\layerlens}[\layer-1] - 1)}

            \ifnum\n=0
                \def\nodes{1,2,".","n"}
            \else
                \def\nodes{1}
                \foreach \d in {2,...,\n} {
                    \xappto{\nodes}{,\d}
                }
            \fi

            \pgfmathsetmacro{\yoffset}{0}
            \ifnum\layer=\ll
            \else
                \pgfmathsetmacro{\nnodes}{1}
            \fi
            \global\let\nnodes\nnodes
            \pgfmathsetmacro{\nnodesd}{1}
            \global\let\nnodesd\nnodesd
            \getlen{\nodes}\pgfmathsetmacro{\nnodesdef}{\lenarray}
            \foreach \i in {1,...,\nnodesdef}
            {
                \pgfmathsetmacro{\nd}{{\nodes}[\i-1]}
                \ifthenelse{\equal{\nd}{.}}{
                    \ifthenelse{\equal{\labeldef}{.}}{
                    }{
                        \pgfmathsetmacro{\lastnode}{\intcalcSub{\nnodes}{1}}
                        \xappto{\dotsdefs}{{"\layer-\lastnode", "\layer-\nnodes", "\verticaldots"},}
                    }
                    \pgfmathsetmacro{\yoffset}{\yoffset + 0.5}
                }{
                    \ifthenelse{\equal{\labeldef}{.}}{
                        \pgfmathsetmacro{\nextlayer}{\intcalcAdd{\layer}{1}}
                        \xappto{\dotsdefs}{{"\layer-\nnodesd", "\nextlayer-\nnodesd", "\horizontaldots"},}
                        \pgfmathsetmacro{\yoffset}{\yoffset + 0.5}
                        \pgfmathsetmacro{\nnodesd}{\intcalcAdd{\nnodesd}{1}}
                        \global\let\nnodesd\nnodesd
                    }{
                        \pgfmathparse{dim({\nodedef}[\j-1])}
                        \ifthenelse{\equal{\labeldef}{-}}{
                            \def\nodelabel{}
                        }{
                            \def\nodelabel{\labeldef_{\nd}}
                            \ifnum\pgfmathresult>2
                                \pgfmathsetmacro{\labeldefsup}{{\nodedef}[\j-1][2]}
                                \ifthenelse{\equal{\labeldefsup}{-}}{
                                    \def\labeldefsup{}
                                }{}
                                \ifnum\pgfmathresult=4
                                    \pgfmathsetmacro{\labeldefsub}{{\nodedef}[\j-1][3]}
                                    \def\labeldefsubn{\labeldefsub}
                                \else
                                    \def\labeldefsub{\nd}
                                    \pgfmathsetmacro{\labeldefsubn}{"\labeldefsub^{\labeldefsup}"}
                                \fi
                                \ifthenelse{\equal{\nd}{n}}{
                                    \def\nodelabel{\labeldef_{\labeldefsubn}^{\labeldefsup}}
                                }{
                                    \def\nodelabel{\labeldef_{\nd}^{\labeldefsup}}
                                }
                            \fi
                        }
                        \ifthenelse{\equal{\fcvertical}{true}}{
                            \pgfmathsetmacro{\xtrueoffset}{\yoffset + \layeryoffset}
                            \pgfmathsetmacro{\ytrueoffset}{\xoffset}
                        }{
                            \pgfmathsetmacro{\xtrueoffset}{\xoffset}
                            \pgfmathsetmacro{\ytrueoffset}{\yoffset + \layeryoffset}
                        }
                        \node
                        [pic actions, inner sep=0pt, circle, minimum size=\fccirclescale cm] (-node-\layer-\nnodes)
                        at
                        ({\fchscale * \xtrueoffset * \fcscale}, {-\fcvscale * \ytrueoffset * \fcscale})
                        {}
                        node at (-node-\layer-\nnodes) [align=center, anchor=center] {$\nodelabel$};
                        \ifnum\lastnnodes=0
                        \else
                            \pgfmathsetmacro{\l}{\intcalcSub{\lastnnodes}{1}}
                            \foreach \k in {1,...,\l}{
                                \draw [pic actions] (-node-\ll-\k) -- (-node-\layer-\nnodes);
                            }
                        \fi
                        \pgfmathsetmacro{\yoffset}{\yoffset + 1}
                        \pgfmathsetmacro{\nnodes}{\intcalcAdd{\nnodes}{1}}
                        \global\let\nnodes\nnodes
                    }
                }
                \global\let\yoffset\yoffset
            }
            \ifthenelse{\equal{\labeldef}{.}}{
                \pgfmathsetmacro{\xoffset}{\xoffset + 0.5}
            }{
                \pgfmathsetmacro{\xoffset}{\xoffset + 1}
            }
            \global\let\xoffset\xoffset
            \global\let\lastnnodes\nnodes
        }
        \foreach \dd in \dotsdefs
        {
            \ifthenelse{\equal{\dd}{}}{}{
                \pgfmathsetmacro{\drawstart}{{\dd}[0]}
                \pgfmathsetmacro{\drawend}{{\dd}[1]}
                \pgfmathsetmacro{\drawtext}{{\dd}[2]}
                \node [align=center, anchor=center, yshift={\fcscale * \fchscale * 0.1 cm}] at ($(-node-\drawstart)!0.5!(-node-\drawend)$) {$\drawtext$};
            }
        }
    },
    /fc/.search also={/tikz},
    /fc/.cd,
    scale/.store in=\fcscale,
    hscale/.store in=\fchscale,
    vscale/.store in=\fcvscale,
    circlescale/.store in=\fccirclescale,
    nodedef/.store in=\nodedef,
    scale=1.0,
    hscale=1.0,
    vscale=1.0,
    circlescale=0.6,
    nodedef={{0, "x"}, {0, "z", "(1)"}, {0, "."}, {0, "z", "(l)"}, {3, "y"}},
    vertical/.store in=\fcvertical,
    vertical/.default = true,
    vertical=false,
}

\newcommand{\rotateRPY}[4][0/0/0]
{   \pgfmathsetmacro{\rollangle}{#2}
    \pgfmathsetmacro{\pitchangle}{#3}
    \pgfmathsetmacro{\yawangle}{#4}

    \pgfmathsetmacro{\newxx}{cos(\yawangle)*cos(\pitchangle)}
    \pgfmathsetmacro{\newxy}{sin(\yawangle)*cos(\pitchangle)}
    \pgfmathsetmacro{\newxz}{-sin(\pitchangle)}
    \path (\newxx,\newxy,\newxz);
    \pgfgetlastxy{\nxx}{\nxy}

    \pgfmathsetmacro{\newyx}{cos(\yawangle)*sin(\pitchangle)*sin(\rollangle)-sin(\yawangle)*cos(\rollangle)}
    \pgfmathsetmacro{\newyy}{sin(\yawangle)*sin(\pitchangle)*sin(\rollangle)+ cos(\yawangle)*cos(\rollangle)}
    \pgfmathsetmacro{\newyz}{cos(\pitchangle)*sin(\rollangle)}
    \path (\newyx,\newyy,\newyz);
    \pgfgetlastxy{\nyx}{\nyy}

    \pgfmathsetmacro{\newzx}{cos(\yawangle)*sin(\pitchangle)*cos(\rollangle)+ sin(\yawangle)*sin(\rollangle)}
    \pgfmathsetmacro{\newzy}{sin(\yawangle)*sin(\pitchangle)*cos(\rollangle)-cos(\yawangle)*sin(\rollangle)}
    \pgfmathsetmacro{\newzz}{cos(\pitchangle)*cos(\rollangle)}
    \path (\newzx,\newzy,\newzz);
    \pgfgetlastxy{\nzx}{\nzy}

    \foreach \x/\y/\z in {#1}
    {   \pgfmathsetmacro{\transformedx}{\x*\newxx+\y*\newyx+\z*\newzx}
        \pgfmathsetmacro{\transformedy}{\x*\newxy+\y*\newyy+\z*\newzy}
        \pgfmathsetmacro{\transformedz}{\x*\newxz+\y*\newyz+\z*\newzz}
        \xdef\savedx{\transformedx}
        \xdef\savedy{\transformedy}
        \xdef\savedz{\transformedz}
    }
}
\tikzset{RPY/.style={x={(\nxx,\nxy)},y={(\nyx,\nyy)},z={(\nzx,\nzy)}}}

\tikzset{
    reverseclip/.style={
        clip,
        insert path={{
            (-16383pt, -16383pt) rectangle (16383pt, 16383pt)
        }}
    },
}

\NiceMatrixOptions{
    custom-line = {
        letter = : ,
        command = hdashline,
        ccommand = chdashline,
        tikz = dashed
    }
}

  \chapter{Introduction}
\label{c:intro}

Because people rely heavily on their vision, eyes and their movements are an important part of human communication. As a result, \et\ as a technique for measuring and recording eye movements, gaze direction, and thus attention, has been used in many different fields, including psychology and medicine, as well as human-computer interaction and usability research.~\cites[\ppno\,3--13]{Duchowski2017}{Hansen2010}{Punde2017}{Stember2019}.

Machine Learning algorithms, and in particular Deep Artificial Neural Networks, have become increasingly popular in recent years. They are widely used and can be applied to many different tasks including image classification, object detection, audio processing, and even text generation.

The realm of \et\ or \gt, which was previously dominated by analytical methods, as shown in~\cites[\ppno\,49-57]{Duchowski2017}{FunesMora2014b}{Hansen2010}{Punde2017}{Xiong2014}, has also received a significant boost with the advent of Deep Learning~\cite{Salman2022}. Especially vision-based \gt\ with sensors distant from the person's eyes has benefited from the use of state-of-the-art \acp{CNN}. Previous research by \citeauthor*{Ansari2021}~\cite{Ansari2021}, \citeauthor*{He2015}~\cite{He2015}, \citeauthor*{Krafka2016}~\cite{Krafka2016}, \citeauthor*{Salman2022}~\cite{Salman2022}, and many more proposed different approaches to the task of \gt\ using \acp{CNN}. An overview of the different techniques -- including non-vision-based methods -- used for \gt\ is given in \Cref{c:relwork} and~\cites{Hansen2010}{Punde2017}.

Transformers~\cite{Vaswani2017} are a recent architecture for \acp{NN} with great success in the field of \ac{NLP}. However, they are also applicable to other domains, such as \ac{CV}, as shown in~\cite{Dosovitskiy2020}. There have been attempts to use Transformers for \gt\ as well~\cites{Cheng2021}{Li2023}{Nagpure2023}, but they have only been applied to RGB images so far. In this thesis, we will investigate the use of Transformers for \gt\ on RGBD images and compare them to an RGB-only baseline to demonstrate the improvements by using depth information in the \ges\ process.

\pagebreak
\section{Problem Statement and Objectives}
\label{s:intro:problem}

The task of \gt\ can be defined as the estimation of the direction of a person's gaze in the 3D space. This is usually done by estimating the gaze angle, i.e., the angle between the person's line of sight and the optical axis of the camera. Subsequently, the gaze point, e.g., on a computer screen, can be calculated by intersecting the estimated gaze ray with the screen plane. The approach of predicting gaze angles is widely used:~\cites{Chen2019}{Chen2020}{Cheng2020}{FunesMora2014b}{Gudi2020}{Li2023}{Liu2019}{Zhang2015}. Previous research partly used other approaches to calculate the gaze point, e.g., in~\cite{Lian2019} and~\cite{Salman2022} the authors trained their \ac{CNN} to predict 2D screen coordinates and the region in a $4\times4$ grid, respectively. In this thesis, we will use a combination of \acp{CNN} and Transformers to estimate the gaze angles.

The task of this thesis is divided into three components:
\begin{itemize}[labelwidth=\widthof{\bfseries Pipeline:}]
    \item[\textbf{Dataset:}] Create a dataset of RGBD images for \gt\ to train and evaluate \dl\ models.
    \item[\textbf{Model:}] Design, train and evaluate models combining \acp{CNN} and Transformers for the \ges\ task on RGBD images.
    \item[\textbf{Pipeline:}] Implement a pipeline that uses a trained model to estimate the gaze angles and gaze points on a computer screen in real-time.
\end{itemize}

Parts of the application pipeline can be used to create the labels for the dataset. In particular, the preprocessing steps of the pipeline are also required for dataset creation and labeling. Therefore, this thesis will focus on the processing sequence (\Cref{s:processing:sequence}) first and then on the data collection process (\Cref{s:processing:datacollection}). We present the application of our pipeline in~\Cref{s:evaluation:pipeline}.

\section{Structure of the Thesis}
\label{s:intro:structure}

The remainder of this thesis is structured as follows: \Cref{c:relwork} provides an overview of related work in the field of \et, existing datasets for \gt, \dl\ architectures and previous approaches to \gt\ using \dl\ and \ml\ methods. In~\Cref{c:processing}, we describe the processing steps for our real-time gaze point estimation pipeline. This includes a detailed description of the model architecture, the data collection process, and the creation of our own dataset. In~\Cref{c:evaluation}, we present the evaluation results of the models on the \xgaze\ dataset~\cite{Zhang2020}, the \stg\ dataset~\cite{Lian2019}, and our own dataset. Furthermore, we describe our gaze estimation pipeline that facilitates the trained models. Finally, we conclude with a summary of the results and an outlook on future work as well as improvements in~\Cref{c:conclusion}.
  \chapter{Related Work}
\label{c:relwork}

In this chapter, we present four aspects of related work in the fields of \gt\ and \dl. First, we provide an introduction to \gt\ and classify the focus of this thesis, vision-based \gt, into the general topic. Then, we give an overview of some existing datasets for the \ges\ task and discuss the need for a custom dataset. In the following section, we introduce some of the \dl\ architectures that form the basis of the models used in the literature and in this thesis. Finally, we present some previous approaches to vision-based \gt\ using \dl\ and \ml\ methods, and discuss their similarities and differences.

\section{Gaze Tracking Approaches}
\label{s:relwork:gaze}

\et\ or \gt\ techniques can be categorized in several ways. We found that there are four main ways in which \gt\ methods differ:
\begin{itemize}[labelwidth=\widthof{\bfseries Estimation Process:}]
    \item[\textbf{Sensor Placement:}] head-mounted~\cite[\ppno\,49--54]{Duchowski2017},\\table-mounted with chin rest~\cite[\ppno\,54--56]{Duchowski2017}, or\\remote sensors~\cite{Punde2017}.
    \item[\textbf{Sensor Type:}] eyeball movement tracking~\cite[\ppno\,51\psq]{Duchowski2017},\\(near-)infrared-based pupil center/corneal reflection tracking \cites[\ppno\,52\psqq]{Duchowski2017}{Hansen2010}{Punde2017}, or\\appearance-based tracking using RGB(D)\,cameras\,\cite[\ppno\,52\psq]{Duchowski2017}.
    \item[\textbf{Estimation Process:}] landmark-based analytical methods using geometric models of faces and eyes~\cites[\ppno\,49--57]{Duchowski2017}{FunesMora2014b}{Hansen2010}{Xiong2014}{Zhang2017}, or\\learning-based methods using \acp{NN}~\cites{Zhang2015}{Zhang2017} (see~\Cref{s:intro:problem}).
    \item[\textbf{Estimation Target:}] gaze angles in 3D space~\cites{Hansen2010}{Zhang2015}{Zhang2020}{Zhang2017},\\gaze points on a 2D screen~\cites{Krafka2016}{Lian2019}{Salman2022}{Zhang2017},\\gaze regions in the subject's field of view~\cite{Harms2021}, or\\in-frame or in-scene objects~\cite{Tu2022}.
\end{itemize}

In this thesis, we develop a vision-based \gt\ system using an RGBD camera and \dl\ models. Therefore, we can categorize our approach as follows:
\begin{itemize}
    \item The sensor (camera) is placed remote, below a computer screen.
    \item We use a single RGBD camera for appearance-based \et.
    \item We use learning-based models built upon \acp{CNN} and Transformers.
    \item The model estimates 3D gaze angles and we then calculate the 2D gaze points.
\end{itemize}

In order to get to know the different techniques for \gt\ better, we will discuss them briefly. Furthermore, we will glance over the advantages and disadvantages of each method. \Cref{t:relwork:gaze:methods} gives an overview of the different \gt\ techniques, their characteristics, and their typical accuracy ranges.

\begin{table}[htbp]
    \centering
    \renewcommand{\arraystretch}{1.2}
    \begin{tabular}{>{\centering\arraybackslash}m{3.47cm}@{\hspace{3.5mm}}>{\centering\arraybackslash}m{3.1cm}@{}>{\centering\arraybackslash}m{1.6cm}@{}>{\centering\arraybackslash}m{3cm}@{\hspace{3.5mm}}>{\centering\arraybackslash}m{3.7cm}}
        \textbf{Method} & \textbf{Sensor} & \textbf{Invasive} & \textbf{Estimation} & \textbf{Accuracy} \\
        \hline\hline
        Eyeball Movement & Contact Lens & \cmark & Analytical & \ang{0.11}~\cite{Massin2020} \\\hline
        Pupil Center/ Corneal Reflection & Infrared Light +~Camera & \xmark & Analytical & \ang{0.15}~to~\ang{1.2} \cites{SRResearch2023,Zhang2019} \\\hline
        \multirow{2}{*}{Appearance-based} & \multirow{2}{*}{RGB(D) Camera} & \multirow{2}{*}{\xmark} & Analytical & \ang{2}~to~\ang{4} \cite{Zhang2019} \\\cline{4-5}
        & & & Learning-based & \ang{3}~to~\ang{10} \cites{Cai2019,Zhang2020} \\
    \end{tabular}
    \caption{Overview of different \gt\ techniques.}
    \label{t:relwork:gaze:methods}
\end{table}

The \textbf{eyeball movement tracking} technique requires special contact lenses which are connected to a measurement device either physically or by the use of electromagnetic fields~\cites[\ppno\,51]{Duchowski2017}{Massin2020}. Although this technique is very accurate, its range is limited to a few degrees due to the lens slipping and it is a very invasive method~\cite[\ppno\,51]{Duchowski2017}. \Citeauthor*{Massin2020}~\cite{Massin2020} reported an accuracy value of \ang{0.11} for a newly developed scleral contact lens incorporating photoreceptors.

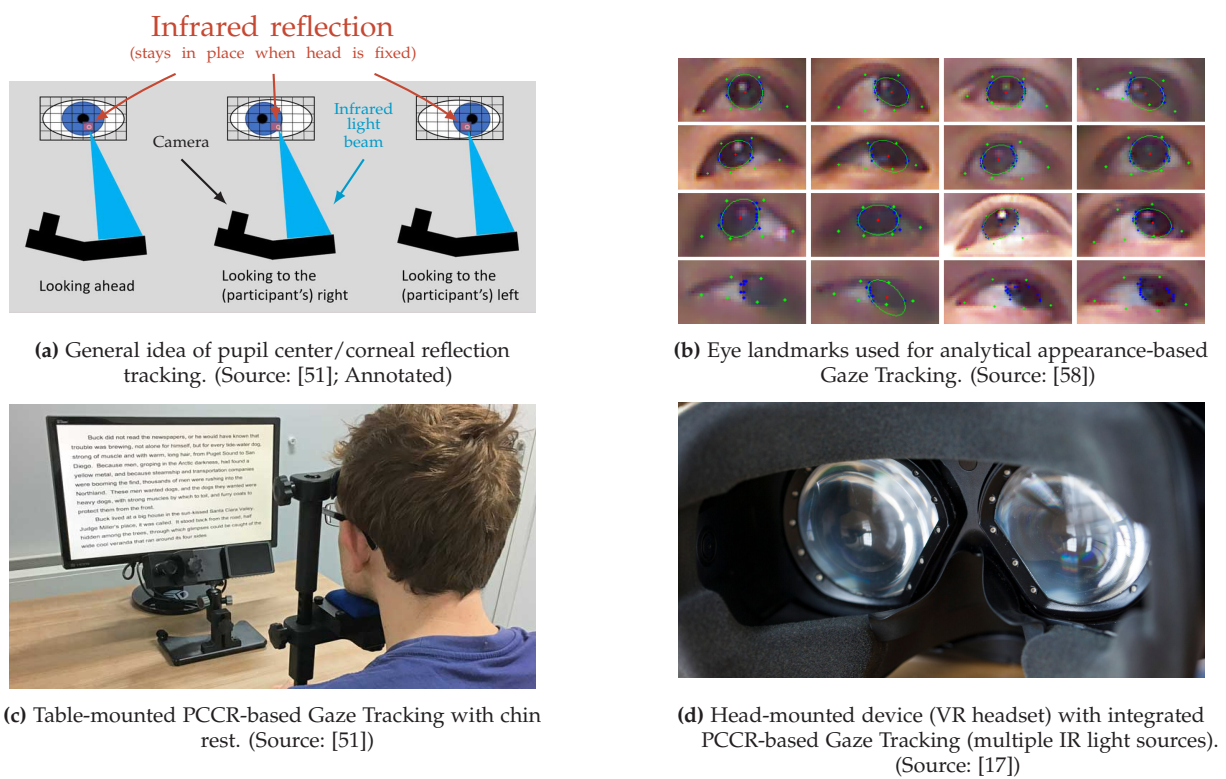
\begin{figure}[htb]
    \vspace*{5mm}
    \centering
    \begin{subfigure}[t]{0.45\linewidth}
        \centering
        \begin{tikzpicture}
            \node at (0, 0) {\includegraphics[width=0.96\linewidth]{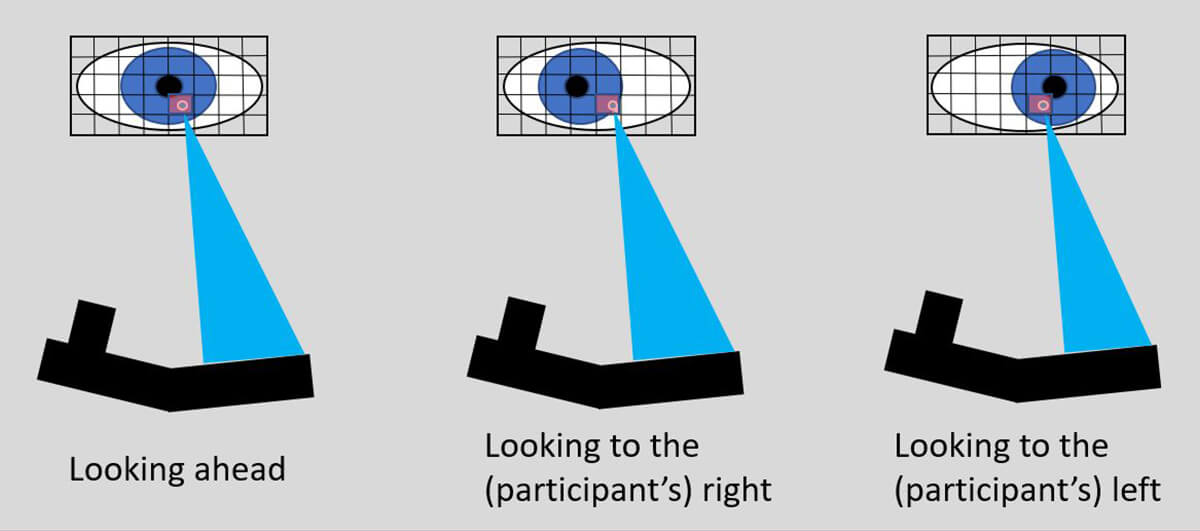}};
            \draw [arrowr, cyan!80!black] (0.8, -0.1) -- (1.2, 0.5) node [above=0cm, text width=1cm, align=center] {\tiny\baselineskip=0pt Infrared light beam\par};
            \draw [arrowr, black] (-0.6, -0.1) -- (-1.2, 0.5) node [above=0cm, text width=1cm, align=center] {\tiny\baselineskip=0pt Camera\par};
            \node [text=red!80!black, text width=4cm, align=center] (reflection) at (0, 2.1) {\small\baselineskip=3mm Infrared reflection\\{\tiny (stays in place when head is fixed)}\par};
            \draw [arrow, red!80!black] (reflection) to [bend right=1] (0.05, 1);
            \draw [arrow, red!80!black] (reflection) to [bend left=5] (2.45, 1);
            \draw [arrow, red!80!black] (reflection) to [bend right=5] (-2.35, 1);
        \end{tikzpicture}
        \caption{General idea of pupil center/corneal reflection tracking. (Source:~\cite{SRResearch2023}; Annotated)}
        \label{f:relwork:gaze:methods:pcr}
    \end{subfigure}
    \hfill
    \begin{subfigure}[t]{0.45\linewidth}
        \centering
        \includegraphics[width=0.96\linewidth]{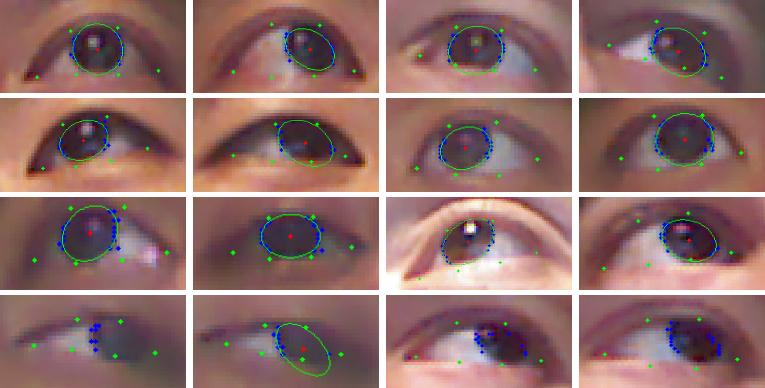}
        \caption{Eye landmarks used for analytical appearance-based \gt. (Source:~\cite{Xiong2014})}
        \label{f:relwork:gaze:methods:elm}
    \end{subfigure}
    \begin{subfigure}[t]{0.45\linewidth}
        \centering
        \includegraphics[width=0.96\linewidth,clip,trim={5cm 0cm 3cm 0cm}]{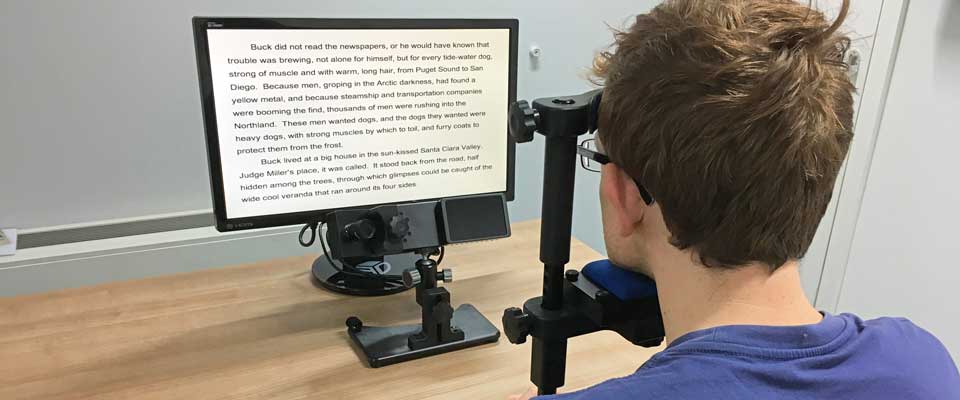}
        \caption{Table-mounted PCCR-based \gt\ with chin rest. (Source:~\cite{SRResearch2023})}
        \label{f:relwork:gaze:methods:chinrest}
    \end{subfigure}
    \hfill
    \begin{subfigure}[t]{0.45\linewidth}
        \centering
        \begin{tikzpicture}
            \node at (0, 0) {\includegraphics[width=0.96\linewidth,clip,trim={0cm 5cm 0cm 5cm}]{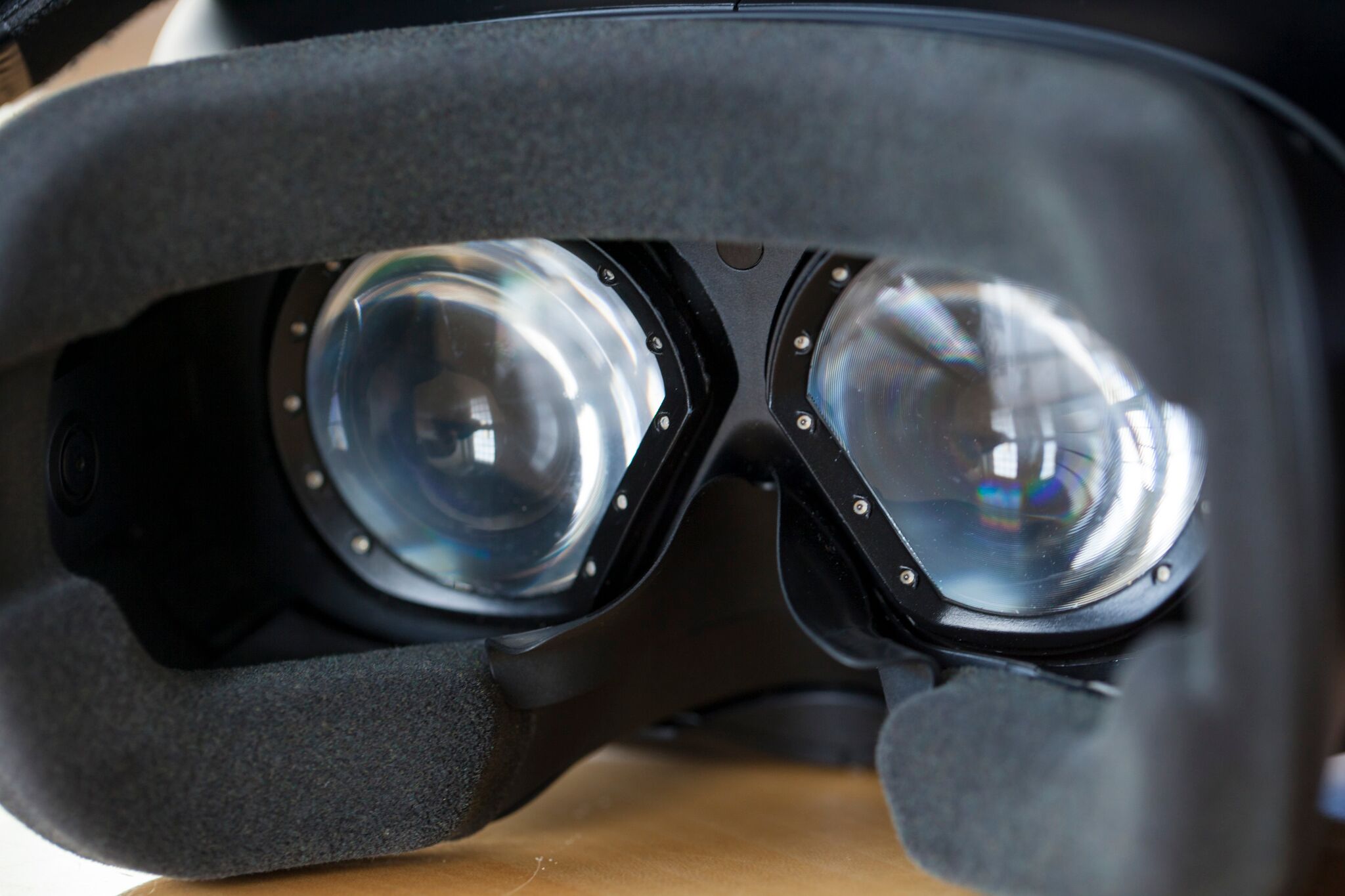}};
        \end{tikzpicture}
        \caption{Head-mounted device (VR headset) with integrated PCCR-based \gt\ (multiple IR light sources). (Source:~\cite{Forrest2017})}
        \label{f:relwork:gaze:methods:hmd}
    \end{subfigure}

    \caption{Illustration of different \et\ techniques.}
    \label{f:relwork:gaze:methods}
\end{figure}

A less intrusive \et\ technique is the method of \textbf{pupil center/corneal reflection tracking} using (near-)infrared light. It offers high accuracy and is commercially available for researchers as well as end users. It works by shining infrared light onto the subject's eyes and simultaneously recording the eyes using an appropriate camera. Then, the difference between the position of the pupil center and the corneal reflection can be utilized to calculate the gaze direction. \Cref{f:relwork:gaze:methods:pcr} visualizes this approach.

Commercial vendors claim an accuracy of down to \ang{0.15} from typically \ang{0.50}~\cite{SRResearch2023}, but (free) head movement is still a challenge for this method. \Citeauthor*{Zhang2019}~\cite{Zhang2019} reported an average angular error of \ang{0.8} to \ang{1.2}. Head-mounted versions of this method offer better accuracy because they keep the relative position between the camera and the eyes. Alternatively, a subject's head can be fixed, e.g., with a chin rest. \Cref{f:relwork:gaze:methods:chinrest} shows a table-mounted \et\ device with a subject resting on the chin rest. An~implementation of \et\ technology into a head-mounted device is depicted in \Cref{,f:relwork:gaze:methods:hmd}. Due to the reliance on the infrared light source, this method is not suited for outdoor lighting conditions as direct and indirect sunlight also affects the (near-)infrared spectrum~\cites{Duchowski2017}{Hansen2010}{Punde2017}{SRResearch2023}.

A generally cheaper and, thus, more accessible method of \et\ is \textbf{appearance-based tracking} using standard RGB or RGBD cameras. With this method, the gaze direction is estimated by analyzing the appearance of the eyes and the face. This method is non-intrusive as it is completely passive unless an illuminating RGBD camera is used. However, the accuracy of this method is lower than the accuracy of the other methods, ranging anywhere from \ang{2} to \ang{4} in experiments with perfect lighting conditions~\cite{Zhang2019}.

Appearance-based tracking can be further divided into two subcategories: \textbf{analytical} and \textbf{learning-based} methods. Typically, analytical methods use geometric models of faces and eyes and try to find within-eye landmarks, such as iris and pupil, to fit these geometric models to the image, as shown in \Cref{f:relwork:gaze:methods:elm}~\cites{Hansen2010}{Xiong2014}. Learning-based methods employ \ml\ or \dl\ techniques to estimate the gaze direction instead of computing it analytically. Due to the wide variety of datasets, the reported accuracy values of learning-based methods vary greatly. Depending on the dataset and model, the mean angular error ranges from about \ang{3}~\cite{Cai2021} up to \ang{10}~\cite{Zhang2020}.

In this thesis, we develop a learning-based \gt\ system and use publicly available datasets to train and evaluate our \dl\ models. Since the reported accuracy values are highly dependent on the dataset, we will compare our model with other models that were trained on the same dataset.

\section{Gaze Tracking Datasets}
\label{s:relwork:datasets}

There are a variety of datasets available for the task of \ges. In this section, we will present some of the recently introduced datasets and discuss their features and limitations. We will only consider publicly available datasets that fit our task of \ges\ using a single remote sensor for appearance-based \et. However, we include datasets containing RGB images only, as there is no dataset available that contains high-resolution RGBD images and gaze angle labels. This is why we have to create our own dataset, which is described in~\Cref{s:processing:dataset}.

\subsection{\ed}
\label{s:relwork:datasets:ed}

The \ed\ dataset was published in 2014 by \citeauthor*{FunesMora2014a}~\cite{FunesMora2014a}. It consists of 94 session videos (2--3 minutes each) of 16 subjects (12 male, 4 female) in up to two lighting conditions. The subjects were asked to perform three different tasks: discrete screen target (random points on screen), continuous screen target (point moving along path on screen) and 3D floating target (floating ball). Each of these tasks was performed in two different conditions: with and without head movement. The dataset provides 2D and 3D gaze target coordinates.

In this thesis, we will not use the \ed\ dataset, as it is not suitable for our task. The images are of low resolution ($640\times480$ pixels) and require additional synchronization and calibration steps due to the capture device used. In addition, \citeauthor*{Zhang2015} noted that \ed\ \enquote{does not cover the range of gaze directions that can occur during laptop interactions}~\cite{Zhang2015}, which indicates that the dataset is not suitable for training a gaze angle estimation model for (larger) computer screens, too.

\subsection{\mpii}
\label{s:relwork:datasets:mpii}

\Citeauthor*{Zhang2015}~\cite{Zhang2015} introduced \mpii\ in 2015. The authors created an in-the-wild dataset which images were recorded outside of controlled conditions in order to capture a wider variety of lighting situations, head poses, and potential backgrounds. In total, 213,659 images of 15 subjects were recorded using different laptop models. The authors took care of the different laptop models and provided additionally to the 2D on-screen coordinates the 3D coordinates of the gaze target.

The dataset contains RGB images with a resolution of $1280\times720$ pixels and no depth information. The published version of the dataset contains only crops to the eye region of each face, effectively removing the background and the rest of the face. Thus, the dataset is not suitable for our task because we need full frames or at least face crops to train our model. Furthermore, the images are normalized in a different way than in \xgaze~\cite{Zhang2020}, which is the normalization process that is used in this thesis (see~\Cref{s:relwork:ml:normalization,s:processing:sequence}).

\subsection{\xgaze}
\label{s:relwork:datasets:xgaze}

In 2020, \citeauthor*{Zhang2020}~\cite{Zhang2020} presented the \xgaze\ dataset consisting of 1,083,492 images. 110 subjects (47 female, 63 male) were recorded using 18 high-resolution cameras in a controlled environment with a fixed head pose. There are different, controlled lighting conditions. The dataset is available in different versions: the original version with full frames and the face-normalized version containing only face crops. The authors provided gaze angles in the camera coordinate system as labels. Furthermore, the authors divided the dataset into three disjoint subsets: a training set (80 subjects), a test set (15 subjects), and a test set for person-specific evaluation (15 subjects). Apart from 200 calibration samples for the person-specific test set, the authors released no ground truth gaze angles for any sample of the two test sets.

In this thesis, we use the face-normalized version with face crops of $448\times448$ pixels due to the enormous full-frame dataset size. The dataset does not contain any depth information, which limits the application to our task. Nevertheless, using the dataset's high-resolution and high-quality RGB images enables us to compare the results of our model on our dataset with the results of our model on the \xgaze\ dataset. Because the subjects had a relatively fixed head pose during the recording, the dataset might not be suitable for training a model to be used in a real-world scenario with free and continuous head movement.

\subsection{\stg}
\label{s:relwork:datasets:stg}

The \stg\ dataset was introduced by \citeauthor*{Lian2019}~\cite{Lian2019} in 2019 and consists of 165,231 RGBD images from 218 subjects (77 female, 141 male). The authors used an Intel RealSense SR300 RGBD camera, which was placed below a computer screen, to capture the images. No specific restrictions on head pose were enforced, but no instructions were given to move the head into as many positions as possible, either. The dataset is not face-normalized, but rather the authors took crops of the faces and resized them to a resolution of $224\times224$ pixels. To overcome missing data artifacts in the depth images, the authors proposed a special \dl\ model architecture which will be described later in \Cref{s:relwork:ml:stg}. The authors also provided a breakdown into a training subset (159 subjects) and a test subset (59 subjects). A limitation of this dataset are the labels provided by the authors. Only 2D on-screen coordinates (metric scale) are included and there is no usable information about the camera intrinsic and extrinsic parameters which would allow for a conversion to 3D world coordinates.

In terms of experiment setup, the work of \citeauthor*{Lian2019} is similar to ours. However, the dataset is not face-normalized as in our work, and the labels are suboptimal for our task. In particular, the lack of 3D gaze angles, or at least 3D gaze target coordinates, makes the dataset specifically tied to the authors' experimental setup. Therefore, the models trained on this dataset are inapplicable to other scenarios, such as different displays, different cameras, different camera placement, etc.

\subsection{Summary of Existing Datasets}
\label{s:relwork:datasets:summary}

The following \Cref{t:relwork:datasets} summarizes the characteristics of the datasets presented in this section. The table shows that there is no dataset available that fits our requirements completely as \ed\ offers low-resolution images only, \mpii\ lacks depth information and full frames or at least face crops, \xgaze\ does not provide depth information either, and \stg\ only offers 2D screen coordinates as labels.

\begin{table}[htbp]
\centering
\begin{tabular}{l@{\hspace{3.1mm}}*{4}{c@{\hspace{3.1mm}}}r@{$\times$}l@{\hspace{3.1mm}}c}
    \textbf{Dataset} & \textbf{Year} & \textbf{Subj.} & \textbf{Samples} & \textbf{Depth} & \multicolumn{2}{c}{\textbf{Resolution}} & \textbf{Labels}\,\textsuperscript{a} \\
    \hline
    \ed~\cite{FunesMora2014a} & 2014 & 16 & 94~videos & \cmark & 640 & 480 & C2D \& C3D \\
    \mpii~\cite{Zhang2015} & 2015 & 15 & 213,659 & \xmark & 1280 & 720\,\textsuperscript{b} & C2D \& C3D \\
    \xgaze~\cite{Zhang2020} & 2020 & 110 & 1,083,492 & \xmark & 6000 & 4000\,\textsuperscript{c} & A3D \\
    \stg~\cite{Lian2019} & 2019 & 218 & 165,231 & \cmark & 224 & 224\,\textsuperscript{d} & C2D \\

    \hline
    \multicolumn{8}{p{0.95\linewidth}}{\small
    \textsuperscript{a} C2D: 2D screen coordinates (either in pixels or in meters; screen coordinate system),\par
    \hspace{\widthof{\textsuperscript{a}}} C3D: 3D screen coordinates (in meters; camera coordinate system), \par
    \hspace{\widthof{\textsuperscript{a}}} A3D: 3D gaze angles (in radians; camera coordinate system) \par
    \textsuperscript{b} No full frames available as they are cropped to the eye region \par
    \textsuperscript{c} In this thesis, we use the face-normalized version with $448\times448$ face crops of \xgaze \par
    \textsuperscript{d} No full frames available, only resized crops of the face region}
\end{tabular}

\caption{Overview of existing datasets for \gt.}
\label{t:relwork:datasets}
\end{table}

To overcome the limitations of the existing datasets, we create our own dataset featuring high-resolution RGBD images and 3D gaze angle labels. We will describe the data collection procedure and the resulting dataset in detail in \Cref{s:processing:datacollection,s:processing:dataset}, respectively.

\section{Deep Learning Architectures}
\label{s:relwork:dl}

The realm of \dl\ has expanded dramatically in the last decade, and new architectures have been introduced with great success in many different fields. In this section, we will present some of the most important \dl\ architectures that form the building blocks of the models for vision-based \ges\ proposed in the literature and in this thesis.

\subsection{\aclp{GAN}}
\label{s:relwork:dl:gan}

In 2014,~\citeauthor*{Goodfellow2014}~\cite{Goodfellow2014} proposed a novel \dl\ architecture named \ac{GAN}. \Acp{GAN} consist of two \acp{NN} that are trained simultaneously: a generator G and a discriminator D. G takes a random noise vector as input and generates a sample that is supposed to be indistinguishable from real samples. The task of D is to classify samples as real or fake. The training process is a minimax game between the adversaries G and D, where G tries to fool D and D tries to distinguish real from fake samples. By doing so, the outputs of G are forced to conform to the distribution of the target dataset. As the model training does not require any labels, it is considered unsupervised.

The authors formulated the following objective function for the training process:
\[ \min_G \max_D V(D, G) = \mathbb{E}_{{x}\sim p_{\text{data}}({x})}[\log D({x})] + \mathbb{E}_{{z}\sim p_{{z}}({z})}[\log(1 - D(G({z})))] \]
where $p_{\text{data}}({x})$ is the distribution of the real data, $p_{{z}}({z})$ is the distribution of the noise vector, and $G({z})$ is the generated sample~\cite{Goodfellow2014}.

\Citeauthor*{Radford2015}~\cite{Radford2015} applied the concept of \acp{GAN} to the task of image generation using \acp{CNN}. The authors called their approach \ac{DCGAN} and proposed a set of architectural constraints for the generator and discriminator networks. These constraints include the use of the \ac{ReLU} activation function in the generator network, the removal of fully-connected layers, and the use of \bn\ layers in both G~and~D.

When using \acp{GAN} for image-to-image translation tasks, the generator architecture must be changed to allow for the input of an image instead of a noise vector. \Citeauthor*{Isola2016}~\cite{Isola2016} applied an encoder-decoder architecture to the generator network and proposed a new discriminator architecture called PatchGAN. This discriminator classifies whether each $N\times N$ patch of an image is real or fake instead of classifying the whole image at once, thus, allowing for a smaller discriminator networks and arbitrary input image sizes.

In this thesis, we will use an image-to-image translating \ac{DCGAN} for the task of artifact removal in depth images like \citeauthor*{Lian2019}~\cite{Lian2019} did in their work (see~\Cref{s:relwork:ml:stg}). As the authors did not specify their discriminator architecture, we found that using a PatchGAN discriminator is a suitable choice for our task. We will describe the architecture of our model in detail in \Cref{s:processing:models}.

\subsection{Residual Networks}
\label{s:relwork:dl:resnet}

\Citeauthor*{He2015}~\cite{He2015} proposed residual learning for deep \acp{NN} in 2015. The authors noted that deeper networks are harder to train and can therefore lead to worse performance than shallower networks. They introduced so-called skip-connections, which allow the network to learn residual functions instead of learning the desired underlying mapping directly. The authors showed that their proposed architecture, called ResNet, can be successfully trained with deeper configurations and performs better than previous architectures. To apply these skip-connections to a \dl\ model, the connected layers must have the same feature map size. If only the number of channels, i.e., a single dimension, differ, one can perform a linear projection to match the sizes. However, the authors noted that applying skip-connections to only the dimension-matching parts of a deep \ac{NN} seems sufficient.

A key benefit of ResNet is that it does not increase the number of parameters of a model compared to a similar model without skip-connections~\cite{He2015}. Furthermore, common \dl\ frameworks, such as PyTorch~\cite{Paszke2019}, offer models that were pre-trained on large datasets, e.g., ImageNet, which can be used as a starting point for transfer learning and fine-tuning. In fact, some approaches to \gt\ use these pre-trained models as backbones for their own models (see~\Cref{s:relwork:ml}).

\subsection{Transformer}
\label{s:relwork:dl:transformer}

One of the more recent breakthroughs in the field of \dl\ model architecture are Transformers, introduced in 2017 by \citeauthor*{Vaswani2017}~\cite{Vaswani2017}. The Transformer architecture employs multi-head self-attention layers instead of recurrent or convolutional layers, which helps to overcome training limitations as the new model architecture allows for efficient training on sequence-to-sequence tasks, e.g., machine translation and other \ac{NLP} tasks. Each Transformer block consists of a multi-head self-attention layer and a feed-forward layer with a residual connection and \bn\ layers in between. To enable the model to process sequences and retain positional information, the authors proposed the use of positional encoding vectors that are added to the input embeddings.

\Citeauthor*{Vaswani2017}~\cite{Vaswani2017} formulated the attention functions as follows:
\begin{align*}
    \mathrm{Attention}(Q, K, V) &= \mathrm{softmax}\!\left(\frac{QK^T}{\sqrt{d_k}}\right)\!V \\
    \mathrm{MultiHead}(Q, K, V) &= \mathrm{Concat}\left(\mathrm{head}_1,\dots,\mathrm{head}_h\right)W^O \\
    \mathrm{head}_i &= \mathrm{Attention}\left(QW_i^Q, KW_i^K, VW_i^V\right)
\end{align*}
where $Q, K, V \in\mathbb{R}^{l \times d_\mathrm{model}}$ are the query, key, and value matrices, respectively, and\\
where $W_i^Q, W_i^K\in\mathbb{R}^{d_\mathrm{model}\times d_k}, W_i^V\in\mathbb{R}^{d_\mathrm{model}\times d_v}$ are input projection matrices, and\\
where $ W^O\in\mathbb{R}^{hd_v\times d_\mathrm{model}}$ is the output projection matrix, and\\
where $d_k$ is the dimension of the query and key vectors, and\\
where $d_v$ is the dimension of the value vector, and\\
where $d_\mathrm{model}$ is the dimension of the token vectors, and\\
where $l$ is the sequence length, i.e., the number of tokens, and\\
where $h$ is the number of heads.

The Transformer architecture consists of an encoder and a decoder part. In the Transformer blocks forming the encoder part, the query, key, and value matrices are all identical, i.e., $Q=K=V$ and $d_k=d_v$. In the decoder part, each block contains two multi-head self-attention layers: a masked one and a regular one. The masked attention layer prevents the model from attending to future tokens, which is necessary for sequence-to-sequence tasks~\cite{Vaswani2017}.

{\spaceskip=3.4pt plus 1pt minus 1.5pt 
In the realm of image classification, often only the encoder part is used, which simplifies the architecture to a single stack of Transformer blocks. In 2020, \citeauthor*{Dosovitskiy2020}~\cite{Dosovitskiy2020} introduced \ac{ViT} applying this simplified Transformer architecture to the task of image classification. Each input image gets split into patches of size ${16\times16}$~pixels, each patch gets flattened and enriched with the positional encoding, and then fed into the \ac{ViT}. The authors added a learnable class token to the input embeddings and used the corresponding output of the \ac{ViT} to predict the class of the input image.
}

There are hybrid approaches that combine \acp{CNN} and Transformers as proposed in~\cite{Dosovitskiy2020}. For example, \citeauthor*{Cheng2021}~\cite{Cheng2021} compared the performance of a pure \ac{ViT} with a hybrid approach for the task of \ges. \Citeauthor*{Harms2021}~\cite{Harms2021} used a hybrid approach without a \ac{CNN} backbone for the task of egocentric gaze forecasting and \citeauthor*{Tu2022}~\cite{Tu2022} used a \ac{CNN}-backbone-based hybrid Transformer using both encoder and decoder parts for the task of in-scene human-object \gt.

In this thesis, we will use a hybrid approach and feed the output of multiple \acp{CNN} as well as other features as input tokens to a Transformer encoder. We will describe the architecture of our model in detail in \Cref{s:processing:models}.

\section{Machine Learning Approaches for Gaze Tracking}
\label{s:relwork:ml}

The task of \gt\ can be approached using either analytical or learning-based methods as stated in \Cref{s:relwork:gaze}. As this thesis proposes a learning-based method, we will exclude analytical methods from this section and instead focus on vision-based \gt\ methods using \dl\ and \ml\ approaches.

\subsection{Off-the-Shelf Face Landmark Detection Models}
\label{s:relwork:ml:landmarks}

An essential preprocessing step for \gt\ in general is the detection of faces in an image. Furthermore, to apply additional preprocessing steps, such as face normalization (see~\Cref{s:relwork:ml:normalization}), not only faces must be detected but also facial landmarks. These landmarks indicate the 2D position of certain facial features, such as the eyes, nose, and mouth, within the image.

There are pre-trained off-the-shelf state-of-the-art face landmark detection models available. We will glance over two of them: dlib~\cite{King2009} and yolov7-face~\cite{Delong2023}.

\textbf{dlib} is a \ml\ toolkit which offers a pre-trained face detection and face landmark detection model. To predict the 68 facial landmarks, one has to perform a two-step process. First, the dlib face detection algorithm is applied to the input image, which returns a bounding box for each detected face. Then, the face landmark detection algorithm is applied to each bounding box to predict the 68 facial landmarks~\cite{King2009}.

A common problem with dlib is that its face detection algorithm is both not robust for non-frontal faces and very slow, rendering it unusable for real-time applications. \Citeauthor*{Zhang2019}~\cite{Zhang2019} noted that they were only able to achieve a frame rate of 17\,fps using the dlib face detection algorithm. This is in line with our own observations. Despite the fact that there are methods to increase the frame rate by skipping the face detection step for a few frames, we will use a different face detection algorithm in this thesis.

\textbf{yolov7-face} is based on the \textit{yolov7} architecture. The authors proposed multiple versions of their model, differing in the amount of model parameters and therefore inference speed. Larger models offer typically better accuracy at the cost of slower inference. The models predict faces and 5 facial landmarks per face in one processing step~\cite{Delong2023}.

In this thesis, we aim for real-time performance and therefore use the second-smallest model available: \textit{yolov7-lite-s}. This model achieves a frame rate of more than 30\,fps in our tests and is generally more robust to non-frontal faces as well as rotated faces than the dlib implementation.

\subsection{Face-Normalization for Gaze Tracking}
\label{s:relwork:ml:normalization}

{\spaceskip=3.4pt plus 1pt minus 1.5pt 
In 2018, \citeauthor*{Zhang2018}~\cite{Zhang2018} proposed a new data normalization method for \gt. It involves warping the input image in a way that the face appears straight and in a fixed distance to the virtual camera. This means that the rotation around the roll axis is canceled and no further scaling must be applied to the resulting face patch. The method involves calculating the actual distance between the camera and the face, which is done using the facial landmarks. When comparing the calculated distances with those obtained from the depth map, we found little to no difference. Therefore, we will~use~the calculated distances in this thesis in order to remain compatible with RGB-only datasets.

An important difference to a previously proposed normalization method by \citeauthor*{Sugano2014}~\cite{Sugano2014} is the handling of the gaze direction. \Citeauthor*{Zhang2018}~\cite{Zhang2018} proposed not to scale the gaze direction according to the image warping, but to apply only the rotation vector.~They claim that this reduces the angular error significantly compared to the previous method.

Previous works either did not use normalization at all, e.g., in~\cites{Arakawa2022}{Krafka2016}{Li2023}{Lian2019}, or used the method of \citeauthor*{Sugano2014}, e.g., in~\cites{Cheng2020}{Zhang2015}{Zhang2017}. The datasets used in this thesis also apply different normalization methods. The \xgaze\ dataset is available in a normalized version using the method of \citeauthor*{Zhang2018}~\cites{Zhang2018}{Zhang2020}, while the authors of the~\stg\ dataset used simple crops to the face region~\cite{Lian2019}. Since there is no version of the \stg\ dataset available that contains both full-frame images and 3D gaze target coordinates or gaze angles, we cannot generate face-normalized images for this dataset afterwards. Therefore, we are limited to using the provided dataset as it is.

\Cref{f:relwork:normalization} illustrates the need for face normalization. The input image in the upper left corner shows a face rotated around the roll axis. For this example, we used the \textit{yolov7-face} face landmark detection model as described in \Cref{s:relwork:ml:landmarks}. The 5 facial landmarks are marked as green dots in the input image. If no data normalization process is applied, the resulting face patch will be rotated as well. To demonstrate this effect, we crop the input image to the person's right eye as shown in the upper right image. It is clearly visible that the eye is not horizontally aligned, making the \ges\ task more difficult. The lower left image shows the result of the normalization process proposed by \citeauthor*{Zhang2018} with the facial landmarks marked in blue. Finally, the lower right image shows the resulting patch after cropping to the right eye. Now the eye is aligned and a \gt\ model does not need to learn to compensate for the rotation.
}

\begin{figure}[htb]
    \centering%
    \vspace{-1mm}%
    \input{c2_relwork_face_norm.tex}%
    \caption{Visualization of the face normalization process.}%
    \label{f:relwork:normalization}%
    \vspace{-6mm}%
\end{figure}

\subsection{Gaze Tracking using a \acs{CNN} and single Patches}
\label{s:relwork:ml:cnn}

{\spaceskip=3.4pt plus 1pt minus 1.5pt 
One of the first to apply learning-based methods to the visual \gt\ task were \citeauthor*{Zhang2015}~\cite{Zhang2015} in 2015. They proposed a \ac{CNN} architecture consisting of two convolutional layers, two pooling layers, and two fully-connected layers. The input to the model is a single gray-scale eye patch of size $60\times36$ pixels and the head pose information, which is concatenated to the hidden layer. The model outputs two gaze angles, pitch and yaw. The authors trained the model using a \ac{MSE} loss. They showed that their model was able to outperform basic \ml\ methods, such as Random Forest, $k$-Nearest Neighbors, Adaptive Linear Regression and Support Vector Regression, on their \mpii\ and on the \ed\ dataset.
}

Alongside the \xgaze\ dataset, \citeauthor*{Zhang2020}~\cite{Zhang2020} published a simple \ac{CNN} architecture as a baseline model for future comparison. Their model is an off-the-shelf ResNet-50 model, which takes a single RGB image of size $224\times224$ pixels as input and outputs two gaze angles. In contrast to the above approach, the model takes face patches instead of eye patches. As a comparison the authors reported \ang{6.5} and \ang{4.5} as mean angular errors on the \ed\ and their dataset, respectively~\cite{Zhang2020}.

\Citeauthor*{Ansari2021}~\cite{Ansari2021} proposed three different \ac{CNN} architectures, two of which taking a single face or eye patch as input and outputting a region on a $5\times4$ grid. Their preprocessing of the eye patches includes the application of an elliptical vignette mask. The authors stated that this is done to \enquote{reduce the influence of meaningless pixels surrounding the eye as much as possible}~\cite{Ansari2021}. This is in stark contrast to \citeauthor*{Zhang2017}~\cite{Zhang2017} who used a learnable spatial weighting approach to force the model to emphasize relevant regions of the face patch. They argued that \enquote{activations from other facial regions are expected to [make] subtle [contributions]}~\cite{Zhang2017}. In their work, the authors proposed a \ac{CNN} architecture that takes a single face patch as input and outputs either a 2D on-screen gaze point or a 3D gaze angle depending on the task~\cite{Zhang2017}.

A similar attention-based approach was proposed by \citeauthor*{Arakawa2022}~\cite{Arakawa2022} in 2022. The authors used two \acp{CNN} and fed the RGB component of the face patch into the first \ac{CNN} and the depth component into the second one. They performed spatial weighting on each of the resulting feature maps and concatenated them afterwards. The resulting feature map is then fed into multiple fully-connected layers which output the 2D~on-screen gaze point.

The literature still differs on the decision which patch to use. \Citeauthor*{Salman2022}~\cite{Salman2022} fed single eye patches as inputs to a model that predicts a region on a $4\times4$ grid. Using a differential approach to \ges, \citeauthor*{Liu2018}~\cites{Liu2019}{Liu2018} utilized eye patches only. Their special approach is described later in \Cref{s:relwork:ml:diff}. On the other hand, \citeauthor*{Gudi2020}~\cite{Gudi2020} reported that using a face patch outperforms single eye patches as well as patches spanning both eyes. However, they also noted that their model using face patches is larger and therefore slower than the two-eyes model. They achieved real-time performance (15\,fps) only with their smallest model using single-eye patches~\cite{Gudi2020}.

The trend, however, is towards using multiple patches as input to a model. Such proposed models and their differences are described in the following \Cref{s:relwork:ml:multicnn}.

\subsection{Gaze Tracking using multiple \acsp{CNN} and Patches}
\label{s:relwork:ml:multicnn}

Typically, three patches are used as input to a model: one for each eye and one for the face. However, the size of the patches varies greatly, as does the image preprocessing.

In 2016, \citeauthor*{Krafka2016}~\cite{Krafka2016} proposed a back then novel multi-patch \dl\ architecture consisting of two \acp{CNN}. The first one takes a single eye patch as input and is applied to both eye patches independently. The two outputs are then concatenated and fed into a fully-connected layer. The second \ac{CNN} takes the face patch as input. The output of the face-\ac{CNN} is fed into two fully-connected layers. All RGB patches are sized $224\times224$ pixels. As no data normalization (see~\Cref{s:relwork:ml:normalization}) is applied, the authors introduced a so-called face grid of size $25\times25$ boolean pixels which indicates the position and dimensions of the face within the original image. The face grid is fed into two fully-connected layers as well. Then, the outputs of the three branches are concatenated and fed into two final fully-connected layers that predict the 2D on-screen gaze point as described in~\cite{Krafka2016}.

{\spaceskip=3.4pt plus 1pt minus 1.5pt 
\Citeauthor*{Chen2019}~\cite{Chen2019} proposed a similar architecture without the face grid in 2019. The authors used dilated convolutional layers in the eye-\ac{CNN} to increase the receptive field of the model. The authors used the first four convolutional layers of a pre-trained VGG-16 network as the starting point of their eye-\ac{CNN}. The image is normalized using the method of \citeauthor*{Sugano2014}~\cite{Sugano2014} (see~\Cref{s:relwork:ml:normalization}), then converted to gray-scale, and histogram-equalized. \Citeauthor*{Chen2019}~\cite{Chen2019} reported better performance when both the face and eye patches are used as input to the model than when only a single eye patch is used. This supports the argument that multiple patches lead to better model performance.
}

\begin{figure}[htb]
    \centering
    \input{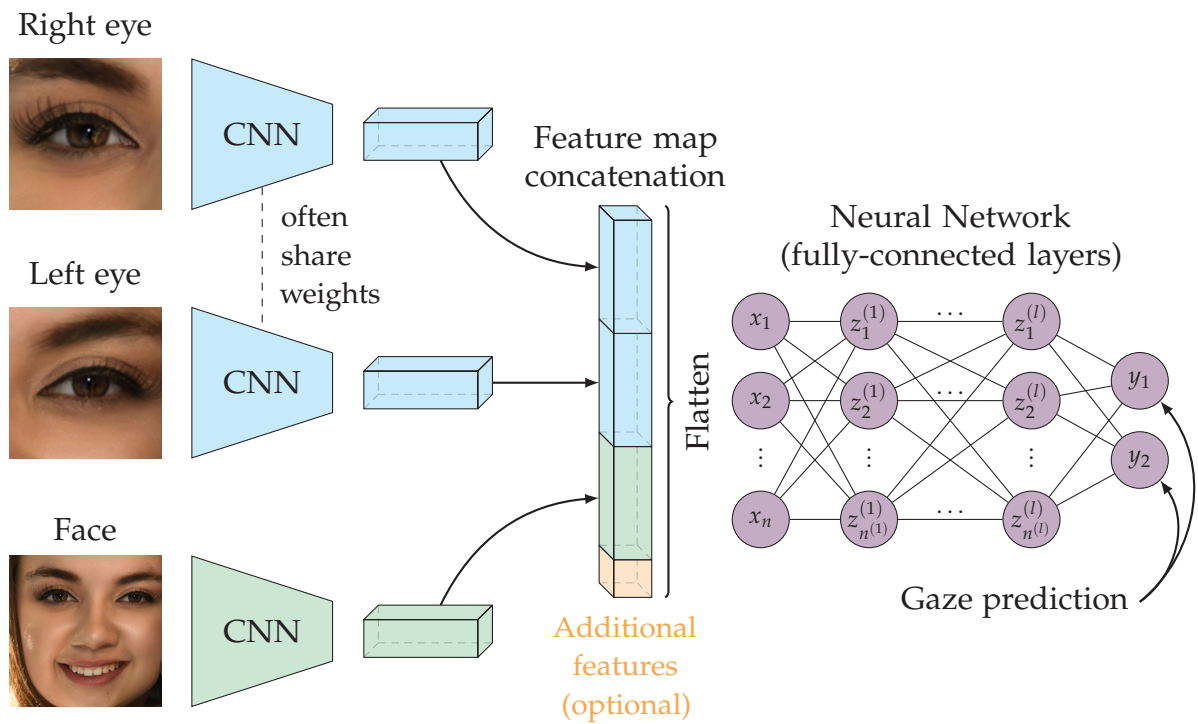}%
    \caption{Basic structure of a multi-patch \ac{CNN} architecture.}
    \label{f:relwork:multicnn}
\end{figure}

The general pattern is to feed the patches into separate \acp{CNN} and then concatenate the outputs, e.g., as shown in~\cites{Chen2019}{Chen2020}{Cheng2020}{Krafka2016}{Li2023}{Lian2019}{Murthy2021}. \Cref{f:relwork:multicnn} illustrates this general architecture using multiple fully-connected layers for the final gaze prediction. Typically, there is only a single \ac{CNN} for both eyes and a separate one for the face patch. As an exception, \citeauthor*{Ansari2021}~\cite{Ansari2021} proposed a model with two separate \acp{CNN} for each eye patch, omitting the face-\ac{CNN} altogether.

When concatenating the outputs of the \acp{CNN}, one can enrich the feature maps with additional information as shown in \Cref{f:relwork:multicnn}. For example, \citeauthor*{Krafka2016}~\cite{Krafka2016} added the face grid. When using RGBD images for \gt, there are multiple ways to use the depth channel of the input image. \Citeauthor*{Lian2019}~\cite{Lian2019} added the position in the original image ($x$ and $y$ coordinate) as well as the depth value for each eye. This helps presumably to compensate the lacking normalization preprocessing step. This is in contrast to \citeauthor*{Arakawa2022}~\cite{Arakawa2022} who used the depth channel as input to a separate \ac{CNN} and not using pixel coordinates at all.

\Citeauthor*{Cheng2020}~\cite{Cheng2020} proposed an attention-based approach to \gt\ in 2020. The authors first divided the gaze direction into a coarse-grained direction and a residual fine-grained one. The coarse-grained direction is predicted by the face-\ac{CNN} and the fine-grained direction is predicted by the eye-\ac{CNN}. The authors proposed an attention component to shift attention between the outputs of the two eye-\acp{CNN} using features from the face-\ac{CNN}.

A different attention-based approach was proposed by \citeauthor*{Murthy2021}~\cite{Murthy2021} in 2021. Their model consists of three \acp{CNN}, one for the face patch and two for the eye patches. In contrast to \citeauthor*{Ansari2021} (see above), both eye patches are fed into the two eye-\acp{CNN} proposed by \citeauthor*{Murthy2021}. The first eye-\ac{CNN} is a standard \ac{CNN} backbone using, inter alia, dilated convolutional layers based on the works of \citeauthor*{Chen2019}. The second eye-\ac{CNN} acts as an attention branch and allows the model to dynamically weigh the features of the first eye-\ac{CNN}~\cite{Murthy2021}.

\subsection{\acsp{GAN} for Depth Image Artifact Removal}
\label{s:relwork:ml:stg}

To the best of our knowledge, there is little public research that addresses the task of \gt\ based on RGBD images using \dl\ methods. Most previous work involving RGBD images has focused on analytical solutions as in \cites{FunesMora2014b}{FunesMora2014a}{Xiong2014}. In particular, we found two works that use \dl\ methods for this specific task: \citeauthor*{Arakawa2022}~\cite{Arakawa2022} and \citeauthor*{Lian2019}~\cite{Lian2019}. However, only \citeauthor*{Lian2019} reported artifacts in the depth channel and proposed the use of a \ac{GAN} to mitigate them.

There is a follow-up work by \citeauthor*{Zhang2020a}~\cite{Zhang2020a} based on the same dataset as the work by \citeauthor*{Lian2019}. In the more recent work, the authors proposed a different method to remove the artifacts in the depth channel. They did this by fitting a 3D face model to the RGB face landmarks and then optimizing the 3D positions using the depth channel.

Comparing the two proposals, the more recent one seems to outperform the older one on their own dataset, \stg~\cites{Lian2019}{Zhang2020a}. However, the authors did not provide a direct comparison with the older method in their work~\cite{Zhang2020a}. Furthermore, the authors did not provide any code samples or pre-trained models for their method, which makes it difficult to reproduce their results. Therefore, we will focus on the method proposed by \citeauthor*{Lian2019} in 2019.

\Citeauthor*{Lian2019}~\cite{Lian2019} argued that a shared architecture between the generator encoder network and the head pose feature extraction network is beneficial for the task of \gt. They proposed a \ac{DCGAN} architecture where the encoder network is used for both the decoder and the head pose feature extraction network. During the prediction phase, the full \ac{GAN} is utilized to remove artifacts from the depth channel. This synthesized depth map is then used to extract the 3D position of the eyes. The authors noted that the depth map is particularly prone to missing data in the eye region. They stated that this is due to the underlying depth sensor technology, which has difficulty measuring the depth of reflective surfaces such as eyeglasses and even the eyes themselves.

\begin{figure}[htb]
    \centering
    \begin{tikzpicture}
    \tikzmath{\scale=0.63;}
    \tikzset{every node/.style={scale=\scale}}
    \pic [pic=rgbenc0, fill=grn20] at (0, 0) {cuboid={width=6, height=112, depth=112, scale=0.02}};
    \pic [pic=rgbenc1, fill=grn20, right=\scale * 5mm of rgbenc0] {cuboid={width=32, height=112, depth=112, scale=0.02}};
    \pic [pic=rgbenc2, fill=grn20, right=\scale * 5mm of rgbenc1] {cuboid={width=64, height=56, depth=56, scale=0.02}};
    \pic [pic=rgbenc3, fill=grn20, right=\scale * 10mm of rgbenc2] {cuboid={width=96, height=28, depth=28, scale=0.02}};
    \pic [pic=rgbenc4, fill=grn20, right=\scale * 13mm of rgbenc3] {cuboid={width=128, height=14, depth=14, scale=0.02}};
    \node [above=\scale * 1cm of rgbenc3, align=center, text=green!80!black, text width=\scale * 8cm] {RGB Encoder\\(Conv2D + Pooling)};

    \pic [pic=depenc0, fill=blu20, below=\scale * 2.5cm of rgbenc0] {cuboid={width=3, height=112, depth=112, scale=0.02}};
    \pic [pic=depenc1, fill=blu20, right=\scale * 5mm of depenc0] {cuboid={width=32, height=112, depth=112, scale=0.02}};
    \pic [pic=depenc2, fill=blu20, right=\scale * 5mm of depenc1] {cuboid={width=64, height=56, depth=56, scale=0.02}};
    \pic [pic=depenc3, fill=blu20, right=\scale * 10mm of depenc2] {cuboid={width=96, height=28, depth=28, scale=0.02}};
    \pic [pic=depenc4, fill=blu20, right=\scale * 13mm of depenc3] {cuboid={width=128, height=14, depth=14, scale=0.02}};
    \node [below=\scale * 1cm of depenc3, align=center, text=blu80, text width=\scale * 8cm] {Depth Encoder\\(Conv2D + Pooling)};

    \node [below=\scale * 0.05cm of $(rgbenc0.south)!0.5!(rgbenc1.south)$, align=center, text=green!80!black] {\tiny$\left\{3, 64\right\}\times224\times224$};
    \node [above=\scale * 0.05cm of $(depenc0.north)!0.5!(depenc1.north)$, align=center, text=blu80] {\tiny$\left\{1, 64\right\}\times224\times224$};
    \node [below=\scale * 0.1cm of rgbenc2, align=center, text=green!80!black] {\tiny$128\times112\times112$};
    \node [above=\scale * 0.1cm of depenc2, align=center, text=blu80, xshift=\scale * 5mm] {\tiny$128\times112\times112$};
    \node [below=\scale * 0.1cm of rgbenc3, align=center, text=green!80!black]{\tiny$256\times56\times56$};
    \node [above=\scale * 0.1cm of depenc3, align=center, text=blu80] {\tiny$256\times56\times56$};
    \node [below=\scale * 0.1cm of rgbenc4, align=center, text=green!80!black] {\tiny$512\times28\times28$};
    \node [above=\scale * 0.1cm of depenc4, align=center, text=blu80] {\tiny$512\times28\times28$};

    \pic [pic=rgbconcat, fill=gry20, right=\scale * 0cm of $(rgbenc4)!0.5!(depenc4)$] {cuboid={width=128, height=14, depth=14, scale=0.02}};
    \pic [pic=depconcat, fill=gry40, right=\scale * 12.4mm of rgbconcat] {cuboid={width=128, height=14, depth=14, scale=0.02}};
    \draw (rgbenc4.south east) edge [out=290, in=70, arrow] (rgbconcat.north);
    \draw (depenc4.north east) edge [out=20, in=250, arrow] (depconcat.south);
    \node [below=\scale * 0.1cm of $(rgbconcat.south)!0.5!(depconcat.south)$, align=center, text=black] {\tiny$1024\times28\times28$};

    \pic [pic=dec5, fill=pur20, right=\scale * 16.5cm of rgbenc4, xscale=-1] {cuboid={width=3, height=112, depth=112, scale=0.02}};
    \pic [pic=dec4, fill=pur20, left=\scale * 5mm of dec5, xscale=-1] {cuboid={width=32, height=112, depth=112, scale=0.02}};
    \pic [pic=dec3, fill=pur20, left=\scale * 5mm of dec4, xscale=-1] {cuboid={width=64, height=112, depth=112, scale=0.02}};
    \pic [pic=dec2, fill=pur20, left=\scale * 10mm of dec3, xscale=-1] {cuboid={width=96, height=56, depth=56, scale=0.02}};
    \pic [pic=dec1, fill=pur20, left=\scale * 13mm of dec2, xscale=-1] {cuboid={width=128, height=28, depth=28, scale=0.02}};
    \pic [pic=dec0, fill=pur20, left=\scale * 26mm of dec1, xscale=-1] {cuboid={width=256, height=14, depth=14, scale=0.02}};
    \node [above=\scale * 1cm of dec2, align=center, text=purple!80!black, text width=\scale * 10cm] {Generator Decoder\\(ConvTranspose2D + Conv2D)};

    \node [below=\scale * 0.1cm of $(dec3.south)!0.5!(dec5.south)$, align=center, text=purple!80!black] {\tiny$\left\{128, 64, 1\right\}\times224\times224$};
    \node [below=\scale * 0.1cm of dec2, align=center, text=purple!80!black] {\tiny$256\times112\times112$};
    \node [below=\scale * 0.1cm of dec1, align=center, text=purple!80!black] {\tiny$512\times56\times56$};
    \node [below=\scale * 0.1cm of dec0, align=center, text=purple!80!black] {\tiny$1024\times28\times28$};

    \pic [pic=hpe1, fill=ora20, right=\scale * 10.945cm of depenc4, xscale=-1] {cuboid={width=256, height=5, depth=5, scale=0.02}};
    \pic [pic=hpe0, fill=ora20, left=\scale * 26mm of hpe1, xscale=-1] {cuboid={width=256, height=10, depth=10, scale=0.02}};
    \node [below=\scale * 1cm of hpe1, align=center, text=orange!80!black, text width=\scale * 10cm] {Head Pose Feature Extractor\\(Conv2D + Pooling)};

    \node [above=\scale * 0.1cm of hpe0, align=center, text=orange!80!black] {\tiny$1024\times7\times7$};
    \node [above=\scale * 0.1cm of hpe1, align=center, text=orange!80!black] {\tiny$1024\times1\times1$};

    \node [above=\scale * -0.2cm of depconcat.north west, fill=white, fill opacity=0.95, align=center, circle, draw] (concat) {+};
    \draw [arrow] (concat.east) -| ($(hpe0.north west) + (\scale * 5 mm, 0cm)$) node [pos=0.25, above, text width=\scale * 5cm, align=center] {Intermediate feature map};
    \draw [arrow] (concat.east) -| ($(dec0.south west) + (\scale * 5 mm, 0cm)$);
\end{tikzpicture}%
    \caption{Architecture of the generator and head pose feature extraction network proposed by \citeauthor*{Lian2019}. (Own figure based on~\cite{Lian2019})}
    \label{f:relwork:stg:gan}
\end{figure}
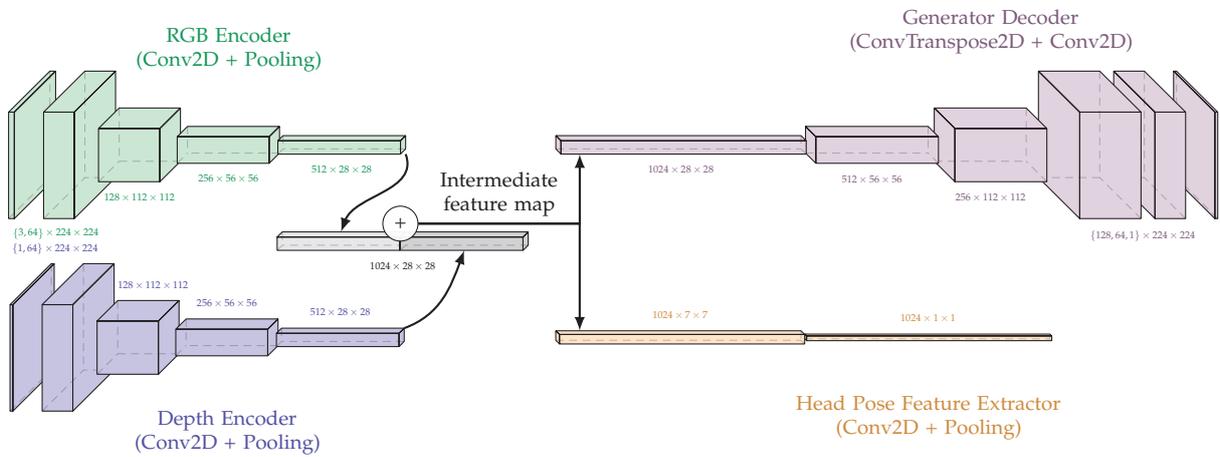

In~\cite{Lian2019}, the generator network G takes as input the RGB face patch and the depth map for the face patch and outputs a synthesized depth map. When using the whole network for \gt, the head pose feature extractor takes the intermediate feature map of G as input and outputs the head pose features. The architecture of G and its interaction with the head pose feature extraction network is illustrated in \Cref{f:relwork:stg:gan}. The discriminator network D takes either the synthesized depth map or a real depth map as input and outputs a probability whether the input is real or fake.

The authors did not specify the architecture for D, only that it \enquote{consists of simple convolution layers}~\cite{Lian2019}. In this thesis, we will use a PatchGAN discriminator as described in \Cref{s:relwork:dl:gan}. During training, the authors enforced the \ac{GAN} loss (see~\Cref{s:relwork:dl:gan}) and an additional L1-loss as a reconstruction loss for the non-missing parts of the depth map~\cite{Lian2019}. The authors formulated this additional loss as follows:
\[ \mathcal{L}_{l1} = \frac{1}{M} \sum_{i=1}^{M} \left\Vert G\left(I_i^\mathrm{D}(\Omega), I_i^{\mathrm{RGB}}\right) - I_i^\mathrm{D}(\Omega) \right\Vert_1 \]
where $I_i^\mathrm{RGB}, I_i^\mathrm{D}$ is the $i$-th RGBD image pair ($i=1,...,M$), $\Omega = \left\{(x,y)\mid I_i^\mathrm{D}(x, y) \ne \mathcal{O} \right\}$ is the set of non-missing pixels in the depth map, and $\mathcal{O}$ is the missing data indicator~\cite{Lian2019}.

In essence, the authors defined an autoencoder architecture where the encoder part consists of not one but two input image encoder networks. The authors argued that this model architecture can \enquote{infer partial structure from the content of the RGB face}~\cite{Lian2019} and, thus, be able to fill in missing data in the depth map. Their goal was to achieve better 3D eye position estimation using the synthesized depth map. Furthermore, the authors argued that the intermediate feature map of G contains information about the head pose and, thus, can be used to extract head pose features. The authors support their statement with an state-of-the-art result on the \ed\ dataset, where their model achieved a mean angular error of \ang{4.8}~\cite{Lian2019}.

In this thesis, we will investigate the effectiveness of using such \ac{GAN} architecture to remove artifacts from the depth channel and to extract head pose features. We will conduct experiments on both the \stg\ dataset and our own dataset. This is described in more detail in \Cref{s:evaluation:stg,s:evaluation:oge}.

\subsection{Gaze Tracking using Transformers}
\label{s:relwork:ml:transformer}

Attention-based \dl\ architectures have been applied to the task of \gt\ before the Transformer architecture was introduced to image processing in 2020 by \citeauthor*{Dosovitskiy2020}~\cite{Dosovitskiy2020}. For example, \Citeauthor*{Zhang2017}~\cite{Zhang2017} proposed the use of spatial weights to attend to different face regions, \citeauthor*{Cheng2020}~\cite{Cheng2020} proposed an attention-based approach for weighting the outputs of the two eye-\acp{CNN}, and \citeauthor*{Murthy2021}~\cite{Murthy2021} proposed two different networks, one of which used a second \ac{CNN} as an attention branch for the eye pose features.

A key difference introduced by the Transformer architecture was multi-head self-attention. One of the first to apply the Transformer architecture, namely the \ac{ViT} architecture proposed by \citeauthor*{Dosovitskiy2020}, to the task of \gt\ were \citeauthor*{Cheng2021}~\cite{Cheng2021} in 2021. The authors used two models, which they called Pure Transformer and Hybrid Transformer. The former is a \ac{ViT} model that takes as input a single RGB face patch that was sliced into patches of $16\times16$ pixels and outputs the gaze angles using a two-layer \ac{MLP}. The latter model consists of a preceding ResNet-18 \ac{CNN} that extracts features from the single RGB face patch. These features are then fed into a Transformer encoder similar to the first model.

The authors added a learnable class token to the input of the Transformer encoder~\cite{Cheng2021}. The corresponding output token is then used as input to an \ac{MLP} which predicts the gaze angles. This approach was previously proposed by \citeauthor*{Dosovitskiy2020}~\cite{Dosovitskiy2020}, and the intuition behind it is that the class token can aggregate information from all patches without sacrificing a specific patch token for it. Alternatively, one could use all output patch tokens and instead apply a global average-pooling operation, as described by \citeauthor*{Dosovitskiy2020}~\cite{Dosovitskiy2020}.

\Citeauthor*{Cheng2021}~\cite{Cheng2021} compared their approach on four datasets, including \ed. They reported that their Hybrid Transformer model outperformed the Pure Transformer model on all datasets. They also found that their hybrid model achieved state-of-the-art performance on all datasets, achieving an average angular error of \ang{5.17} on the \ed\ dataset. The authors also conducted tests on a version of their hybrid model predicting 2D on-screen coordinates. They reported that their model outperformed an existing state-of-the-art model on both datasets tested.

In their work, \citeauthor*{Cheng2021}~\cite{Cheng2021} conducted an ablation study to demonstrate the effectiveness of their proposed Hybrid Transformer model. They showed that the self-attention mechanism improves performance but this is highly dependent on the dataset. In particular, the prediction of more extreme gaze angles benefits from the model's ability to self-attend. However, they showed that the prepended \ac{CNN} is crucial for the performance of their model, independent of the dataset. The authors emphasized the need for pre-training on a large-scale dataset, such as \xgaze, to improve model performance. They stated that a Transformer-based architecture benefits more from pre-training than previous \ac{CNN}-based models~\cite{Cheng2021}.

In 2023, \citeauthor*{Nagpure2023}~\cite{Nagpure2023} proposed a multi-resolution fusion architecture based on Transformers. They extended a previous proposal by \citeauthor*{Ding2021}~\cite{Ding2021} with a Transformer-based feature fusion network to apply self-attention to the extracted features spanning multiple resolutions, and applied it to the task of \ges. \Citeauthor*{Nagpure2023} also modified the internal architecture of a so-called search block, introducing a squeeze-and-excitation block (see~\cite{Hu2017}) and adding a residual connection within the search block. The authors reported that their model outperforms the pre-trained Hybrid Transformer model of \citeauthor*{Cheng2021} on all four datasets tested, including \ed. The authors reported a mean angular error of \ang{5.00} on the \ed\ dataset~\cite{Nagpure2023}.

In their work, \citeauthor*{Nagpure2023} and \citeauthor*{Ding2021} used a \ac{NAS}, which is similar to the progressive shrinking strategy described in~\cite{Cai2019}. This strategy helps to reduce the computational cost of a model by decreasing the number of parameters and the number of operations when the model gets applied. The model size reduction is done progressivley during training and is based on the relative importance of weights as described in~\cites{Cai2019}{Ding2021}{Nagpure2023}.

Due to the limited scope of this thesis, we will not cover the topic of \ac{NAS} in the following chapters. However, when face and eye patches are used as input to a model, it can extract features from different resolutions. Since we will be using a Transformer-based feature fusion stage in our proposed model (see~\Cref{s:processing:models}), one can argue that we are also performing a form of multi-resolution feature fusion using Transformers.

The work by \citeauthor*{Li2023}~\cite{Li2023} featured a more traditional architecture, following the general pattern illustrated in \Cref{f:relwork:multicnn}. Instead of the fully-connected layers for feature fusion, the authors used a multi-head self-attention fusion block. They also added a learnable class token to the input of the Transformer encoder. Since the authors focused on video-based \gt, they fed the output of the fusion block into a novel temporal differential network based on \ac{LSTM} (see~\cite{Sherstinsky2018}) cells. The authors reported that their model outperformed state-of-the-art models on two datasets tested, including \ed\ where their model achieved a mean angular error of \ang{5.02}. The ablation study conducted in their work shows that both the self-attention mechanism and the use of eye patches as input to the model are important for the performance of their model~\cite{Li2023}.

In this thesis, we will not focus on video-based \gt. However, we will use basic filtering techniques to improve the accuracy of our model predictions when applied in the real-time pipeline. The filtering will also help to reduce the noise introduced by the preprocessing steps. This is described in \Cref{s:evaluation:pipeline}.

In 2021, \citeauthor*{Cai2021}~\cite{Cai2021} combined four different \ac{CNN}- and Transformer-based architectures for \gt\ as an ensemble model. The authors used a Transformer-based variation of the model proposed by \citeauthor*{Krafka2016}~\cite{Krafka2016}, a model featuring ResNet-Bottlenecks (see~\cite{He2015}) as well as Transformer-Bottlenecks, a HRNet model based on the work by \citeauthor*{Ding2021}~\cite{Ding2021}, and a ResNeSt~(see~\cite{Zhang2020b}) model. The authors used three configurations of the HRNet model. By weighing the outputs of the six models, the authors reported that their ensemble model achieved state-of-the-art performance on the \xgaze\ dataset with a mean angular error of \ang{3.11}~\cite{Cai2021}.

Unfortunately, the authors did not report the performance of their model on the \ed\ dataset, making it difficult to compare their model to other state-of-the-art models. However, when comparing the performance of their model to the baseline model provided by \citeauthor*{Zhang2020}~\cite{Zhang2020} alongside their \xgaze\ dataset (see~\Cref{s:relwork:datasets:xgaze}), one can calculate a relative improvement of $1-\frac{\ang{3.11}}{\ang{4.5}}\approx \SI{30.9}{\percent}$. This is better than other models' improvements on the \ed\ dataset. For example, \citeauthor*{Nagpure2023}~\cite{Nagpure2023} reported a \ang{5.00} mean angular error, resulting in an improvement of $1-\frac{\ang{5.00}}{\ang{6.5}}\approx \SI{23.1}{\percent}$ over the baseline model of \citeauthor*{Zhang2020}.

\subsection{Differential Gaze Tracking and Offset Calibration}
\label{s:relwork:ml:diff}

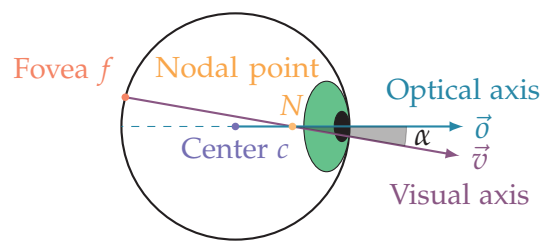
\begin{figure}[htb]
    \centering
    \begin{tikzpicture}
    \node (eye) at (0, 0) {};
    \draw [thick] (eye) circle (1.5cm);
    \coordinate (C) at (eye);
    \coordinate (N) at ($(eye) + (0:0.75cm)$);
    \coordinate (F) at ($(eye) + (165:1.5cm)$);
    \draw [fill=grn60] (eye) + (0:1.2cm) coordinate (lens) ellipse (0.3cm and 0.6cm);
    \draw [fill=black] (eye) + (0:1.4cm) ellipse (0.1cm and 0.2cm);

    \draw [arrow, purple] (F.center) -- (N.center) -- ($(F.center)!2!(N.center)$) coordinate (V) node [below right=-3mm and 0mm] {$\vec{v}$} node [below=2mm] {\small Visual axis};
    \draw [dashed, cyan!60!black] (C.center) -- (180:1.5cm);
    \draw [arrow, cyan!60!black] (C.center) -- (lens.center) -- ($(eye.center)!2.5!(lens.center)$) coordinate (O) node [right] {$\vec{o}$} node [above=1mm] {\small Optical axis};
    \pic [draw, fill=black, opacity=0.33, text opacity=1, "$\alpha$", angle radius=1.5cm, angle eccentricity=1.15] {angle = V--N--O};

    \fill [ora60] (N) circle (0.5mm) node [above, text=ora80, text width=2.3cm, align=right, anchor=south east, xshift=5mm] {\small Nodal point\\$N$\hspace*{2mm}};
    \fill [blu60] (C) circle (0.5mm) node [below, text=blu60] {\small Center $c$};
    \fill [red60] (F) circle (0.5mm) node [above left, text=red60] {\small Fovea $f$};
\end{tikzpicture}%
    \caption{Illustration of the eye geometry with optical and visual axes. (Own figure based on~\cite{FunesMora2014b})}
    \label{f:relwork:ml:diff:eye}
\end{figure}

Many \ges\ models suffer from poor performance when applied to a previously unseen subject. This is mainly due to inter-person variations in the appearance of the eyes and the face~\cites{Chen2020}{FunesMora2014b}{Liu2019}{Liu2018}. In particular, the angle $\alpha$ -- the angular offset between the optical and visual axes -- shown in \Cref{f:relwork:ml:diff:eye} varies from person to person~\cite{FunesMora2014b}. Because this angle cannot be estimated from appearance alone, \citeauthor*{Liu2019} concluded that \enquote{gaze can not be fully predicted from the visual appearance}~\cites{Liu2019}{Liu2018}. To address this issue, \citeauthor*{Chen2020} categorized two calibration methods: differential and adaptation-based approaches~\cite{Chen2020}. In this section, we will discuss both.

{\spaceskip=3.4pt plus 1pt minus 1.5pt 
\textbf{Differential approach.} In 2018, \citeauthor*{Liu2018}~\cite{Liu2018} proposed a novel approach to \gt, which they later extended in 2019~\cite{Liu2019}: a differential network that predicts the difference of gaze angles between two eye images of the same subject. This allows the model to focus on the appearance difference between different gaze angles of the same eye instead of the appearance difference between different eyes. The authors trained two independent models for the left and right eyes, respectively. It is important to note that there are no subject-specific models. \Citeauthor*{Liu2019}~\cites{Liu2019}{Liu2018} only enforced that the two input~images to the differential network were from the same subject and the same eye.
}

The authors formulated the loss function of their model as follows:
\begin{align*}
    \mathcal{L} &= \sum_{I, J \in \mathcal{D}^k} \left\Vert \hat{d}(I, J) - d(I, J) \right\Vert _1 \\
    d(I, J) &= g_\mathrm{gt}(I) - g_\mathrm{gt}(J)
\end{align*}
where $\mathcal{D}^k$ is the set of all same-eye images of the $k$-th subject, $\hat{d}(I, J)$ is the predicted gaze angle difference between the two images $I$ and $J$, and $g_\mathrm{gt}(I)$ is the ground truth gaze angle of the image $I$~\cites{Liu2019}{Liu2018}.

To apply the model, the proposed method needs at least one calibration image from the subject. These calibration images $\mathcal{D}_c$ are then used to predict the absolute gaze angles of a given sample. The two works by \citeauthor*{Liu2018} differ mainly in the calculation of the absolute gaze. In their prior work, they proposed to use a simple average of the predicted gaze angles~\cite{Liu2018}:
\[ \hat{g}(I) = \frac{1}{\left\vert \mathcal{D}_c \right\vert} \sum_{F\in \mathcal{D}_c} \left( g_\mathrm{gt} (F) + \hat{d}(I, F) \right) \]
In their later work, the authors proposed a weighted average of the predicted gaze angles~\cite{Liu2019}:
\begin{align*}
    \hat{g}(I) &= \frac{\sum\limits_{F\in \mathcal{D}_c}\omega \left(\hat{d}(I, F) \right) \cdot \left( g_\mathrm{gt} (F) + \hat{d} (I, F) \right)}{\sum\limits_{F\in \mathcal{D}_c}\omega \left(\hat{d}(I, F) \right)} \\
    \omega (x) &= \mathcal{N}(0, \sigma^2)(x) = \frac{1}{\sqrt{2\pi\sigma^2}} \exp\left( -\frac{x^2}{2\sigma^2} \right)
\end{align*}
where $\omega(\cdot)$ is a Gaussian weighting function with a standard deviation of $\sigma$. The authors used $\sigma=0.1~[\mathrm{rad}] \approx \ang{5.7}$~\cites{Liu2019}.

In addition to the updated averaging function, \citeauthor*{Liu2019}~\cite{Liu2019} proposed two other changes to their network. First, they performed fine-tuning of the differential network for each subject using the calibration images. This resulted in a subject-specific differential network that performed better for the given subject. Second, they proposed to use a pre-trained VGG-16 network as the backbone of their differential network.

The authors reported in~\cite{Liu2019} that their model outperformed state-of-the-art models on three datasets tested, including \ed. They reported a mean angular error of \ang{3.36} when using their unweighted averaging function and no fine-tuning. When using the weighted averaging function, the error decreased to \ang{3.23} and further to \ang{3.00} when using fine-tuning. The VGG-16 backbone achieved an average error of \ang{3.12} without fine-tuning.

\textbf{Adaptation-based approach.} The second approach to calibration is to define some model parameters as subject-dependent. \citeauthor*{Chen2020}~\cite{Chen2020} proposed such adaptation-based calibration by decomposing the gaze into the subject-independent term $\hat{t}$ describing the optical axis $\vec{o}$, and a subject-dependent bias term $\hat{b}_i$ describing the angular offset $\alpha$:
\[ \hat{g}_{i,j} = \hat{t}\left( X_{i,j}; \Phi \right) + \hat{b}_i \]
where $X_{i,j}$ is the $j$-th image of the $i$-th subject, $\Phi$ is the set of all model parameters, and $\hat{g}_{i,j}$ is the predicted gaze angle~\cite{Chen2020}.

The authors used a standard \ac{CNN} architecture as the backbone of their model as shown in \Cref{f:relwork:multicnn}. The key difference is the addition of a subject-dependent bias term $\hat{b}_i$ to the output of the \ac{CNN}. In order to center all learnable bias terms around zero, the authors added a regularization term to the loss function:
\[ \mathcal{L} = \sum_{i,j} \left\Vert g_{i,j} - \hat{t}\left( X_{i,j}; \Phi \right) - \hat{b}_i \right\Vert _2^2 + \lambda \left\vert \sum_{i} \hat{b}_i \right\vert \]
where $g_{i,j}$ is the ground truth gaze and $\lambda$ is a hyperparameter~\cite{Chen2020}.

Such model can now be applied to a previously unseen subject with and without calibration images. When no calibration images are available, the bias term is assumed to be zero, i.e., $\hat{b}_m = 0$. When calibration images are available, the bias term is calculated as follows:
\[ \hat{b}_m = \frac{1}{\left\vert \mathcal{D}_m \right\vert} \sum_{\left( X_{m,j}, g_{m,j} \right) \in \mathcal{D}_m} \left( g_{m,j} - \hat{t} \left( X_{m,j}; \Phi \right) \right) \]
where $\mathcal{D}_m$ is the set of calibration images of the subject $m$~\cite{Chen2020}.

{\spaceskip=3.4pt plus 1pt minus 1.5pt 
\citeauthor*{Chen2020}~\cite{Chen2020} emphasized that no computationally expensive fine-tuning is required for their model. They also noted that the calibration images used to determine $\hat{b}_m$ should resemble the target head pose, illumination, and gaze variance as closely as possible.
}

In their comparison to other models, the authors reported that their model outperformed state-of-the-art models on two datasets tests, including \ed. In particular, their method's performance is dependent on the number of calibration samples used. The authors reported a mean angular error of \ang{4.7}, \ang{3.4}, \ang{3.1}, and \ang{3.0} when using no calibration, four calibration images, nine calibration images, and 25 calibration images, respectively~\cite{Chen2020}. The mean angular error increased by about \ang{0.5} if only calibration images were used where a single target is looked at, supporting their earlier statement that the target variance should be captured~\cite{Chen2020}.

In this thesis, we will use the adaptation-based approach proposed by \citeauthor*{Chen2020} to design and calibrate our model. This is for two reasons: first, no architectural overhaul is required to implement the calibration method and the training process is straightforward, and second, the calibration can be done afterwards with minimal computational effort and does not require any fine-tuning of the network, although this is possible. Our proposed model architecture is described in more detail in \Cref{s:processing:models}.

  \chapter{Processing Sequence and Data Collection}
\label{c:processing}

\vspace{-4.6mm} 
In this chapter, we will tackle the three components of this thesis' task. First, we present the processing sequence, which is the pipeline of our proposed \gt\ system. Next, we introduce the models used in the processing sequence and discuss their architecture and training process. Then, we describe the data collection process, and finally, we present our custom \gt\ dataset, which is used to train our models. The application of our \gt\ system is described in \Cref{c:evaluation}.

\section{Processing Sequence}
\label{s:processing:sequence}

\begin{figure}[!htb]
    \centering
    \input{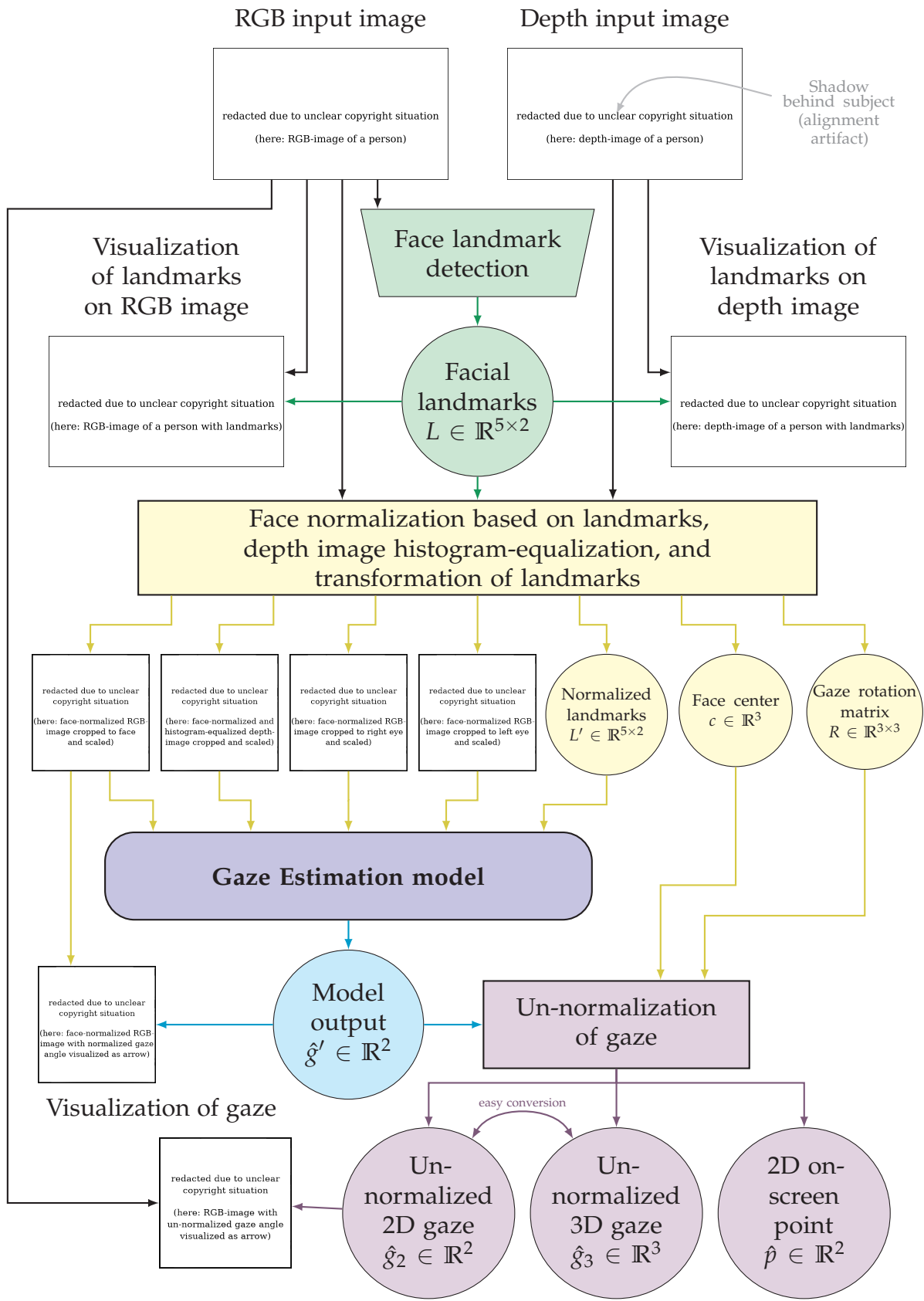}
    \caption{Processing sequence of our proposed \gt\ system.}
    \label{f:processing:sequence}
\end{figure}

This section describes the processing sequence of our proposed \gt\ system. We implemented it as a pipeline of several processing steps and, thus, we will use the terms processing sequence and pipeline interchangeably. The processing sequence is illustrated in \Cref{f:processing:sequence} and consists of four major steps. In brackets, we list the corresponding color in the figure. The processing sequence is as follows:

\begin{enumerate}[itemsep=-1.2mm]
    \item Face Landmark Detection (\textcolor{green!80!black}{green}): \Cref{s:processing:sequence:landmark}
    \item Face Normalization (\textcolor{yellow!80!black}{yellow}): \Cref{s:processing:sequence:normalization}
    \item Gaze Estimation (\textcolor{blue!80!black}{blue}): \Cref{s:processing:sequence:estimation}
    \item Gaze Un-Normalization and Gaze Point Estimation (\textcolor{purple!80!black}{purple}): \Cref{s:processing:sequence:unnorm}
\end{enumerate}

The input to our pipeline consists of two images: an RGB image and its corresponding depth map, which contains the z-axis components of the pixels. In this thesis, we use an Intel RealSense D435 RGBD camera to capture these images. It is necessary to align the depth map to the RGB image because their perspective usually differs. Furthermore, the image resolution gets adjusted to match the RGB image size. In \Cref{f:processing:sequence}, the shadow behind the subject is an artifact of this alignment process caused by the different perspectives of the RGB sensor and the infrared sensors. In essence, the color sensor can see pixels which are occluded for at least one of the infrared sensors. The alignment is done with a built-in function of the Intel\,RealSense\,SDK, which can be accessed in Python using the \package{pyrealsense2} package~\cite{IntelCorporation2018}.

\clearpage 

\subsection{Face Landmark Detection}
\label{s:processing:sequence:landmark}

Face detection in the RGB image is the first step in our pipeline. For this task, we use a pre-trained yolov7-face~\cite{Delong2023} \dl\ model as described in \Cref{s:relwork:ml:landmarks}. We feed the RGB image to the model and directly obtain facial landmarks of all detected faces. In case of multiple detected faces, we compute the bounding boxes of all landmarks per face and select the one with the largest area. We assume that the largest face in the image is the face of interest. Alternatively, we could use the depth map to determine the distance of each detected face to the camera and select the closest one. However, we did not implement this approach because it is not necessary for our purposes. The modular design of our pipeline allows for easy replacement or modification of the face detection and face landmark detection steps.

In \Cref{f:processing:sequence}, the landmarks are visualized as circles in both the RGB and depth images. It is important to point out that the face landmark detection step relies only on the RGB input image and does not use the depth map. Because of the image alignment process, the landmarks can also be applied directly to the depth image. Each of the five landmarks is a 2D image coordinate. We denote the set of all landmarks as $L \in \mathbb{R}^{5 \times 2}$. They are used in the next step to perform the normalization process.

\subsection{Face Normalization}
\label{s:processing:sequence:normalization}

In order to allow the \ges\ model to focus on the appearance difference between different gaze angles rather than head poses, we employ a face normalization process. It is based on the work of \citeauthor*{Zhang2018}~\cite{Zhang2018} and was described in \Cref{s:relwork:ml:normalization}.

The face normalization process uses the facial landmarks $L$ and a generic 3D model of a human face. It consists of three steps: first, we estimate the head pose by fitting the generic model to the facial landmarks found in the RGB input image. Next, we calculate the face center $c$ and the gaze rotation matrix $R$ based on the estimated head pose. Finally, we apply a transformation matrix to warp both the RGB and depth images as well as the landmarks: $L' = T(L)$. In case of a previously known gaze angle or gaze point, e.g., during data collection or calibration, we can also apply $R$ to the gaze vector $g_3 \in \mathbb{R}^3$ to obtain the normalized gaze vector $g'_3 = Rg_3$.

During the face normalization process, the images are warped so that the face is always centered and the roll axis rotation is removed. In this thesis, we use the same generic 3D model as \citeauthor*{Zhang2020}~\cite{Zhang2020}. We also set the same focal length of the virtual camera and the same normalized distance: \SI{960}{\milli\meter} and \SI{300}{\milli\meter}, respectively. The face patch size is set to $448 \times 448$ pixels. The warped RGB face patch is then used to crop the eye images of size $112 \times 112$ pixels based on the warped landmarks $L'$.

As an additional preprocessing step, we histogram-equalize the face depth patch. This allows the \ges\ model to focus on the relative depth differences between face regions rather than the absolute depth values. \Cref{f:processing:sequence} shows the outputs of the face normalization process. The face normalization process is implemented in Python using the packages \package{NumPy}~\cite{Harris2020} and OpenCV (\package{cv2})~\cite{Bradski2000}. We modified the implementation provided by \citeauthor*{Zhang2020}~\cite{Zhang2020} to fit our needs.

\subsection{Gaze Estimation}
\label{s:processing:sequence:estimation}

\renewcommand{\thefootnote}{$\left(\ast\right)$}
The core of our \gt\ system is the \ges\ model. It takes the normalized face patches, eye crops, and landmarks as inputs and outputs gaze angles in the normalized camera coordinate system. Formally, we denote the inputs to the \ges\ model as $I^\mathrm{FC}, I^\mathrm{FD}, I^\mathrm{REC}, I^\mathrm{LEC}$, and $L'$\,\footnote{~FC = Face Color (RGB),~~~FD = Face Depth,~~~REC/LEC = Right/Left Eye Color (RGB)}. We denote the output as $\hat{g}' \in \mathbb{R}^2$, the predicted gaze angles pitch and yaw. For the \stg\ dataset, where the gaze angles are not available, the model output is denoted as $\hat{p} \in \mathbb{R}^2$ instead to indicate that the prediction is a 2D on-screen point.

The \ges\ models are implemented in Python using the PyTorch~(\package{torch}) package~\cite{Paszke2019}. The architectures and training processes of the \dl\ models are described in \Cref{s:processing:models}. In \Cref{f:processing:sequence}, the output of the \ges\ model is visualized in the normalized face patch as a cyan arrow.

\subsection{Gaze Un-Normalization and Gaze Point Estimation}
\label{s:processing:sequence:unnorm}

In order to obtain a 3D gaze vector in the (un-normalized) camera coordinate system, we first transform the two angles $\hat{g}' = \left(\begin{matrix}p & y\end{matrix}\right)^T$ into a 3D vector:
\[ \hat{g}'_3 = \left(\begin{matrix} -\cos p \cdot \sin y \\ \sin p \\ \cos p \cdot \cos y \end{matrix}\right) \in \mathbb{R}^3 \]
Then, we can transform the vector into the camera coordinate system using the inverse of the gaze rotation matrix $R$: $\hat{g}_3 = R^{-1} \hat{g}'_3$. The transformation between $\hat{g}_3$ and $\hat{g}_2$ can be done using the formula above.

To estimate the gaze point from the obtained 3D gaze vector, we need the 3D face center $c$ and a plane representing the screen in the camera coordinate system. In this thesis, we use the camera extrinsic parameters $E \in \mathbb{R}^{4 \times 4}$ to transform $c$ and $\hat{g}_3$ into the world coordinate system, denoted by the asterisk. Then, we can calculate the intersection point $\hat{p}^* \in \mathbb{R}^3$ between the gaze vector and the screen plane $S^* \in \mathbb{R}^4$. Finally, we project $\hat{p}^*$ onto the screen coordinate system to obtain $\hat{p} \in \mathbb{R}^2$ in pixels.
\begin{align*}
    \begin{aligned}[c]
        c^* &= \left[E^{-1} \left(\begin{matrix} c \\ 1 \end{matrix}\right)\right]_{1:3} \\
        \hat{g}_3^* &= \left[E^{-1} \left(\begin{matrix} c + \hat{g}_3 \\ 1 \end{matrix}\right)\right]_{1:3}
    \end{aligned}
    \qquad\vline\qquad
    \begin{aligned}[c]
        \hat{p}^* &= \mathrm{intersect}\left(\mathrm{ray}\left(c^*, \hat{g}_3^*\right), S^*\right) \\
        \hat{p} &= \mathrm{project}\left(\hat{p}^*\right)
    \end{aligned}
\end{align*}
where $\left[v\right]_{1:3}$ denotes the first three elements of a vector $v$.

To reduce the noise in the gaze point estimation pipeline, we apply up to three Kalman filters within: one for the landmarks, one for the predicted gaze angles, and one for the gaze point. The Kalman filters are implemented in Python and model 2D objects with position and velocity. The implementation of our pipeline allows each filter to be individually enabled or disabled.

\addtocontents{toc}{\protect\newpage} 
\section{Models}
\label{s:processing:models}

In this section, we will describe our proposed model architecture. We will first introduce an overview and then dive into the components. In the following we call our model \textit{RGBDTr}, since it estimates the gaze angles using RGBD images and employs a feature fusion step based on the Transformer architecture. We first focus on the \ac{GAN} part of our model, since it is based on the work of \citeauthor*{Lian2019}~\cite{Lian2019} (see~\Cref{s:relwork:ml:stg}).

In the following subsections, we describe our proposed model architecture in more detail. We list all model and training parameters in \Cref{t:appendix:default-config} for a quick overview. The parameters are introduced in the corresponding subsections below and will be used in the experiments in \Cref{c:evaluation}.

\subsection{General Architecture Overview}
\label{s:processing:models:overview}

\begin{figure}[htb]
    \centering
    \input{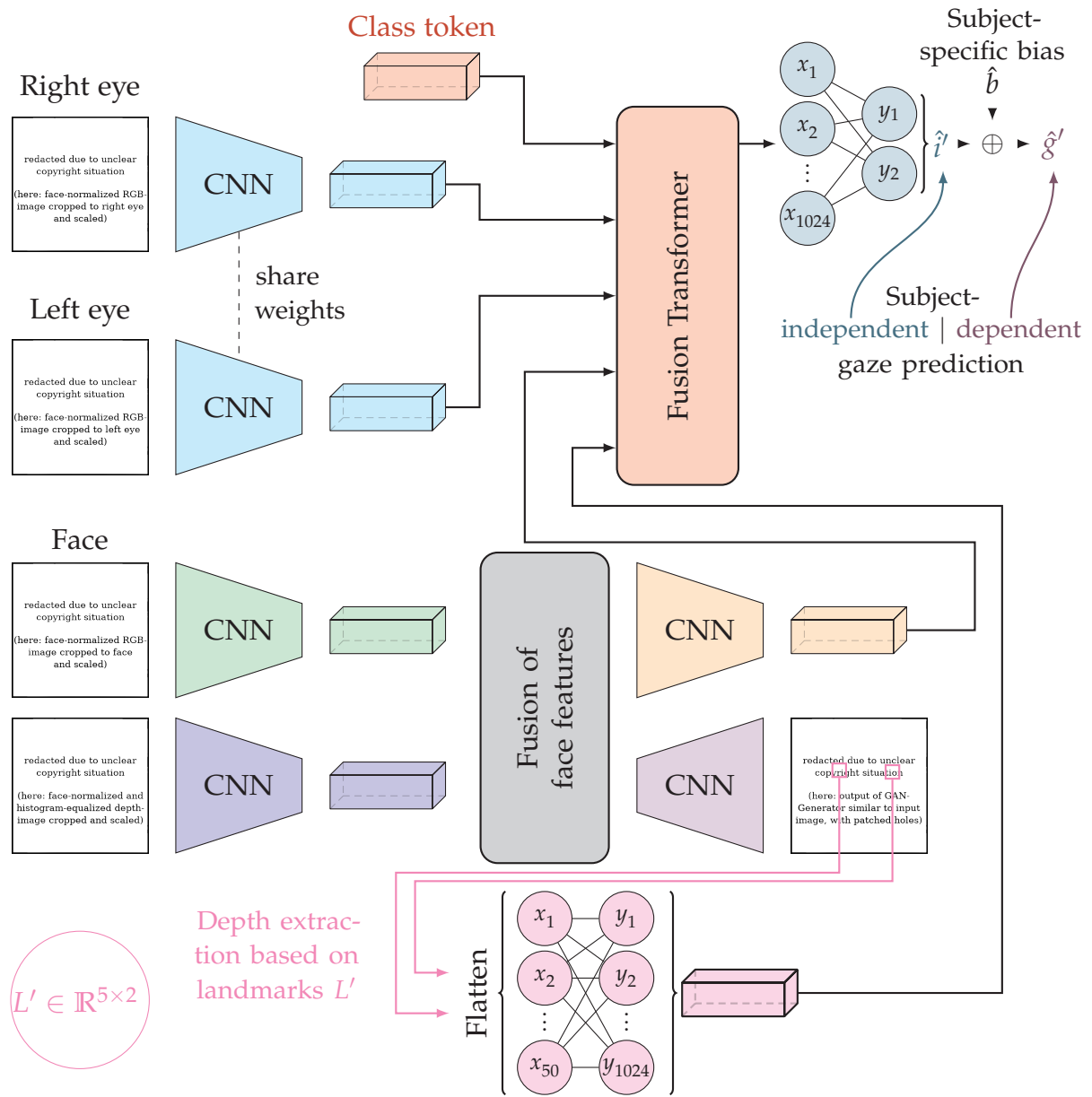}
    \caption{Overview of our proposed model architecture.}
    \label{f:processing:models:overview}
\end{figure}

An overview of our proposed model architecture is shown in \Cref{f:processing:models:overview}. The basic structure is similar to the work of \citeauthor*{Lian2019}~\cite{Lian2019}. However, instead of fully-connected layers, we propose the use of a Transformer to perform the final feature fusion step. The architecture consists of many parts which are colored differently for better distinction. In the following subsections, we will describe each part in detail:

\begin{table}[htbp]
\centering
\begin{tabular}{lrc}
    \textbf{Part} & \textbf{Color} & \textbf{Section} \\
    \hline\hline
    Generator Encoder (RGB)     & \textcolor{green!80!black}{green}         & \multirow{4}{*}{\ref{s:processing:models:gan}}            \\
    Generator Encoder (Depth)   & \textcolor{blue!80!black}{blue}           & \\
    Generator Fusion            & \textcolor{blk80}{gray}                   & \\
    Generator Decoder           & \textcolor{purple!80!black}{purple}       & \\ \hline
    Head Pose Feature Extractor & \textcolor{orange!80!black}{orange}       & \multirow{2}{*}{\ref{s:processing:models:headpose}}       \\
    Depth Extractor             & \textcolor{magenta!80!black}{magenta}     & \\ \hline
    Eye Pose Feature Extractor  & \textcolor{cyan!80!black}{cyan}           & \ref{s:processing:models:eyepose}     \\ \hline
    Feature Fusion Transformer  & \textcolor{red!80!black}{red}             & \multirow{2}{*}{\ref{s:processing:models:transformer}}    \\
    Gaze Estimator              & \textcolor{teal!80!black}{teal}           & \\
\end{tabular}
\caption{Overview of architecture parts.}
\label{t:processing:models:parts}
\end{table}

In \Cref{f:processing:models:overview}, the inputs to the model are shown on the left side: the normalized eye patches, the normalized face patches (RGB+D), and the normalized landmarks. The outputs are shown in the upper right corner: two gaze angles, i.e., pitch and yaw. Alternatively, the model can also output a 2D on-screen point without any modification.

Since we incorporate the offset calibration method of \citeauthor*{Chen2020}~\cite{Chen2020} (see~\Cref{s:relwork:ml:diff}), our model uses a learnable subject-specific bias term~$\hat{b}$, which is combined with the subject-independent gaze prediction~$\hat{i}'$ to obtain the final gaze prediction~$\hat{g}'$. It is important to note that the model then needs a subject label as input. In this thesis, we simply use the subject's number as label. For previously unseen subjects, we set the bias term to zero as proposed by \citeauthor*{Chen2020}~\cite{Chen2020}.

\begin{figure}[htb]
    \centering
    \begin{tikzpicture}
        \node (inputreal) at (0, 0) {\includegraphics[height=2cm]{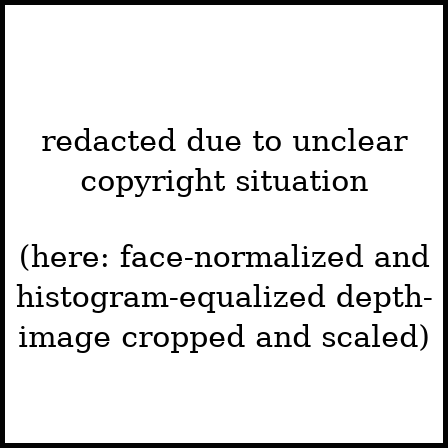}};
        \node [below=0cm of inputreal] (inputfake) {\includegraphics[height=2cm]{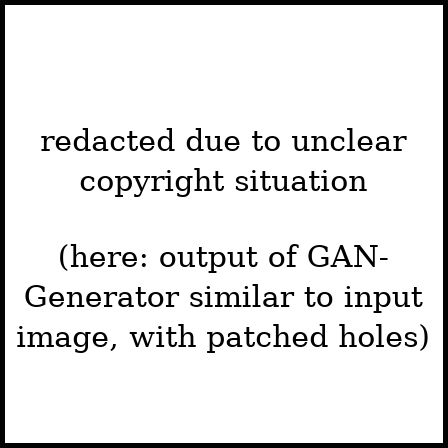}};
        \node (labelreal) [left=0.1cm of inputreal.west, rotate=90, anchor=center] {Real};
        \node (labelfake) [left=0.1cm of inputfake.west, rotate=90, anchor=center] {Fake};

        \pic [pic=dis0, fill=vio20, right=2cm of $(inputreal.east)!0.5!(inputfake.east)$] {cuboid={width=2, height=56, depth=56, scale=0.05}};
        \pic [pic=dis1, fill=vio20, right=2mm of dis0] {cuboid={width=16, height=36, depth=36, scale=0.05}};
        \pic [pic=dis2, fill=vio20, right=8mm of dis1] {cuboid={width=32, height=18, depth=18, scale=0.05}};
        \pic [pic=dis3, fill=vio20, right=12mm of dis2] {cuboid={width=48, height=10, depth=10, scale=0.05}};
        \pic [pic=dis4, fill=vio20, right=0mm of dis3] {cuboid={width=2, height=6, depth=6, scale=0.05}};

        \node [below=0.05cm of dis0, align=center, text=violet!80!black] {\tiny$1\times448\times448$};
        \node [below=0.15cm of dis1, align=center, text=violet!80!black] {\tiny$64\times224\times224$};
        \node [below=0.05cm of dis2, align=center, text=violet!80!black] {\tiny$128\times112\times112$};
        \node [below=0.05cm of dis3, align=center, text=violet!80!black] {\tiny$256\times56\times56$};
        \node [below=0.05cm of dis4, align=center, text=violet!80!black, xshift=4mm] {\tiny$1\times28\times28$};
        \node [right=1cm of dis4, anchor=west] (question) {Real or fake?};
        \draw [arrow] (inputreal.east) -- ++(0.5cm, 0) |- ($(dis0.west)!0.1!(dis0.south west)$);
        \draw [arrow] (inputfake.east) -- ++(0.5cm, 0) |- ($(dis0.west)!0.3!(dis0.south west)$);
        \draw [arrow] (dis4.east) -- (question);
    \end{tikzpicture}
    \caption{One architecture of the discriminator.}
    \label{f:processing:models:discriminator}
\end{figure}
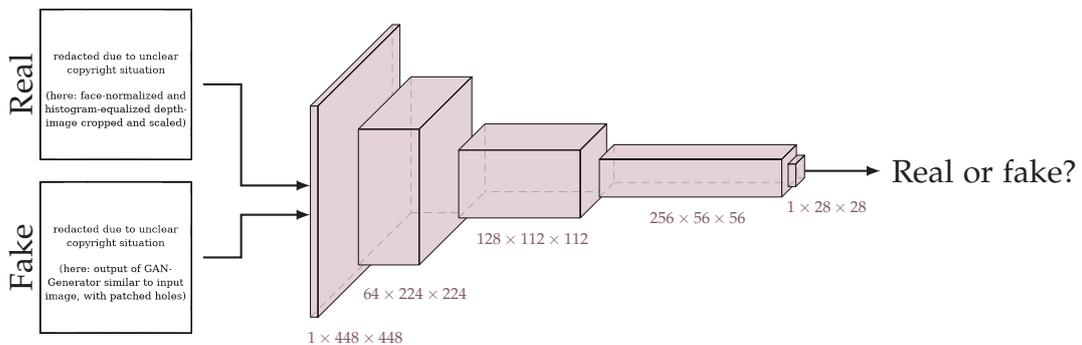

\clearpage 

\subsection{Depth Reconstruction \acs{GAN}}
\label{s:processing:models:gan}

The depth maps captured by the Intel RealSense D435 camera contain minor artifacts and noise in the form of holes in the image. We use a different camera system than \citeauthor*{Lian2019} although ours relies on the same functional principle. In our dataset, the amount and size of the missing data patches are smaller than in the \stg\ dataset (see~\Cref{s:processing:dataset}). However, the artifacts often occur at the face boundary and in the eye region for subjects wearing glasses. The former does not affect our \ges\ model much, but the latter does, because the eye region is used as input in the depth extractor module (see~\Cref{s:processing:models:headpose}). In order to improve the depth map, we employ a \ac{GAN} as proposed by \citeauthor*{Lian2019}~\cite{Lian2019}.

One of the implemented architectures of the discriminator is shown in \Cref{f:processing:models:discriminator}. There are four layers, all of which are convolutional layers with a kernel size of $4 \times 4$ and a stride of 2, with no additional pooling layers. Between the convolutional layers, we add \bn\ layers and use \lrelu\ with a slope of $0.2$ as the activation function. This approach follows the recommendations of \citeauthor*{Radford2015}~\cite{Radford2015}. The final activation function is \sigm. We found that more layers would make the discriminator too strong, thus degrading the performance of the generator G.

\begin{figure}[htb]
    \centering
    \begin{tikzpicture}
    \tikzmath{\scale=0.66;}
    \tikzset{every node/.style={scale=\scale}}
    \pic [pic=rgbenc0, fill=grn20] at (0, 0) {cuboid={width=6, height=224, depth=224, scale=0.02}};
    \pic [pic=rgbenc1, fill=grn20, right=\scale * 0mm of rgbenc0] {cuboid={width=32, height=112, depth=112, scale=0.02}};
    \pic [pic=rgbenc2, fill=grn20, right=\scale * 5mm of rgbenc1] {cuboid={width=64, height=56, depth=56, scale=0.02}};
    \pic [pic=rgbenc3, fill=grn20, right=\scale * 9mm of rgbenc2] {cuboid={width=96, height=28, depth=28, scale=0.02}};
    \pic [pic=rgbenc4, fill=grn20, right=\scale * 12.5mm of rgbenc3] {cuboid={width=128, height=14, depth=14, scale=0.02}};

    \node [above=\scale * 1cm of rgbenc3, align=center, text=green!80!black, text width=\scale * 8cm] {RGB Encoder\\(Conv2D + Pooling)};

    \pic [pic=depenc0, fill=blu20, below=\scale * 4.5cm of rgbenc0] {cuboid={width=3, height=224, depth=224, scale=0.02}};
    \pic [pic=depenc1, fill=blu20, right=\scale * 0mm of depenc0] {cuboid={width=32, height=112, depth=112, scale=0.02}};
    \pic [pic=depenc2, fill=blu20, right=\scale * 5mm of depenc1] {cuboid={width=64, height=56, depth=56, scale=0.02}};
    \pic [pic=depenc3, fill=blu20, right=\scale * 9mm of depenc2] {cuboid={width=96, height=28, depth=28, scale=0.02}};
    \pic [pic=depenc4, fill=blu20, right=\scale * 12.5mm of depenc3] {cuboid={width=128, height=14, depth=14, scale=0.02}};

    \node [below=\scale * 1cm of depenc3, align=center, text=blu80, text width=\scale * 8cm] {Depth Encoder\\(Conv2D + Pooling)};

    \node [below=\scale * -0.8cm of rgbenc0.south, align=center, text=green!80!black, rotate=45, xshift=-5mm] {\tiny$3 \times 448 \times 448$};
    \node [above=\scale * -0.8cm of depenc0.north, align=center, text=blu80, rotate=45] {\tiny$1 \times 448 \times 448$};
    \node [below=\scale * 0.1cm of rgbenc1, align=center, text=green!80!black, xshift=\scale * 5mm] {\tiny$64 \times 224 \times 224$};
    \node [above=\scale * 0.1cm of depenc1, align=center, text=blu80, xshift=\scale * 12mm] {\tiny$64 \times 224 \times 224$};
    \node [below=\scale * 0.1cm of rgbenc2, align=center, text=green!80!black] {\tiny$128 \times 112 \times 112$};
    \node [above=\scale * 0.1cm of depenc2, align=center, text=blu80, xshift=\scale * 6mm] {\tiny$128 \times 112 \times 112$};
    \node [below=\scale * 0.1cm of rgbenc3, align=center, text=green!80!black]{\tiny$256\times56\times56$};
    \node [above=\scale * 0.1cm of depenc3, align=center, text=blu80] {\tiny$256\times56\times56$};
    \node [below=\scale * 0.1cm of rgbenc4, align=center, text=green!80!black] {\tiny$512\times28\times28$};
    \node [above=\scale * 0.1cm of depenc4, align=center, text=blu80] {\tiny$512\times28\times28$};

    \path let \p1=($(rgbenc4.south east) - (depenc4.north west)$) in node at ($(rgbenc4.south)!0.5!(depenc4.north)$) [inner sep=0pt, box={0.7 * \y1 / \scale}{2.25 * \x1 / \scale}, rounded corners=0.25cm, fill=blk20, anchor=west] (ffusion) {};

    \path let \p1=($(ffusion.west)!0.175!(ffusion.east)$), \p2=($(ffusion.north)!0.675!(ffusion.south)$) in pic [pic=depconcat, fill=gry40, rotate=90] at (\x1 + 1mm, \y2) {cuboid={width=128, height=14, depth=14, scale=0.02}};
    \pic [pic=rgbconcat, fill=gry20, above=7.75mm of depconcat, rotate=90] {cuboid={width=128, height=14, depth=14, scale=0.02}};
    \pic [pic=fus1, fill=gry60, right=\scale * 1.2cm of $(rgbconcat.east)!0.5!(depconcat.east)$, rotate=90] {cuboid={width=256, height=14, depth=14, scale=0.02}};
    \pic [pic=fus2, fill=gry80, right=\scale * 1.2cm of fus1, rotate=90] {cuboid={width=256, height=14, depth=14, scale=0.02}};

    \draw ($(rgbenc4.south west)!0.2!(rgbenc4.south east)$) edge [out=270, in=180, arrow] (rgbconcat.west);
    \draw ($(depenc4.north west)!0.2!(depenc4.north east)$) edge [out=90, in=180, arrow] (depconcat.west);
    \node [left=\scale * 1.8cm of fus1, align=center, text=black, rotate=90, anchor=center] {\tiny$1024 \times 28 \times 28$};
    \node [below=0cm of $(ffusion.north)!0.9!(ffusion.south)$, align=center, text=black] {Fusion of face features};

    \node [fill=white, fill opacity=0.95, align=center, circle, draw] (concat) at ($(rgbconcat)!0.5!(depconcat)$) {+};
    \draw [arrow] (concat) -- (fus1);
    \draw [arrow] (fus1) -- (fus2);

    \pic [pic=dec3, fill=pur20, right=\scale * 11cm of depenc4, xscale=-1] {cuboid={width=3, height=224, depth=224, scale=0.02}};
    \pic [pic=dec2, fill=pur20, left=\scale * 5mm of dec3, xscale=-1] {cuboid={width=64, height=112, depth=112, scale=0.02}};
    \pic [pic=dec1, fill=pur20, left=\scale * 9mm of dec2, xscale=-1] {cuboid={width=96, height=56, depth=56, scale=0.02}};
    \pic [pic=dec0, fill=pur20, left=\scale * 12.5mm of dec1, xscale=-1] {cuboid={width=128, height=28, depth=28, scale=0.02}};
    \node [below=\scale * 1cm of $(dec0.south)!0.5!(dec1.south)$, align=center, text=purple!80!black, text width=\scale * 10cm] {Generator Decoder\\(ConvTranspose2D)};

    \node [above=\scale * -0.8cm of dec3, align=center, text=purple!80!black, rotate=-45] {\tiny$1 \times 448 \times 448$};
    \node [above=\scale * 0.1cm of dec2, align=center, text=purple!80!black, xshift=\scale * -6mm] {\tiny$128 \times 224 \times 224$};
    \node [above=\scale * 0.1cm of dec1, align=center, text=purple!80!black,  xshift=\scale * -3mm] {\tiny$256 \times 112 \times 112$};
    \node [above=\scale * 0.1cm of dec0, align=center, text=purple!80!black] {\tiny$512 \times 56 \times 56$};

    \pic [pic=hpe2, fill=ora20, right=\scale * 7.425cm of rgbenc4, xscale=-1] {cuboid={width=32, height=5, depth=5, scale=0.02}};
    \pic [pic=hpe1, fill=ora20, left=\scale * 5mm of hpe2, xscale=-1] {cuboid={width=64, height=10, depth=10, scale=0.02}};
    \pic [pic=hpe0, fill=ora20, left=\scale * 9mm of hpe1, xscale=-1] {cuboid={width=96, height=14, depth=14, scale=0.02}};

    \pic [pic=hpenn, fill=ora20, draw=black, right=\scale * 1.5cm of hpe2] {fc={
        nodedef={{0, "x", "-", 3136}, {0, "y", "-", 1024}},
        scale=0.75,
        circlescale=1.1,
        hscale=2.5,
        vscale=1.4,
    }};
    \draw [decoration={brace, raise=5pt}, decorate, thick] (hpenn.south west) -- (hpenn.north west) node [midway, left=6mm, text width=2cm, vertical centered] {Flatten};

    \node [above=\scale * 1cm of $(hpe0)!0.5!(hpe1)$, align=center, text=orange!80!black, text width=\scale * 10cm] {Head Pose Feature Extractor\\(Conv2D + Pooling + FC)};

    \node [below=\scale * 0.15cm of hpe0, align=center, text=orange!80!black] {\tiny$256 \times 28 \times 28$};
    \node [below=\scale * 0.15cm of hpe1, align=center, text=orange!80!black] {\tiny$128 \times 14 \times 14$};
    \node [below=\scale * -0.01cm of hpe2, align=center, text=orange!80!black] {\tiny$64 \times 7 \times 7$};

    \draw [arrow] (fus2.east) -| ($(hpe0.south west) + (\scale * 2mm, 0cm)$);
    \draw [arrow] (fus2.east) -| ($(dec0.north west) + (\scale * 4mm, 0cm)$);
    \node [text width=\scale * 5cm, align=center, right=\scale * 1.1cm of fus2] {Intermediate feature map};

\end{tikzpicture}%
    \caption{Architecture of the generator.}
    \label{f:processing:models:generator}
\end{figure}
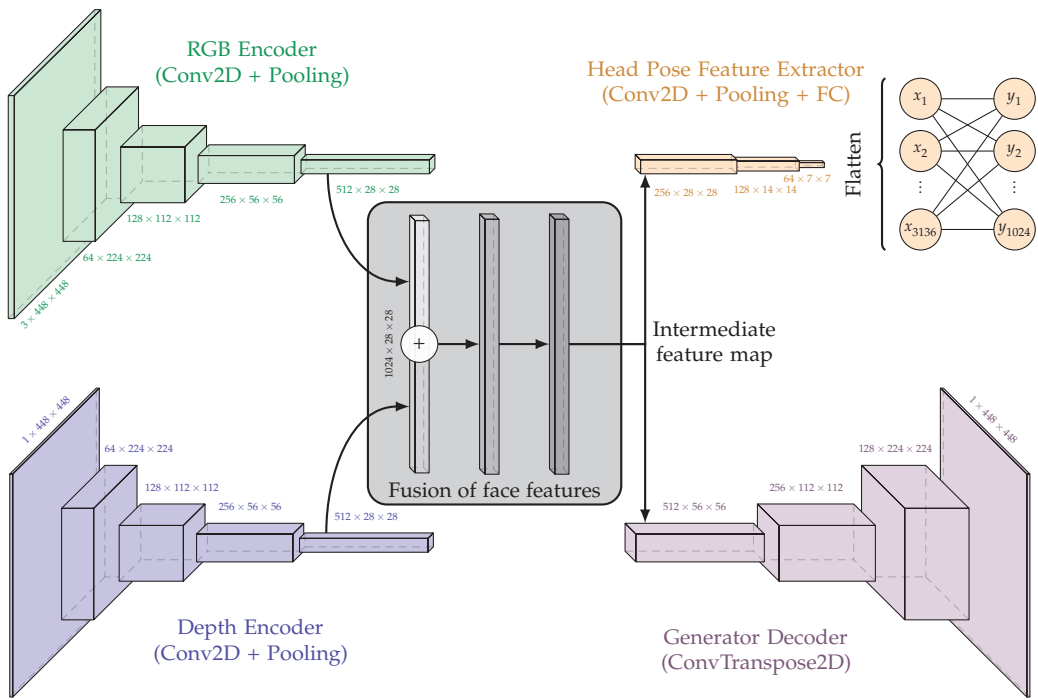

The architecture of G is more complex due to its multi-task design. We modify the architecture of \citeauthor*{Lian2019}~\cite{Lian2019} (see~\Cref{s:relwork:ml:stg} and~\Cref{f:relwork:stg:gan}) to accommodate the larger images we use. In addition, we introduce a fusion block that is shared between the decoder and the head pose feature extractor. The architecture of G is shown in \Cref{f:processing:models:generator}. The inputs to G are two tensors of size $3 \times 448 \times 448$ and $1 \times 448 \times 448$, respectively. These are the two normalized face patches (RGB and depth). In the fusion block, we concatenate the feature maps of the two encoder networks.

The encoder networks are identical and each block in \Cref{f:processing:models:generator} shows the output feature map size after applying a convolutional layer with a kernel size of $3 \times 3$ and a stride of~1, a \bn\ layer, the \ac{ReLU} activation function, and a \mpool\ layer. The fusion block consists of two convolutional layers with a kernel size of $3 \times 3$ and a stride of 1, a \bn\ layer, and the \ac{ReLU} activation function. The decoder network consists of four transposed convolutional layers with a kernel size of $3 \times 3$ and a stride of \sfrac{1}{2}. After each of these layers, there is a \bn\ layer, and the activation function is \ac{ReLU}. The final layer uses $\tanh$ as the activation function to output the depth map in the range $[-1, 1]$ as recommended by \citeauthor*{Radford2015}~\cite{Radford2015}.

If no depth map is available, such as when using the \xgaze\ dataset, the depth encoder and the decoder part of G are removed. The fusion block remains in place but does not concatenate any feature maps because there is only one. In this case, the output of G is the intermediate feature map of the fusion block, which is then fed into the head pose feature extractor (see~\Cref{s:processing:models:headpose}).

We have implemented our model in a way that the architecture of the \ac{GAN} is easily adaptable. For example, the number of downsampling encoder layers $n_e$ can be changed with a single parameter. The same applies to the number of upsampling decoder layers $n_d$. It is possible to change these two parameters independently of each other. In case $n_e \ne n_d$, the synthesized depth map is of different size than the input depth map. In our experiments, we use $n_e = n_d = 4$ and $n_e = n_d = 3$. The size ($w_\mathrm{out} \times h_\mathrm{out}$ pixels) of the resulting synthesized depth map can be calculated as follows: $\left( \begin{matrix} w_\mathrm{out} & h_\mathrm{out} \end{matrix} \right) = \left( \begin{matrix} w_\mathrm{in} & h_\mathrm{in} \end{matrix} \right) \cdot 2^{n_d - n_e}$, with $w_\mathrm{in}$ and $h_\mathrm{in}$ being the input image width and height, respectively.

It is also possible to adjust the number of fusion layers $n_f$ independently of $n_e$ and $n_d$. In our experiments, we use $n_f = 2$. Furthermore, it is possible to change the number of channels in the feature maps. The parameters $c_g$ and $c_d$ control the number of channels in the first feature maps of the generator encoders and the discriminator, respectively. We use in our experiments $c_g = c_d = 64$. All subsequent feature maps have twice as many channels as the previous ones. The generator decoder halves the number of channels in each layer, starting with $c_f = \min \left\{ c_g \cdot 2^{n_e - 1}, 512 \right\} \cdot n_g$ with $n_g$ being the number of encoder networks, i.e., 2~for RGBD datasets and~1~for RGB datasets. We denote the number of channels in the fusion block as $c_f$. We limit the number of channels to 512 per encoder network to reduce the number of total parameters when using $n_e > 4$ in our hyperparameter tuning experiments (see~\Cref{s:evaluation:hyperparams:enc}).

\clearpage 
\subsection{Head Pose Feature Extractor and Depth Extractor}
\label{s:processing:models:headpose}

The head pose feature extraction network also operates on the intermediate feature map and is depicted in orange in \Cref{f:processing:models:generator}. Its architecture differs from the one proposed by \citeauthor*{Lian2019}~\cite{Lian2019}: instead of preserving the number of channels in the feature map, we first reduce it to 256 using a convolution with kernel size $1 \times 1$ and stride 1. We then apply two convolutional layers with kernel size $3 \times 3$ and stride 2, followed by a \bn\ layer and the \ac{ReLU} activation function. Next, we flatten the feature map and apply a fully-connected layer with 1024 output neurons and the \ac{ReLU} activation function to obtain the head pose token.

In order to obtain the depth token, we take the output of G, the synthesized depth map, and crop small patches of size $5 \times 5$ pixels around the two eye landmarks. We then undo the histogram-equalization process to retrieve the true depth value of these patches. Finally, we flatten the patches and feed them into a fully-connected layer with 1024 output neurons and the \ac{ReLU} activation function. The output of said layer is the depth token, which is then fed into the Fusion Transformer (see~\Cref{s:processing:models:transformer}).

The parameter $c_h$ controlling the channel reduction at the beginning of the head pose feature extraction network can be changed independently. In our experiments, we use $c_h = 256$. In the depth extractor module, it is possible to change the region size around the eye landmarks. In our experiments, we set this parameter to $r = 5$. It may be advisable to change this parameter according to the size of the synthesized depth map.

\subsection{Eye Pose Feature Extractor}
\label{s:processing:models:eyepose}

To extract eye pose features, we employ a ResNet-18 pre-trained on the ImageNet dataset as a backbone. We then replace the last fully-connected layer with a new one that has 1024 output neurons. During training, we freeze the weights of the backbone and only train the new layer. The output of the new layer is the eye pose token, which is then fed into the Fusion Transformer (see~\Cref{s:processing:models:transformer}). The eye pose feature extractor net is shared between the left and right eye as shown in \Cref{f:processing:models:overview}.

It is also possible to change the backbone architecture via a parameter to ResNet-34. In our experiments, we use ResNet-18. Due to the pre-training on ImageNet, the backbone expects input images of size $224 \times 224$ pixels. We resize the eye crops to this size before feeding them into the backbone. Alternatively, it is possible to use a the backbone without any pre-trained weights and train it from scratch. In this case, the input images do not need to be resized, saving some computation time during inference. The resize factor $f$ can be changed via a parameter. We propose to use $f = 0.5$ and $f = 0.25$ for pre-trained and non-pre-trained backbones, respectively.

Another parameter controls whether the image of the left eye is flipped horizontally before feeding it into the backbone. This allows the model to learn features that are independent of the eye side. \Citeauthor*{Sugano2014}~\cite{Sugano2014} noted that mirroring one eye image allows for a single estimator. This approach was later used by \citeauthor*{Zhang2018}~\cite{Zhang2018} and \citeauthor*{Bao2021}~\cite{Bao2021}. In our experiments, we use this parameter and flip the left eye image.

\subsection{Feature Fusion Transformer}
\label{s:processing:models:transformer}

In the previous subsections, we introduced three feature extractor networks. Each of them outputs a token per input, resulting in a total of four tokens. The size of these tokens is $d_\mathrm{model} = 1024$ in our experiments. Our Fusion Transformer takes these four tokens and an additional class token as input and fuses them using the multi-head self-attention mechanism. It outputs the transformed class token that aggregates the information from the other input tokens. This output is then fed into a fully-connected layer with two output neurons and no activation function to regress the subject-independent gaze angles $\hat{i}'$ in the normalized coordinate system.

The sequence length of our proposed Fusion Transformer is fixed and therefore we can employ a learnable positional encoding as introduced by \citeauthor*{Vaswani2017}~\cite{Vaswani2017} and proposed by \citeauthor*{Dosovitskiy2020}~\cite{Dosovitskiy2020} and \citeauthor*{Cheng2021}~\cite{Cheng2021}. The class token is also learnable and is initialized with zeros. The input to the Fusion Transformer, $z \in \mathbb{R}^{n_t \times d_\mathrm{model}}$, can be formally described as follows:
\[ z = \left[ t_\mathrm{class}; t_\mathrm{righteye}; t_\mathrm{lefteye}; t_\mathrm{headpose}; t_\mathrm{depth} \right] + z_\mathrm{pos} \]
where $t_* \in \mathbb{R}^{d_\mathrm{model}}$ are the $n_t$ input tokens, and\\
where $z_\mathrm{pos} \in \mathbb{R}^{n_t \times d_\mathrm{model}}$ is the learnable positional encoding.

The number of input tokens depends on the dataset. For RGB datasets, $n_t = 4$ since there is no depth map and therefore no $t_\mathrm{depth}$. Futhermore, if we omit the \ac{GAN} part of our model, there is also no $t_\mathrm{headpose}$, resulting in $n_t = 3$. For RGBD datasets and our full model, $n_t = 5$.

Our proposed Fusion Transformer consists of $n_l$ identical encoder blocks. There are two different encoder architectures: Pre-LN and Post-LN. They differ in the arrangement of the layer normalization, the multi-head self-attention, and the addition operations within an encoder block. \Citeauthor*{Vaswani2017}~\cite{Vaswani2017} originally proposed a Post-LN architecture, which was also used by \citeauthor*{Nagpure2023}~\cite{Nagpure2023}. Other work such as \citeauthor*{Dosovitskiy2020}~\cite{Dosovitskiy2020}, \citeauthor*{Cheng2021}~\cite{Cheng2021}, and \citeauthor*{Li2023}~\cite{Li2023} used a Pre-LN architecture.

In 2022, \citeauthor*{Takase2022}~\cite{Takase2022} proposed a new encoder architecture with a so-called bottom-to-top (B2T) connection. It is based on a Post-LN encoder and adds a residual connection from the input to the output of the feed-forward \ac{MLP}. The three encoder architectures are described in \Cref{t:processing:models:encoderarchitectures}.
\begin{table}[htbp]
\centering
\begin{tabular}{*{2}{>{\centering\arraybackslash}p{0.285\textwidth}} >{\centering\arraybackslash}p{0.33\textwidth}}
    \textbf{Pre-LN} & \textbf{Post-LN} & \textbf{B2T}~\cite{Takase2022} \\
    \hline\\[-3mm]
    {$\!\begin{aligned}
        x' &= \mathrm{MHSA}(\mathrm{LN}(x)) + x \\
        y  &= \mathrm{MLP}(\mathrm{LN}(x')) + x' \\
    \end{aligned}$} & {$\!\begin{aligned}
        x' &= \mathrm{LN}(\mathrm{MHSA}(x) + x) \\
        y  &= \mathrm{LN}(\mathrm{MLP}(x') + x') \\
    \end{aligned}$} & {$\!\begin{aligned}
        x' &= \mathrm{LN}(\mathrm{MHSA}(x) + x) \\
        y  &= \mathrm{LN}(\mathrm{MLP}(x') + x' + x) \\
    \end{aligned}$} \\[5mm]
    \hline \\[-4mm]
    \multicolumn{3}{p{0.95\linewidth}}{
        where $x$ is the input, $y$ is the output, and \par
        where $\mathrm{LN}$ is the layer normalization, and \par
        where $\mathrm{MHSA}$ is the multi-head self-attention, and \par
        where $\mathrm{MLP}$ is the feed-forward \ac{MLP}.
    }
\end{tabular}
\caption{Encoder block architectures.}
\label{t:processing:models:encoderarchitectures}
\end{table}

We implemented all three encoder block architectures, but use the B2T-connection type in our experiments as \citeauthor*{Takase2022} showed that it improves Transformer performance as the number of encoder blocks increases~\cite{Takase2022}. The number of encoder blocks $n_l$ and the number of self-attention heads $n_h$ can be changed via a parameter. In our experiments, we use $n_l = 6$ and $n_h = 8$.

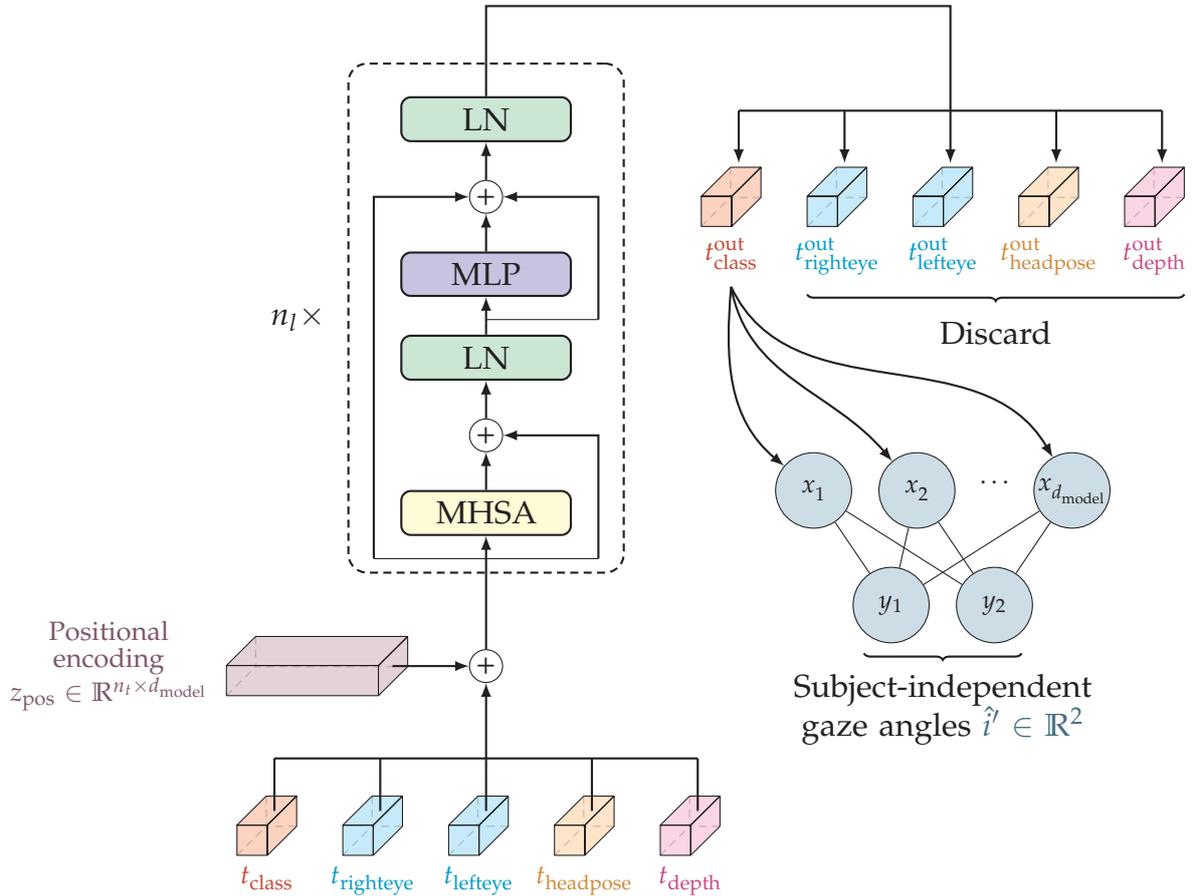
\begin{figure}[htb]
    \centering
    \begin{tikzpicture}
    \tikzmath{\scale=1.0;}
    \tikzset{every node/.style={scale=\scale}}

    \pic [pic=tcls, fill=red20] at (0, 0) {cuboid={width=4, height=4, depth=10}};
    \pic [pic=tre, fill=cya20] [right=\scale * 1cm of tcls] {cuboid={width=4, height=4, depth=10}};
    \pic [pic=tle, fill=cya20] [right=\scale * 1cm of tre] {cuboid={width=4, height=4, depth=10}};
    \pic [pic=thp, fill=ora20] [right=\scale * 1cm of tle] {cuboid={width=4, height=4, depth=10}};
    \pic [pic=td, fill=mag20] [right=\scale * 1cm of thp] {cuboid={width=4, height=4, depth=10}};

    \node [below=0cm of tcls, align=center, text=red!80!black] {\footnotesize $t_\mathrm{class}$};
    \node [below=0cm of tre, align=center, text=cyan!80!black] {\footnotesize $t_\mathrm{righteye}$};
    \node [below=0cm of tle, align=center, text=cyan!80!black] {\footnotesize $t_\mathrm{lefteye}$};
    \node [below=0cm of thp, align=center, text=orange!80!black] {\footnotesize $t_\mathrm{headpose}$};
    \node [below=0cm of td, align=center, text=magenta!80!black] {\footnotesize $t_\mathrm{depth}$};

    \coordinate (tconcat1) at ($(tcls.north) + (\scale * 1mm, \scale * 0.5cm)$);
    \coordinate (tconcat2) at ($(td.north) + (\scale * 1mm, \scale * 0.5cm)$);

    \draw [thick] (tconcat1) -- (tconcat2);
    \foreach \t in {tcls, tre, tle, thp, td} {
        \coordinate (temp) at ($(\t.center)!0.5!(\t.north) + (\scale * 1mm, 0)$);
        \draw [thick] (temp) -- (temp |- tconcat1);
    }
    \draw [arrow] ($(tconcat1)!0.5!(tconcat2)$) -- ++(0, \scale * 1.0cm) coordinate (tconcat3);

    \node (addpe) [above=0cm of tconcat3, align=center, circle, draw, inner sep=1pt] {+};

    \pic [pic=pe, fill=vio20] [left=\scale * 2cm of addpe] {cuboid={width=20, height=4, depth=10}};
    \node [left=0cm of pe, align=center, anchor=east, text=violet!80!black, text width=\scale * 2.8cm] {\footnotesize\baselineskip=10pt Positional encoding $z_\mathrm{pos} \in \mathbb{R}^{n_t \times d_\mathrm{model}}$\par};
    \draw [arrow] ($(pe.east) - (\scale * 2mm, 0)$) -- (addpe);

    \path let \p1=($(tconcat2) - (tconcat1)$), \p2=(\x1 * 0.35, \scale * 0.5cm) in
    node (enc) [above=\scale * 1cm of addpe, align=center, rounded corners=\scale * 0.25cm, box={\x1 * 0.6}{\scale * 6.75cm}, densely dashed] {}
    node (mhsa) [above=\scale * 0.5cm of enc.south, rounded corners=\scale * 0.125cm, box={\x2}{\y2}, fill=yel20] {MHSA}
    node (addmhsa) [above=\scale * 0.5cm of mhsa, align=center, circle, draw, inner sep=1pt] {+}
    node (ln1) [above=\scale * 0.5cm of addmhsa, rounded corners=\scale * 0.125cm, box={\x2}{\y2}, fill=grn20] {LN}
    node (ff) [above=\scale * 0.5cm of ln1, rounded corners=\scale * 0.125cm, box={\x2}{\y2}, fill=blu20] {\ac{MLP}}
    node (addff) [above=\scale * 0.5cm of ff, align=center, circle, draw, inner sep=1pt] {+}
    node (ln2) [above=\scale * 0.5cm of addff, rounded corners=\scale * 0.125cm, box={\x2}{\y2}, fill=grn20] {LN};

    \draw [arrow] (addpe) -- (mhsa);
    \coordinate (encleft) at ($(enc.west)!0.5!(mhsa.west)$);
    \coordinate (encright) at ($(enc.east)!0.5!(mhsa.east)$);
    \path [draw] let \p1=($(mhsa.west) - (encleft)$), \p2=($(encright) - (mhsa.east)$), \p3=($(mhsa.east)-(mhsa.west)$), \p5=(-\x3 * 0.5 - \x1, 0), \p6=(\x3 * 0.5 + \x2, 0) in
    ($(mhsa.south)!0.6!(enc.south)$) -- ++(\x6, 0) edge[arrow, to path={|- (\tikztotarget) \tikztonodes}] (addmhsa.east)
    ($(mhsa.south)!0.6!(enc.south)$) -- ++(\x5, 0) edge[arrow, to path={|- (\tikztotarget) \tikztonodes}] (addff.west)
    ($(ff.south)!0.6!(ln1.north)$)  -- ++(\x6, 0) edge[arrow, to path={|- (\tikztotarget) \tikztonodes}] (addff.east);
    \draw [arrow] (mhsa) -- (addmhsa);
    \draw [arrow] (addmhsa) -- (ln1);
    \draw [arrow] (ln1) -- (ff);
    \draw [arrow] (ff) -- (addff);
    \draw [arrow] (addff) -- (ln2);

    \node [left=0cm of enc, align=center, anchor=east, text width=\scale * 1cm] {$n_l \times$};

    \pic [pic=ocls, fill=red20] [right=\scale * 3cm of addff] {cuboid={width=4, height=4, depth=10}};
    \pic [pic=ore, fill=cya20] [right=\scale * 1cm of ocls] {cuboid={width=4, height=4, depth=10}};
    \pic [pic=ole, fill=cya20] [right=\scale * 1cm of ore] {cuboid={width=4, height=4, depth=10}};
    \pic [pic=ohp, fill=ora20] [right=\scale * 1cm of ole] {cuboid={width=4, height=4, depth=10}};
    \pic [pic=od, fill=mag20] [right=\scale * 1cm of ohp] {cuboid={width=4, height=4, depth=10}};

    \node [below=0cm of ocls, align=center, text=red!80!black] {\footnotesize $t^{\mathrm{out}}_\mathrm{class}$};
    \node [below=0cm of ore, align=center, text=cyan!80!black] {\footnotesize $t^{\mathrm{out}}_\mathrm{righteye}$};
    \node [below=0cm of ole, align=center, text=cyan!80!black] {\footnotesize $t^{\mathrm{out}}_\mathrm{lefteye}$};
    \node [below=0cm of ohp, align=center, text=orange!80!black] {\footnotesize $t^{\mathrm{out}}_\mathrm{headpose}$};
    \node [below=0cm of od, align=center, text=magenta!80!black] {\footnotesize $t^{\mathrm{out}}_\mathrm{depth}$};

    \coordinate (oconcat1) at ($(ocls.north) + (\scale * 1mm, \scale * 0.75cm)$);
    \coordinate (oconcat2) at ($(od.north) + (\scale * 1mm, \scale * 0.75cm)$);

    \draw [thick] (oconcat1) -- (oconcat2);
    \foreach \o in {ocls, ore, ole, ohp, od} {
        \coordinate (temp) at ($(\o.center)!0.5!(\o.north) + (\scale * 1mm, \scale * 2.5mm)$);
        \draw [arrowr] (temp) -- (temp |- oconcat1);
    }
    \draw [thick] (ln2) -- ++(0, \scale * 1.5cm) -| ($(oconcat1)!0.5!(oconcat2)$);

    \draw [decoration={brace, raise=\scale * 9mm}, decorate, thick] (od.south east) -- (ore.south west) node [midway, below=\scale * 11mm, text width=\scale * 4cm, align=center] {Discard};

    \pic [pic=ge, fill=tea20, draw=black] (ge) at ($(ocls.south)!0.5!(od.south)$) [yshift=\scale * -3.5cm] {fc={scale=0.8, circlescale=1.25, vscale=1.9, hscale=1.7, vertical, nodedef={{0, "x", "-", "d_\mathrm{model}"}, {2, "y"}}}};

    \foreach \i in {1,...,3}{
        \draw [arrow] ($(ocls.south) + (0, \scale * -8mm)$) to [out=250 + 15 * \i, in=150 - 10 * \i] (ge-node-1-\i);
    }

    \draw [decoration={brace, raise=\scale * 8pt}, decorate, thick] (ge-node-2-2.south east) -- (ge-node-2-1.south west) node [midway, below=\scale * 4mm, text width=\scale * 4cm, align=center] {Subject-independent gaze angles \textcolor{teal!80!black}{$\hat{i}' \in \mathbb{R}^2$}};

\end{tikzpicture}
    \caption{Architecture of the Fusion Transformer and the gaze estimator.}
    \label{f:processing:models:transformer}
\end{figure}

In \Cref{f:processing:models:transformer}, we show the architecture of our proposed Fusion Transformer as described above. The depicted encoder block shows the B2T-connection. The output of the Fusion Transformer are the transformed input tokens. We discard all but the transformed class token as proposed by \citeauthor*{Dosovitskiy2020}~\cite{Dosovitskiy2020} and \citeauthor*{Cheng2021}~\cite{Cheng2021}. The transformed class token is then fed into a fully-connected layer with two output neurons. There is no final activation function, and the outputs are the predicted gaze angles $\hat{i}'$ in radians. To calculate the final subject-specific gaze angles $\hat{g}'$, we extend the proposal of \citeauthor*{Chen2020}~\cite{Chen2020}: we do not only add a learnable offset term $\hat{b}_\mathrm{offset}$ but also a learnable scaling factor $\hat{b}_\mathrm{scale}$ to our model. We denote all subject-specific components as $\hat{b}$ and calculate the final gaze angles as follows:
\[ \hat{g}' = \hat{i}' \cdot \hat{b}_\mathrm{scale} + \hat{b}_\mathrm{offset} = \left( \begin{matrix}
    \hat{i}'{^\mathrm{pitch}} \cdot \hat{b}_\mathrm{scale}^\mathrm{pitch}  + \hat{b}_\mathrm{offset}^\mathrm{pitch} \\[4mm]
    \hat{i}'^{\mathrm{yaw}} \cdot \hat{b}_\mathrm{scale}^\mathrm{yaw} + \hat{b}_\mathrm{offset}^\mathrm{yaw}
\end{matrix} \right) \]
We implemented $\hat{b}_\mathrm{scale}$ to be technically centered around zero to allow for an easy initialization and to enable future weight decay or other regularization techniques. For each subject, the four values of $\hat{b}$ are initialized with zeros and are learned during training. For previously unseen subjects, we set all values to zero or estimate them using calibration images as proposed by \citeauthor*{Chen2020}~\cite{Chen2020}. Because our approach includes both a subject-specific offset and a subject-specific scaling factor, we use a Least Squares Linear Regression to estimate the values of $\hat{b}$ from the calibration images. The bias terms for pitch and yaw are estimated separately. We use the Python package scikit-learn (\package{sklearn})~\cite{Pedregosa2011} to perform the regression.

The extension of the proposal of \citeauthor*{Chen2020} to include a scaling factor $\hat{b}_\mathrm{scale}$ is motivated by the results of \citeauthor*{Liu2019}~\cites{Liu2019}{Liu2018}. Their work shows not only a subject-specific offset, but also a subject-specific scaling factor that differs between the two gaze angles~\hbox{(see~\cite[Figure~5]{Liu2019})}. Their results are consistent with our results on the \xgaze\ dataset shown in \Cref{f:appendix:xgaze-parameter-distribution:learned}. The four plots in \Cref{f:appendix:xgaze-parameter-distribution} show the distribution of the subject-specific offsets and scaling factors for the two gaze angles. The data is aggregated over all 5-fold cross-validation splits of the 80 training subjects of the \xgaze\ dataset. For better intuition, we show the true scaling factors centered around one instead of the technical scaling factors centered around zero.

To validate our approach of using learnable bias terms, we also show the distribution of the same terms that were estimated using a Linear Regression model on all images. When comparing \Cref{f:appendix:xgaze-parameter-distribution:learned} and \Cref{f:appendix:xgaze-parameter-distribution:linreg} we see similar distributions for both $\hat{b}_\mathrm{offset}$ and $\hat{b}_\mathrm{scale}$ and both gaze angles, which validates our approach.

The plots show that the offset distribution varies more for pitch than for yaw with standard deviations of $\sigma_\mathrm{offset}^\mathrm{pitch} = \SI{2.10}{\degree}$ and $\sigma_\mathrm{offset}^\mathrm{yaw} = \SI{0.75}{\degree}$. The distributions of the scaling factors are similar for pitch and yaw and show that there exists a subject-specific scaling factor with a standard deviation of $\sigma_\mathrm{scale}^\mathrm{pitch} \approx \sigma_\mathrm{scale}^\mathrm{yaw} \approx 0.05$. We therefore conclude that the inclusion of a subject-specific scaling factor improves the performance of our model as it allows for a more fine-grained calibration.

However, the plots also show that the offset distribution for the pitch angle is not centered around zero. Furthermore, both scaling factor distributions are not centered around one, but rather around $0.98$. The distributions' means are expected to be zero and one, respectively. A possible explanation could be that the different camera angles in the \xgaze\ dataset are not centered and therefore exert a shift in the predicted gaze angles. Alternatively, there is the possibility that training without weight decay or a loss term to address this issue (see \Cref{s:processing:models:training}) leads to this behavior. Nevertheless, additional research is needed to substantiate or disprove these conjectures and to explain our findings.

\clearpage

\subsection{Training Process}
\label{s:processing:models:training}

{\spaceskip=3.4pt plus 1pt minus 1.5pt 
In order to train our proposed model, we follow the training process of \citeauthor*{Lian2019}~\cite{Lian2019}. First, we train the \ac{GAN} part of our model for $n_\mathrm{gan}$ epochs. Then, we freeze the weights of the \ac{GAN} and optionally the pre-trained backbone of the eye pose feature extractor (see~\Cref{s:processing:models:eyepose}) and train the rest of the model for $n_\mathrm{rgbdtr}$ epochs. Finally, we fine-tune the overall network for $n_\mathrm{ft}$ epochs. In our experiments, we use $n_\mathrm{gan} = 100$, $n_\mathrm{rgbdtr} = 25$ and $n_\mathrm{ft} = 10$. We use the Adam optimizer~\cite{Kingma2014} with $\beta_1 = 0.9$ and $\beta_2 = 0.999$ and no weight decay, which are the default settings provided by the PyTorch framework~\cite{Paszke2019}. In this subsection, we will describe the training process of the \ac{GAN} and the RGBDTr model. Dataset-specific details are described in their respective sections in \Cref{c:evaluation}.
}

\bigskip

\textbf{\acs{GAN} training.} Training a \ac{GAN} is an unsupervised process where we provide a set of input images and a set of desired target images. In our case, the input and target sets are based on the same dataset. The input set consists of the normalized RGB and depth face patches, and the latter is also the target set. \Citeauthor*{Lian2019} argued that this is a reasonable approach to remove artifacts from the depth map~\cite{Lian2019}. However, we found that this approach does not work well for removing artifacts in the eye region. The generator G has no incentive to fill the holes in the depth map because the target dataset contains these artifacts, too. Without modifying of the target dataset, we cannot tell the discriminator D to penalize G for artifacts in the eye regions. In fact, using the original approach by \citeauthor*{Lian2019}, G is trained to be a very good autoencoder, ignoring the RGB input altogether.

\begin{figure}[htb]
    \centering
    \includegraphics[width=0.12\textwidth, clip, trim={0cm 0cm 8cm 0cm}]{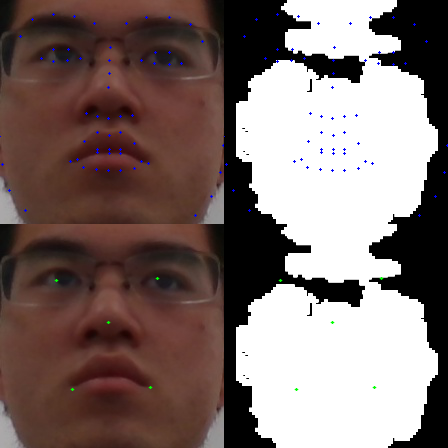}%
    \includegraphics[width=0.12\textwidth, clip, trim={0cm 0cm 8cm 0cm}]{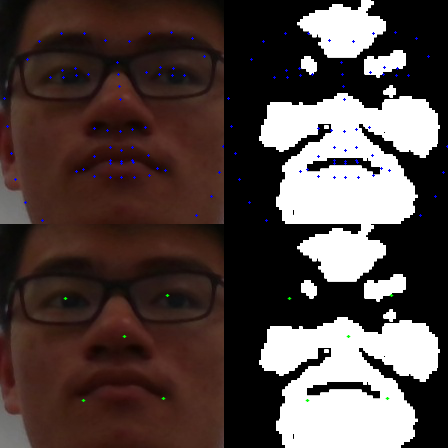}%
    \includegraphics[width=0.12\textwidth, clip, trim={0cm 0cm 8cm 0cm}]{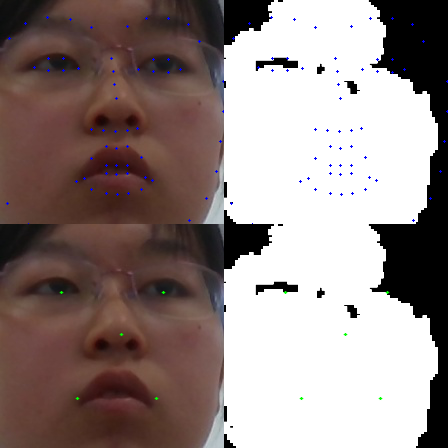}%
    \hfill
    \includegraphics[width=0.12\textwidth, clip, trim={0cm 0cm 8cm 0cm}]{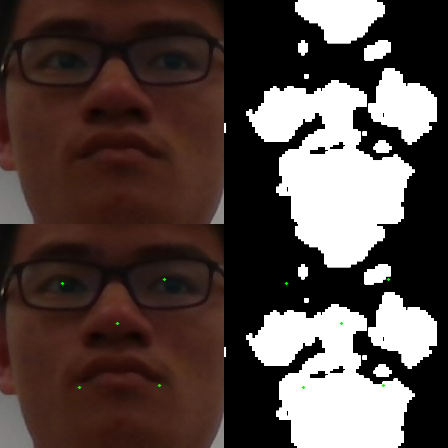}%
    \includegraphics[width=0.12\textwidth, clip, trim={0cm 0cm 8cm 0cm}]{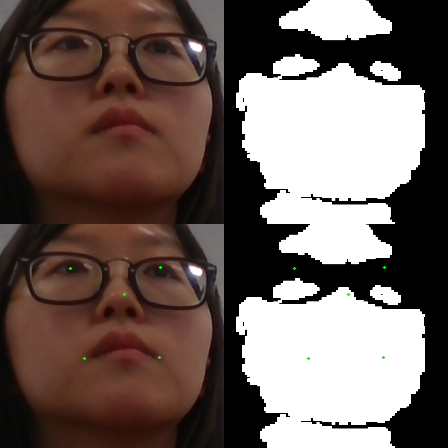}%
    \hfill
    \includegraphics[width=0.12\textwidth, clip, trim={0cm 0cm 8cm 0cm}]{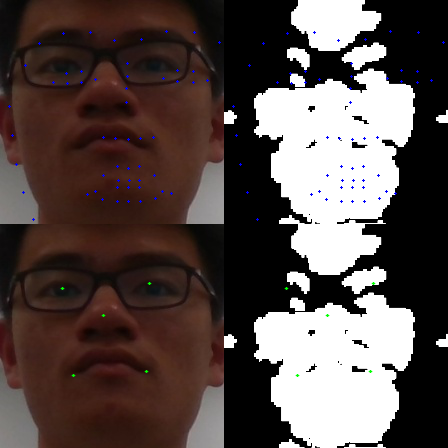}%
    \includegraphics[width=0.12\textwidth, clip, trim={0cm 0cm 8cm 0cm}]{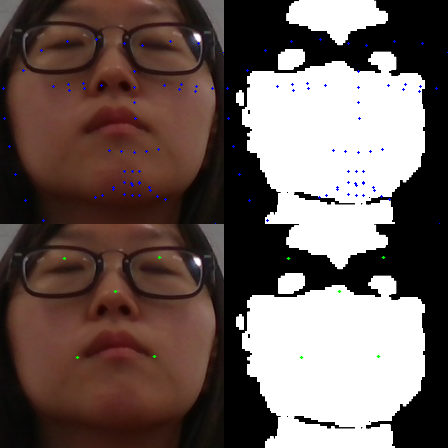}%
    \includegraphics[width=0.12\textwidth, clip, trim={0cm 0cm 8cm 0cm}]{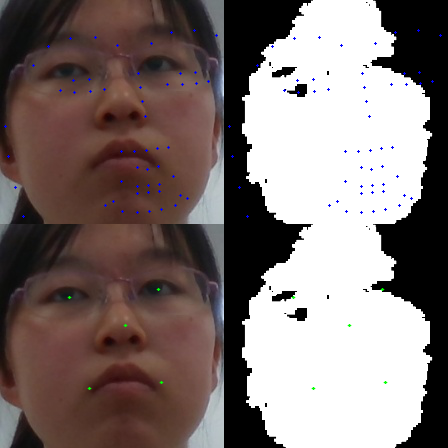}%
    \caption{Landmarks provided by the \stg\ dataset authors (top half) vs.\, landmarks detected by yolov7-face (bottom half). From left to right: $3 \times$ good, $2 \times$ missing, $3 \times$ bad landmarks provided with the dataset.}
    \label{f:processing:models:stglandmarks}
\end{figure}

Therefore, we propose the use of a different target set for the \ac{GAN} training. To obtain it, we first applied the landmark detection model used in our pipeline, yolov7-face, to the \stg\ dataset, because the landmarks provided by the authors were partially missing or positioned completely wrong as shown in \Cref{f:processing:models:stglandmarks}. The landmark detection model yolov7-face failed in 5 images, which is negligible.

\begin{figure}[htb]
    \centering%
    \includegraphics[width=\textwidth]{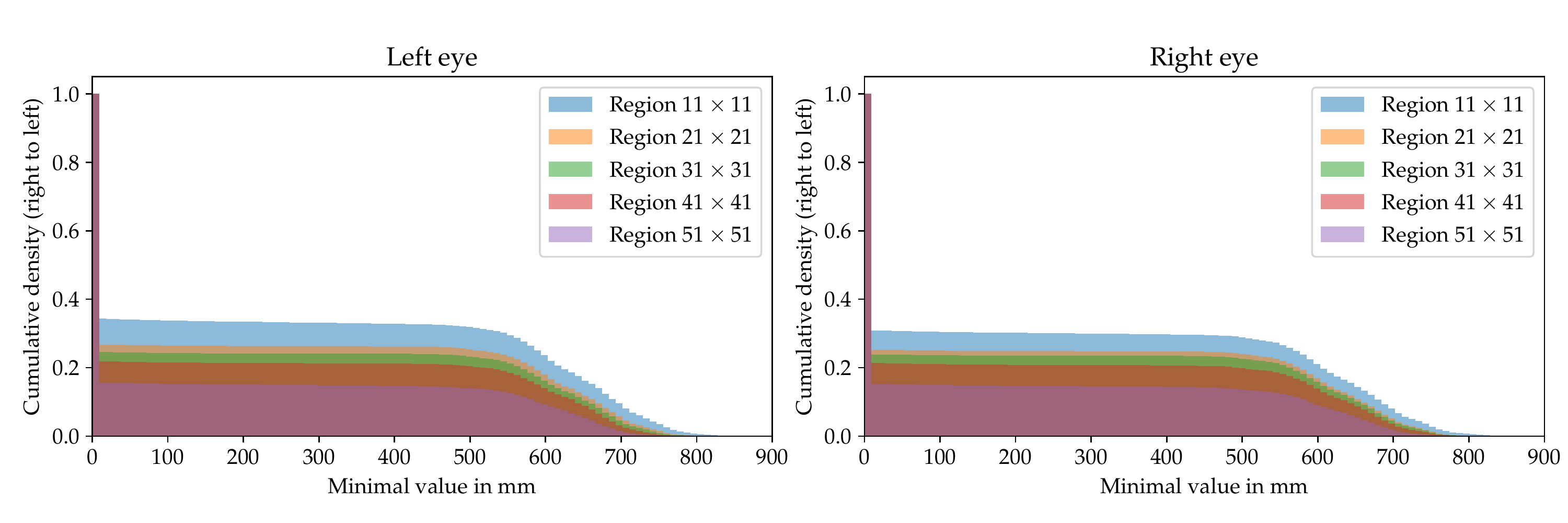}%
    \caption{Cumulative histogram of the minimum depth values in the eye regions of the \stg\ dataset.}%
    \label{f:processing:models:stghistogram}%
    \vspace*{-1mm}%
\end{figure}

We then used the detected landmarks to examine the depth maps in the eye regions. \Cref{f:processing:models:stghistogram} shows the cumulative distribution of the minimal depth values in various sized, square regions around each eye. The cumulation was performed from right to left in order to obtain a better visual representation. It is important to note that the \stg\ dataset defines missing depth data as zero, i.e., $\mathcal{O} = 0$.

\clearpage 

\begin{figure}[htb]
    \centering
    \begin{subfigure}{0.49\textwidth}%
        \includegraphics[width=\textwidth]{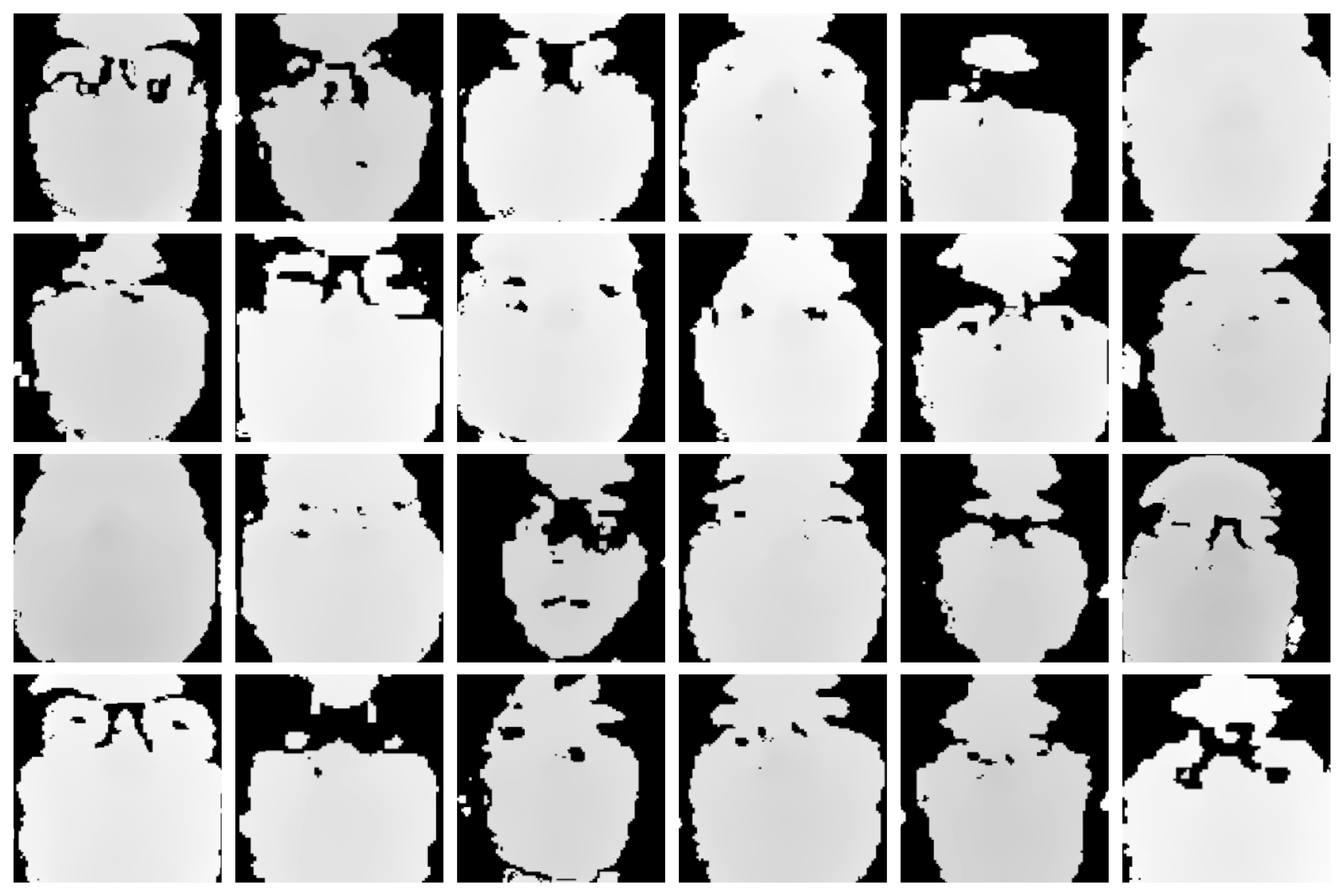}%
        \caption{Example depth maps from the input set $\upsilon$.}%
    \end{subfigure}%
    \hfill%
    \begin{subfigure}{0.49\textwidth}%
        \includegraphics[width=\textwidth]{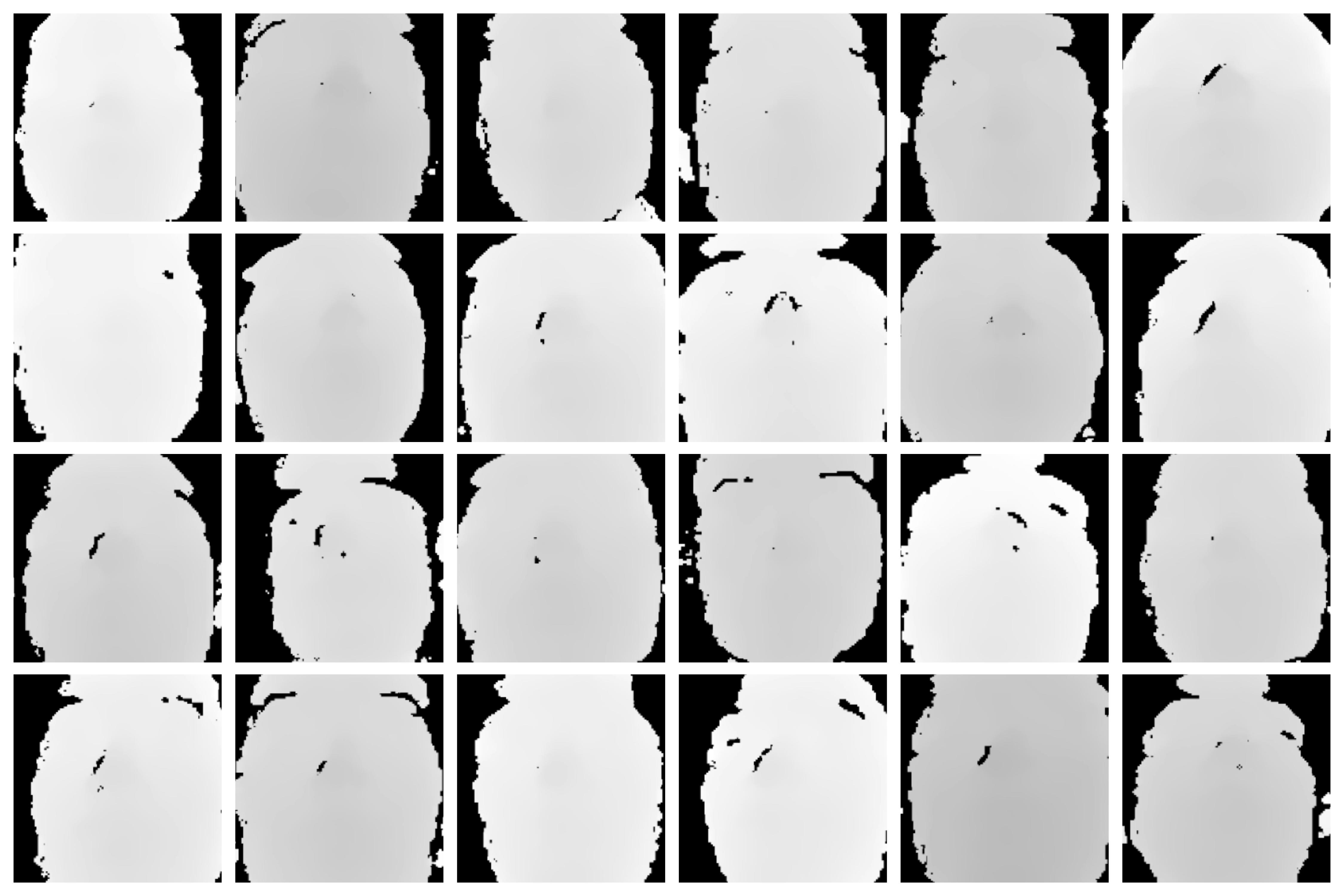}%
        \caption{Example depth maps from the target set $\tau$.}%
    \end{subfigure}%
    \caption{Example depth maps from the input and target sets of the \ac{GAN} training on the \stg\ dataset.}%
    \label{f:processing:models:stgsets}%
\end{figure}

{\spaceskip=3.4pt plus 1pt minus 1.5pt 
Next, we defined the region size and the threshold to be $41 \times 41$ and \SI{200}{\milli\meter}, respectively. This enables us to vary the parameter $r$ of the depth extractor module (see~\Cref{s:processing:models:headpose}) for future experiments. We then removed all images from the \stg\ dataset where the minimal depth value in at least one of the two eye regions was below the threshold. This resulted in a new target set for the \ac{GAN} training with $25,664$ images, i.e., \SI{15.5}{\percent} of the original dataset. The input set remains unchanged. We denote the input set as $\upsilon$ and the new target set as $\tau \subset \upsilon$. In \Cref{f:processing:models:stgsets}, random samples from both sets are shown, illustrating the difference between the two sets.
}

Selecting this subset for the discriminator may not be representative of the entire dataset. For example, it is possible that subjects with glasses appear less frequent in $\tau$ because the depth map artifacts are more severe on reflective surfaces. However, not selecting a subset for the discriminator results in the undesired behaviors of G and D described above. We therefore argue that our approach is a reasonable compromise.

In \Cref{f:appendix:stg-depth-distribution}, we compare $\upsilon$ and $\tau$ in terms of the depth value distribution. It shows that the average depth values are similar, but only when considering all pixels. When considering the nonzero pixels, i.e., ignoring the missing data, the average depth value of $\tau$ is lower than that of $\upsilon$. The same can be observed for the median and the maximum depth values, regardless of whether zero values are considered or not. This means that images with the subject further away from the camera are more likely to be excluded from $\tau$ due to missing data artifacts.

However, the \ac{GAN} does not take the raw depth map as input, but rather the histogram-equalized version. The last two rows in \Cref{f:appendix:stg-depth-distribution} show the average and the median of the histogram-equalized depth values. In these cases, the average and median values of $\tau$ are higher than those of $\upsilon$ when considering all pixels. This is expected as $\tau$ contains fewer regions with artifacts. When ignoring the missing data, both distributions are very similar to each other, indicating that our approach using histogram-equalization is both reasonable and a best effort to make the two sets similar in terms of depth value distribution.

\bigskip

We define the loss functions of the \ac{GAN} training as follows:
\begin{align*}
    \mathcal{L}_\mathrm{D} &= \frac{1}{N} \sum_{j=1}^N \log \left[ D \left(J^\mathrm{FD}_{j}\right) \right] + \frac{1}{M} \sum_{i=1}^M \log \left[ 1 - D \left( G \left( \mathcal{I}^\mathrm{FC}_i, \mathcal{I}^\mathrm{FD}_i \right) \right) \right] \\
    \mathcal{L}_\mathrm{G} &= \mathcal{L}_\mathrm{G_\mathrm{adv}} + \lambda_\mathrm{L1} \cdot \mathcal{L}_\mathrm{G_\mathrm{L1}} \\
    \mathcal{L}_\mathrm{G_\mathrm{adv}} &= - \frac{1}{M} \sum_{i=1}^M \log \left[ D \left( G \left( \mathcal{I}_i^\mathrm{FC}, \mathcal{I}^\mathrm{FD}_i \right) \right) \right] \\
    \mathcal{L}_\mathrm{G_\mathrm{L1}} &= \frac{1}{M} \sum_{i=1}^M \left\Vert G \left( \mathcal{I}_i^\mathrm{FC}, \mathcal{I}_i^\mathrm{FD} \right) \left( \Omega \right) - I_i^\mathrm{FD} \left( \Omega \right) \right\Vert_1 \\
    \Omega &= \left\{ (x, y) \mid \nexists (x_0, y_0) \in I_i^\mathrm{FD}: d \leq r_\mathrm{d} \land I_i^\mathrm{FD} (x_0, y_0) \in \left[ \mathcal{O} - r_\mathcal{O} ; \mathcal{O} + r_\mathcal{O} \right] \right\} \\
    d &= \sqrt{(x - x_0)^2 + (y - y_0)^2}
\end{align*}
where $I_i^\mathrm{FC}, I_i^\mathrm{FD}$ are the $i$-th RGB and depth image pair in $\upsilon$ ($i = 1, ..., M$), and\\
where $\mathcal{I}$ is an augmented version of $I$, and\\
where $J_j^\mathrm{FD}$ is the $j$-th depth image in $\tau$ ($j = 1, ..., N$), and\\
where $\lambda_\mathrm{L1}$ is a hyperparameter controlling the influence of the L1 loss, and\\
where $r_\mathrm{d}$ is a radius parameter, and\\
where $\mathcal{O}$ is the missing data indicator, and\\
where $r_\mathcal{O}$ is used to define a range of values indicating missing data.

The depth images are already histogram-equalized to the range~$\left[-1; 1\right]$. We redefine~$\Omega$ here to exclude not only pixels with missing data, but also pixels within a radius~$r_\mathrm{d}$ around them. In practice, we first create a binary mask that excludes all pixels with missing data. A pixel is defined as missing data if its value is within a range of~$\mathcal{O} \pm r_\mathcal{O}$. In this mask, a zero indicates a missing pixel and a one indicates a valid pixel. We then erode the mask with a circular kernel of size $\left(2 r_\mathrm{d} + 1 \right) \times \left(2 r_\mathrm{d} + 1 \right)$ pixels. This results in a mask where all pixels within a radius of~$r_\mathrm{d}$ around a missing pixel are also excluded. In our experiments, we use~$r_\mathrm{d} = 12$, $\mathcal{O} = -1$, and $r_\mathcal{O} = 0.01$. The L1 loss is applied only to the valid pixels, i.e., pixels that are not excluded by $\Omega$, and is weighted by~$\lambda_\mathrm{L1}$, which we set to $10$ in our experiments.

\begin{figure}[htb]
    \vspace*{-2mm}%
    \centering\captionsetup{width=0.95\textwidth}%
    \begin{subfigure}[t]{0.245\textwidth}%
        \includegraphics[width=\textwidth]{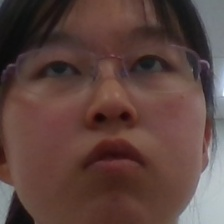}%
        \caption{RGB image $I^\mathrm{FC}$.}%
    \end{subfigure}\hfill%
    \begin{subfigure}[t]{0.245\textwidth}%
        \includegraphics[width=\textwidth]{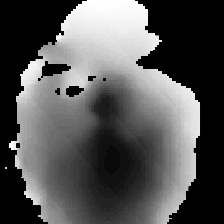}%
        \caption{Depth image $I^\mathrm{FD}$.}%
    \end{subfigure}\hfill%
    \begin{subfigure}[t]{0.245\textwidth}%
        \includegraphics[width=\textwidth]{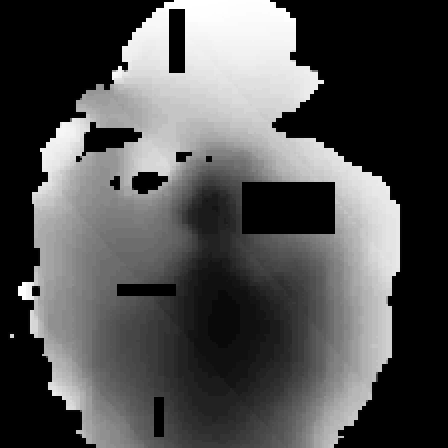}%
        \caption{Augmented depth image~$\mathcal{I}^\mathrm{FD}$.}%
    \end{subfigure}\hfill%
    \begin{subfigure}[t]{0.245\textwidth}%
        \includegraphics[width=\textwidth]{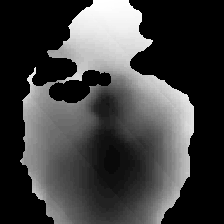}%
        \caption{Binary mask on depth image~$I^\mathrm{FD}\left(\Omega\right)$.}%
    \end{subfigure}%
    \caption{Visualization of the input and target images for \ac{GAN} training on the \stg\ dataset.}%
    \label{f:processing:models:stgmask}%
    \vspace*{-4mm}%
\end{figure}

During training, we augment the input depth images by randomly adding rectangular patches of missing data, thus, creating $\mathcal{I}^\mathrm{FD}_i$ from $I^\mathrm{FD}_i$. The amount and size of these patches are randomly chosen. We use a maximum of 1 big patch and 10 small patches per image. Since we add these patches after creating the binary mask, we can apply a L1-reconstruction-loss to them, forcing the generator to fill them. Another augmentation we apply is a random horizontal flip, where both $I^\mathrm{FC}$ and $I^\mathrm{FD}$ get flipped together. In \Cref{f:processing:models:stgmask}, we show an example of this augmentation process and visualize the binary mask $\Omega$.

The discriminator D is trained on the target set $\tau$, which contains fewer samples than the input set $\upsilon$. We therefore repeat the samples in $\tau$ to match the size of $\upsilon$ during training. The \ac{GAN} is trained for $n_\mathrm{gan} = 100$ epochs using the Adam optimizer and an initial learning rate of $2 \cdot 10^{-4}$. The learning rate gets halved every 10 epochs. We use a batch size of 32 on a single NVIDIA~A100~GPU with \SI{40}{\giga\byte} of memory. The training depends on the dataset size and takes about \qtyrange{30}{40}{\hour}.

The described approach for \ac{GAN} training using the \stg\ dataset is applied to our own dataset as well (see~\Cref{s:processing:dataset} and specifically~\Cref{s:processing:dataset:preprocessing}).

\bigskip\smallskip\smallskip\smallskip

\textbf{RGBDTr training.} After training the \ac{GAN}, we freeze its weights and then train our RGBDTr model in a supervised process. When using a pre-trained backbone for the eye pose feature extractor, we also freeze its weights. We define the loss function of the RGBDTr training as a \ac{MSE} loss:
\begin{align*}
    \mathcal{L} &= \frac{1}{M} \sum_{i=1}^M \left\Vert \hat{g}'_i - g'_i \right\Vert _2^2 \\
    \hat{g}'_i &= \mathrm{RGBDTr} \left( \mathcal{I}_i^\mathrm{FC}, \mathcal{I}_i^\mathrm{FD}, \mathcal{I}_i^\mathrm{REC}, \mathcal{I}_i^\mathrm{LEC}, L', s_i \right)
\end{align*}
where $g'_i$ is the (normalized) ground truth gaze angle of the $i$-th sample, and\\
where $\hat{g}'_i$ is the predicted gaze angle of the $i$-th sample, and\\
where $M$ is the number of samples in the training set, and\\
where $\mathcal{I}$ is an augmented version of $I$, and\\
where $L'$ are the normalized facial landmarks, and\\
where $s_i$ is the subject ID of the $i$-th sample used for the subject-specific bias terms.

If a dataset contains no gaze angles, e.g., the \stg\ dataset, we apply the same loss function, but use the ground truth gaze point coordinates $p$ and the predicted gaze point coordinates $\hat{p}$ instead. The subject-specific bias terms remain unchanged in this case.

During the training of the RGBDTr model, we apply no horizontal flips to the input images. This is because we cannot simply flip the ground truth gaze angles as this would result in wrong labels. Therefore, we only apply random brightness, contrast, saturation, and hue augmentations to the input images. It is important to note that we apply these augmentations on all input images of a sample in the same way. This ensures that the input images of a sample are still consistent with each other. Furthermore, we normalize all RGB images using the ImageNet statistics.

We train the RGBDTr model for $n_\mathrm{rgbdtr} = 25$ epochs using the Adam optimizer and an initial learning rate of $5\cdot 10^{-5}$. The learning rate gets multiplied by $0.1$ every 10 epochs. We use a batch size of 48 for RGBD datasets and 64 for RGB datasets on a single NVIDIA~A100~GPU with \SI{40}{\giga\byte} of memory. The training takes about \qtyrange{7.5}{35}{\hour} depending on the dataset size.

Our RGBDTr model expects eye patches and facial landmarks as input. We already created them for the \stg\ dataset for creation of the discriminator target set~$\tau$. For our own dataset, we can use the landmarks that we obtain during the image preprocessing. However, the \xgaze\ dataset provides only raw facial landmarks detected by dlib on the original images. We therefore have to either convert the existing landmarks or to obtain them using, e.g., the yolov7-face landmark detector. We tried both approaches and describe them more detailed in \Cref{s:evaluation:xgaze:preprocessing}.

\bigskip\bigskip

\textbf{Fine-tuning.} After training both the \ac{GAN} and the RGBDTr model, we fine-tune the entire model in a multi-task learning process similar to the work of \citeauthor*{Lian2019}~\cite{Lian2019}. We reuse the loss functions of both prior training stages and perform three gradient updates per iteration: first the discriminator, second the generator and third the RGBDTr model. This also means that the shared weights of the generator encoders and the fusion block will get updated twice. We use the same Adam optimizer as before, but with an initial learning rate of $5\cdot 10^{-7}$. The learning rate gets multiplied by $0.9$ every epoch and we fine-tune for $n_\mathrm{ft} = 10$ epochs. We use a batch size of 32 on a single NVIDIA~A100~GPU with \SI{40}{\giga\byte} of memory. The fine-tuning takes about \qtyrange{10}{15}{\hour}.

We perform the final fine-tuning only on RGBD datasets, i.e., the \stg\ dataset and ours. For the \xgaze\ dataset, there is no \ac{GAN} training and therefore no need for a fine-tuning stage as the RGBDTr model does not freeze the weights of the generator encoder and the fusion block during training.

\clearpage
\section{Data Collection}
\label{s:processing:datacollection}

As part of this thesis, we collected samples for our own dataset. In this section, we describe the technical setup, the camera calibration process, and the data collection process in detail. The resulting dataset is described in the next section. The data collection process is separate from our pipeline, but uses the same underlying functions, just not in real-time. All scripts used for data collection and the pipeline were written in Python using the packages OpenCV (\package{cv2})~\cite{Bradski2000}, {NumPy}~\cite{Harris2020} and \package{pyrealsense2}~\cite{IntelCorporation2018}.

After the data collection process, we apply the first steps of our processing pipeline to the collected data. This includes the face landmark detection and the face normalization (see~\Cref{s:processing:sequence:landmark,s:processing:sequence:normalization}). This allows us to use the same functions as in the pipeline and preparing the collected data for model training, ensuring that the training and test data is processed in the same way as the live data during inference.

\subsection{Experimental Setup}
\label{s:processing:datacollection:setup}

\begin{figure}[htb]
    \centering
    \begin{tikzpicture}
    \pic [pic=monitor] at (0, 0, 0) {cuboid={width=2, height=22, depth=40, scale=0.25, /tikz/every edge/.style={}}};
    \pic [right=-19mm of monitor, yshift=0.9mm, pic=screen] {cuboid={width=0, height=18, depth=36, scale=0.25, /tikz/every edge/.style={}}};
    \node [above=-1.5cm of monitor.north, rotate=45, align=center, xshift=-3mm] {\SI{27}{\inch} Screen};
    \node [above=-3.35cm of monitor.north, rotate=45, align=center, xshift=8mm] {\scriptsize ($3840 \times 2160$ pixels)};

    \node [below=-2.125cm of monitor.south, rotate=45, align=center, xshift=3mm] (camera) {};
    \draw [rotate=45] ($(camera) + (-0.2, 0)$) ellipse (0.04 and 0.06);
    \draw [rotate=45] ($(camera) + (0.2, 0)$) ellipse (0.04 and 0.06);
    \draw [rotate=45] ($(camera) + (0.4, 0)$) ellipse (0.02 and 0.03);
    \draw [rotate=45] ($(camera) + (-0.02, 0.07)$) -- ++(-0.05, 0) arc [start angle=90, end angle=270, x radius=0.04, y radius=0.07] -- ++(0.1, 0) arc [start angle=-90, end angle=90, x radius=0.04, y radius=0.07] -- cycle;
    \draw [rotate=45] ($(camera) + (0.4, 0.11)$) coordinate (ftr) -- ($(camera) + (-0.25, 0.11)$) coordinate (ftl) arc [start angle=90, end angle=250, x radius=0.08, y radius=0.11] coordinate (fbl) -- ($(camera) + (0.4, -0.11)$) coordinate (fbr) arc [start angle=-90, end angle=90, x radius=0.08, y radius=0.11];
    \draw [rotate=45] ($(ftr) + (0.04, -0.01)$) -- ++(-0.15, 0.05) coordinate (btr) -- ++($(ftl) - (ftr)$) coordinate (btl) arc [start angle=90, end angle=270, x radius=0.10, y radius=0.13] -- (fbl);
    \draw [rotate=45] (ftl) -- (btl);

    \node [below=0cm of camera, rotate=45, align=center] {\scriptsize RealSense D435};

    \rotateRPY[0.25cm/0cm/-0.25cm]{0}{-25}{0}
    \csvreader[head to column names]{setup-face_model_all.csv}{}{
        \fill [cyan!80!black, RPY] (\x * 0.01cm + 0.25cm, \y * 0.01cm, \z * 0.01cm - 0.25cm) circle (0.035cm);
    }

    \coordinate (face) at (\savedx, \savedy, \savedz);
    \coordinate (fc) at ($(face) + (-0.9cm, -0.1cm)$);
    \coordinate (lefteye) at ($(face) + (-1.45cm, 0.5cm)$);
    \coordinate (righteye) at ($(face) + (-0.15cm, 0.22cm)$);

    \coordinate (onscreen) at ($(screen) + (0.5, 0.1, 0)$);

    \draw [thick, ora80] (lefteye) -- (onscreen);
    \draw [thick, ora80] (righteye) -- (onscreen);

    \filldraw [blue] (fc) circle (0.1cm);
    \node [below=0.2cm of fc, anchor=north west, align=left, text=blue, inner sep=2pt, fill=white] {\footnotesize\baselineskip=10pt Face center $c^*$\par};
    \draw [arrow, blue] (fc) -- (onscreen) node [below, midway, sloped] {3D gaze vector $g_3^*$};

    \filldraw [red!80!black] (onscreen) circle (0.05cm);
    \node [above=0.1cm of onscreen, text width=2.5cm, align=center, rotate=45, text=red!80!black] {\scriptsize\baselineskip=10pt Gaze point $p \in \mathbb{R}^2 \sim p^*_3 \in \mathbb{R}^3$\par};

\end{tikzpicture}%
    \caption{Schematic structure of our data acquisition system.}
    \label{f:processing:datacollection:setup}
\end{figure}
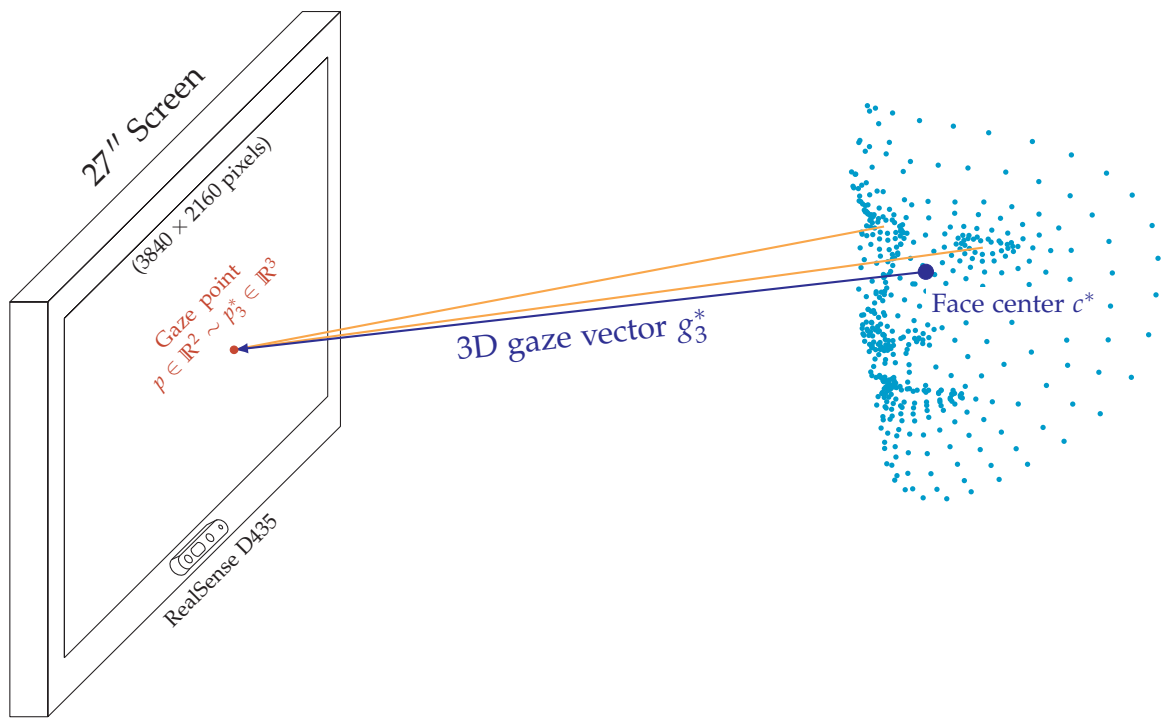

Our experimental setup is similar to the work of \citeauthor*{Lian2019}~\cite{Lian2019}. We use a \SI{27}{\inch} monitor with a resolution of $3840 \times 2160$ pixels, placed between \SIrange{50}{90}{\centi\meter} in front of the subject. The subject is seated on a chair which they can adjust without restrain. The data acquisition is done by a Python script running on the computer that is connected to both the Intel RealSense D435 RGBD camera and the monitor. We describe the data collection process in detail in \Cref{s:processing:datacollection:process}. Our experimental setup is depicted in \Cref{f:processing:datacollection:setup}. The RGBD camera is placed below the monitor display and angled upwards to be able to capture the subject's face.

We define the face center as the center point of the two eyes and the nose. This is similar to the work of \citeauthor*{Zhang2020}~\cite{Zhang2020}. The 3D gaze vector $g_3^*$ is defined as the vector from the face center $c^*$ to the gaze point $p_3^*$. Because we calculate the face center based on the image, we obtain $c$ in the camera coordinate system. However, the 3D gaze vector we want to obtain is in the normalized camera coordinate system: $g'_3$.

In order to un-project the 2D gaze point $p \in \mathbb{R}^2$ (in pixels) in the screen coordinate system to a 3D gaze point $p^*_3 \in \mathbb{R}^3$ (in \unit{\milli\meter}) in the world coordinate system, we need information about the monitor used. The Python package \package{screeninfo}~\cite{Kurczewski2022} provides access to the monitor's physical dimensions as well as its resolution. We assume square pixels and a rectangular monitor that is not curved or bent in any way. We define the origin of the world coordinate system to be in the center of the screen, i.e., $0^*_3 = \left(\begin{matrix} 0 & 0 & 0 \end{matrix} \right)^T \sim \left(\begin{matrix} \sfrac{w}{2} & \sfrac{h}{2} \end{matrix}\right)^T$, where $w$ and $h$ are the width and height of the screen in pixels, respectively. The screen is defined to be in the world's $xy$-plane. To calculate the 3D gaze point $p^*_3$ in the world coordinate system as well as $p_3$ and the 3D gaze vector $g_3$ in the camera coordinate system, we use the following equations:
\begin{align*}
    p^*_3 &= \left[\left(\begin{matrix} p_x \\ p_y \\ 0 \end{matrix}\right) - \left(\begin{matrix} \sfrac{w}{2} \\ \sfrac{h}{2} \\ 0 \end{matrix}\right)\right] \cdot \left(\begin{matrix} \sfrac{W}{w} \\ \sfrac{H}{h} \\ 1 \end{matrix}\right) \\
    p_3 &= \left[E \left(\begin{matrix} p^*_3 \\ 1 \end{matrix}\right)\right]_{1:3} \\
    g_3 &= p_3 - c \\
    g'_3 &= Rg_3
\end{align*}
where $p_x$ and $p_y$ are the $x$- and $y$-coordinates of $p$, respectively, and\\
where $W$ and $H$ are the width and height of the screen in \unit{\milli\meter}, respectively, and\\
where $\left[v\right]_{1:3}$ denotes the first three elements of the vector $v$, and\\
where $E \in \mathbb{R}^{4 \times 4}$ is the extrinsic camera matrix, i.e., the transformation matrix from the world coordinate system to the camera coordinate system, and\\
where $R \in \mathbb{R}^{3 \times 3}$ is the rotation matrix calculated during the face normalization process (see~\Cref{s:processing:sequence:normalization}).

The 3D gaze angle $g'_3$ in the normalized camera coordinate system is then converted to the two gaze angles pitch and yaw as follows:
\[ g' = \left(\begin{matrix} p \\ y \end{matrix}\right) = \left(\begin{matrix} \arcsin\left(\left[g'_3\right]_2\right) \\[2mm] -\arctan\left(\frac{\left[g'_3\right]_1}{\left[g'_3\right]_3}\right) \end{matrix}\right) \]
where $\left[v\right]_i$ denotes the $i$-th element of the vector $v$.

This is the inverse operation of the transformation shown in \Cref{s:processing:sequence:unnorm}. It is important to note that $g'_3$ must be normalized to a length of one before applying the trigonometric functions.

\clearpage 
\subsection{Camera Calibration}
\label{s:processing:datacollection:calibration}

{\spaceskip=3.35pt plus 1pt minus 1.5pt 
In the previous formulas, we used the extrinsic camera matrix $E \in \mathbb{R}^{4 \times 4}$. We determined this matrix using a camera calibration method based on mirrors. The method was proposed by \citeauthor*{Takahashi2016}~\cite{Takahashi2016} in 2016. They also provided code for the calibration process written in MATLAB, which we adapted to be used with the Open-Source-Software GNU~Octave~\cite{Eaton2023}\,\footnote{~We discovered that the code does not work properly when executed on a Linux-Fedora system. Neither the RPM nor the Flathub version worked correctly. However, we found that on both Windows and Linux-Manjaro the code works as expected. We assume that this is due to different compile configurations of GNU~Octave's underlying libraries.}. We create the necessary files for the extrinsic camera matrix estimation in Python, then execute the Octave code, and finally read the results back into Python. The calibration should be done once for each camera and monitor setup.
}

The method by \citeauthor*{Takase2022} is especially useful in our case because it allows the calculation of the extrinsic camera matrix using a single mirror and a fixed camera position. In practice, we display a checkerboard pattern on the screen with a known tile size and take multiple images of the screen through the mirror. As the mirror is the only object in the scene that is not fixed, we have to move it to different positions and orientations for the method to work. We found that the less distance between the mirror and the camera/screen, the better the calibration results are. Furthermore, a greater angle variation usually yields better results, too. Our observations match the findings by \citeauthor*{Hesch2009}~\cite{Hesch2009}. However, it is necessary for the camera to capture the whole screen in one image so that the checkerboard detection algorithm can work. In this thesis, we use a $10 \times 5$ checkerboard with a tile size of \SI{50}{\milli\meter}. Our implementation relies on algorithms provided by the OpenCV (\package{cv2}) package~\cite{Bradski2000}.

\begin{figure}[htb]
    \centering
    \begin{subfigure}[t]{0.47\textwidth}%
        \includegraphics[width=\textwidth]{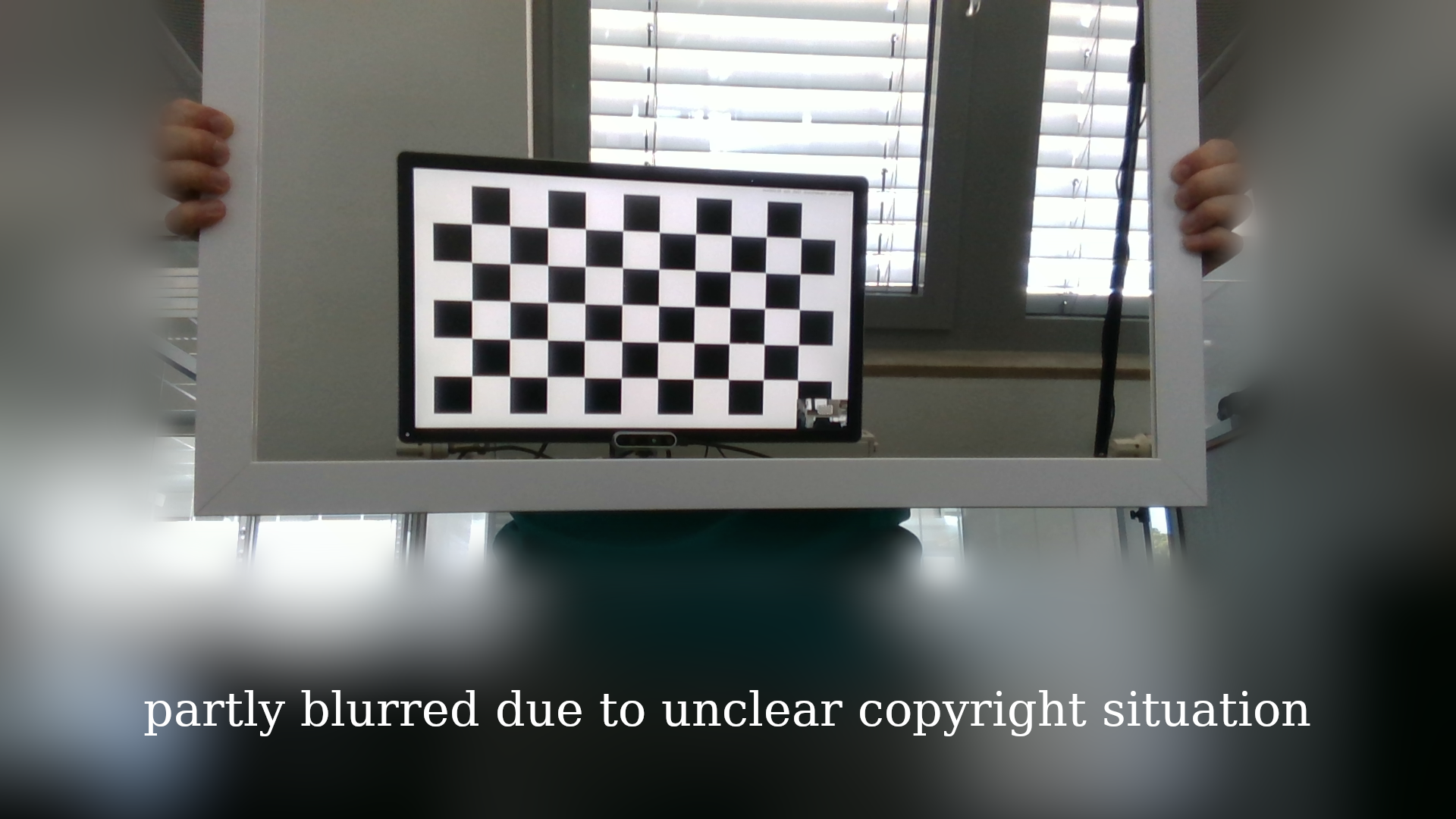}%
        \caption{Mirror far from camera and screen.}%
        \label{f:processing:datacollection:calibration:away}
    \end{subfigure}%
    \hfill
    \begin{subfigure}[t]{0.47\textwidth}%
        \includegraphics[width=\textwidth]{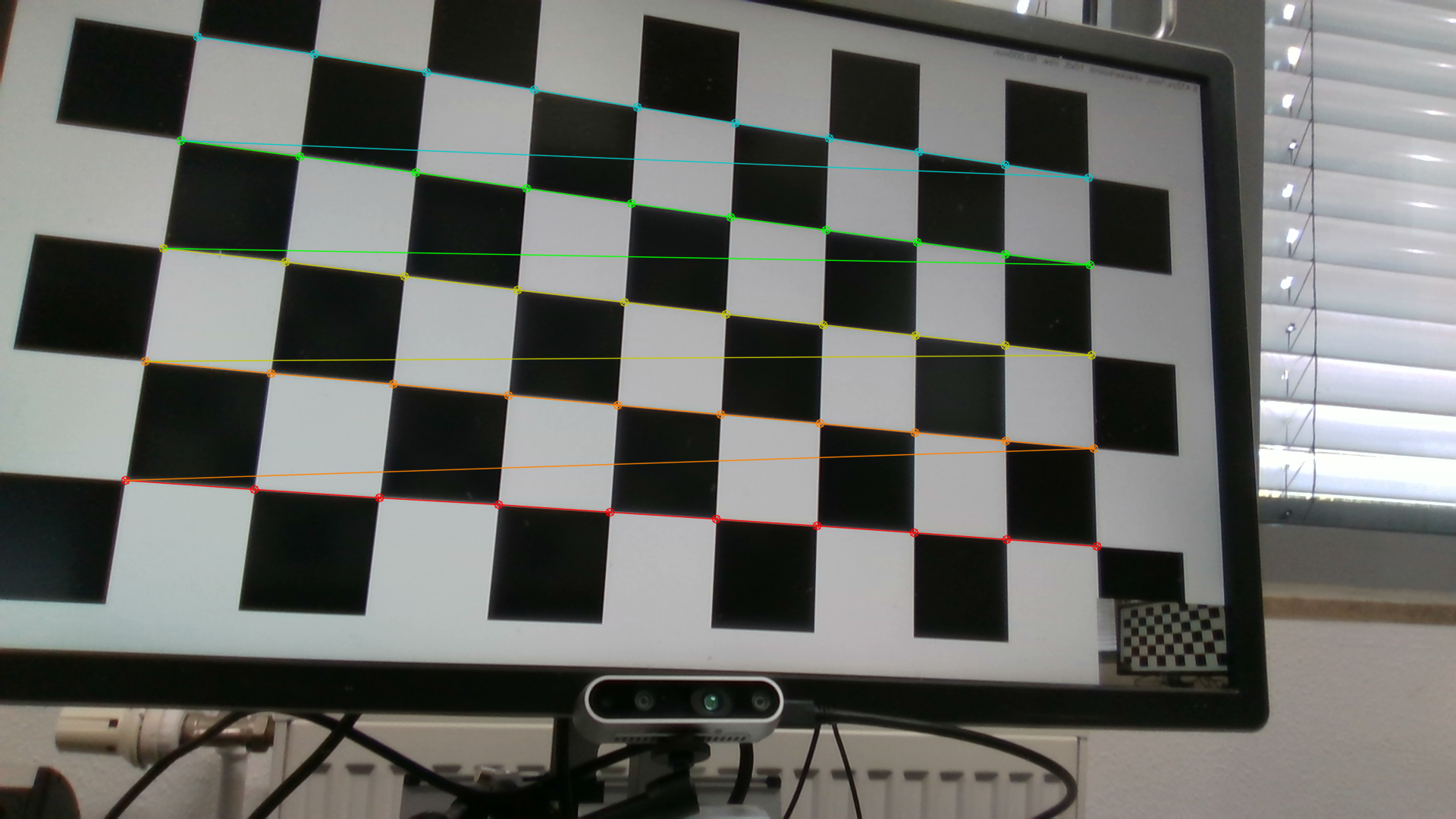}%
        \caption{Mirror close to camera and screen,\\the mirror plane covers the entire image,\\checkerboard pattern found and visualized.}%
        \label{f:processing:datacollection:calibration:near}
    \end{subfigure}%
    \caption{Calibration images for the mirror-based calibration method.}
    \label{f:processing:datacollection:calibration}
\end{figure}

In \Cref{f:processing:datacollection:calibration}, we show two images taken by the RGB sensor of our Intel RealSense D435 camera during the calibration process. The first image shows the mirror far away from the camera and the screen to illustrate the general procedure. In \Cref{f:processing:datacollection:calibration:near}, the mirror was moved closer to the camera, filling the entire image, such that it appears to be taken from the mirror's point-of-view. The checkerboard pattern is clearly visible in the mirror and is detected using OpenCV's algorithms. For illustration purposes, we visualize the detected checkerboard pattern in this image.

\subsection{Data Collection Process}
\label{s:processing:datacollection:process}

Our data collection script consists of three consecutive phases, in which we collect gaze data: the single-point-single-sample phase, the single-point-continuous-sample phase, and the moving-point-continuous-sample phase. In the following, we will refer to a run through as \textit{recording session}. Each recording session consists of the three aforementioned phases in this exact order. During a recording session multiple thousands of samples are collected. Each sample consists of
\begin{itemize}
    \item the RGB image $I^\mathrm{FC}$,
    \item the corresponding aligned depth image $I^\mathrm{FD}$ (not histogram-equalized yet),
    \item the screen coordinates $p$ of the gaze point,
    \item the screen coordinates $p_\mathrm{click}$ of the click point (if available),
    \item the screen coordinates $p_\mathrm{mouse}$ of the mouse cursor (if available),
    \item a boolean value indicating whether the gaze point was on a grid or not.
\end{itemize}
All screen coordinates are in pixels. Depending on the phase, the click point and the mouse cursor are not available.

\begin{figure}[htb]
    \vspace{5mm}\centering%
    \hspace*{\fill}
    \begin{subfigure}{0.4\textwidth}%
        \includegraphics[width=\textwidth]{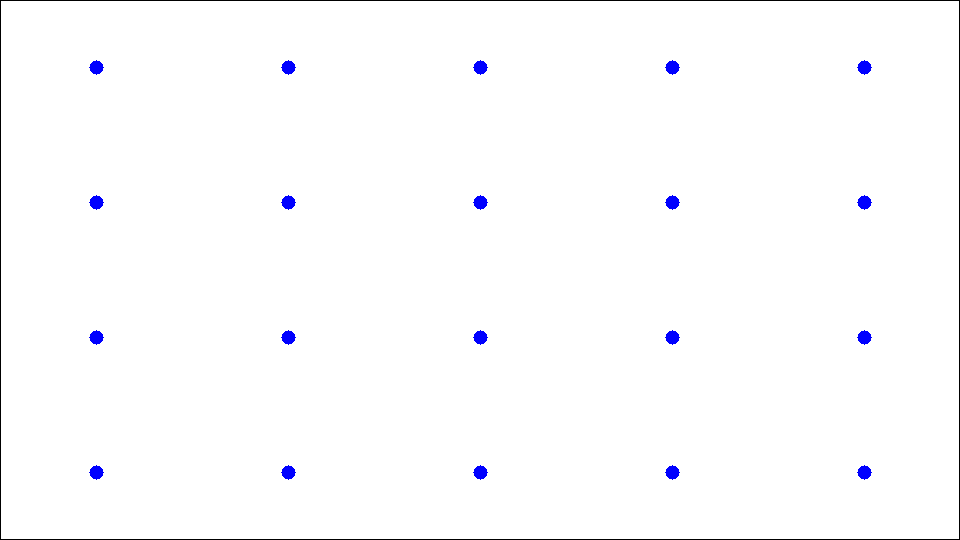}%
        \caption{$5 \times 4$ grid.}\label{f:processing:datacollection:process:grid:1}
    \end{subfigure}%
    \hspace*{\fill}
    \begin{subfigure}{0.4\textwidth}%
        \includegraphics[width=\textwidth]{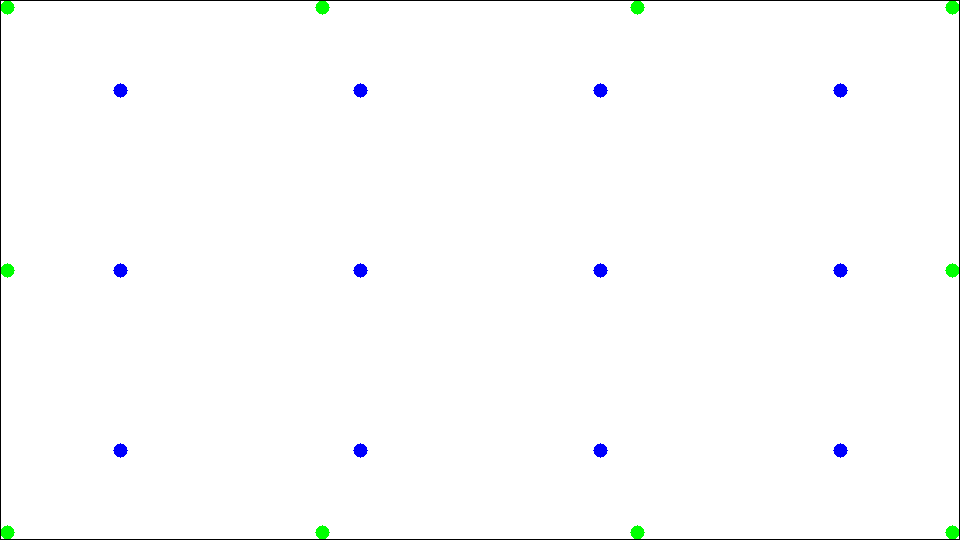}%
        \caption{$4 \times 3$ grid (blue) and border (green).}\label{f:processing:datacollection:process:grid:2}
    \end{subfigure}%
    \hspace*{\fill}
    \caption{Visualization of the grid points used in the single-point-single-sample (\subref{f:processing:datacollection:process:grid:1}) and single-point-continuous-sample (\subref{f:processing:datacollection:process:grid:2}) phases.}
    \label{f:processing:datacollection:process:grid}
    \vspace{5mm}
\end{figure}

\textbf{Single-Point-Single-Sample phase.} In the first phase, subjects are required to look at 100 points and click them using a mouse. 80 points where sampled randomly and uniformly across the entire screen. They differ from recording session to recording session. The remaining 20 points are sampled from a $5 \times 4$ grid spanning $\sfrac{4}{5}$ and $\sfrac{3}{4}$ of the screen's width and height, respectively. It is centered in the middle of the screen, such that the left and right margins as well as the top and bottom margins are equal. The grid points are visualized in \Cref{f:processing:datacollection:process:grid:1}. The 100 total points are shuffled to avoid a bias emerging from, e.g., fatigue. Each point on the screen has a diameter of approximately \SI{2.2}{\milli\meter} and is required to be clicked at within its dimensions. The points are shown one at a time to avoid confusion and to allow for a simple correlation between an image and it gaze point. For each successful click, a single sample is recorded. The sample contains the optional $p_\mathrm{click}$ parameter. The subject is allowed to mis-click a point with no penalties. Usually, the second click is successful.

\textbf{Single-Point-Continuous-Sample phase.} In this phase, we generate a total of 22 points located on two grids. The first one is a $4 \times 3$ grid spanning $\sfrac{3}{4}$ and $\sfrac{2}{3}$ of the screen's width and height, respectively. This yields 12 points, whereas the remaining 10 points are located on a $4 \times 3$ perimeter grid spanning \SI{98}{\percent} of the screen's width and height. All 22 points are visualized in \Cref{f:processing:datacollection:process:grid:2} and they differ from the ones in the first phase. During the recording session, the points are shuffled analogous to the previous phase. Similar to the first phase, there is no more than a single point visible at a time. The subject is asked to fixate the shown point for \SI{10}{\second} and moving their head and upper body as much as possible, generating a wide variety of head and eye poses. During the fixation, the samples are recorded as fast as possible, targeting 30\,fps, i.e., the camera's maximum framerate at full resolution. The subject starts the recording by pressing the space bar and can abort it, if necessary, e.g., when they lost focus. Each sample contains none of the optional parameters $p_\mathrm{click}$ and $p_\mathrm{mouse}$ because the mouse is not used in this phase.

\textbf{Moving-Point-Continuous-Sample phase.} The task of the third and last phase is to follow a moving point with the eyes. A total of 10 points are shown following a randomly generated circular path. The path is visible throughout the sampling process to allow the subject to better focus on the moving point. To ensure that the subjects gaze at the moving point, they are required to follow its trajectory with the mouse cursor. We tried differently shaped paths, such as randomly generated Bézier curves and straight lines, but they suffered from non-uniform movement speed and were generally too hard to track with the mouse cursor. This is why we chose circular paths and a recording duration of \SI{20}{\second} each. Furthermore, the moving point has a bigger diameter of \SI{4.7}{\milli\meter}. Like in the previous phase, the samples are recorded as fast as possible, and the subject starts and may abort the recording. Each sample contains the optional $p_\mathrm{mouse}$ parameter. To ensure that the subject follows the moving point, we require the mouse cursor to be close to the moving point. The distance threshold is set to about \SI{4.7}{\milli\meter}, i.e., double the point radius. If a sample exceeds this threshold, the distance gets added to a penalty score. If the penalty score exceeds a threshold of \SI{350}{\milli\meter} during a recording, it gets aborted automatically and the subject is asked to restart this particular path.

\bigskip

To enhance the sampling performance, the script caches the samples during the continuous recording phases in memory and writes them to disk after each recording. This results in a delay before the subject can start the next recording, but the subject always gets direct feedback on the state of the script, i.e., if it is sampling, writing, or ready. However, because of the used Python language and the underlying libraries, we were not able to hit the target framerate of 30\,fps during the continuous recording phases. This was discovered after the data collection process was finished and we believe that it is due to the framerates of the camera and the monitor. During the development of the script, we were able to hit the target framerate consistently using a monitor with a refresh rate of \SI{165}{\hertz}. However, the monitor used during the data collection process had a refresh rate of \SI{60}{\hertz}, creating longer delays between the displayed frames. This led to a reduced overall sampling rate, which is non-consistent across the dataset.

\section{Our Gaze Tracking Dataset: \oge}
\label{s:processing:dataset}

In this section, we describe the dataset we collected as part of this thesis, which we call the \oge\ dataset. We start by describing the data collection setup and the structure of the resulting dataset. Then, we evaluate the distribution of our dataset and define the training, validation and test subsets. Finally, we describe further preprocessing steps we apply to the dataset before using it for model training.

\subsection{Data Collection Setup and Dataset Structure}
\label{s:processing:dataset:setup}

The dataset was collected on two dates in May and June 2023 in the facilities of the Ostbayerische Technische Hochschule Amberg-Weiden. A total of 12 subjects were recorded in 14 recording sessions. Two subjects participated in two recording sessions each. On the first date in May, two recording sessions were completed successfully, and the remaining 12 recording sessions were conducted in June. We performed the camera calibration process described in \Cref{s:processing:datacollection:calibration} twice on both recording dates.

The experiment setup was described in \Cref{s:processing:datacollection:setup}. The room was lit by two ceiling lights and by indirect sunlight. We used blinds to block direct sunlight facing the subjects, as shown in \Cref{f:processing:datacollection:calibration}. We used the camera's automatic exposure and contrast settings to accommodate for the lighting situation. The subjects were seated on a chair and were allowed to adjust it to their liking. Since the subjects could start each recording on their own, they were allowed to take small breaks in between. The subjects were not paid for their participation.

We collected a total of 136,234 samples resulting in \SI{396}{\giga\byte} of raw data. After the face landmark detection, 136,122 samples remained. We then manually removed all samples in which the subject was blinking, i.e., the eyes were closed, or in which the subject was not looking at the gaze target, i.e., when there was a distraction. We removed a total of 3148~samples (\SI{2.3}{\percent}), with 2105~samples containing eye blinking and 1043~samples being unusable due to distraction, wrong face detection, and other reasons. This resulted in a total of 132,974 samples, which form the \oge\ dataset.

\begin{figure}[htb]
    \centering%
    \captionsetup{width=1\textwidth}
    \includegraphics[width=\textwidth]{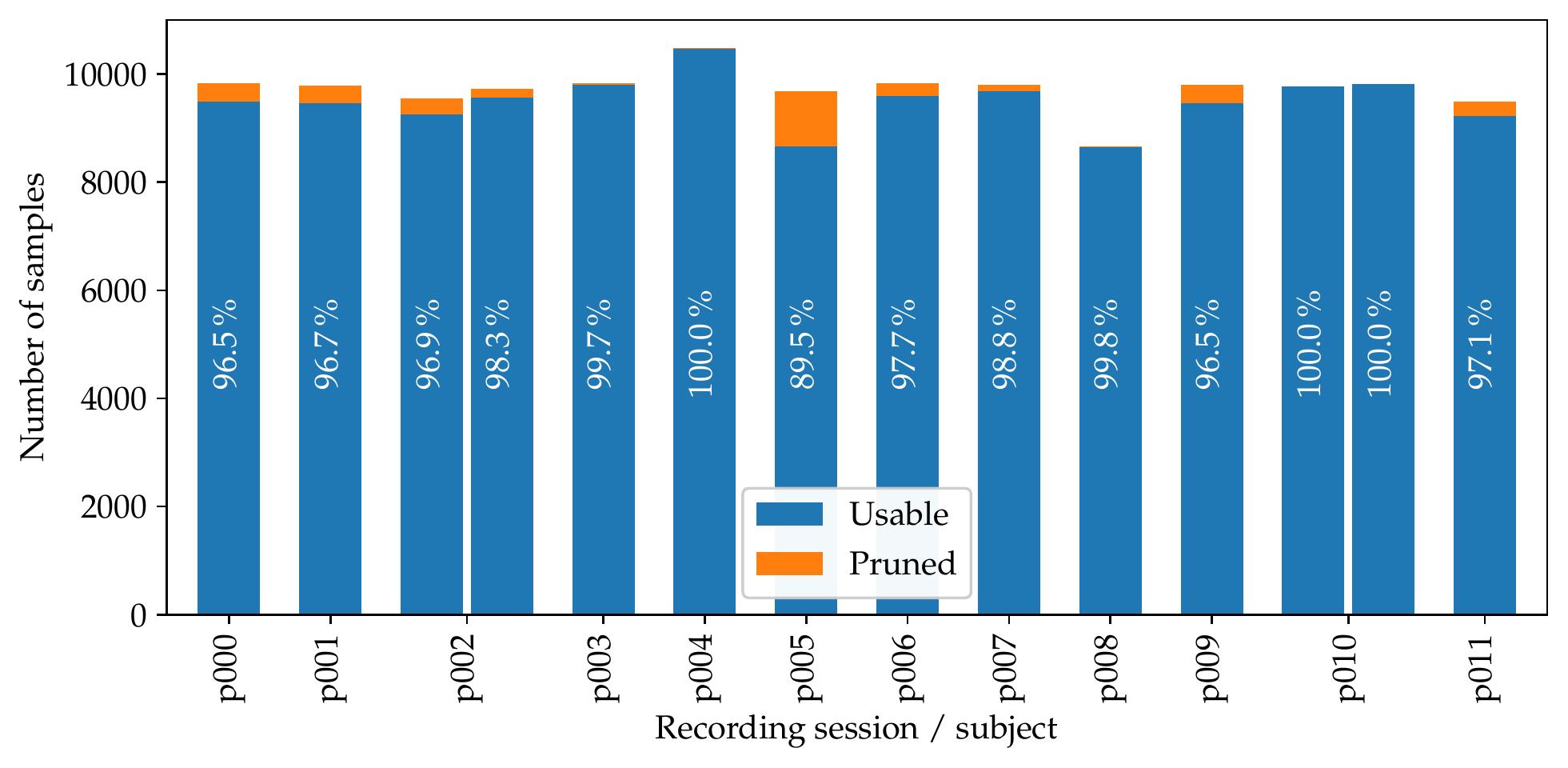}%
    \caption{Number of samples per recording session. (Raw data in \Cref{t:appendix:oge-samples-per-session})}%
    \label{f:processing:dataset:setup:sessions}
\end{figure}

Our dataset consists of 12 HDF(5)\,\footnote{~Hierarchial Data Format; File extension is usually \texttt{.h5}} files, one for each subject. They are named \texttt{p000.h5} to \texttt{p011.h5} and the structure of each HDF file is described below in more detail. \Cref{f:processing:dataset:setup:sessions} shows the number of samples recorded and pruned per recording session. Subjects \texttt{p002} and \texttt{p010} participated in two recording sessions each and have therefore two bars in the plot. The number of samples varies between 8665 and 10,480, which is due to the difficulties during the data collection process (see~\Cref{s:processing:datacollection:process}). The number of samples recorded, pruned, and remaining is listed in \Cref{t:appendix:oge-samples-per-session}. Our \oge\ dataset with its 12 HDF files takes a total of \SI{150}{\giga\byte} of disk space being between the \stg\ dataset with \SI{53.7}{\giga\byte} and the \xgaze\ dataset with \SI{576}{\giga\byte} (face-normalized version with $448 \times 448$ face crops).

\clearpage 

\vspace{-4mm} 
To interact with the HDF files, we use the Python package \package{h5py}~\cite{Collette2023}. We decided to group the collected samples per subject. This allows for easy separation of the dataset into various subsets, e.g., a test set and multiple $k$-fold training and validation subsets. Furthermore, if a subject decides to withdraw their data, it is easy to remove all samples of that subject by simply deleting the corresponding HDF file. In order to accommodate for multiple recording sessions per subject, we store the extrinsic camera matrices $E$ and other parameters in each HDF file, making it self-contained. However, we do not use any of the hierarchial features of the HDF format as we opted for simpler data access. The elements of each HDF file are listed in \Cref{t:processing:dataset:setup:elements}.

Technically, each element in a HDF file is a NumPy-array, with an added dimension compared to the specifications in \Cref{t:processing:dataset:setup:elements}. For example, the array retrieved by the key \texttt{face\_color} is of size $n \times 448 \times 448 \times 3$ with $n$ being the number of samples in that HDF file. The same applies to the other arrays in the upper part of the table. The two arrays in the lower part, namely \texttt{extrinsics} and \texttt{monitor}, are of size $m \times 3 \times \cdot$ with $m$ being the number of recording sessions of that subject, i.e., $m = \max\left(i_s\right) + 1 \in \left\{1, 2\right\}$. We opted for this approach as their values stay the same throughout a recording session and we can therefore save memory by not storing them for each sample. The correlation between a sample and the recording-session-specific parameters is given by the key \texttt{recording\_session}.

The monitor matrix $M$ is defined as follows:\vspace{-2mm}
\[ M = \left(\begin{matrix} w & h \\ \mathrm{offset}_x & \mathrm{offset}_y \\ W & H \end{matrix}\right) \]
where $w$ and $h$ are the width and the height of the screen in pixels, respectively, and
where $W$ and $H$ are the width and the height of the screen in \unit{\milli\meter}, respectively, and
where $\mathrm{offset}_x$ and $\mathrm{offset}_y$ are the horizontal and vertical offset of the screen in pixels. The offsets are only used in multi-monitor setups and are zero in our dataset.

\clearpage

\begin{table}[htbp]
    \centering
    \begin{tabular}{>{\ttfamily}rp{10.5cm}}
        \rmfamily\textbf{Key}   & \bfseries Value \\
        \hline\hline\\[-3mm]
        face\_center            & $c \in \mathbb{R}^3$ in camera coordinate system \\
        face\_color             & $I^\mathrm{FC} \in \mathbb{Z}^{448 \times 448 \times 3} \mathrm{~with~} I^\mathrm{FC}_* \in \left\{0, \dots, 255\right\}$ \\
        face\_depth             & $I^\mathrm{FD} \in \mathbb{Z}^{448 \times 448}  \mathrm{~with~} I^\mathrm{FD}_* \in \left\{0, \dots, 65535\right\}$ \\
        face\_landmarks         & $L' \in \mathbb{R}^{5 \times 2} \mathrm{~with~} L'_* \in \left[0; 448\right]$ \\
        face\_transformation    & $R \in \mathbb{R}^{3 \times 3}$ \\
        gaze                    & $g' \in \mathbb{R}^2$ normalized gaze angles \\
        gaze\_point             & $p_2 \in \mathbb{R}^2$ in screen coordinate system / pixels \\
        head\_rot\_norm         & $hr' \in \mathbb{R}^3$ normalized head rotation angles \\
        in\_recording\_index    & $i_i \in \left\{0, \dots, \#\left(\mathrm{samples\ in\ recording\ \mathnormal{i_r}\ of\ session\ \mathnormal{i_s}}\right) - 1\right\}$\\
        left\_eye\_color        & $I^\mathrm{LEC} \in \mathbb{Z}^{112 \times 112 \times 3} \mathrm{~with~} I^\mathrm{LEC}_* \in \left\{0, \dots, 255\right\}$ \\
        left\_eye\_depth        & $I^\mathrm{LED} \in \mathbb{Z}^{112 \times 112} \mathrm{~with~} I^\mathrm{LED}_* \in \left\{0, \dots, 65535\right\}$ \\
        mouse\_distance         & $d_\mathrm{mouse} \in \mathbb{R}$ \\
        on\_grid                & $b_\mathrm{grid} \in \left\{0, 1\right\}$ \\
        recording\_index        & $i_r \in \left\{0, \dots, 131\right\}$ \\
        recording\_session      & $i_s \in \left\{0, \dots, \#\left(\mathrm{recording\ sessions\ of\ subject}\right) - 1\right\}$ \\
        right\_eye\_color       & $I^\mathrm{REC} \in \mathbb{Z}^{112 \times 112 \times 3} \mathrm{~with~} I^\mathrm{REC}_* \in \left\{0, \dots, 255\right\}$ \\
        right\_eye\_depth       & $I^\mathrm{RED} \in \mathbb{Z}^{112 \times 112} \mathrm{~with~} I^\mathrm{RED}_* \in \left\{0, \dots, 65535\right\}$ \\
        \\[-4mm]\hline\\[-4mm]
        extrinsics              & $\mathcal{E} \in \mathbb{R}^{3 \times 4} = \left[E\right]_{1:3,:}$ \\
        monitor                 & $M \in \mathbb{Z}^{3 \times 2}$ \\
        \hline\\[-3mm]
        \multicolumn{2}{p{0.95\linewidth}}{\small
        where $X_* \in \mathbb{A}$ denotes that all elements $x \in X$ are in the interval or set $\mathbb{A}$, and\par
        where $\left[X\right]_{1:3,:}$ denotes the first three rows of the matrix $X$, and\par
        where the extrinsic camera matrix $E$ is defined as $E = \left(\begin{matrix} \multicolumn{4}{c}{\mathcal{E}} \\ 0 & 0 & 0 & 1 \end{matrix}\right) \in \mathbb{R}^{4 \times 4}$} \\
    \end{tabular}
    \caption{Elements of the HDF files in the \oge\ dataset.}
    \label{t:processing:dataset:setup:elements}
\end{table}

The RGB and depth images are stored as \texttt{uint8} and \texttt{uint16} arrays, respectively. The face landmarks are stored as \texttt{float32} arrays because the landmark detection model returns floating point values which we further transform during the face normalization process. However, we convert them to regular integers during model training and inference. The gaze point $p$ is also stored as \texttt{float32} array. This is because of the third phase (see~\Cref{s:processing:datacollection:process}) where the gaze point is moving along a circular trajectory. The array \texttt{mouse\_distance} contains the distance of the mouse position to the displayed point in pixels. This value is only available for the first and the third phase and may be used to determine a quality score. In this thesis, we do not use this value during training or evaluation. The boolean value \texttt{on\_grid} indicates whether the gaze point was on a grid or not.

The recording index $i_r$ can be used to determine the phase -- and therefore task -- a given sample was collected: $0 \le i_r < 100$ indicates the single-point-single-sample phase, $100 \le i_r < 122$ indicates the single-point-continuous-sample phase, and $122 \le i_r < 132$ indicates the moving-point-continuous-sample phase. The index $i_i$ indicates the index of the sample within a recording $i_r$. The tuple $\left(i_s, i_r, i_i\right)$ is unique within a subject's HDF file.

\subsection{Data Distribution and Train-Val-Test-Split}
\label{s:processing:dataset:distribution}

The subjects of our \oge\ dataset can be categorized by gender and if they wore glasses during the data collection process. We then divide our dataset~$\mathcal{D}$ into two disjoint sets based on these two features: the training set~$\mathcal{D}_\mathrm{tr} \subset \mathcal{D}$ and the test/evaluation set~$\mathcal{D}_\mathrm{te} \subset \mathcal{D}$, with $\mathcal{D}_\mathrm{tr} \cap \mathcal{D}_\mathrm{te} = \emptyset$ and $\mathcal{D}_\mathrm{tr} \cup \mathcal{D}_\mathrm{te} = \mathcal{D}$. We assign \SI{25}{\percent} of the subjects to $\mathcal{D}_\mathrm{te}$ as shown in \Cref{t:processing:dataset:distribution:subjects}.

We tried to make the split as fair as possible, but since there is only a single female subject wearing no glasses, we assigned her to the training set. Furthermore, we assigned the two subjects participating in two recording sessions each to the training set in order to increase the number of training samples. The remaining subjects were assigned randomly to the training and test sets, resulting in a total of 105,218 and 27,756 samples in $\mathcal{D}_\mathrm{tr}$ and $\mathcal{D}_\mathrm{te}$, respectively.

\begin{table}[htb]
    \centering
    \begin{tabular}{>{\ttfamily}ccccl@{\hspace{1mm}}r}
        \rmfamily \textbf{Subject} & \textbf{Gender} & \textbf{Glasses} & \textbf{Set} & \multicolumn{2}{c}{\textbf{Samples}} \\
        \hline\hline
        p000 & Male     & \xmark    & Test  &&   9487 \\\hline
        p001 & Female   & \xmark    & Train &&   9464 \\\hline
        p002 & Male     & \cmark    & Train && 18,833 \\\hline
        p003 & Female   & \cmark    & Train &&   9812 \\\hline
        p004 & Male     & \cmark    & Train && 10,477 \\\hline
        p005 & Female   & \cmark    & Test  &&   8666 \\\hline
        p006 & Male     & \cmark    & Test  &&   9603 \\\hline
        p007 & Male     & \xmark    & Train &&   9688 \\\hline
        p008 & Male     & \xmark    & Train &&   8652 \\\hline
        p009 & Male     & \xmark    & Train &&   9469 \\\hline
        p010 & Male     & \cmark    & Train && 19,593 \\\hline
        p011 & Male     & \xmark    & Train &&   9230 \\\hline\hline
        \multirow{2}{*}{\rmfamily\textbf{Total}} & \textbf{F: 3} & \textbf{\cmark: 6} & \textbf{Train: 9} & \textbf{Train:} & \textbf{105,218} \\
        & \textbf{M: 9} & \textbf{\xmark: 6} & \textbf{Test: 3} & \textbf{Test:} & \textbf{27,756} \\
    \end{tabular}
    \caption{Classification of the subjects in the \oge\ dataset.}
    \label{t:processing:dataset:distribution:subjects}
\end{table}

\begin{figure}[htb]
    \centering
    \begin{subfigure}{0.49\textwidth}%
        \includegraphics[width=\textwidth]{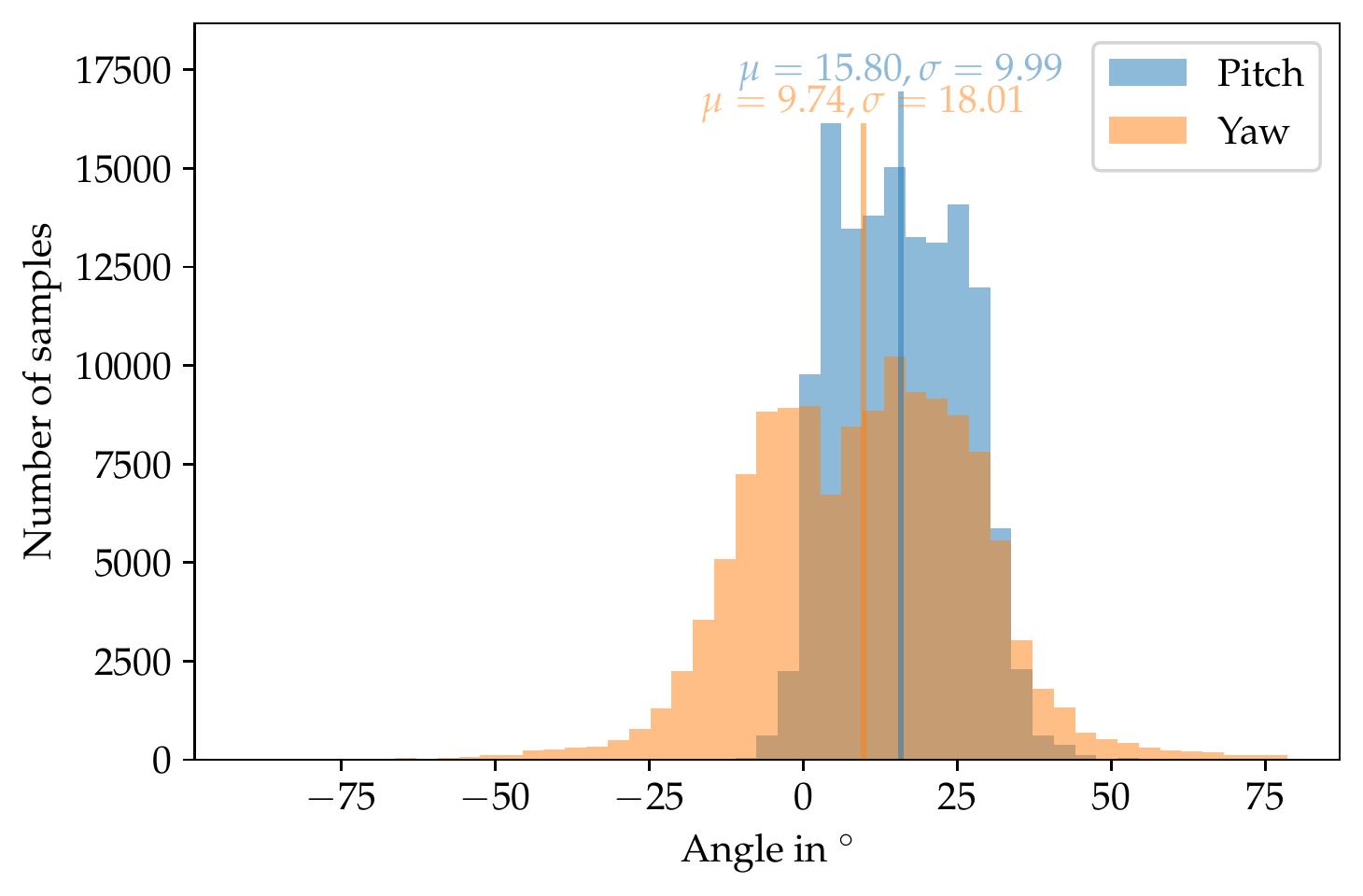}%
        \vspace*{-2mm}%
        \caption{Histogram of the normalized gaze angles.}%
        \label{f:processing:dataset:distribution:hist:1d-gaze}%
        \vspace*{2mm}%
    \end{subfigure}%
    \hfill%
    \begin{subfigure}{0.49\textwidth}%
        \includegraphics[width=\textwidth]{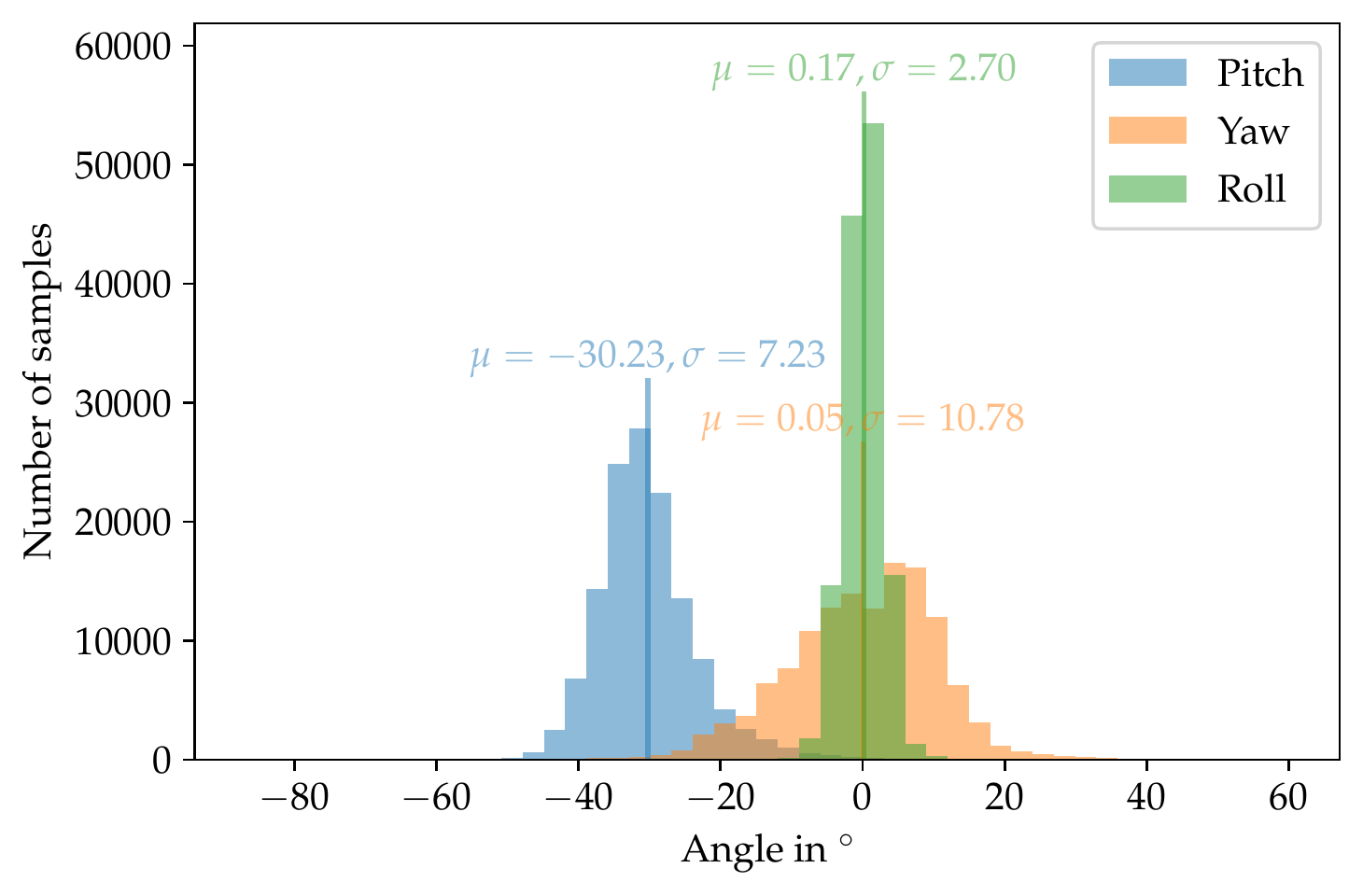}%
        \vspace*{-2mm}%
        \caption{Histogram of the normalized head rotation angles.}%
        \label{f:processing:dataset:distribution:hist:1d-head}%
        \vspace*{2mm}%
    \end{subfigure}
    \begin{subfigure}{\textwidth}%
        \includegraphics[width=\textwidth]{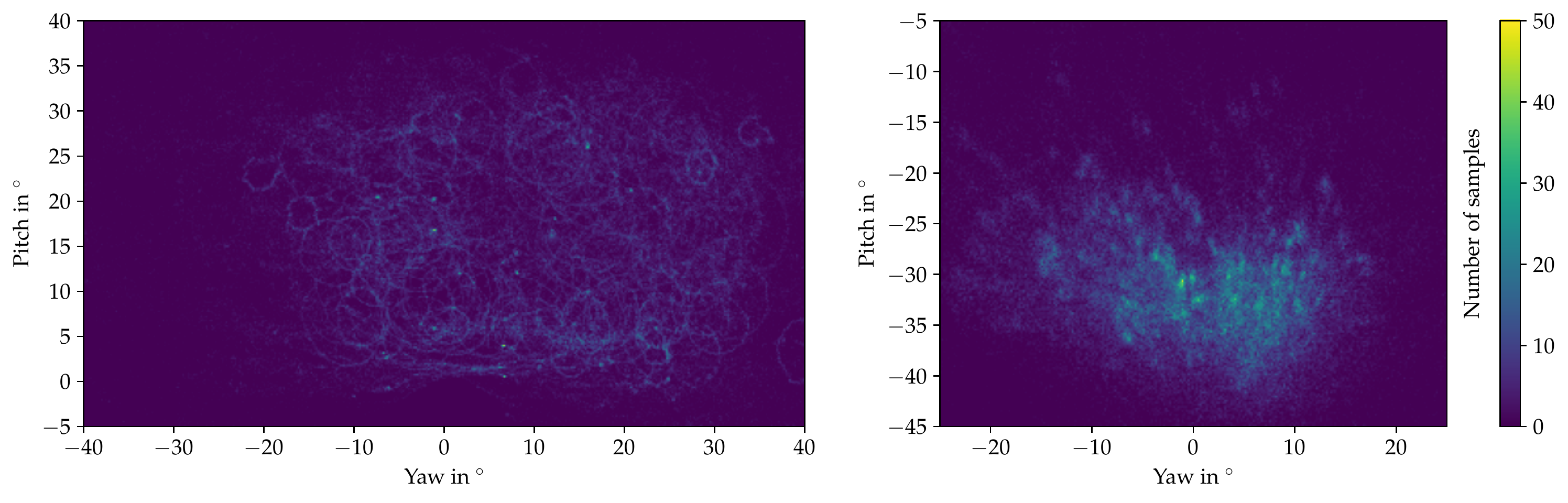}%
        \vspace*{-2mm}%
        \caption{2D-histograms of the normalized gaze (left) and head rotation angles (right).}%
        \label{f:processing:dataset:distribution:hist:2d}%
        \vspace*{2mm}%
    \end{subfigure}%
    \caption{Histograms of the normalized gaze and head rotation angles in the \oge\ dataset.}%
    \label{f:processing:dataset:distribution:hist}%
\end{figure}

In \Cref{f:processing:dataset:distribution:hist}, we show histograms of the normalized gaze angles $g'$ and the normalized head rotation angles $hr'$. The pitch angle of the normalized gaze angles shows a bias towards positive values with a mean value of $\mu = \SI{15.80}{\degree}$ and a standard deviation of $\sigma = \SI{9.99}{\degree}$, which is expected due to the experimental setup. Since the camera is mounted below the screen and the target gaze points are therefore located above the camera, the pitch gaze angle is expected to be positive, i.e., the subjects are looking upwards from the camera's perspective. This is visualized in \Cref{f:processing:dataset:distribution:hist:1d-gaze}. However, the yaw gaze angle also shows a -- albeit less pronounced -- bias towards positive values with $\mu = \SI{9.74}{\degree}$ and $\sigma = \SI{18.01}{\degree}$, which indicates an asymmetry in our gaze acquisition system. This is clearly visible in the 2D-histogram shown in \Cref{f:processing:dataset:distribution:hist:2d} on the left.

\newpage 

The distribution of the normalized head rotation angles is visualized in \Cref{f:processing:dataset:distribution:hist:1d-head}. The roll angle is distributed close around zero with $\mu = \SI{0.17}{\degree}$ and $\sigma = \SI{2.70}{\degree}$. This is expected since the figure shows the normalized head rotation angles where rotations around the roll axis should have been counteracted. With $\mu = \SI{-30.23}{\degree}$ and $\sigma = \SI{7.23}{\degree}$, the pitch angle is distributed symmetrically around its mean value. The yaw angle has a mean value of $\mu = 0.05$ but is distributed asymmetrically with $\sigma = \SI{10.78}{\degree}$ being larger than the standard deviation of the pitch angle. This is also visible in the 2D-histogram shown in \Cref{f:processing:dataset:distribution:hist:2d} on the right.

After splitting the dataset $\mathcal{D}$ into the training set $\mathcal{D}_\mathrm{tr}$ and test set $\mathcal{D}_\mathrm{te}$ as described above, we further divide the training set into training and validation subsets. We use a $k$-fold cross-validation approach with $k = 5$ and therefore split $\mathcal{D}_\mathrm{tr}$ into five pairs of subsets $\left(\mathcal{D}_\mathrm{tr}^{(1)}, \mathcal{D}_\mathrm{val}^{(1)}\right), \dots, \left(\mathcal{D}_\mathrm{tr}^{(5)}, \mathcal{D}_\mathrm{val}^{(5)}\right)$ with $\mathcal{D}_\mathrm{tr}^{(i)} \cap \mathcal{D}_\mathrm{val}^{(i)} = \emptyset$ and $\mathcal{D}_\mathrm{tr}^{(i)} \cup \mathcal{D}_\mathrm{val}^{(i)} = \mathcal{D}_\mathrm{tr}$ for all $i \in \left\{1, \dots, 5\right\}$. We assign subjects to the training and validation subsets randomly, but we ensure that the two subjects with two recording sessions each do not appear in the same validation subset to avoid a too small training subset. However, we do not take the subject's gender or if they wore glasses into consideration for this $k$-fold split. We ensure only that no subject is assigned to multiple validation subsets, i.e., $\mathcal{D}_\mathrm{val}^{(i)} \cap \mathcal{D}_\mathrm{val}^{(j)} = \emptyset$ for all $i \ne j $ with $i, j \in \left\{1, \dots, 5\right\}$.

With an exception to $\mathcal{D}_\mathrm{tr}^{(1)}$ all training subsets have a similar number of samples at around 86,000. In \Cref{t:processing:dataset:distribution:split}, we show our $k$-fold cross-validation split and the number of samples in each training subset. The validation subsets contain the remaining samples of the original training set, i.e., $\left\vert\mathcal{D}_\mathrm{val}^{(i)}\right\vert = \left\vert\mathcal{D}_\mathrm{tr}\right\vert - \left\vert\mathcal{D}_\mathrm{tr}^{(i)}\right\vert$ with $i \in \left\{1, \dots, 5\right\}$, where $\left\vert\mathcal{D}_\mathrm{tr}\right\vert = 105,218$ as shown in \Cref{t:processing:dataset:distribution:subjects}. It is important to note that in \Cref{t:processing:dataset:distribution:split} all subjects in the test set are not part of the training or validation subsets and therefore do not appear in the table.

\begin{table}[htbp]
    \centering%
    \begin{tabular}{@{}r|*{9}{@{\hspace{3mm}}c}|l@{~$=$~}r@{}}%
        \textbf{\small Split} & \texttt{p001} & \texttt{p002} & \texttt{p003} & \texttt{p004} & \texttt{p007} & \texttt{p008} & \texttt{p009} & \texttt{p010} & \texttt{p011} & \multicolumn{2}{c}{\textbf{\small Samples}} \\\hline
        1 & \textcolor{orange}{Val}   & \textcolor{orange}{Val}   & \textcolor{cyan!60!blue}{Train} & \textcolor{cyan!60!blue}{Train} & \textcolor{cyan!60!blue}{Train} & \textcolor{cyan!60!blue}{Train} & \textcolor{cyan!60!blue}{Train} & \textcolor{cyan!60!blue}{Train} & \textcolor{cyan!60!blue}{Train} & $\left\vert\mathcal{D}_\mathrm{tr}^{(1)}\right\vert$ & 76,921 \\
        2 & \textcolor{cyan!60!blue}{Train} & \textcolor{cyan!60!blue}{Train} & \textcolor{cyan!60!blue}{Train} & \textcolor{orange}{Val}   & \textcolor{cyan!60!blue}{Train} & \textcolor{cyan!60!blue}{Train} & \textcolor{orange}{Val}   & \textcolor{cyan!60!blue}{Train} & \textcolor{cyan!60!blue}{Train} & $\left\vert\mathcal{D}_\mathrm{tr}^{(2)}\right\vert$ & 85,272 \\
        3 & \textcolor{cyan!60!blue}{Train} & \textcolor{cyan!60!blue}{Train} & \textcolor{cyan!60!blue}{Train} & \textcolor{cyan!60!blue}{Train} & \textcolor{orange}{Val}   & \textcolor{orange}{Val}   & \textcolor{cyan!60!blue}{Train} & \textcolor{cyan!60!blue}{Train} & \textcolor{cyan!60!blue}{Train} & $\left\vert\mathcal{D}_\mathrm{tr}^{(3)}\right\vert$ & 86,878 \\
        4 & \textcolor{cyan!60!blue}{Train} & \textcolor{cyan!60!blue}{Train} & \textcolor{orange}{Val}   & \textcolor{cyan!60!blue}{Train} & \textcolor{cyan!60!blue}{Train} & \textcolor{cyan!60!blue}{Train} & \textcolor{cyan!60!blue}{Train} & \textcolor{cyan!60!blue}{Train} & \textcolor{orange}{Val}   & $\left\vert\mathcal{D}_\mathrm{tr}^{(4)}\right\vert$ & 86,176 \\
        5 & \textcolor{cyan!60!blue}{Train} & \textcolor{cyan!60!blue}{Train} & \textcolor{cyan!60!blue}{Train} & \textcolor{cyan!60!blue}{Train} & \textcolor{cyan!60!blue}{Train} & \textcolor{cyan!60!blue}{Train} & \textcolor{cyan!60!blue}{Train} & \textcolor{orange}{Val}   & \textcolor{cyan!60!blue}{Train} & $\left\vert\mathcal{D}_\mathrm{tr}^{(5)}\right\vert$ & 85,625 \\
    \end{tabular}%
    \caption{Train-validation-split of the training subset $\mathcal{D}_\mathrm{tr}$ into five pairs of subsets $\left(\mathcal{D}_\mathrm{tr}^{(1)}, \mathcal{D}_\mathrm{val}^{(1)}\right), \dots, \left(\mathcal{D}_\mathrm{tr}^{(5)}, \mathcal{D}_\mathrm{val}^{(5)}\right)$.}%
    \label{t:processing:dataset:distribution:split}%
\end{table}

In the appendix, we show 2D-histograms of the distributions of the normalized gaze angles for each training and validation subset as well as for the test set. In \Cref{f:appendix:oge-gaze-angles-train-val}, we show the distribution of the training subsets in the left column and the distribution of the validation subsets in the right column. This allows for a visual comparison of the distributions of the training and validation subsets. The distribution of the test set is shown in \Cref{f:appendix:oge-gaze-angles-test}, which contains data from three subjects.

{\spaceskip=3.4pt plus 1pt minus 1.5pt 
The validation subset $\mathcal{D}_\mathrm{val}^{(2)}$ shows a different distribution compared to the other validation subsets as there are speckles in the 2D-histogram. These speckles can be found in all training subsets but $\mathcal{D}_\mathrm{tr}^{(2)}$ and the test set. This is due to the second phase (see~\Cref{s:processing:datacollection:process}) of subject \texttt{p009}. The subject did not move their head continuously during the recordings and therefore an over-proportional number of similar images were taken, resulting in similar normalized gaze angles. This is what produces the speckles in the 2D-histograms. However, we decided to keep the data as it is a valid recording and the subject did alter their facial expressions in some of these recordings. We argue that the variation in the facial expressions is helping the model to generalize better and therefore compensate for the lack of variation in the normalized gaze angles in this subject's recordings.
}

In all 2D-histograms of the normalized gaze angles (\Cref{f:processing:dataset:distribution:hist:2d,f:appendix:oge-gaze-angles-train-val,f:appendix:oge-gaze-angles-test}), circular patterns are clearly visible. Sometimes they are blurry and jittery, but they are always recognizable as oval-shaped patterns. This is a remnant of the third phase (see~\Cref{s:processing:datacollection:process}) since the subjects are focused on the task and move their head only slightly. In contrast, during the second phase, the subjects were moving their head continuously and therefore generate no patterns but rather spread the samples across multiple bins in the 2D-histograms, with the exception being subject \texttt{p009} as described above. The first phase has too few samples to be clearly visible in the 2D-histograms.

\begin{figure}[htb]
    \centering%
    \vspace*{-4mm}%
    \captionsetup{width=1\textwidth}
    \begin{subfigure}{0.5\textwidth}%
        \captionsetup{width=0.75\textwidth}%
        \includegraphics[width=\textwidth]{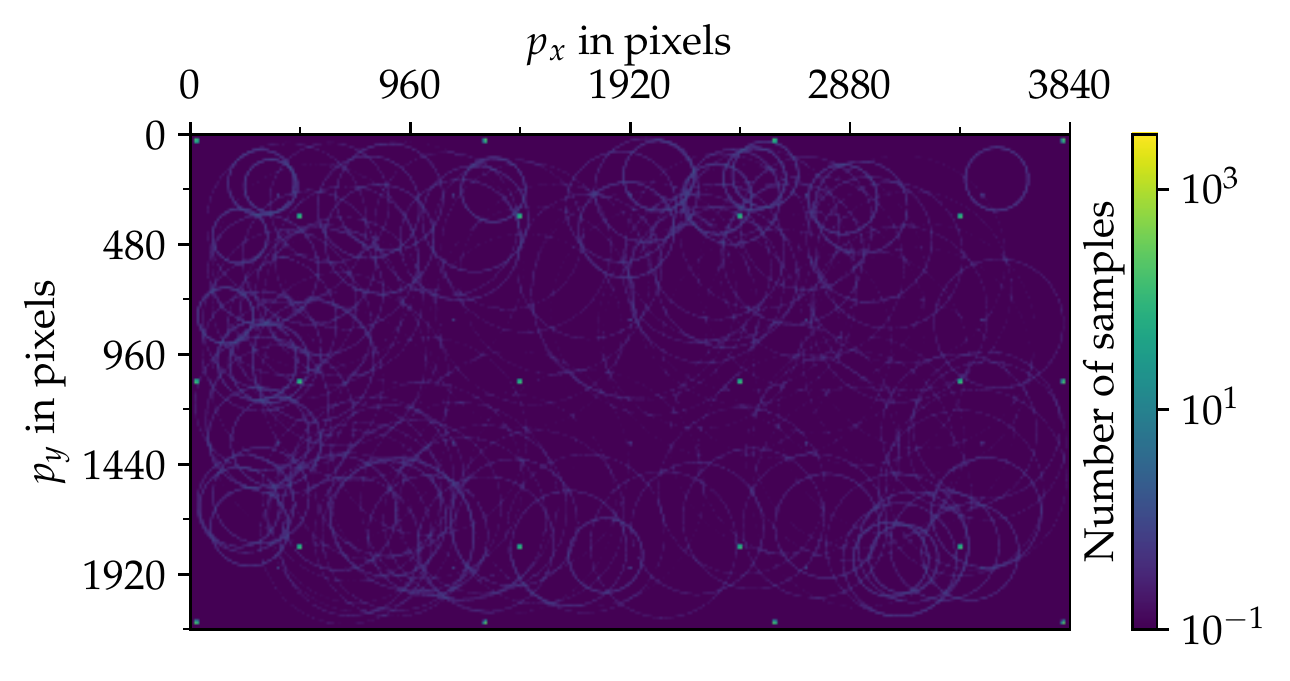}%
        \vspace*{-4mm}%
        \caption{On-screen gaze point distribution with grid~points (logarithmic scale).}%
        \label{f:processing:dataset:distribution:gaze-points:all}%
    \end{subfigure}%
    \hfill%
    \begin{subfigure}{0.5\textwidth}%
        \captionsetup{width=0.75\textwidth}%
        \includegraphics[width=\textwidth]{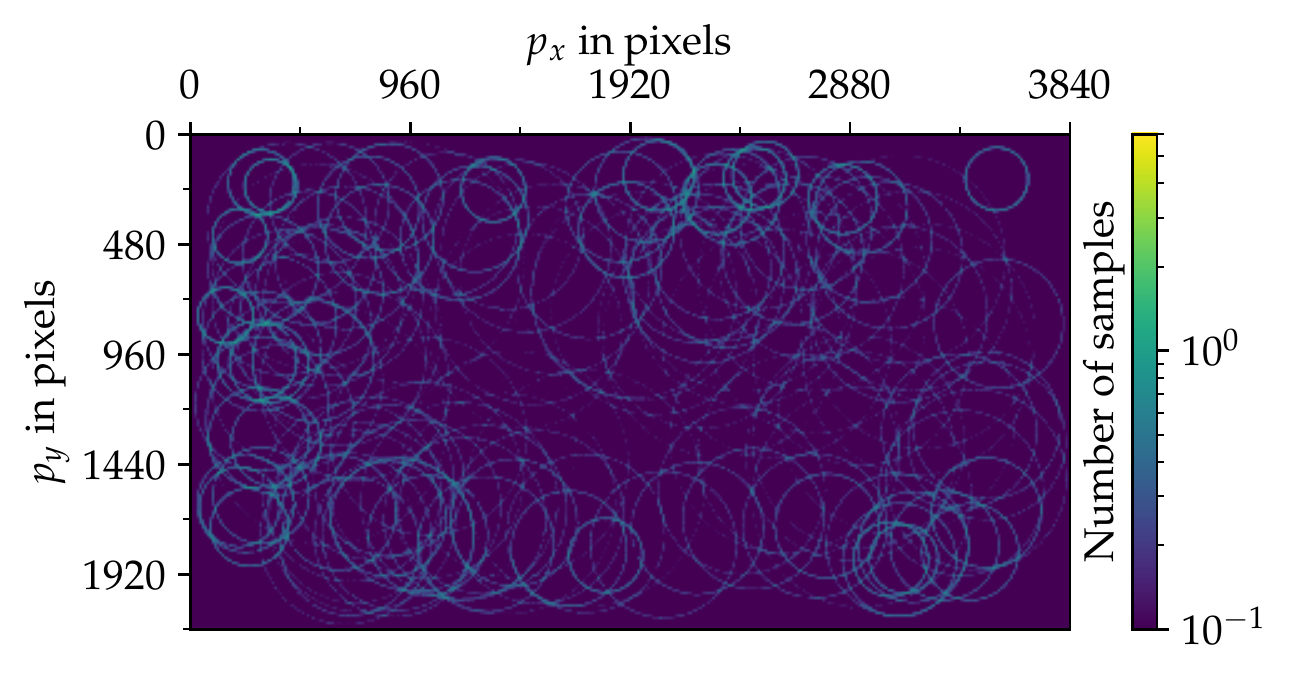}%
        \vspace*{-4mm}%
        \caption{On-screen gaze point distribution without grid~points (logarithmic scale).}%
        \label{f:processing:dataset:distribution:gaze-points:off}%
    \end{subfigure}%
    \caption{On-screen gaze point distribution in the \oge\ dataset.}%
    \label{f:processing:dataset:distribution:gaze-points}%
    \vspace*{-4mm}
\end{figure}

In order to validate the 2D-histograms shown in \Cref{f:processing:dataset:distribution:hist:2d,f:appendix:oge-gaze-angles-train-val,f:appendix:oge-gaze-angles-test}, we can compare them to the distribution of the on-screen gaze points. In \Cref{s:processing:datacollection:process}, we stated that 80 points in the first phase were generated at a random position uniformly from across the screen. Furthermore, all circular paths in the third phase were generated randomly. In contrast, 20 points in the first phase and all 22 points in the second phase were grid points and therefore the same for each subject.

\begin{figure}[htb]
    \vspace*{-2mm}%
    \centering\captionsetup{width=\textwidth} 
    \includegraphics[width=0.9\textwidth]{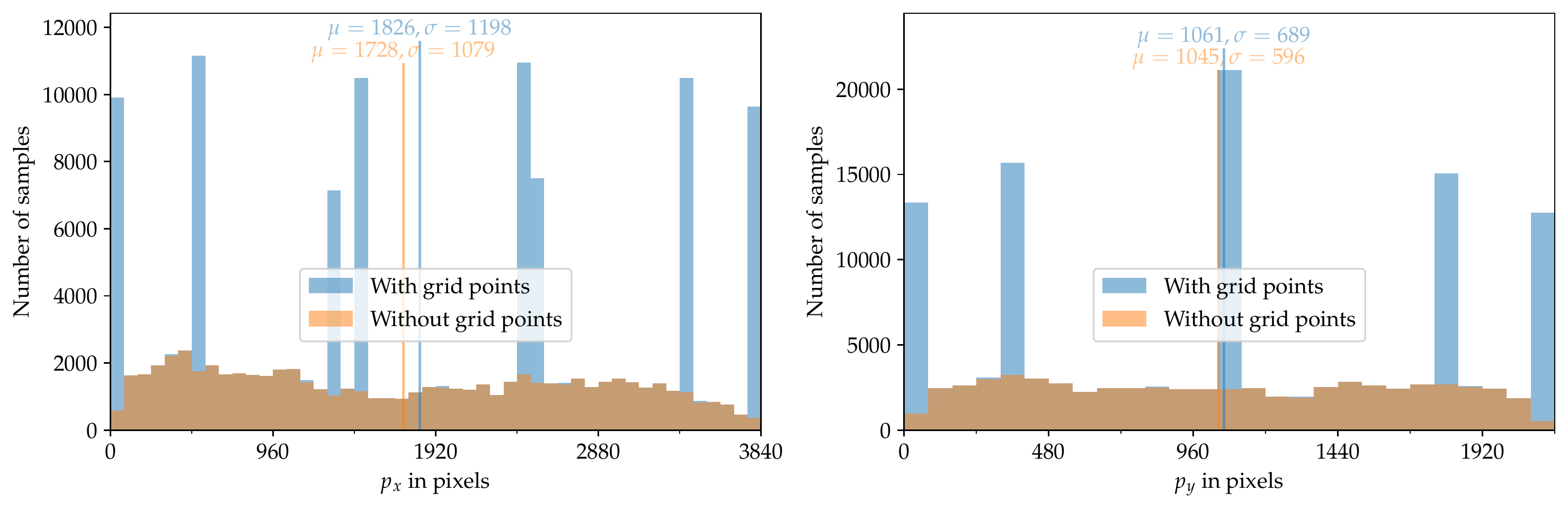}%
    \vspace*{-2mm}%
    \caption{Histogram of the on-screen gaze points in the \oge\ dataset.}%
    \label{f:processing:dataset:distribution:gaze-points-hist}%
    \vspace*{-2mm}
\end{figure}

In \Cref{f:processing:dataset:distribution:gaze-points}, we show two distributions in the form of 2D-histograms: one with all on-screen gaze points (\Cref{f:processing:dataset:distribution:gaze-points:all}) and one without the grid points (\Cref{f:processing:dataset:distribution:gaze-points:off}). Both 2D-histograms are in logarithmic color scale to enhance the visualization because the number of samples of the 22 grid points of the second phase are more than an order of magnitude higher than the number of samples per point on, e.g., the circular paths. To further improve the visibility of the circular paths, \Cref{f:processing:dataset:distribution:gaze-points:off} shows the 2D-histogram without the grid points. The corresponding single-axis histograms are shown in \Cref{f:processing:dataset:distribution:gaze-points-hist} where a slight bias towards the upper left quarter of the screen is visible: the histograms show the distribution along the x-axis and the y-axis, respectively. Both mean values are slightly smaller than expected with $\mu_x = 1826$ and $\mu_y = 1061$. We would expect that $\mu_x = \sfrac{w}{2} = 1920$ and $\mu_y = \sfrac{h}{2} = 1080$. This is due to the randomness of the circular paths in the third phase as the means for all randomly generated points are less than the means over all points including the grid points.

For further comparison, we include the 2D-histograms of the on-screen gaze points for each training and validation subset as well as the test set in the appendix: \Cref{f:appendix:oge-gaze-points-train-val} shows the distributions in the training subsets in the left column and the distributions in the validation subsets in the right column. The distribution of the test set is shown in \Cref{f:appendix:oge-gaze-points-test}. All on-screen gaze point 2D-histograms are in logarithmic color scale.

\subsection{Dataset Preprocessing for \acs{GAN} Training}
\label{s:processing:dataset:preprocessing}

\begin{figure}[htb]
    \centering%
    \includegraphics[width=\textwidth]{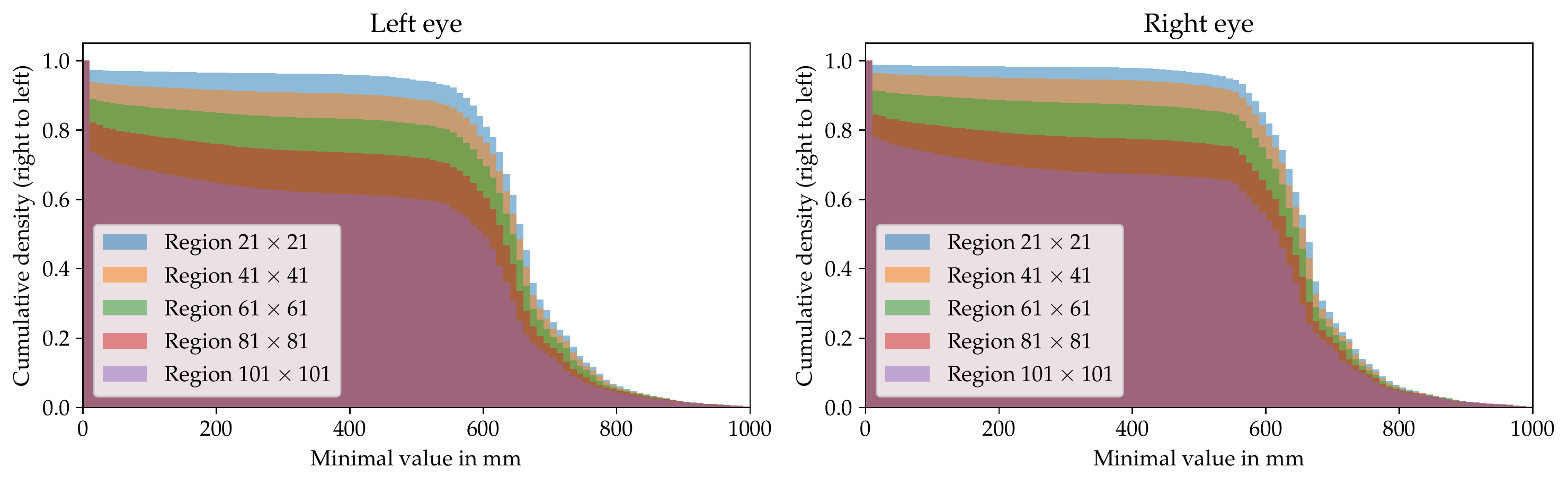}%
    \caption{Cumulative histogram of the minimum depth values in the eye regions of the \oge\ dataset.}%
    \label{f:processing:dataset:preprocessing:ogehistogram}%
\end{figure}

We apply a similar preprocessing procedure to the depth maps as described in \Cref{s:processing:models:training} to obtain the discriminator target set $\tau$ for the \ac{GAN} training. Our \oge\ dataset contains depth maps with a missing depth data value of zero, i.e., $\mathcal{O}=0$, which is identical to the \stg\ dataset. Since our dataset includes the facial landmarks, we can use them to examine the depth maps in the eye regions. \Cref{f:processing:dataset:preprocessing:ogehistogram} shows the cumulative distribution of the minimal depth values in various sized, square regions around each eye. Analogous to \Cref{f:processing:models:stghistogram}, the cumulation was performed from right to left. Because our dataset provides input images of size $448 \times 448$ pixels compared to the $224 \times 224$ pixels of the \stg\ dataset, we increase the size of the examination regions accordingly.

A comparison of the results between our dataset (\Cref{f:processing:dataset:preprocessing:ogehistogram}) and the \stg\ dataset (see \Cref{s:processing:models:training} and \Cref{f:processing:models:stghistogram}) reveals major differences. First, the \oge\ dataset contains substantially less missing data artifacts in the eye regions, which is clearly visible by the plateau at the beginning of the curve being in the range from \SIrange{74}{99}{\percent}. In contrast, the plateau in the \stg\ dataset is in the range from \SIrange{15}{35}{\percent}. Second, the data shows that in the \oge\ dataset, less artifacts occur even in larger areas around the eyes. In \Cref{f:processing:dataset:preprocessing:ogehistogram}, the difference between the various sized regions is less in percent than in \Cref{f:processing:models:stghistogram}.

\begin{figure}[htb]
    \centering
    \begin{subfigure}{0.49\textwidth}%
        \includegraphics[width=\textwidth]{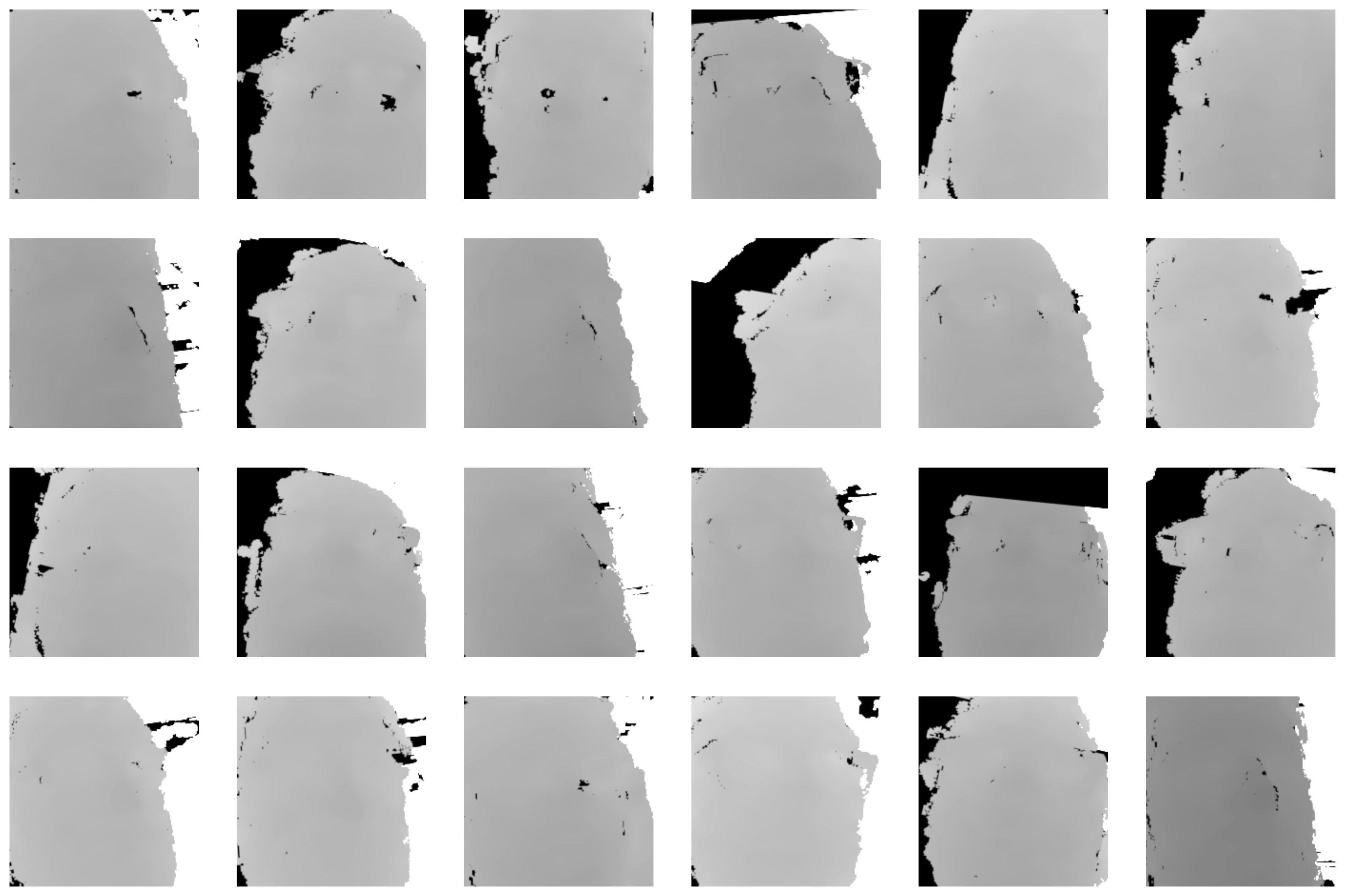}%
        \caption{Example depth maps removed from the input set: $\upsilon \setminus \tau$.}%
        \label{f:processing:dataset:preprocessing:ogesets:removed}%
    \end{subfigure}%
    \hfill%
    \begin{subfigure}{0.49\textwidth}%
        \includegraphics[width=\textwidth]{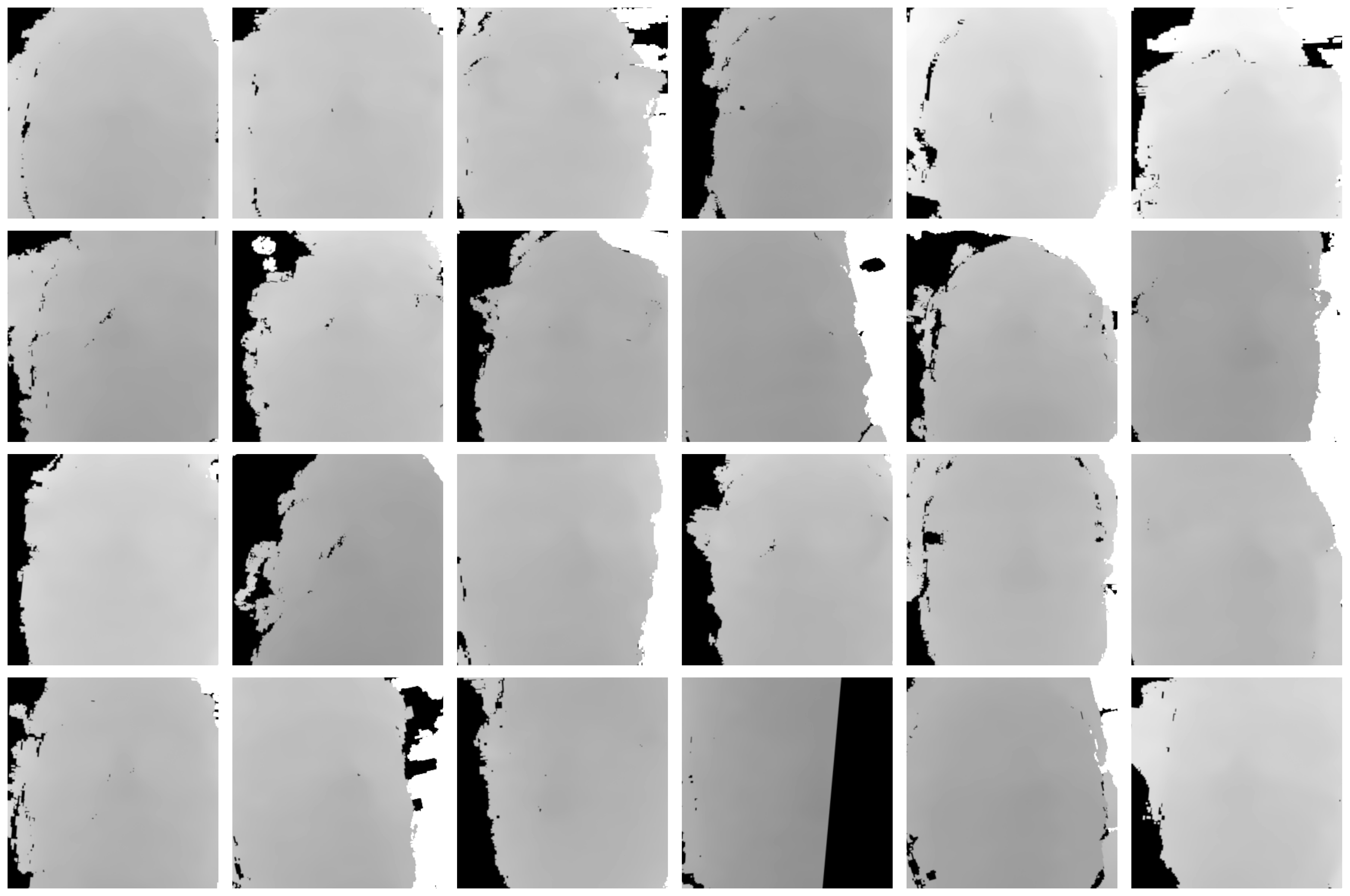}%
        \caption{Example depth maps from the target set $\tau$.}%
        \label{f:processing:dataset:preprocessing:ogesets:discriminator}%
    \end{subfigure}%
    \caption{Example depth maps from the input and target sets of the \ac{GAN} training on the \oge\ dataset.}%
    \label{f:processing:dataset:preprocessing:ogesets}%
\end{figure}

In order to achieve comparable results with our dataset, we opted for the same (adjusted) region size and threshold as for the \stg\ dataset: we defined the region size for the \oge\ dataset to be $81 \times 81$ and the threshold to be \SI{200}{\milli\meter}. We then removed all images from our dataset where the minimal depth value of at least one of the two eye regions was below the threshold. The resulting dataset $\tau$ contains 80,290 samples, i.e., \SI{60.6}{\percent} of the original dataset $\mathcal{D}$. To illustrate the difference between the two sets, random samples are shown in \Cref{f:processing:dataset:preprocessing:ogesets}. Because the majority of samples in the input set $\upsilon$ are also present in $\tau$, \Cref{f:processing:dataset:preprocessing:ogesets:removed} shows samples from the set difference $\upsilon \setminus \tau$ instead of $\upsilon$. However, the samples in \Cref{f:processing:dataset:preprocessing:ogesets:discriminator} are from the target set $\tau$.

\clearpage 

Analogous to the \stg\ dataset, we also compare $\upsilon$ and $\tau$ in terms of depth value distribution and visualize it in \Cref{f:appendix:oge-depth-distribution}. Due to the better quality of the depth maps in our dataset, the distributions of $\upsilon$ and $\tau$ are very similar not only when considering nonzero pixels but also when considering all pixels. Furthermore, the histogram-equalized depth maps of $\upsilon$ and $\tau$ have a similar distribution when considering all pixels and almost match when considering nonzero pixels. This indicates that our approach of removing samples with missing depth data values is both applicable and reasonable for our dataset, too.

  \makeatletter
\let\@savedchapterhead\@makechapterhead
\def\@makechapterhead{\vspace*{-4mm}\@savedchapterhead}
\chapter[Evaluation]{\vspace*{-2mm}Evaluation}
\label{c:evaluation}
\let\@makechapterhead\@savedchapterhead
\makeatother

\vspace*{-5mm} 
{\spaceskip=3.4pt plus 1pt minus 1.5pt 
We conduct a series of experiments to evaluate the performance of our proposed \ges\ model. This chapter is structured as follows: in \Cref{s:evaluation:xgaze}, we evaluate our model on the \xgaze\ dataset and compare it to the baseline results of the original work by \citeauthor*{Zhang2020}~\cite{Zhang2020}. Next, we evaluate our model on the \stg\ dataset and compare the full model approach to one without prior \ac{GAN} training in \Cref{s:evaluation:stg}. In \Cref{s:evaluation:oge}, we evaluate our model on our own \oge\ dataset and compare our findings to the ones on the other two datasets. Then, we optimize some hyperparameters of our model in \Cref{s:evaluation:hyperparams} and conduct an ablation study in \Cref{s:evaluation:ablation}. Finally, we present our real-time gaze point estimation pipeline using previously trained models in \Cref{s:evaluation:pipeline}. \Cref{t:appendix:default-config} shows the default configuration of our model. We use this configuration for all experiments unless stated otherwise.

Depending on the dataset, our model predicts either gaze angles $\hat{g}' \in \mathbb{R}^2$ in the normalized camera space or gaze points $\hat{p} \in \mathbb{R}^2$ in the screen coordinate system. Based on this, we can calculate angular and/or Euclidean errors. \Cref{t:evaluation:metrics} shows which metrics can be used to evaluate \dl\ models on the three datasets. While the \xgaze\ dataset provides information about the on-screen gaze points, there is no easy conversion between them and the gaze angles due to the different camera poses. Therefore, we focus on the angular errors. On the other hand, the \stg\ dataset does not provide information about the gaze angles and therefore only allows for the calculation of Euclidean errors in the screen coordinate system. Our \oge\ dataset provides information to calculate both metrics. In practice, we always predict the gaze angles in the normalized camera space and then convert them to on-screen gaze points. This allows us to calculate both metrics and compare the results.
}

\vspace*{-2mm}
\begin{table}[htbp]
    \centering
    \captionsetup{width=\textwidth}
    \begin{tabular}{@{}r|*{3}{@{\hspace{1.55mm}}c}@{\hspace{1.55mm}}|m{0.44\textwidth}@{}}
        \textbf{Dataset} & \textbf{AE}\,\textsuperscript{a} & \textbf{EE}\,\textsuperscript{b} & \textbf{Reference} & \textbf{Remarks} \\
        \hline\hline
        \xgaze & \cmark & \xmark & \SI{4.2}{\degree}~\cite{Zhang2020} & No screen coordinate system available \par $\rightarrow$ no distance metric \\\hline
        \stg & \xmark & \cmark & \parbox{23.5mm}{\SI{38.7}{\milli\meter}~\cite{Lian2019}\\\SI{32.3}{\milli\meter}~\cite{Zhang2020a}} & No 3D gaze angles available\par $\rightarrow$ no angular metric \\\hline
        \oge & \cmark & \cmark & -- & Both metrics available;\par Model trained only on 3D gaze angles \\\hline
        \multicolumn{5}{p{0.95\linewidth}}{\small
            \textsuperscript{a} AE: Angular Error (in \unit{\degree}) in the normalized camera coordinate system \par
            \textsuperscript{b} EE: Euclidean Error (in \unit{\milli\meter}) in the screen coordinate system
        }
    \end{tabular}
    \vspace*{-1mm}
    \caption{Overview of the metrics used for model evaluation on different datasets.}
    \label{t:evaluation:metrics}
    \vspace*{-5mm}
\end{table}

\section{Experiments on the \xgaze\ Dataset}
\label{s:evaluation:xgaze}

In order to accommodate the specifics of the \xgaze\ dataset, we have to modify our model architecture. We also describe the dataset preprocessing steps and two methods of retrieving the facial landmarks. Then we run different experiments to show the difference between various model architecture configurations and compare our results to the baseline results of the original work by \citeauthor*{Zhang2020}~\cite{Zhang2020}.

\subsection{Model Architecture Changes and Dataset Preprocessing}
\label{s:evaluation:xgaze:preprocessing}

As mentioned in \Cref{s:relwork:datasets:xgaze,s:relwork:datasets:summary}, the \xgaze\ dataset does not include any depth maps. Consequently, we cannot utilize the \ac{GAN} training approach and have to adjust our model architecture to allow for an RGB-only mode of operation. Removing the generator encoder for the depth input, the generator decoder, and the depth extractor modules accomplishes this. In visual terms, we remove the lower third of the model shown in \Cref{f:processing:models:overview}: the \textcolor{blue!80!black}{blue}, \textcolor{purple!80!black}{purple}, and \textcolor{magenta!80!black}{magenta} colored modules. As described in \Cref{s:processing:models:gan}, we then adjust the fusion block to take a single feature map as input. Similarly, the Fusion Transformer module now takes four instead of five tokens as input. The remaining modules are kept unchanged.

To train and evaluate our RGBDTr model, the facial landmarks need to be detected. Although they are not needed as direct input for the model as we removed the depth extractor module, they are still required for the preprocessing step of cropping the eye patches from the original input image. The \xgaze\ dataset provides no processed information about the facial landmarks, i.e., no face-normalized landmark coordinates. However, we found that the authors do provide the raw landmark coordinates in the original images alongside their calculated head rotation and head translation vectors. Therefore, we can convert the raw landmark coordinates to face-normalized landmark coordinates similar to the face normalization preprocessing step in our pipeline (see~\Cref{s:processing:sequence}). Since dlib predicts 68 landmarks, we take the mean of the respective eye corner landmarks to obtain the center eye positions.

Because of the extreme viewing angles, the landmarks provided by the authors are not always correct. We found that our conversion algorithm failed on 11 samples where it calculated clearly incorrect landmark coordinates. These are visualized in \Cref{f:appendix:xgaze-landmarks:dlib-failed} where we show each defective sample and its neighboring samples from the same camera perspective. We noticed that there are two types of errors: first the landmark coordinates are erroneous but the face normalization performed by \citeauthor*{Zhang2020} succeeded, and second both the landmark coordinates and the face normalization failed. Interestingly the former case applies to samples from the test sets and the latter case to samples from the training set. As we cannot repair the failed face normalization process, we decided to discard the six samples from the training set. However, we provided manual eye positions for the five samples from both test sets to ensure that our model can be evaluated on the full test sets.

\clearpage 

\begin{figure}[htb]
    \centering
    \captionsetup{width=\textwidth}
    \begin{tikzpicture}
        \node (img) at (0, 0) {\includegraphics[width=0.934\textwidth]{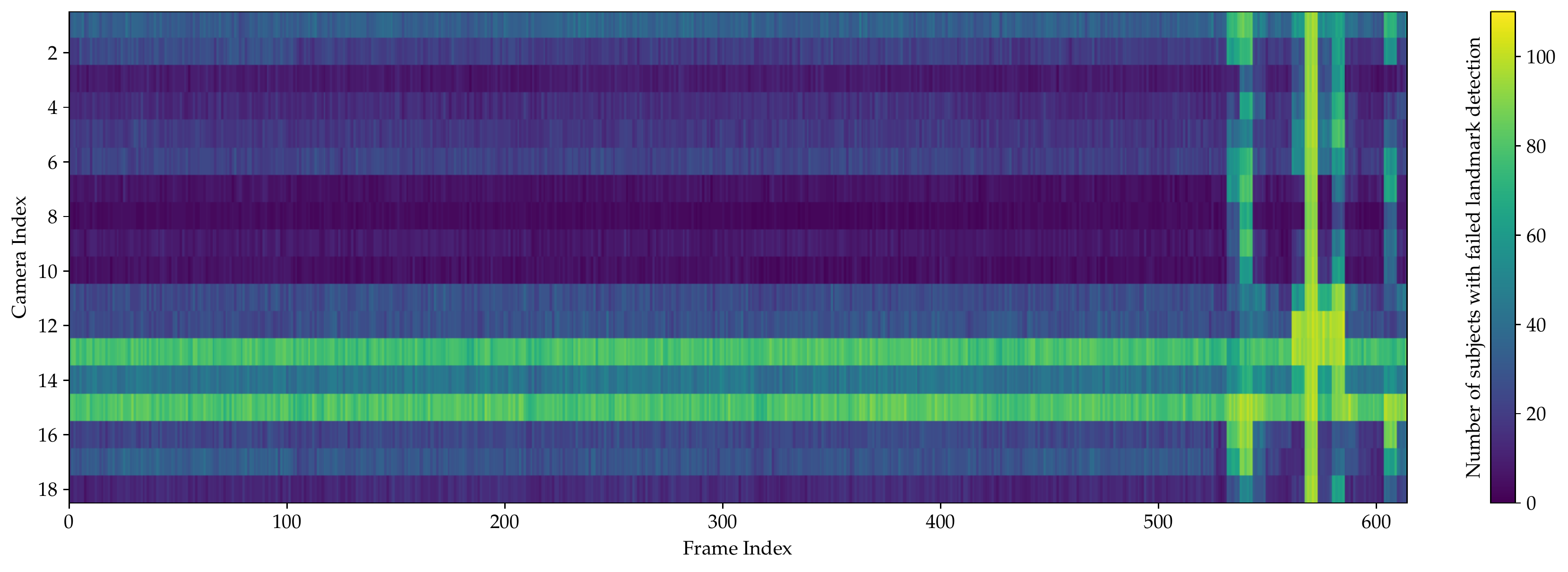}};
        \coordinate (plotne) at ($(img.north east)!0.22!(img.north) + (0, -2.5mm)$);
        \coordinate (plotnw) at ($(img.north west)!0.11!(img.north) + (0, -2.5mm)$);
        \coordinate (plotse) at ($(img.south east)!0.22!(img.south) + (0, 8.5mm)$);
        \coordinate (plotsw) at ($(img.south west)!0.11!(img.south) + (0, 8.5mm)$);

        \coordinate (difflight1) at ($(plotnw)!540/615!(plotne)$);
        \coordinate (difflight2) at ($(plotnw)!570/615!(plotne)$);
        \coordinate (difflight3) at ($(plotnw)!605/615!(plotne)$);

        \coordinate (diffcam13) at ($(plotnw)!12.5/18!(plotsw)$);
        \coordinate (diffcam15) at ($(plotnw)!14.5/18!(plotsw)$);

        \node [above=0.75cm of plotnw, anchor=north west, xshift=0.3cm] (diffcam) {\small Difficult camera positions};
        \node [above=0.75cm of plotne, anchor=north east, xshift=-2cm] (difflight) {\small Difficult lighting conditions};

        \foreach \t in {diffcam13, diffcam15} {
            \draw [arrow] ($(diffcam.west) + (0, -1mm)$) to [in=180, out=210] (\t);
        }
        \foreach \t in {difflight1, difflight2, difflight3} {
            \draw [arrow] ($(difflight.east) + (0, 1mm)$) to [in=90, out=10] (\t);
        }
    \end{tikzpicture}
    \caption{Histogram of the failed landmark detections with yolov7-face on the \xgaze\ dataset.}
    \label{f:evaluation:xgaze:yolo-failed}
\end{figure}
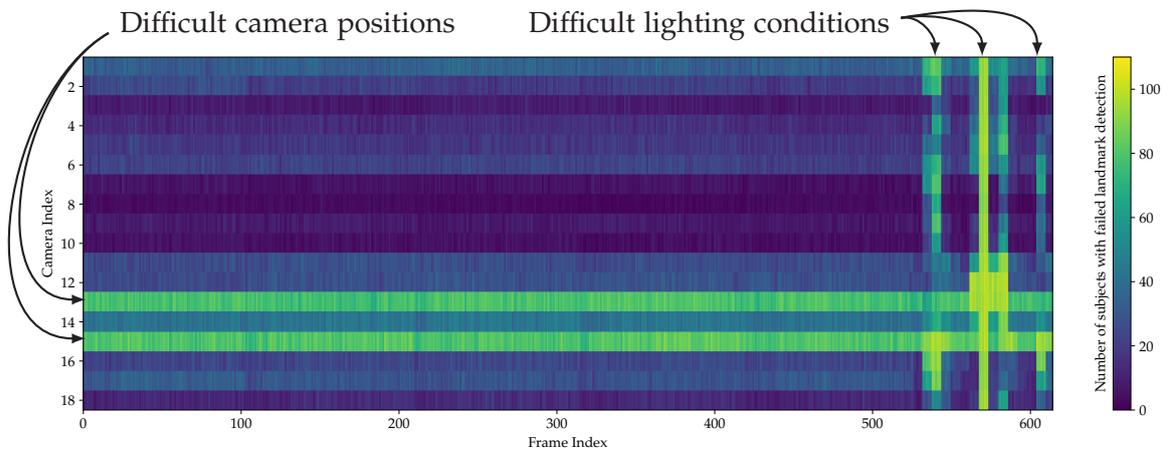

We also tried to apply the yolov7-face model to the \xgaze\ dataset to obtain facial landmarks. This, however, resulted in a much higher failure rate of about \SI{28.3}{\percent}. In total, the landmark detection model failed in 306,898 cases, which we visualize as a 2D-histogram in \Cref{f:evaluation:xgaze:yolo-failed}. We found that the yolov7-face model fails on images with difficult viewing angles and lighting conditions. \Cref{f:evaluation:xgaze:yolo-failed} shows that the failures occur predominantly in two situations: first, camera positions 13 and 15, in which the subject is viewed from a steep angle from below. Second, frame indices around 540, 570 and 605, in which difficult lighting conditions occur. Due to the missing facial landmarks, these images would not be usable for training or evaluation, making the resulting subset of the \xgaze\ dataset unbalanced and therefore unsuitable.

We decided to use the landmarks calculated from the raw dlib landmarks provided by \citeauthor*{Zhang2020} and not use the landmarks obtained with yolov7-face. Furthermore, we employ histogram-equalization on the luma channel of the input images in difficult lighting situations. This is suggested by \citeauthor*{Zhang2020}~\cite{Zhang2020} in their work. In a code sample provided with the dataset, the authors defined all images with a frame index greater than 524 to be modified with luma histogram-equalization. We adopt this approach and apply histogram-equalization on the luma channel on these images.

The authors of the \xgaze\ dataset do not provide ground truth gaze angles for either test set. They only provide an online interface where one can submit predicted gaze angles and then get a score\,\footnote{~\url{https://codalab.lisn.upsaclay.fr/competitions/7423} on CodaLab~\cite{Pavao2022}}. Therefore, we conduct our experiments on the training set, where we employ $k$-fold cross-validation beforehand. Doing so enables us to have at least a validation set to compare against. We use $k = 5$ folds to split the 80 training subjects randomly into five disjoint subsets of 16 subjects each. The splits are the same for all experiments to ensure comparability. We denote the training and validation subsets similarly to our dataset's cross-validation splits as $\mathcal{X}_\mathrm{tr}^{(i)}$ and $\mathcal{X}_\mathrm{val}^{(i)}$ with $i \in \left\{1, \dots, 5\right\}$, respectively.

\clearpage 
\subsection{General Training Process}
\label{s:evaluation:xgaze:training}

\begin{figure}[htb]
    \centering%
    \includegraphics[width=\textwidth]{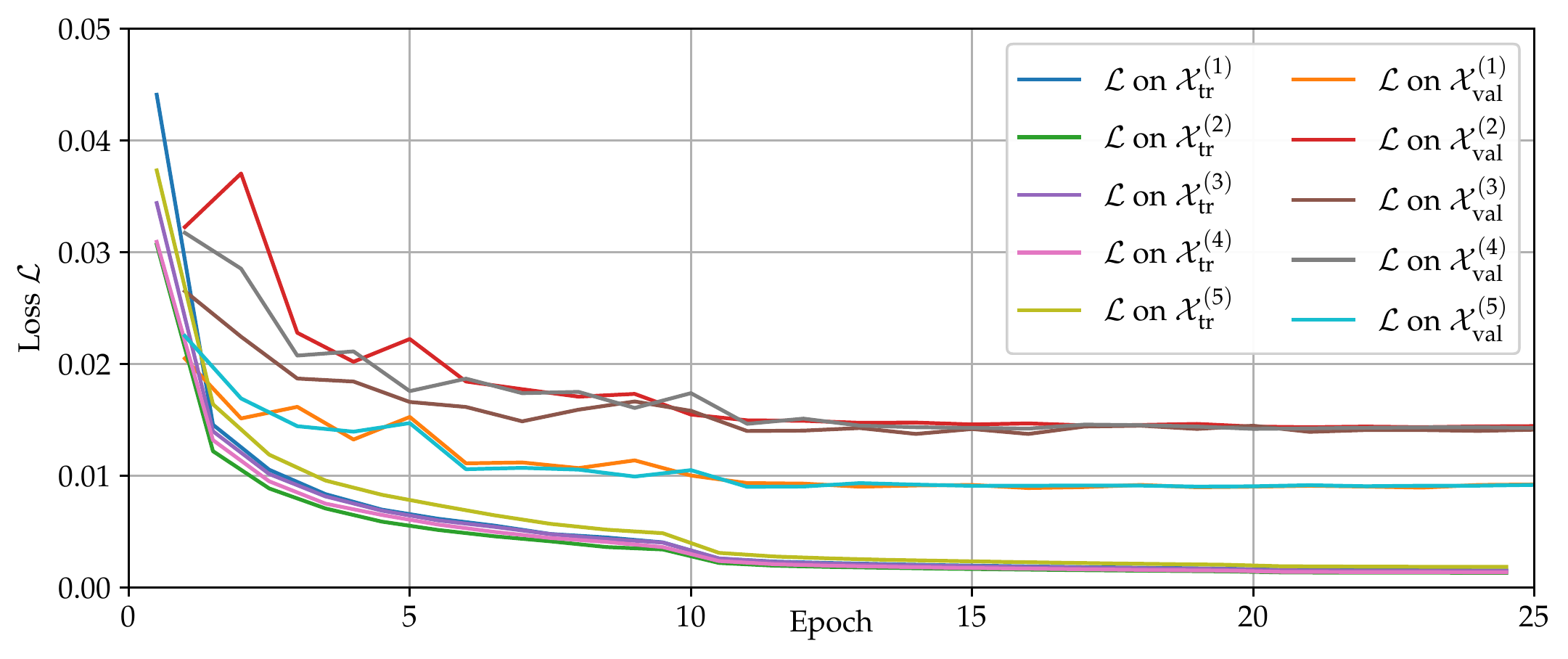}%
    \caption{Typical progression of the loss values during training for each cross-validation split on the \xgaze\ dataset.}%
    \label{f:evaluation:xgaze:losses}%
\end{figure}

In \Cref{f:evaluation:xgaze:losses}, the loss values of a typical training process are depicted. We show the graph for each cross-validation split separately. The training process is very stable after the first few epochs. The graph shows that the loss values do not change much after the 12\textsuperscript{th} epoch on the validation subset. However, the loss on the training subset still declines. This is why we settled for $n_\mathrm{rgbdtr} = 25$ epochs of training. It is important to note that the validation error is calculated without subject calibration, i.e., all subject-specific biases are set to zero. However, the plot still allows us to characterize the training process as the validation error is a clear indicator for the generalization ability of the model on unseen data.

The change in learning rate after the 10\textsuperscript{th} and 20\textsuperscript{th} epoch is clearly visible in the graph, but manifests differently for the training and validation subsets: the loss on the training subsets decreases relatively quickly after the two learning rate changes, while the loss on the validation subsets tends to stabilize and decrease only slightly. Although the loss values on the validation subsets do not change much after the 12\textsuperscript{th} epoch as mentioned above, the increasing stability of these loss values indicates a more robust model. This validates our choice of $n_\mathrm{rgbdtr} = 25$ epochs of training.

It is noticeable that the loss values on the first and last validation subset are significantly lower than on the other three validation subsets even though the loss values on their training subsets are all very similar. This characteristic is visible in all training processes we conducted on the \xgaze\ dataset and is therefore related to the cross-validation splits themselves. However, we did not conduct any further experiments to investigate this phenomenon. In the following subsections, we will aggregate the results of all five cross-validation splits to obtain a more general result.

\clearpage 

\subsection{Weight Decay vs. no Weight Decay}
\label{s:evaluation:xgaze:weight-decay}

\begin{figure}[htb]
    \centering%
    \begin{subfigure}{0.49\textwidth}%
        \includegraphics[width=\textwidth]{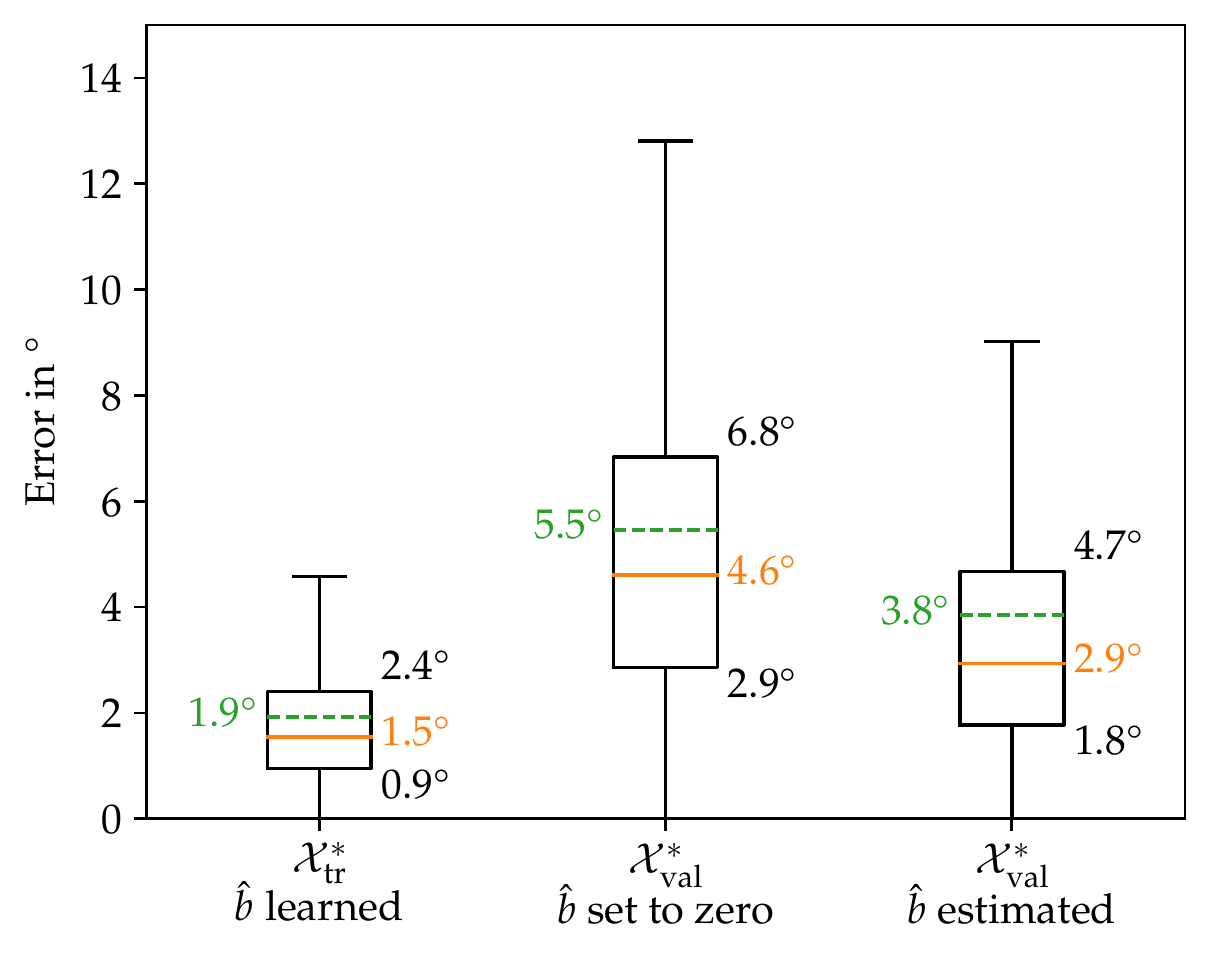}%
        \vspace*{-2mm}%
        \caption{Without weight decay.}%
        \label{f:evaluation:xgaze:wd:without}%
    \end{subfigure}%
    \hfill%
    \begin{subfigure}{0.49\textwidth}%
        \includegraphics[width=\textwidth]{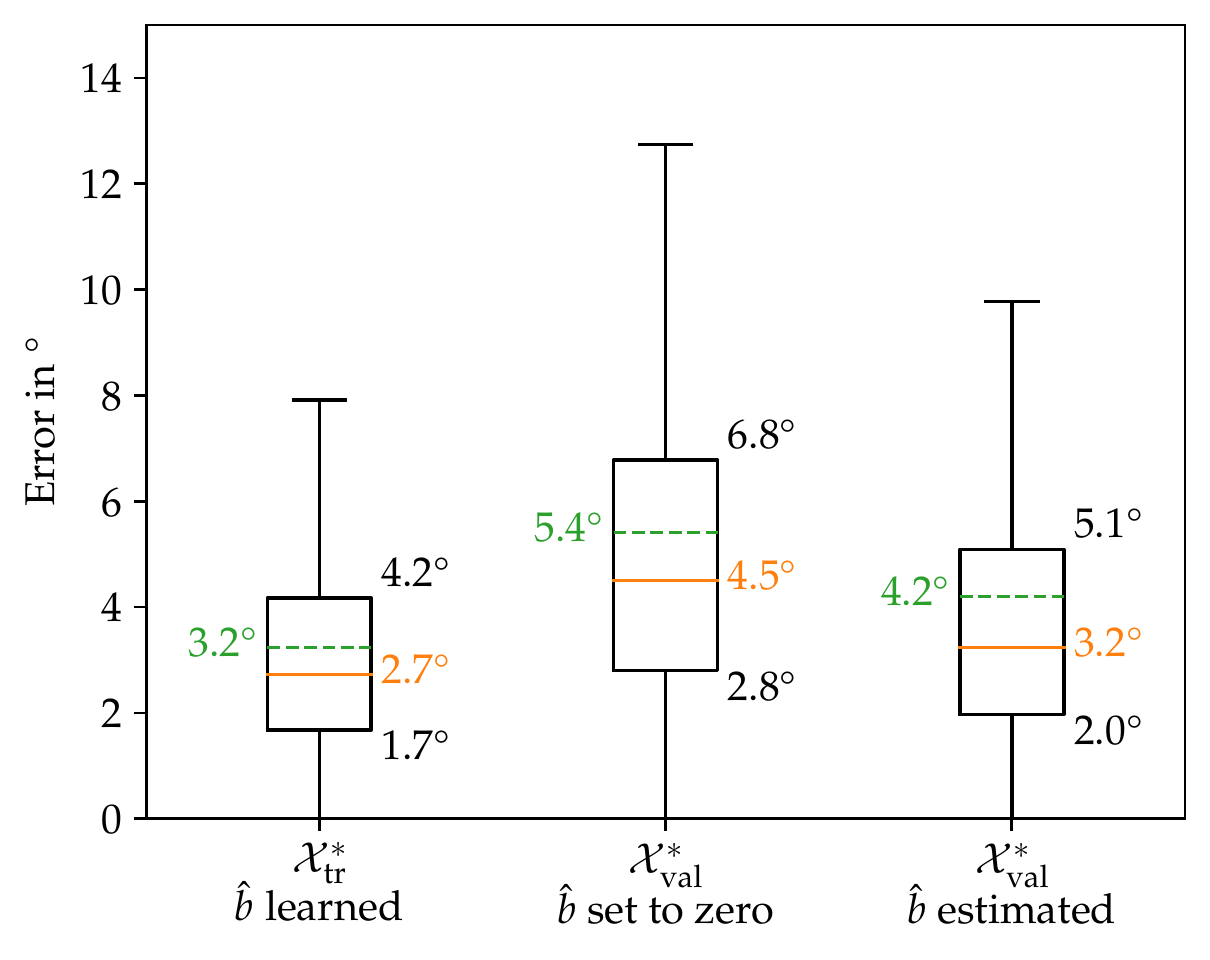}%
        \vspace*{-2mm}%
        \caption{With weight decay of $10^{-3}$.}%
        \label{f:evaluation:xgaze:wd:with}%
    \end{subfigure}%
    \caption{Box-plots of the baseline results of our model in its default configuration on the \xgaze\ dataset trained with and without weight decay.}%
    \label{f:evaluation:xgaze:wd}%
\end{figure}

In order to determine whether a global weight decay affects model performance, we trained two otherwise identical models: one with a weight decay of $10^{-3}$ and one without. The weight decay was applied on all trainable parameters of the model. The results are depicted in \Cref{f:evaluation:xgaze:wd} as box-plots. We aggregated the results for all five cross-validation splits as described above. Each box-plots shows the 25\textsuperscript{th}, the 50\textsuperscript{th} (\textcolor{orange!80!black}{orange}), and the 75\textsuperscript{th} percentiles as well as the mean angular error (\textcolor{green!80!black}{green}), allowing us to compare the angular errors of the models. There are three box-plots per training configuration showing the aggregated results on the training subsets, the validation subsets, and the validation subsets with subject-specific calibration applied. We use $n_\mathrm{cal} = 1000$ samples to estimate $\hat{b}$ for each subject using the method described in \Cref{s:processing:models:transformer}. We denote the aggregated results on the training subsets $\mathcal{X}_\mathrm{tr}^{(1)}, \dots, \mathcal{X}_\mathrm{tr}^{(5)}$ as $\mathcal{X}_\mathrm{tr}^{*}$ and on the validation subsets $\mathcal{X}_\mathrm{val}^{(1)}, \dots, \mathcal{X}_\mathrm{val}^{(5)}$ as $\mathcal{X}_\mathrm{val}^{*}$.

We can observe that the mean angular error on the training subsets~$\mathcal{X}_\mathrm{tr}^{*}$ is significantly higher when applying the weight decay: it increases from \ang{1.9}~(\Cref{f:evaluation:xgaze:wd:without}) to \ang{3.2}~(\Cref{f:evaluation:xgaze:wd:with}). Contrary to that, the mean angular error on the validation subsets~$\mathcal{X}_\mathrm{val}^{*}$ is slightly lower when training with weight decay and applying no subject-specific calibration afterwards: it decreases from \ang{5.5} to \ang{5.4}. However, on the model trained without weight decay, the mean angular error decreases more when applying subject-specific calibration: it decreases from \ang{5.5} to \ang{3.8} in \Cref{f:evaluation:xgaze:wd:without}, i.e., improving by \ang{1.7}, and decreases from \ang{5.4} to \ang{4.2} in \Cref{f:evaluation:xgaze:wd:with}, i.e., improving by \ang{1.2}. This indicates that a model trained without weight decay performs better on both seen and unseen data when subject-specific calibration is applied afterwards. We therefore conduct all further experiments without weight decay.

\clearpage 

To compare our model's performance with the reference in \Cref{s:relwork:datasets:xgaze}, we focus on the mean angular error on the validation subset. \Citeauthor*{Zhang2020} reported a mean angular error of \ang{4.2} on the test set of the \xgaze\ dataset when using input images of size $448 \times 448$ pixels~\cite{Zhang2020}. Our model trained without weight decay and applied without subject-specific calibration achieves \ang{5.5}, which is \ang{1.3} worse than the baseline.

When we apply the subject-specific calibration with $n_\mathrm{cal} = 1000$ randomly chosen samples, our model achieves a mean angular error of \ang{3.8}, which is \ang{0.4} better than the baseline. However, this is still worse than the results reported by \citeauthor*{Zhang2020} on their calibrated model with a mean angular error of \ang{2.0} on the test set for person-specific evaluation~\cite{Zhang2020}.

The comparison between our model and the baseline of \citeauthor*{Zhang2020} is more complicated as the numbers are not directly comparable. We can only evaluate our model on the validation subsets and have to aggregate the results over all five cross-validation splits, while \citeauthor*{Zhang2020} trained their models on the whole training set and evaluated them on the two test sets.

\subsection{Conclusion on the \xgaze\ Dataset}
\label{s:evaluation:xgaze:conclusion}

The experiments on the \xgaze\ dataset serve as a foundation for our future experiments on the other two datasets. Since the focus of this thesis is on RGBD images specifically, we will conduct more experiments on the \stg\ and our \oge\ datasets as they offer depth maps. The training process on the \xgaze\ dataset is very stable and our findings are a good baseline to compare the training processes on the other two datasets to.

In \Cref{s:evaluation:xgaze:weight-decay}, we found that applying weight decay during model training hurts the performance. We therefore conduct all further experiments without weight decay in order to mitigate this issue. Furthermore we already used the \xgaze\ dataset to determine the effectiveness of subject-specific calibration in \Cref{s:processing:models:transformer}.

Although we used the provided dlib landmarks when training on the \xgaze\ dataset, we will use the yolov7-face landmarks for the other two datasets as described above. This is mainly due to runtime performance and the ability to build a real-time gaze point estimation pipeline as stated in \Cref{s:relwork:ml:landmarks}. However, the dlib landmarks seem to be more stable when considering the \xgaze\ dataset. This might open up the possibility for a hybrid approach in the future, where we use the yolov7-face model to detect faces and the dlib model to detect facial landmarks. Since we do not know if the dlib landmarks provided with the \xgaze\ dataset were manually cleansed and/or refined, we cannot make any further assumptions about the performance of the dlib model on other datasets.

\clearpage
\section{Experiments on the \stg\ Dataset}
\label{s:evaluation:stg}

In this section, we present the results of our model on the \stg\ dataset. As this dataset provides depth maps, we can conduct experiments with and without the \ac{GAN} training approach. We also compare our results to the original work by \citeauthor*{Lian2019}~\cite{Lian2019}. First, we describe the dataset preprocessing steps. Then we run different experiments to show the difference between various model architecture configurations.

\subsection{Dataset Preprocessing}
\label{s:evaluation:stg:preprocessing}

The \stg\ dataset is split into two sets by the authors: a training set and a validation set. The authors did not intend to provide a test set. Due to the long training times of multiple days per model, we opted to not split the training set into $k$-fold cross-validation subsets. Instead, we train each model on the training set and evaluate it on the validation set. Our experiments on the \stg\ dataset are primarily to establish a baseline on an publicly available dataset. We do not aim to perform a thorough hyperparameter search on this dataset and therefore having no dedicated test set seems sufficient.

We use our own loss functions for \ac{GAN} training and for the RGBDTr training as described in \Cref{s:processing:models:training}. For the RGBDTr training, our loss function is identical to the one used by \citeauthor*{Lian2019}~\cite[Equation (4)]{Lian2019}. However, training the \ac{GAN} is different as we employ two changes to the process: first, we use a dedicated discriminator training set $\tau$, and second, we apply an erosion kernel to the binary mask used for the L1-loss. Both changes were described in more detail in \Cref{s:processing:models:training}.

\subsection{Discriminator Training}
\label{s:evaluation:stg:discriminator}

Before we can train our RGBDTr model, we need to train the generator for depth reconstruction. The generator architecture was described in \Cref{s:processing:models:gan} and is depicted in \Cref{f:processing:models:generator}. The head pose feature extraction network does not get trained in this first step because it is not needed for depth reconstruction. In order to train the generator network G, we also have to simultaneously train the discriminator D, which general architecture is depicted in \Cref{f:processing:models:discriminator}.

\begin{figure}[htb]
    \centering\captionsetup{width=\textwidth}%
    \includegraphics[width=\textwidth]{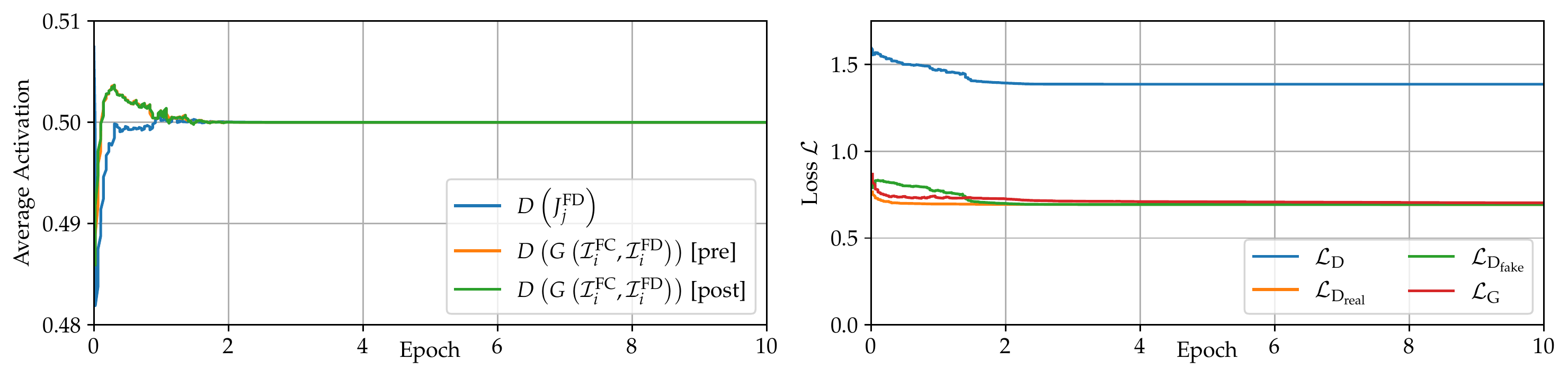}%
    \caption{Progression of the average activation (left) and loss values (right) during training of the generator and discriminator. D has a \bn\ layer before the final \sigm\ activation function, which hurts the performance.}%
    \label{f:evaluation:stg:bn-loss}%
\end{figure}

In our first implementation of D, we used a \bn\ layer before the final \sigm\ activation function. This resulted in a massively degraded performance allowing G to fool D easily. However, this also means that G did not learn the desired mapping function, i.e., the depth reconstruction. The stagnation during training is clearly visible in \Cref{f:evaluation:stg:bn-loss}. After the second epoch, D has lost its ability to distinguish real from fake completely and outputs a steady value of $0.5$ as seen in the left chart. Similarly, the losses of both networks do not change anymore after the collapse, indicating that no further learning is possible. Furthermore, the \bn\ layer before the activation function seems to hinder the discriminator from rejecting images that contain nonzero pixels in the subject's backgrounds. An example of this can be seen in \Cref{f:evaluation:stg:augmentation-removal}.

We also conducted a second experiment without the discriminator target set $\tau$. In this experiment, we used $\tau = \upsilon$, i.e., the full dataset, as images the discriminator has to label as real. The results were interesting: G was able to identify and remove the artificially added rectangular patches of missing data, but did not modify the non-rectangular artifacts. In \Cref{f:evaluation:stg:augmentation-removal}, we visualize four random depth input images with artificial artifacts added and their corresponding output images after applying G. Apparently, G did learn the desired mapping function, but applied it only on the patches created by the data augmentation process. We argue that these finding emphasize the need for a discriminator target set $\tau$ that does not contain artifacts in the eye region in order to incentivize the generator to fill these artifacts, too. As mentioned in the last paragraph, this version of D employs a \bn\ layer before the \sigm\ activation function and also fails to classify nonzero pixels in the subject's backgrounds as fake. All further experiments on the \stg\ dataset use the discriminator target set~$\tau$ as described in \Cref{s:processing:models:training}.

\begin{figure}[!htb]
    \centering
    \input{c4_eval_stg_augmentation_removal.tex}
    \caption{Example depth input and output images of the generator G removing only the artificially added rectangular patches of missing data.}
    \label{f:evaluation:stg:augmentation-removal}
\end{figure}

In our next experiment, we added a fully-connected layer to the output of the convolutional layers of~D. This circumvented the above issue of D outputting a quasi-constant value. The output of this modified architecture is a single value indicating whether D classifies the input image as real or fake. Changing the architecture in this way means that it is no longer a PatchGAN discriminator as it operates on the whole image instead of patches. Therefore, we call it a Full-Image-Discriminator.

\begin{figure}[htb]
    \centering\captionsetup{width=\textwidth}%
    \begin{subfigure}[c]{0.66\textwidth}%
        \begin{subfigure}{\textwidth}%
            \includegraphics[width=\textwidth]{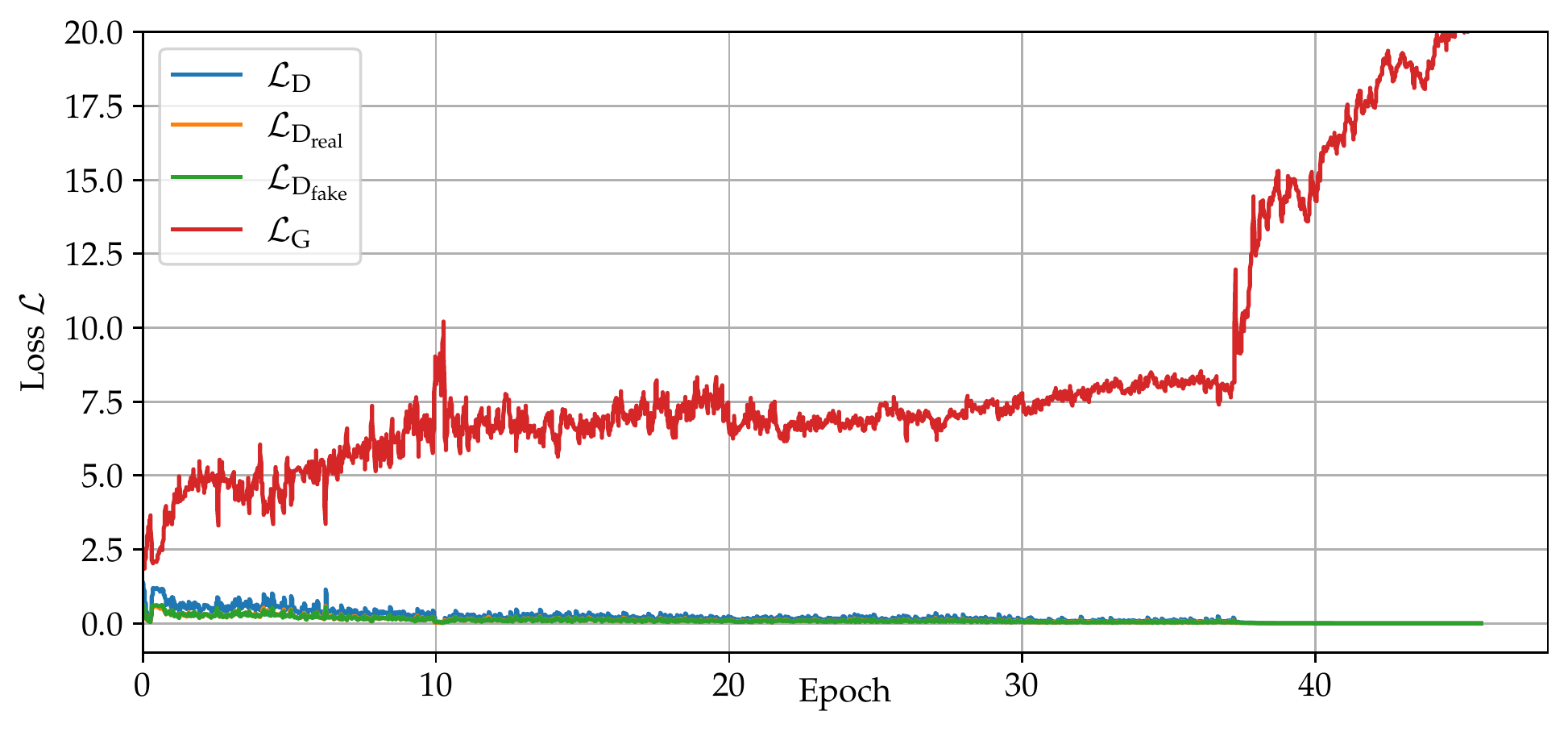}%
            \vspace*{-2mm}%
            \caption{Progression of the average loss values of G and D during training.}%
            \label{f:evaluation:stg:full-image:loss}%
        \end{subfigure}
        \begin{subfigure}{\textwidth}%
            \includegraphics[width=\textwidth]{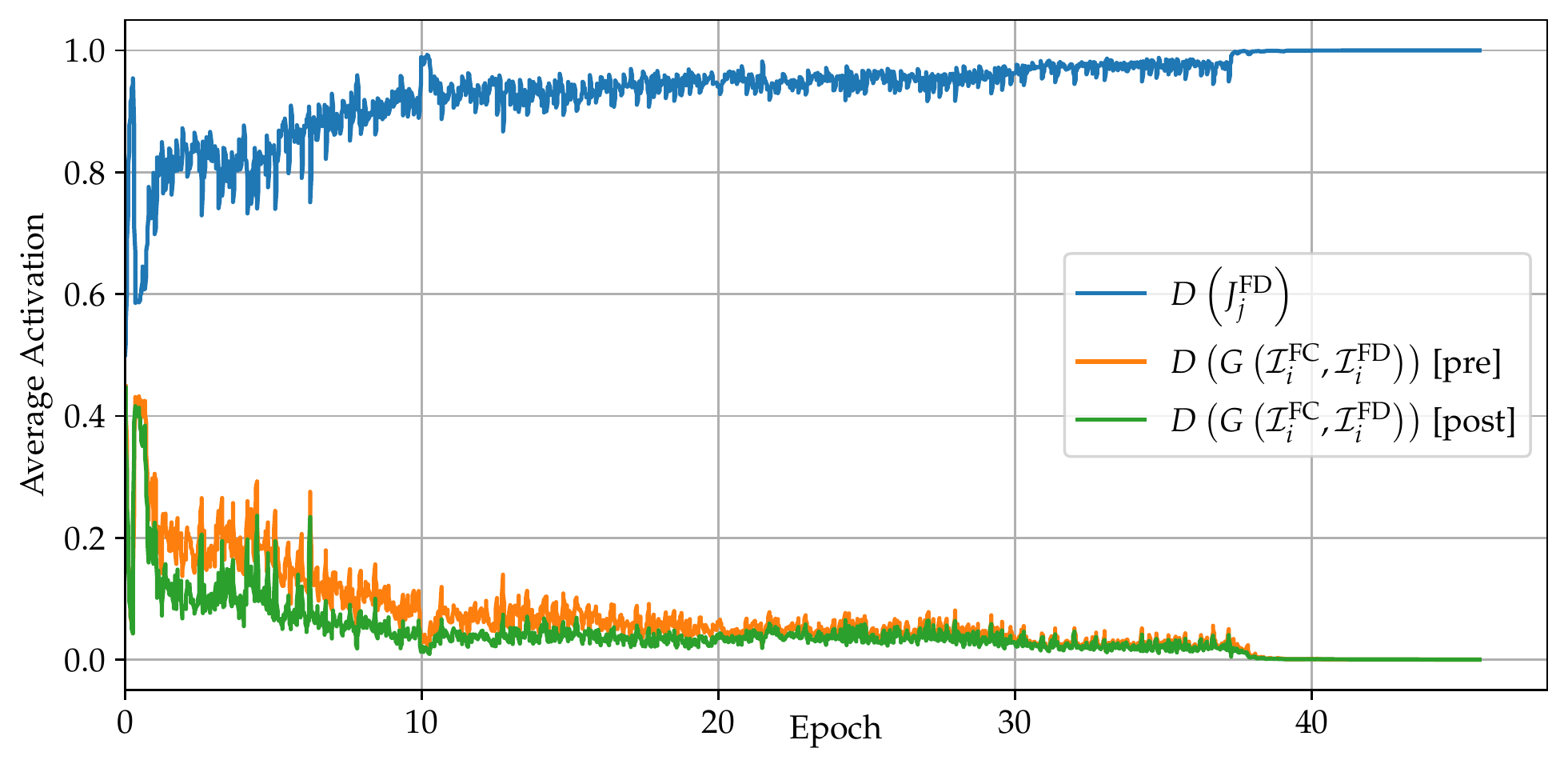}%
            \vspace*{-2mm}%
            \caption{Progression of the average activation values of D during training.}%
            \label{f:evaluation:stg:full-image:activation}%
        \end{subfigure}%
    \end{subfigure}%
    \hfill%
    \begin{subfigure}[c]{0.33\textwidth}%
        \input{c4_eval_stg_mode_collapse.tex}
        \vspace*{-2mm}%
        \caption{G suffers from mode collapse.}%
        \label{f:evaluation:stg:full-image:mode-collapse}%
    \end{subfigure}%
    \caption{Training process of the \ac{GAN} using a Full-Image-Discriminator architecture.}
    \label{f:evaluation:stg:full-image}
\end{figure}

However, this change hugely improved D's ability to distinguish between real and fake images, making it hard for G to fool D. Furthermore, a strong discriminator hurts the training process of the generator, resulting G not learning the desired mapping function or not learning at all. In fact, our experiments show that G suffers from mode collapse when using a Full-Image-Discriminator D. This is visualized in \Cref{f:evaluation:stg:full-image}. The average activation values of D have a steep slope at the beginning of the training as depicted in \Cref{f:evaluation:stg:full-image:activation}. This is a strong indication for a superior discriminator compared to the generator. Although the training process seems to stabilize within the first epochs, the discriminator slowly outperforms the generator's abilities. Eventually, during the 38\textsuperscript{th} epoch, G is unable to learn and suffers from mode collapse, i.e., the network outputs always the (almost) same image regardless of the input images. This hugely increases $\mathcal{L}_\mathrm{G}$, which is visible in \Cref{f:evaluation:stg:full-image:loss}. Three sample images the generator G produces are depicted in \Cref{f:evaluation:stg:full-image:mode-collapse}.

\clearpage 

After the unsuccessful experiment with the Full-Image-Discriminator architecture, we returned to the PatchGAN discriminator architecture and removed the \bn\ layer before the final activation. We tried several configurations differing in the amount of downsampling convolutional layer and in the optional addition of an unpadded convolutional layer with kernel size $3 \times 3$ and a stride of $2$ before the activation function. The addition of this last layer effectively shrinks the output activation map again. The smaller the output activation map is, the bigger are the patches that are classified as real or fake, improving the ability of the discriminator to distinguish correctly. Therefore, we call a discriminator with the addition of this unpadded convolutional layer a BigPatchGAN discriminator.

\begin{figure}[htb]
    \centering\captionsetup{width=0.9\textwidth}%
    \begin{subfigure}{0.49\textwidth}%
        \includegraphics[width=\textwidth]{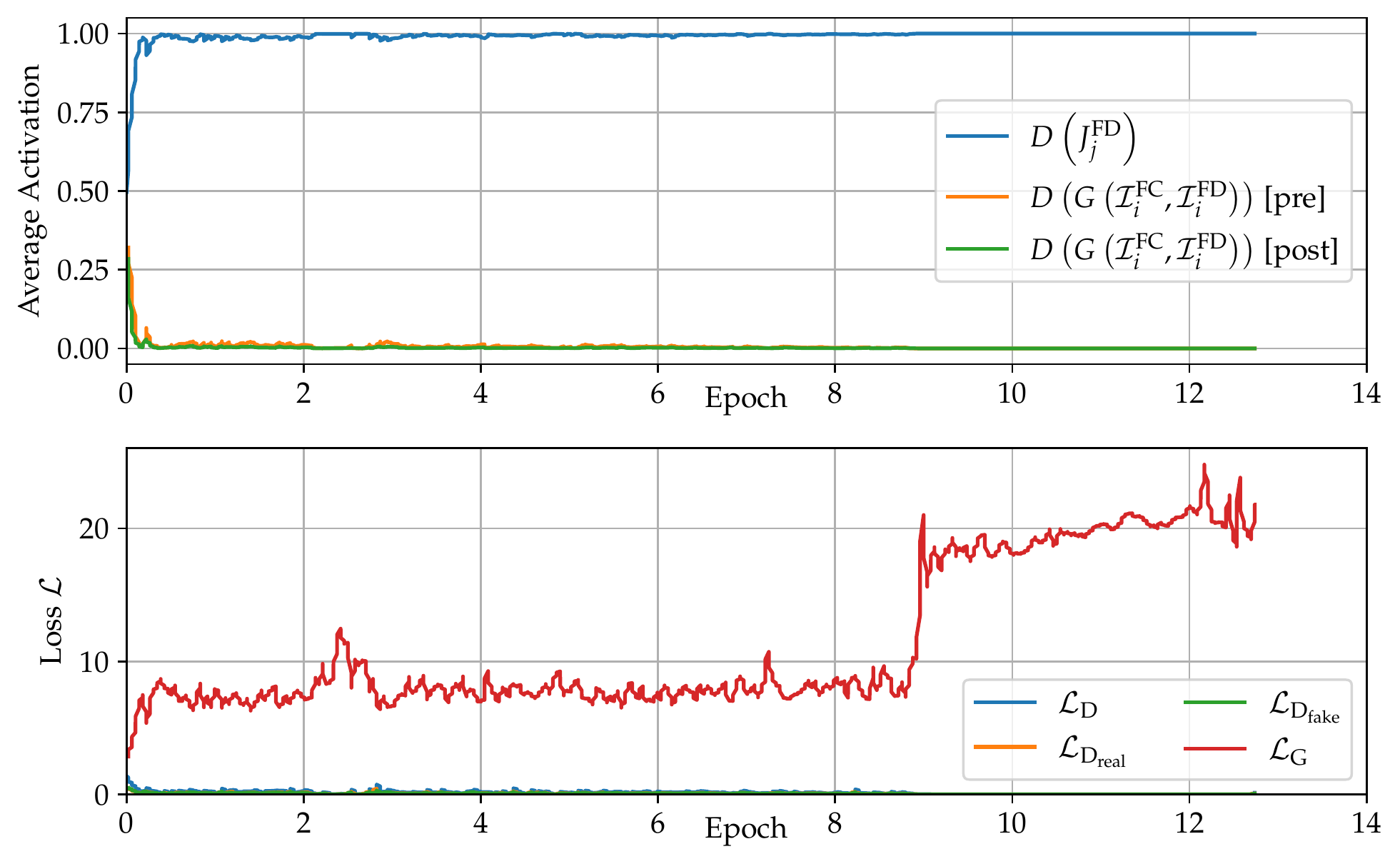}%
        \caption{PatchGAN with 6 layers.}%
        \label{f:evaluation:stg:discriminator:pgan-vs-bpgan:pgan-ndis6}%
    \end{subfigure}%
    \hfill%
    \begin{subfigure}{0.49\textwidth}%
        \includegraphics[width=\textwidth]{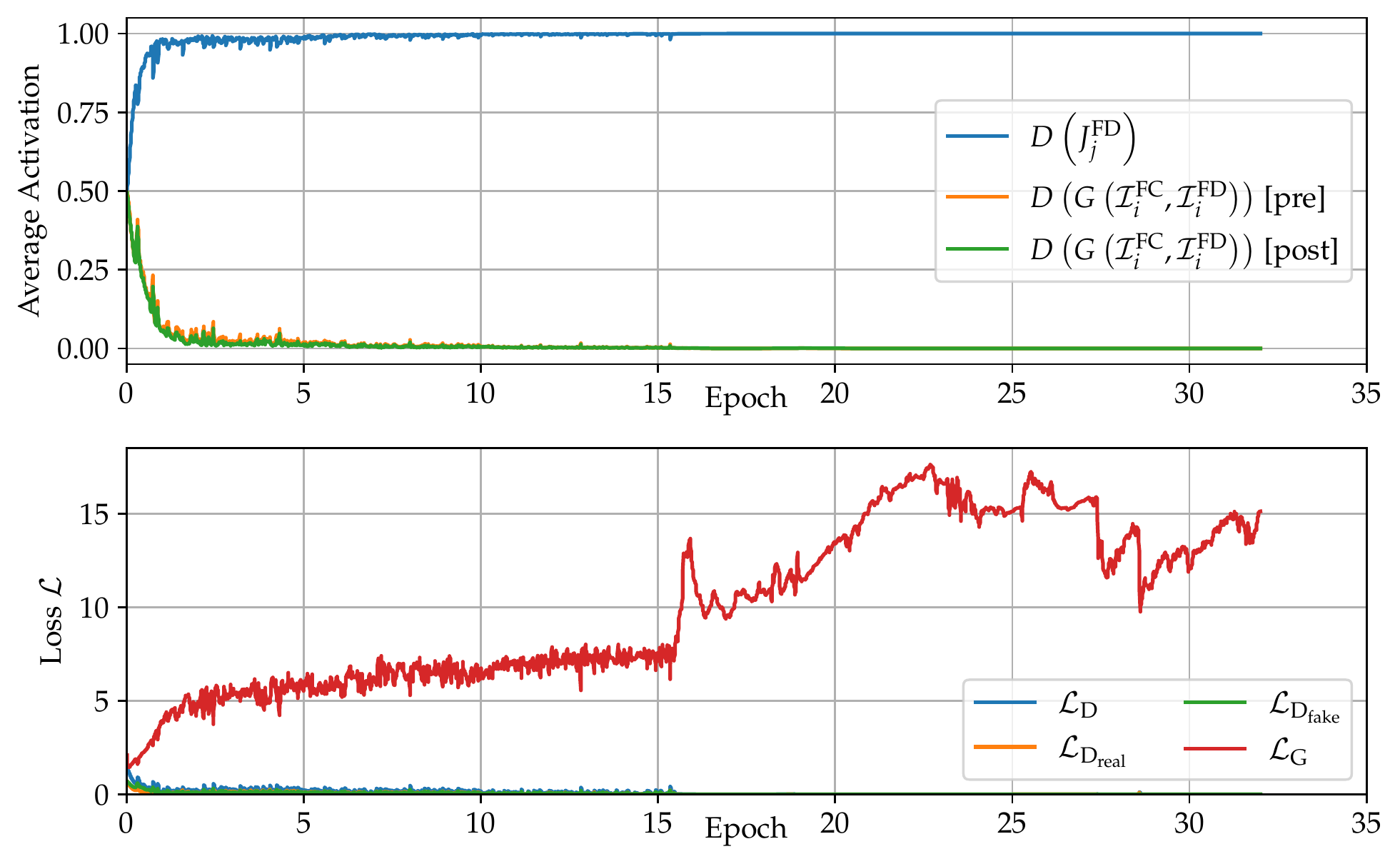}%
        \caption{BigPatchGAN with 4 layers.}%
        \label{f:evaluation:stg:discriminator:pgan-vs-bpgan:bpgan-ndis4}%
    \end{subfigure}%
    \caption{Progression of the average activation and loss values of G and D during training. Both generators suffered from mode collapse.}
    \label{f:evaluation:stg:discriminator:pgan-vs-bpgan}
\end{figure}

\begin{figure}[htb]
    \centering%
    \begin{subfigure}[t]{0.49\textwidth}%
        \includegraphics[width=\textwidth]{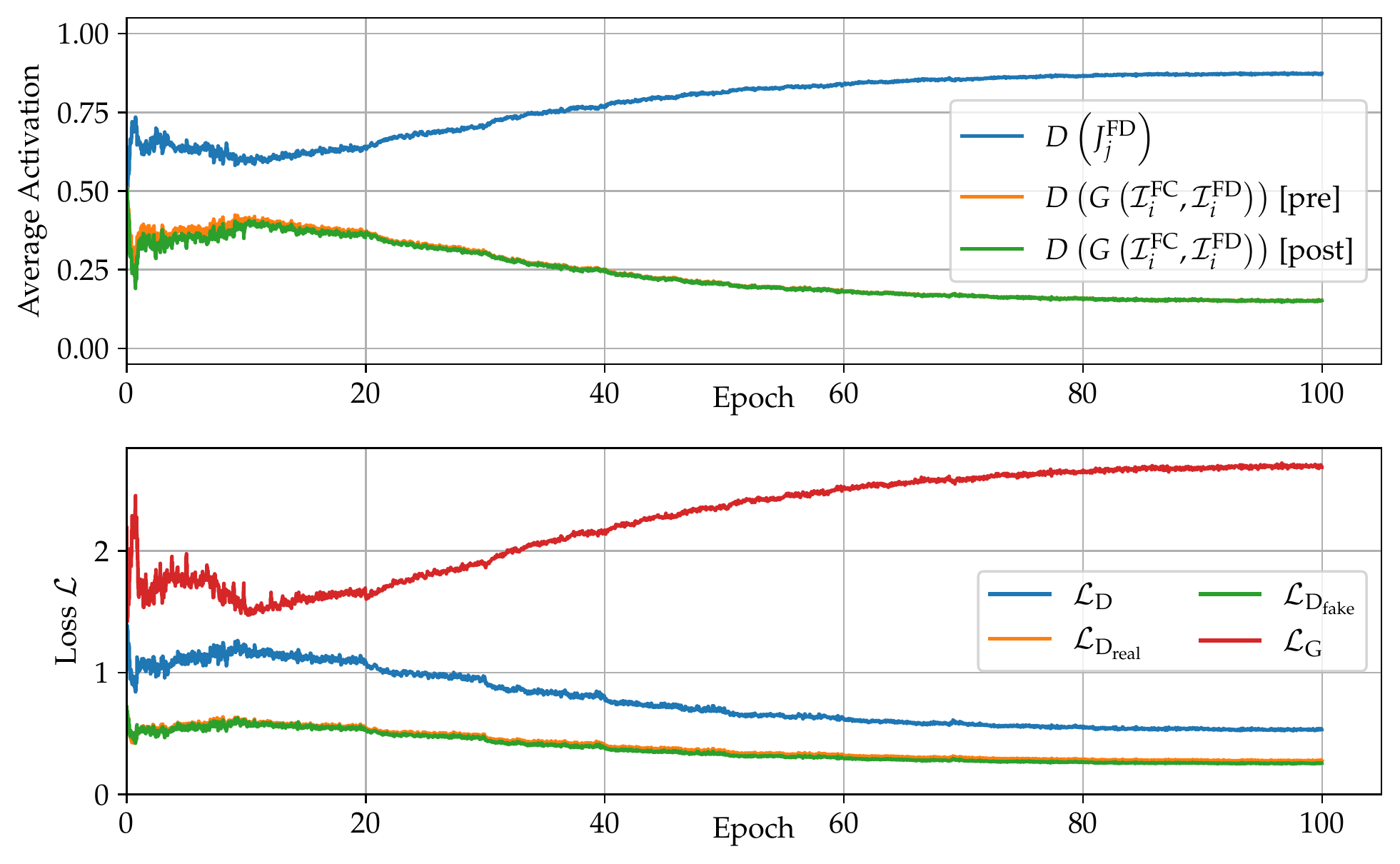}%
        \vspace*{-2mm}%
        \caption{PatchGAN with 4 layers.}%
        \label{f:evaluation:stg:fusion:pgan:ndis4}%
        \vspace*{2mm}%
    \end{subfigure}%
    \hfill%
    \begin{subfigure}[t]{0.49\textwidth}%
        \includegraphics[width=\textwidth]{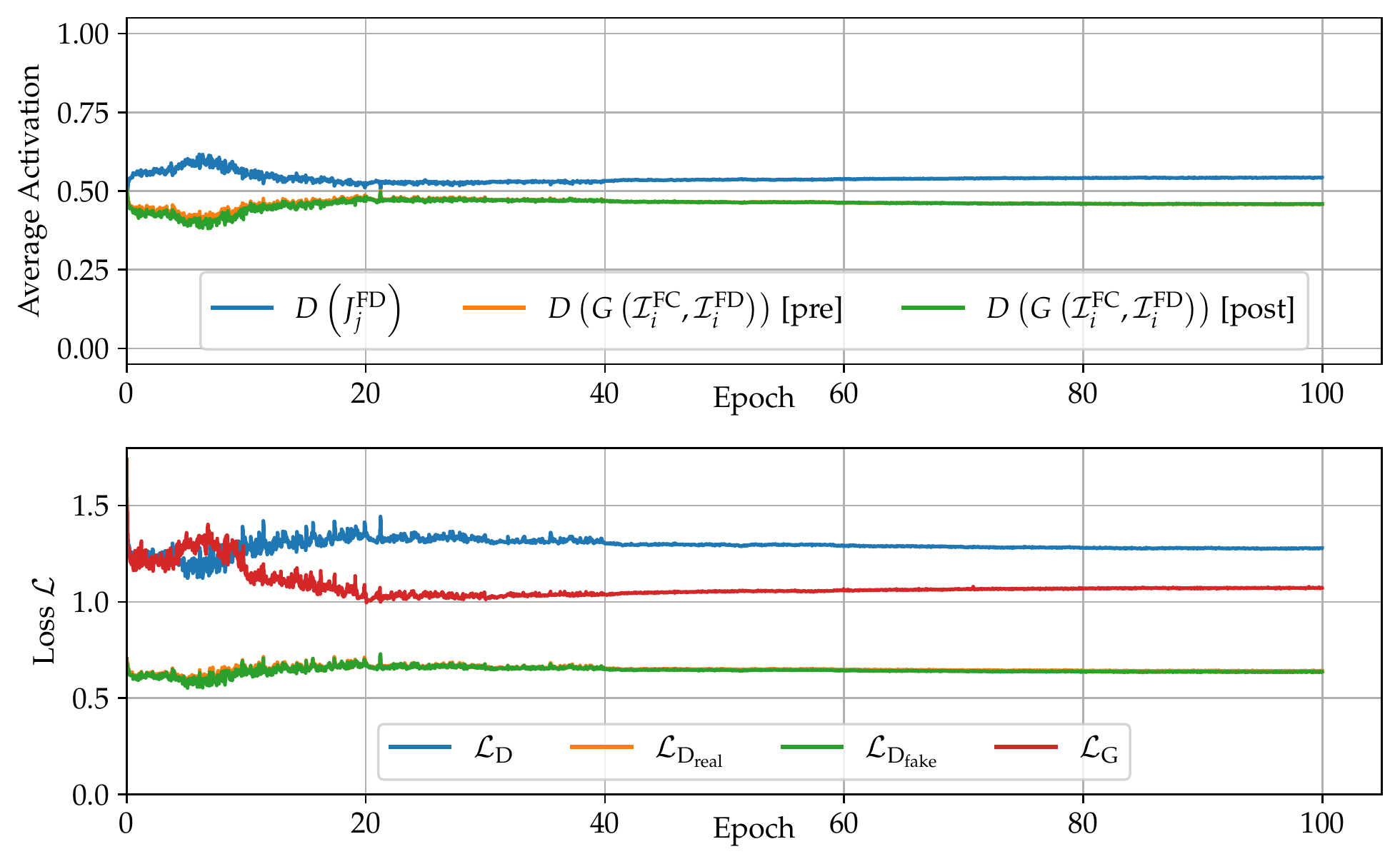}%
        \vspace*{-2mm}%
        \caption{PatchGAN with 3 layers on images of size $224 \times 224$ pixels.}%
        \label{f:evaluation:stg:fusion:pgan:ndis3-i224}%
        \vspace*{2mm}%
    \end{subfigure}

    \begin{subfigure}[t]{0.49\textwidth}%
        \includegraphics[width=\textwidth]{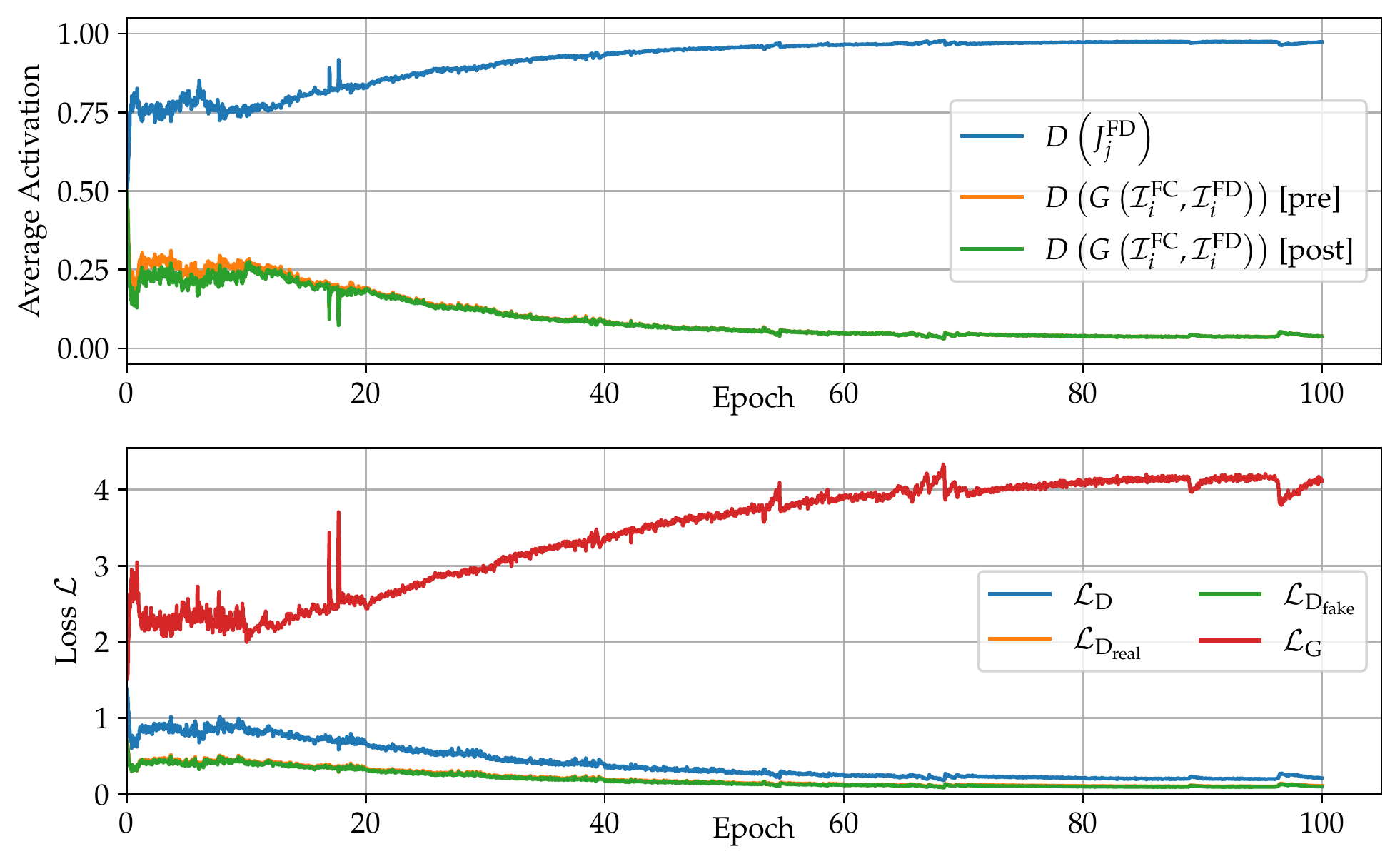}%
        \vspace*{-2mm}%
        \caption{PatchGAN with 4 layers and no fusion block.}%
        \label{f:evaluation:stg:fusion:pgan:ndis4-nofus}%
        \vspace*{2mm}%
    \end{subfigure}%
    \hfill%
    \begin{subfigure}[t]{0.49\textwidth}%
        \includegraphics[width=\textwidth]{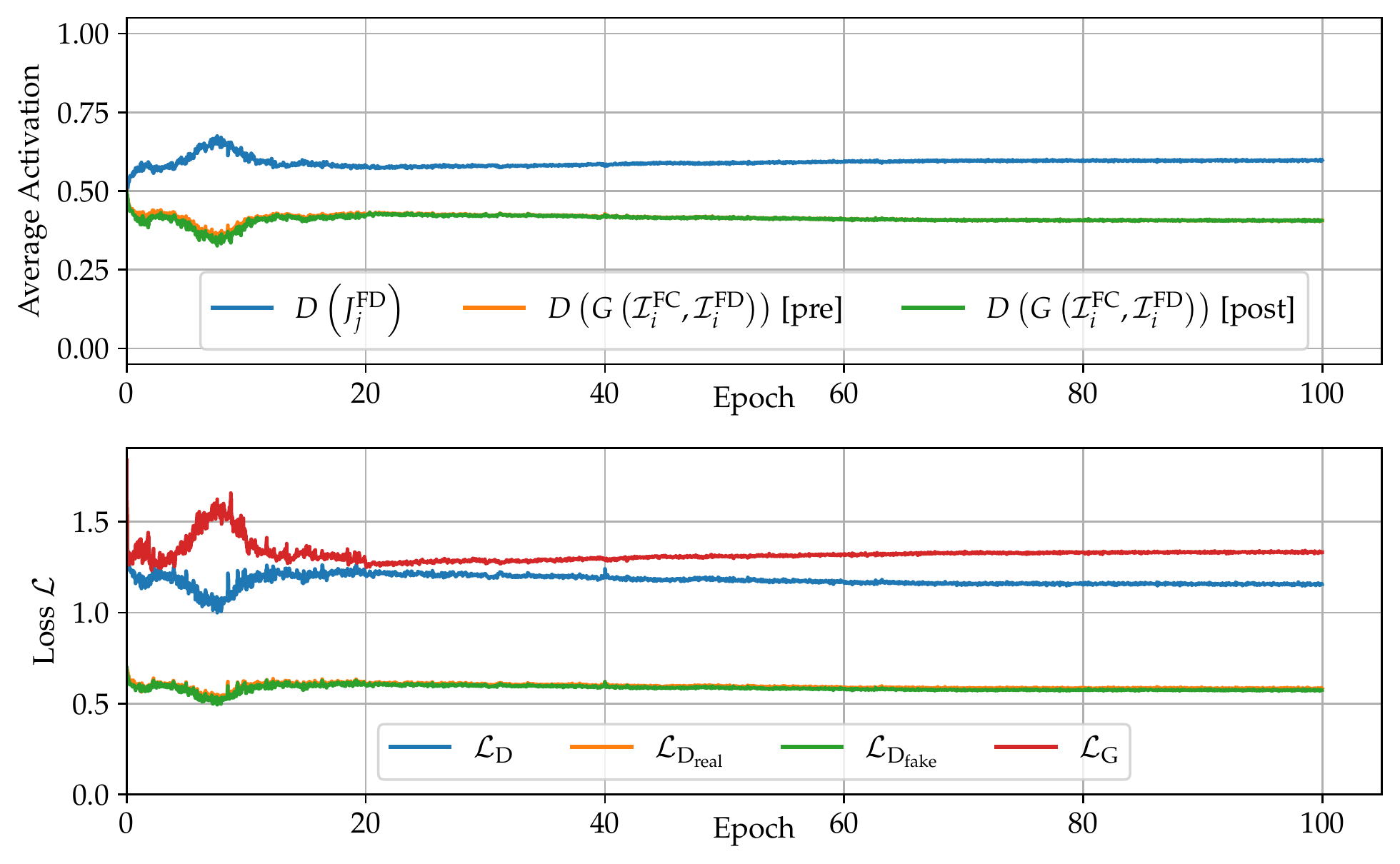}%
        \vspace*{-2mm}%
        \caption{PatchGAN with 3 layers and no fusion block on images of size $224 \times 224$ pixels.}%
        \label{f:evaluation:stg:fusion:pgan:ndis3-i224-nofus}%
        \vspace*{2mm}%
    \end{subfigure}
    \caption{Progression of the average activation and loss values of G and D during a successful training.}
    \label{f:evaluation:stg:fusion:pgan}
\end{figure}

We found that using a six layer PatchGAN discriminator is still too strong for our generator architecture. This is clearly visible on the very steep slope of the average activation values at the beginning of the training process in \Cref{f:evaluation:stg:discriminator:pgan-vs-bpgan:pgan-ndis6}. Similarly a four layer BigPatchGAN also resulted in G to suffer from mode collapse due to a too strong discriminator. This can be seen in \Cref{f:evaluation:stg:discriminator:pgan-vs-bpgan:bpgan-ndis4} where $\mathcal{L}_\mathrm{G}$ suddenly increases during the 16\textsuperscript{th} epoch. However, we found that a four layer PatchGAN works well with input images sized $448 \times 448$ pixels. When using smaller images of size $224 \times 224$ pixels, we used a three layer PatchGAN. A comparison of the training processes of a four layer PatchGAN for normal-sized images and a three layer PatchGAN for smaller images is depicted in \Cref{f:evaluation:stg:fusion:pgan}. Furthermore, the figure compares the training processes for two configurations: with and without a fusion block.

The training process of the four layer PatchGAN is similar regardless of the existence of the fusion block. However, the fusion block seems to stabilize the training process and smoothens the loss curves for both G and D. This is especially visible when comparing the red loss curves $\mathcal{L}_\mathrm{G}$ in \Cref{f:evaluation:stg:fusion:pgan:ndis4,f:evaluation:stg:fusion:pgan:ndis4-nofus}. As the discriminator architecture is identical, D's performance on distinguishing between real and fake images allows us to compare the performance of the two generators. We can see that G is better at fooling D when using a fusion block as the average activation values are closer to $0.5$ in \Cref{f:evaluation:stg:fusion:pgan:ndis4} than in \Cref{f:evaluation:stg:fusion:pgan:ndis4-nofus}. This indicates that the fusion block improves the generator's ability to produce images that are less distinguishable from real images.

Although the training process of the three layer PatchGAN on smaller images is also similar regardless of the existence of the fusion block, the graphs are distinctively different from the four layer PatchGAN training process. The average activation values of the three layer PatchGAN are much closer to $0.5$ in \Cref{f:evaluation:stg:fusion:pgan:ndis3-i224,f:evaluation:stg:fusion:pgan:ndis3-i224-nofus} than the values of the four layer PatchGAN in \Cref{f:evaluation:stg:fusion:pgan:ndis4,f:evaluation:stg:fusion:pgan:ndis4-nofus}. However, in the two training process without a fusion block (\Cref{f:evaluation:stg:fusion:pgan:ndis3-i224-nofus,f:evaluation:stg:fusion:pgan:ndis4-nofus}), D's average activation is more distant to $0.5$ than in the two training processes with a fusion block (\Cref{f:evaluation:stg:fusion:pgan:ndis3-i224,f:evaluation:stg:fusion:pgan:ndis4}). This supports our previous finding that the fusion block improves the generator's ability to fool the discriminator.

{\spaceskip=3.4pt plus 1pt minus 1.5pt 
Interestingly, the $\mathcal{L}_\mathrm{G}$ is less than $\mathcal{L}_\mathrm{D}$ during the training of the three layer PatchGAN with a fusion block on smaller images (\Cref{f:evaluation:stg:fusion:pgan:ndis3-i224}). This is the only configuration out of the four regarded in this section where this is the case. We argue that this is due to the smaller input image size and therefore comparatively larger generator architecture in comparison to the discriminator architecture. This allows G to generate images that resemble the target dataset distribution more closely. However, this is only a hypothesis and we did not conduct any further experiments to investigate this finding due to time constraints. In the next section, we compare the performance of our RGBDTr model that uses these four generator and discriminator configurations as backbone.
}

\clearpage 

\subsection{Generator: Fusion Block vs. no Fusion Block}
\label{s:evaluation:stg:fusion}

In this section, we compare the performance of our RGBDTr model in four configurations. We conduct experiments on the generator backbone with and without a fusion block as well as on two input image sizes: $448 \times 448$ pixels and $224 \times 224$ pixels. This allows us to compare the impact of the fusion block on different input image sizes.

First, we trained four generator and discriminator pairs, one for each combination. In \Cref{f:evaluation:stg:fusion:pgan}, the training process is depicted in terms of average activation and loss values. We used a four layer PatchGAN for the larger input images and a three layer PatchGAN for the smaller input images as these model configurations worked well and did not result in mode collapse. The discriminator training process was already discussed in \Cref{s:evaluation:stg:discriminator}.

\begin{figure}[htb]
    \centering\captionsetup{width=0.95\textwidth}%
    \vspace*{-2mm}%
    \begin{subfigure}{0.49\textwidth}%
        \includegraphics[width=\textwidth]{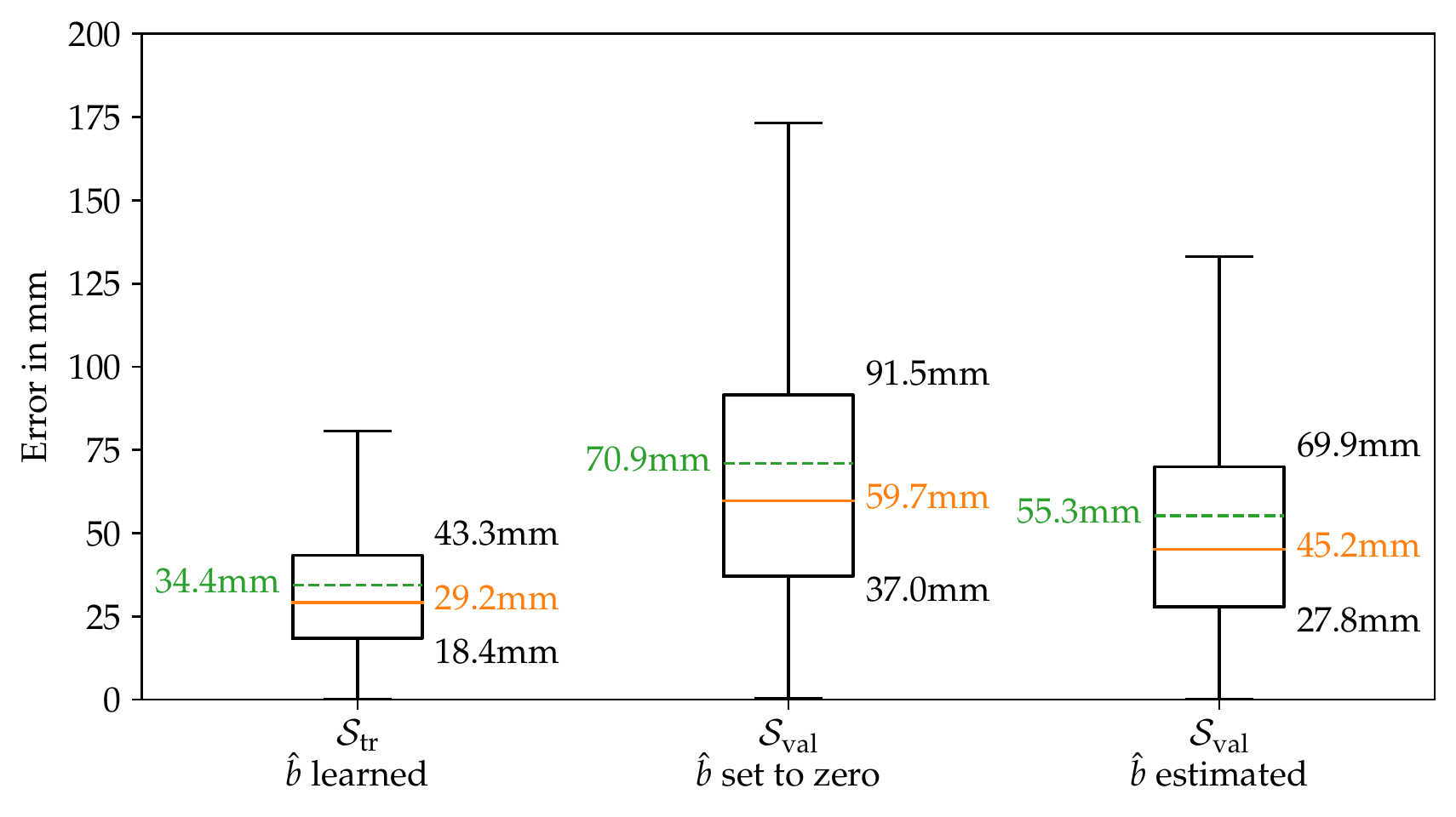}%
        \vspace*{-2mm}%
        \caption{With fusion block on images of size $448 \times 448$ pixels.}%
        \label{f:evaluation:stg:fusion:results:fus-i448}%
        \vspace*{2mm}%
    \end{subfigure}%
    \hfill%
    \begin{subfigure}{0.49\textwidth}%
        \includegraphics[width=\textwidth]{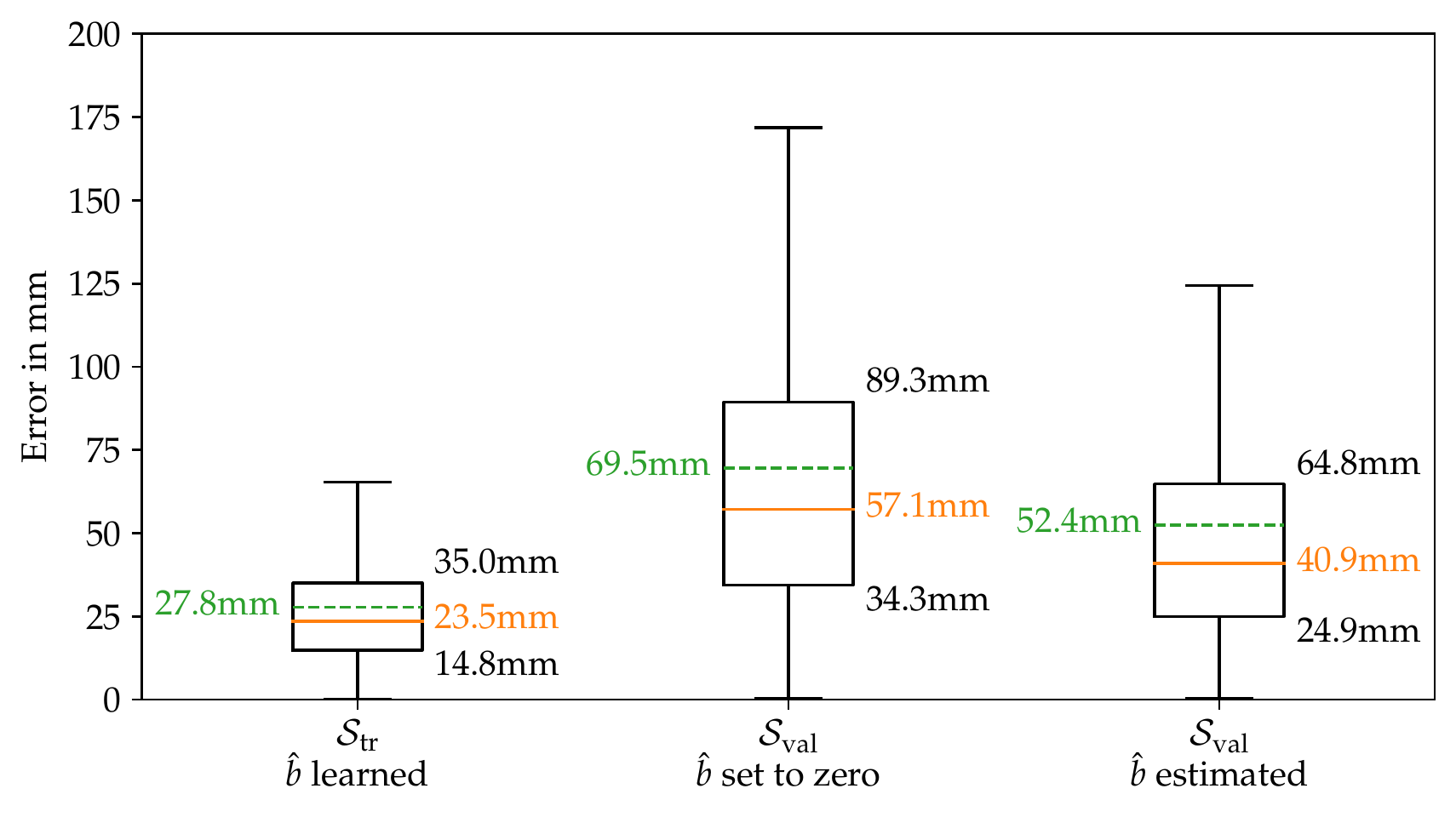}%
        \vspace*{-2mm}%
        \caption{With fusion block on images of size $224 \times 224$ pixels.}%
        \label{f:evaluation:stg:fusion:results:fus-i224}%
        \vspace*{2mm}%
    \end{subfigure}
    \begin{subfigure}{0.49\textwidth}%
        \includegraphics[width=\textwidth]{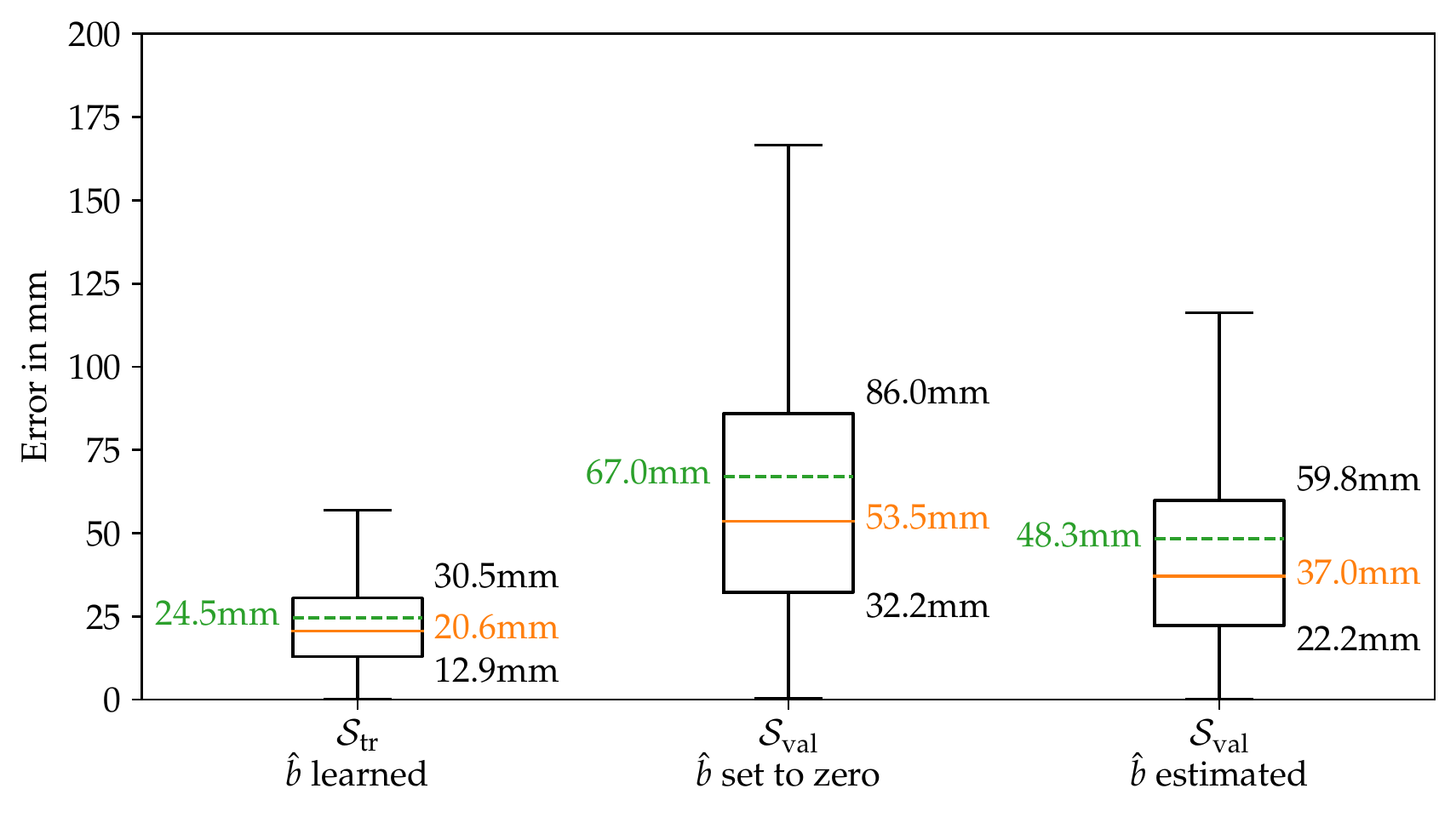}%
        \vspace*{-2mm}%
        \caption{Without fusion block on images of size $448 \times 448$ pixels.}%
        \label{f:evaluation:stg:fusion:results:nofus-i448}%
        \vspace*{2mm}%
    \end{subfigure}%
    \hfill%
    \begin{subfigure}{0.49\textwidth}%
        \includegraphics[width=\textwidth]{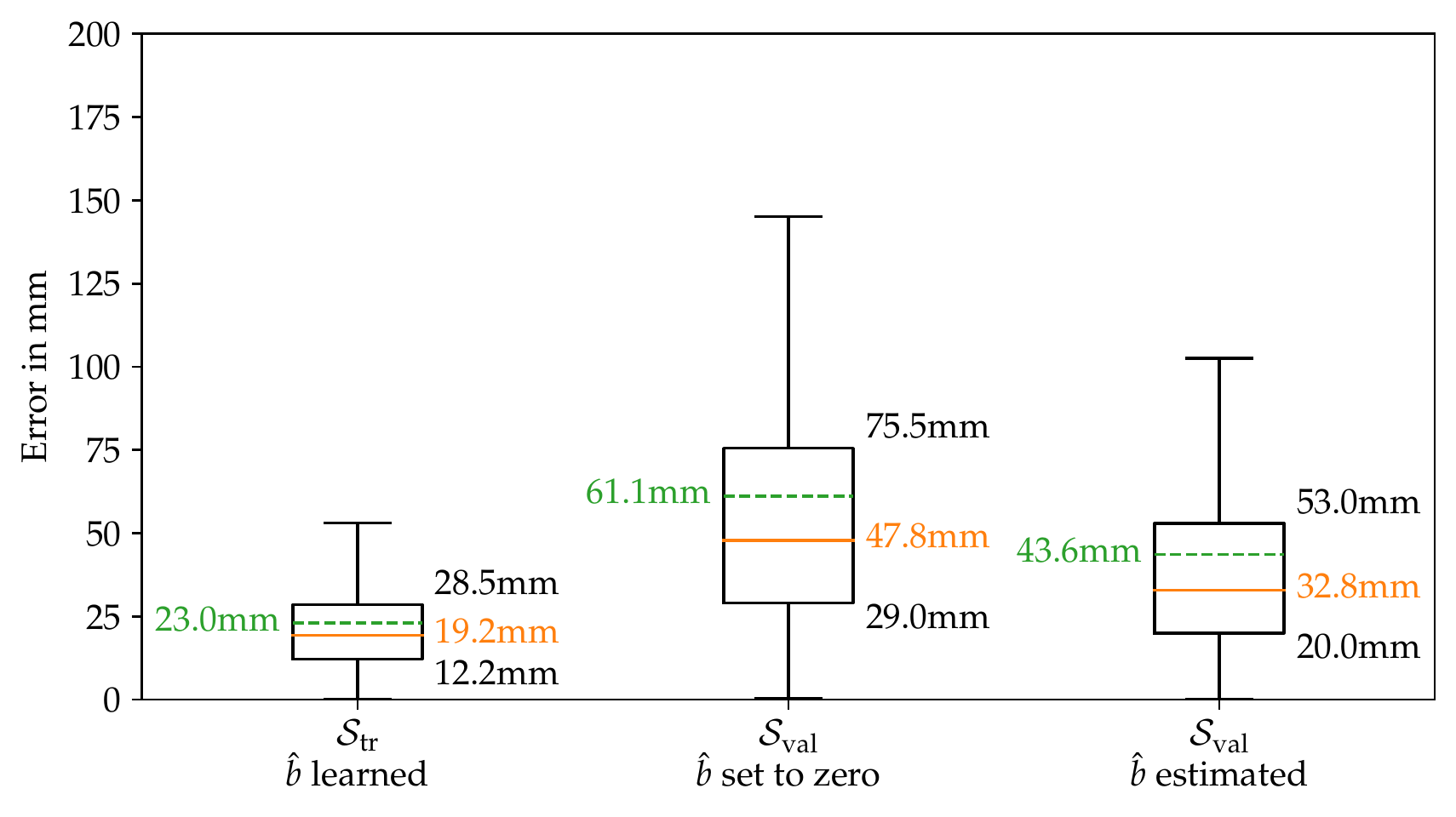}%
        \vspace*{-2mm}%
        \caption{Without fusion block on images of size $224 \times 224$ pixels.}%
        \label{f:evaluation:stg:fusion:results:nofus-i224}%
        \vspace*{2mm}%
    \end{subfigure}%
    \caption{Box-plots of the results of our RGBDTr model on the \stg\ dataset in four different configurations.}%
    \label{f:evaluation:stg:fusion:results}%
    \vspace*{-3mm}%
\end{figure}

When training the full RGBDTr model, we denote the training set as $\mathcal{S}_\mathrm{tr}$ and the validation set as~$\mathcal{S}_\mathrm{val}$. The training process was described in \Cref{s:processing:models:training}. We use $n_\mathrm{gan} = 100$, $n_\mathrm{rgbdtr} = 25$, and $n_\mathrm{ft} = 10$ epochs and no weight decay. In \Cref{f:evaluation:stg:fusion:results}, we show box-plots for each model configuration. Each box-plot shows the median distance error in \textcolor{orange!80!black}{orange} and the mean distance error in \textcolor{green!80!black}{green}. We estimate the subject-specific bias terms $\hat{b}$ using $n_\mathrm{cal} = 50$ randomly chosen samples from the validation set. There are about 800 samples per subject, which means that we use about \qty{6}{\percent} of the data to calibrate. We use the same calibration samples for all four configurations to ensure comparability. As there are no cross-validation splits, we do not have to aggregate the results. The error is calculated as the Euclidean distance between the predicted gaze point $\hat{p}$ and the ground truth gaze point $p$, i.e., $d_i = \left\Vert \hat{p}_i - p_i \right\Vert _2^2$ and $\bar{d} = \frac{1}{M} \sum_{i = 1}^{M} d_i$.

We can observe two phenomenons: first, the models without a fusion block (\Cref{f:evaluation:stg:fusion:results:nofus-i448,f:evaluation:stg:fusion:results:nofus-i224}) perform better than their counterparts with a fusion block (\Cref{f:evaluation:stg:fusion:results:fus-i448,f:evaluation:stg:fusion:results:fus-i224}). This is true for both input image sizes and regardless of the evaluation set. All metrics, including the median distance error and most of the quartiles, are better, i.e., lower, for the model without a fusion block. This indicates that the fusion block does not improve the model's performance.

\begin{figure}[htb]
    \vspace*{-3mm}%
    \centering\captionsetup{width=0.95\textwidth}%
    \includegraphics[width=\textwidth]{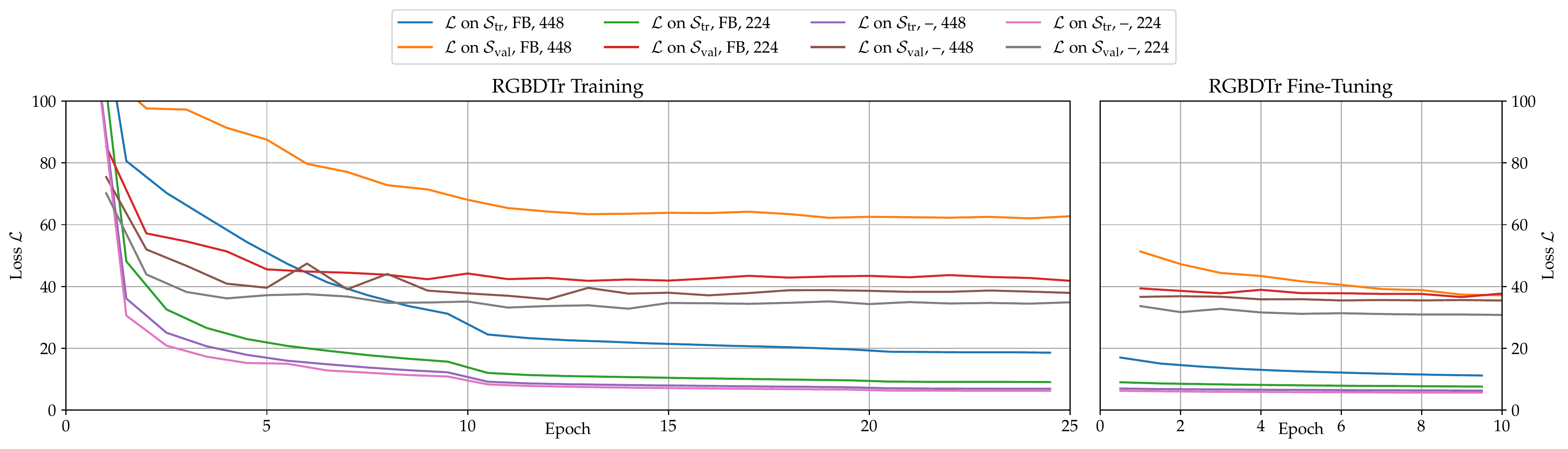}%
    \vspace*{-1mm}%
    \caption{Progression of the loss values during training and fine-tuning of our RGBDTr model on the \stg\ dataset in four different configurations.}%
    \label{f:evaluation:stg:fusion:losses}%
    \vspace*{-3mm}%
\end{figure}

\vspace*{-0.3mm}
{\spaceskip=3.4pt plus 1pt minus 1.5pt 
Second, we can observe that the model without a fusion block trained on smaller input images (\Cref{f:evaluation:stg:fusion:results:nofus-i224}) achieves lower errors than its counterpart trained on larger images (\Cref{f:evaluation:stg:fusion:results:nofus-i448}). This is also true for the model configuration with a fusion block: the model trained on smaller images (\Cref{f:evaluation:stg:fusion:results:fus-i224}) outperforms its counterpart (\Cref{f:evaluation:stg:fusion:results:fus-i448}). This indicates that the performance of the model benefits from both smaller images and the lack of a fusion block on the \stg\ dataset.

\vspace*{-0.3mm}
We validate the results of these four models by interpreting the losses during the training and fine-tuning process. The loss graphs are depicted in \Cref{f:evaluation:stg:fusion:losses}. The chart supports our findings that the models without a fusion block outperform the models with a fusion block. The model with a fusion block trained on the larger input images seems to benefit from fine-tuning the most (\textcolor{blue!80!white}{blue} and \textcolor{orange!80!black}{orange} lines). All other models seem to improve only slightly during fine-tuning. Nevertheless, the model with a fusion block trained on the larger input images performs significantly worse without the fine-tuning process, which indicates that the \ac{GAN} pre-training process may have resulted in a bad local minimum that could only be escaped by fine-tuning the overall model. However, the other models seem to have benefited only slightly from fine-tuning, which indicates that more fine-tuning epochs would not have improved the performance significantly.
}

{\spaceskip=3.4pt plus 1pt minus 1.5pt 
Our best approach is therefore the model without a fusion block trained on images of size $224 \times 224$ pixels. It achieves a median distance error of \qty{32.8}{\milli\meter} and a mean distance error of $\bar{d} = \qty{43.6}{\milli\meter}$. This is \qty{4.9}{\milli\meter} worse than the results reported by \citeauthor*{Lian2019}~\cite{Lian2019} (\qty{38.7}{\milli\meter}). However, our model's generator architecture has only three encoder and three decoder layers while the model by \citeauthor*{Lian2019} has four. Next, our model's eye pose feature extractor is a pre-trained ResNet-18, while the model by \citeauthor*{Lian2019} uses a ResNet-34. Furthermore, we do not feed raw eye coordinates into the model, because our architecture is focused on angular error minimization. We therefore argue that our model's performance is to some extent comparable to the model by \citeauthor*{Lian2019}.
}

\subsection{\acs{GAN} Backbone vs. no Depth Reconstruction}
\label{s:evaluation:stg:gan}

In the previous subsection, we noticed a strong influence of the \ac{GAN} training process on the overall model performance. In this section, we investigate the impact of the \ac{GAN} backbone on the model performance in more detail. To do so, we have trained a total of 16 RGBDTr models on the \stg\ dataset with different backbones. We trained 12 models on a pre-trained \ac{GAN} backbone in different training stages, i.e., after 25, 50, and 100 epochs. Furthermore, we trained four models without the \ac{GAN} backbone, i.e., removing the depth reconstruction part of the model. In visual terms, we remove the \textcolor{purple!80!black}{purple} colored modules of the model shown in \Cref{f:processing:models:overview}. It is important to note that we still use the \textcolor{blue!80!black}{blue} colored depth encoder module and the \textcolor{magenta!80!black}{magenta} colored depth extraction module. The depth extraction is performed on the input depth image without augmentation $I_i^\mathrm{FD}$. The relevant model configurations for this experiment are listed in \Cref{t:evaluation:stg:gan:models}. The models are named according to the following scheme: \texttt{<encoder\,layers>-<decoder\,layers>-<fusion\,block>[g<gan\,epoch>]}. In case of a missing decoder, we denote \texttt{x} as placeholder for \texttt{<decoder\,layers>}.

\renewcommand{\tableHeaderHeight}{30mm}
\begin{table}[htbp]
    \vspace{5mm}
    \centering
    \begin{tabular}{l|c|c|c|c|c|>{\centering\arraybackslash}b{18mm}|r}
        \textbf{Model} & \tableHeaderR{\textbf{\# Enc. Layers}} & \tableHeaderR{\textbf{\# Dec. Layers}} & \tableHeaderR{\textbf{Depth\\Reconstruction}} & \tableHeaderR{\textbf{Fusion Block}} & \tableHeaderR{\textbf{\ac{GAN} Epoch}} & \textbf{Input Image Size}\par\small{(in Pixels)} & \textbf{Remarks} \\
        \hline\hline

        4-4-1g100   & 4 & 4  & \cmark & \cmark & 100 & $448 \times 448$   & Same as in \Cref{s:evaluation:stg:fusion}\\
        4-4-1g50    & 4 & 4  & \cmark & \cmark & 50 & $448 \times 448$    & Same \ac{GAN}, but earlier epoch \\
        4-4-1g25    & 4 & 4  & \cmark & \cmark & 25 & $448 \times 448$    & Same \ac{GAN}, but earlier epoch \\
        4-x-1       & 4 & -- & \xmark & \cmark & -- & $448 \times 448$   & No decoder module \\

        \hline

        4-4-0g100   & 4 & 4  & \cmark & \xmark & 100 & $448 \times 448$   & Same as in \Cref{s:evaluation:stg:fusion}\\
        4-4-0g50    & 4 & 4  & \cmark & \xmark & 50 & $448 \times 448$    & Same \ac{GAN}, but earlier epoch \\
        4-4-0g25    & 4 & 4  & \cmark & \xmark & 25 & $448 \times 448$    & Same \ac{GAN}, but earlier epoch \\
        4-x-0       & 4 & -- & \xmark & \xmark & -- & $448 \times 448$   & No decoder module \\

        \hline

        3-3-1g100   & 3 & 3  & \cmark & \cmark & 100 & $224 \times 224$   & Same as in \Cref{s:evaluation:stg:fusion}\\
        3-3-1g50    & 3 & 3  & \cmark & \cmark & 50 & $224 \times 224$    & Same \ac{GAN}, but earlier epoch \\
        3-3-1g25    & 3 & 3  & \cmark & \cmark & 25 & $224 \times 224$    & Same \ac{GAN}, but earlier epoch \\
        3-x-1       & 3 & -- & \xmark & \cmark & -- & $224 \times 224$   & No decoder module \\

        \hline

        3-3-0g100   & 3 & 3  & \cmark & \xmark & 100 & $224 \times 224$   & Same as in \Cref{s:evaluation:stg:fusion}\\
        3-3-0g50    & 3 & 3  & \cmark & \xmark & 50 & $224 \times 224$    & Same \ac{GAN}, but earlier epoch \\
        3-3-0g25    & 3 & 3  & \cmark & \xmark & 25 & $224 \times 224$    & Same \ac{GAN}, but earlier epoch \\
        3-x-0       & 3 & -- & \xmark & \xmark & -- & $224 \times 224$   & No decoder module \\
    \end{tabular}
    \caption{Overview of the 16 RGBDTr models trained on the \stg\ dataset.}
    \label{t:evaluation:stg:gan:models}
\end{table}
\clearpage 

{\spaceskip=3.4pt plus 1pt minus 1.5pt 
\textbf{Training process.} The progression of the loss values during the training and fine-tuning process for all 16 RGBDTr models is depicted in \Cref{f:appendix:stg-losses}. It is important to note that the four models without the \ac{GAN} backbone -- namely {4-x-1}, {4-x-0}, {3-x-1}, and {3-x-0} -- do not have a fine-tuning process as described in \Cref{s:processing:models:training}. Therefore, the loss values for these models are only shown for the training process. One can clearly see that the models without \ac{GAN} backbone achieve significantly lower losses during training for both the training set $\mathcal{S}_\mathrm{tr}$ and the validation set $\mathcal{S}_\mathrm{val}$. In \Cref{f:appendix:stg-losses:fus-448,f:appendix:stg-losses:nofus-448,f:appendix:stg-losses:fus-224,f:appendix:stg-losses:nofus-224}, the \textcolor{magenta!80!black}{magenta} and \textcolor{gray!80!black}{gray} colored lines depict the loss values of the models without \ac{GAN} backbone. It is important to note that for the 4-x-1 model the loss on the validation set is similar to the other models' loss on the training set (see~\Cref{f:appendix:stg-losses:fus-448}). This indicates that the \ac{GAN} backbone hurts the performance of the RGBDTr model significantly. Furthermore, none of the models with \ac{GAN} backbone achieves a lower loss on the validation set than the models without \ac{GAN} backbone, even after the fine-tuning process. This is true for both input sizes and regardless of the existence of a fusion block.
}

\begin{figure}[htb]
    \centering%
    \includegraphics[width=\textwidth]{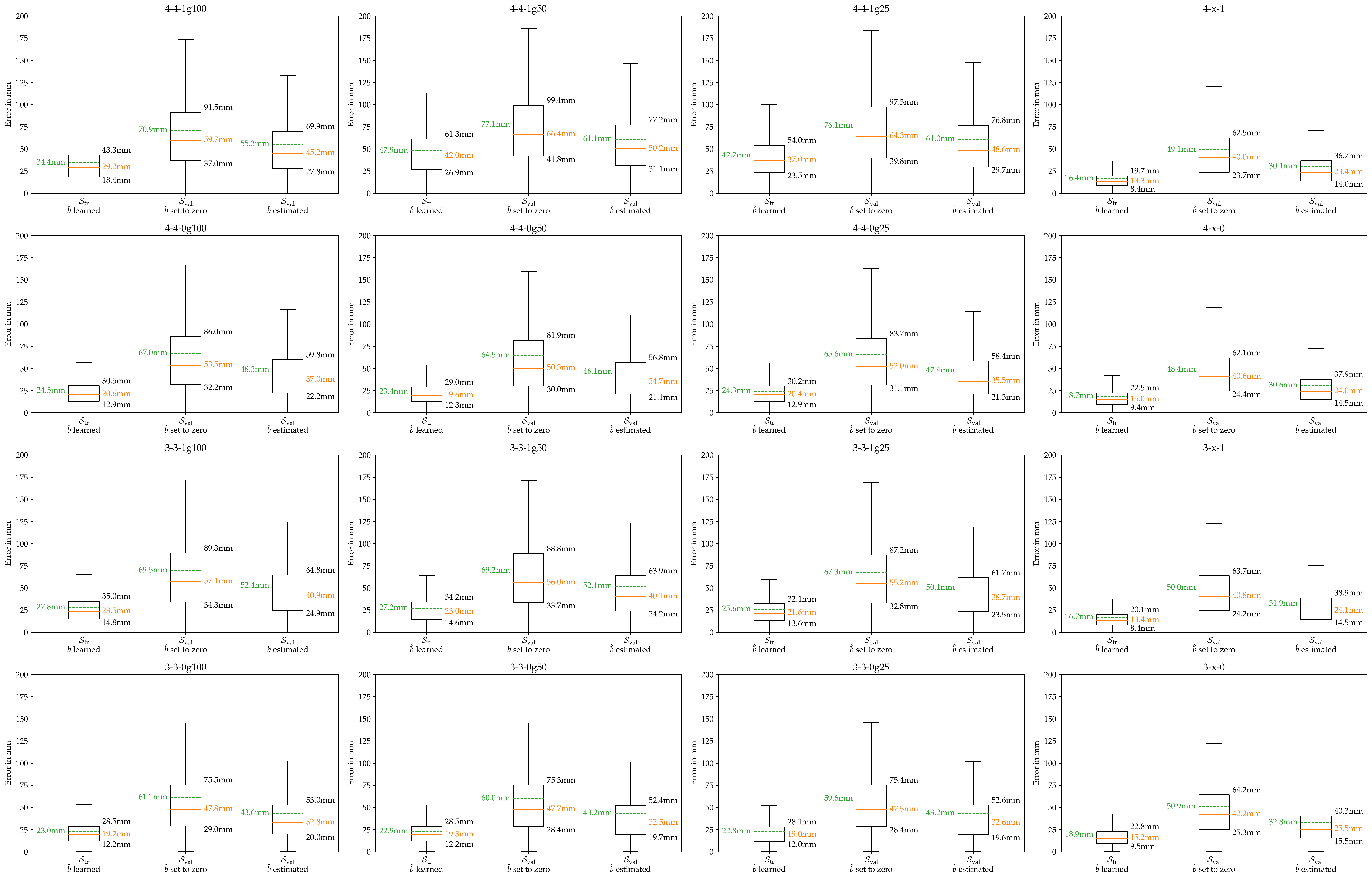}%
    \caption{Box-plots of the results of our RGBDTr model on the \stg\ dataset in 16 different configurations.}%
    \label{f:evaluation:stg:gan:results}%
\end{figure}

\textbf{Evaluation.} We use $n_\mathrm{cal} = 50$ randomly chosen samples for each subject of the validation set to estimate the subject-specific bias terms $\hat{b}$. The calibration samples are the same for all 16 models to ensure comparability. In \Cref{f:evaluation:stg:gan:results}, we show the performance of our RGBDTr models in the form of box-plots. Each box-plot shows the median distance error in \textcolor{orange!80!black}{orange} and the mean distance error in \textcolor{green!80!black}{green}. We calculate the errors as stated in \Cref{s:evaluation:stg:fusion}.

The first column in \Cref{f:evaluation:stg:gan:results} is identical to \Cref{f:evaluation:stg:fusion:results} and shows the results of the models that are based on a fully trained \ac{GAN} backbone. In the second and third column the results of the models that are based on half and quarterly trained \ac{GAN} backbones are shown. The fourth column shows the results of the models without \ac{GAN} backbone. It is clearly visible that these models achieve significantly lower errors than the models with \ac{GAN} backbone. This is true for the training set, the validation set, and the validation set after calibration. The training losses described above validate our findings.

\begin{figure}[htb]
    \vspace*{-2mm}%
    \centering\captionsetup{width=0.95\textwidth}%
    \includegraphics[width=\textwidth]{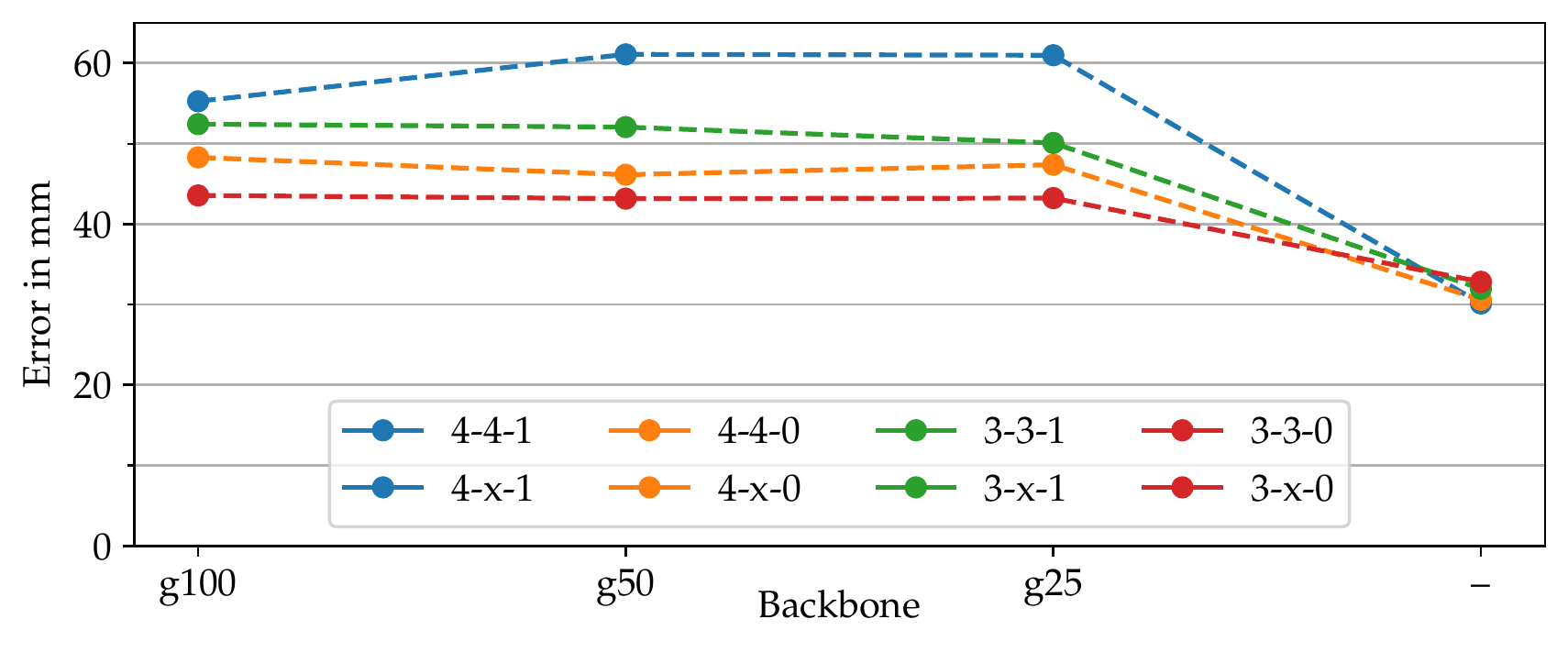}%
    \vspace*{-2mm}%
    \caption{Comparison of the mean distance errors of the 16 configurations of our RGBDTr model on the validation set $\mathcal{S}_\mathrm{val}$ using calibration.}%
    \label{f:evaluation:stg:gan:mean-error-by-backbone}%
\end{figure}

In order to take a closer look at the performance of the models, we compare the mean distance errors on the validation set $\mathcal{S}_\mathrm{val}$ in \Cref{f:evaluation:stg:gan:mean-error-by-backbone}. The figure groups a model family, i.e., all models sharing the same encoder architecture, by color and discriminates between the training stages of the used \ac{GAN} backbones. The overall model performance is similar regardless of the training stage of the \ac{GAN} backbone. One exception is the 4-4-1g100 model that performs better than its siblings 4-4-1g50 and 4-4-1g25. However, the models with no \ac{GAN} backbone perform significantly better than their siblings.

There is an interesting finding regarding the input image size and the impact of the fusion block. While models with \ac{GAN} backbone perform worse on larger input images and when incorporating the fusion block, this trend gets inverted for models without the \ac{GAN} backbone. The four layer encoder models 4-x-1 and 4-x-0 outperform their three layer counterparts 3-x-1 and 3-x-0. Furthermore, the models with a fusion block outperform their counterparts without a fusion block. This is true for both input image sizes. This indicates that the fusion block improves the performance of the model without \ac{GAN} backbone. However, the fusion block hurts the performance of the model with \ac{GAN} backbone as discussed above.

\textbf{Comparison with baseline.} Our best performing model is 4-x-1, which achieves a median distance error of \qty{23.4}{\milli\meter} and a mean distance error of $\bar{d} = \qty{30.1}{\milli\meter}$. This is \qty{8.6}{\milli\meter} better than the results reported by \citeauthor*{Lian2019}~\cite{Lian2019} (\qty{38.7}{\milli\meter}) and \qty{2.2}{\milli\meter} better than the results reported by \citeauthor*{Zhang2020a}~\cite{Zhang2020a} (\qty{32.3}{\milli\meter}).

\textbf{Conclusion.} The experiments in this section show a strong favor for the models that do not use a pre-trained \ac{GAN} as their backbone. Our findings indicate that the multi-task approach of \citeauthor*{Lian2019} to both refine depth images and predict gaze using a partly shared architecture leads to worse model performance compared to a single-task approach without depth reconstruction. Furthermore, we found that the fusion block and more encoder layers improve the performance of the model without \ac{GAN} backbone, making our 4-x-1 model the best performing model on the \stg\ dataset.

\subsection{RGBD Input vs. RGB Input}
\label{s:evaluation:stg:rgbd}

In order to investigate the influence of the depth map as an input to our RGBDTr model, we compare the four best performing models from the previous section with their counterparts that do not use the depth map as input. We therefore train four additional models on the \stg\ dataset with the same basic architectures as before. However, we additionally remove both the depth encoder module (\textcolor{blue!80!black}{blue} in \Cref{f:processing:models:overview}) and the depth extraction module (\textcolor{magenta!80!black}{magenta}) from the model as there is no depth map. These models share the same architecture as the models we trained on the \xgaze\ dataset (see~\Cref{s:evaluation:xgaze}).

\begin{figure}[htb]
    \centering\captionsetup{width=0.95\textwidth}%
    \includegraphics[width=\textwidth]{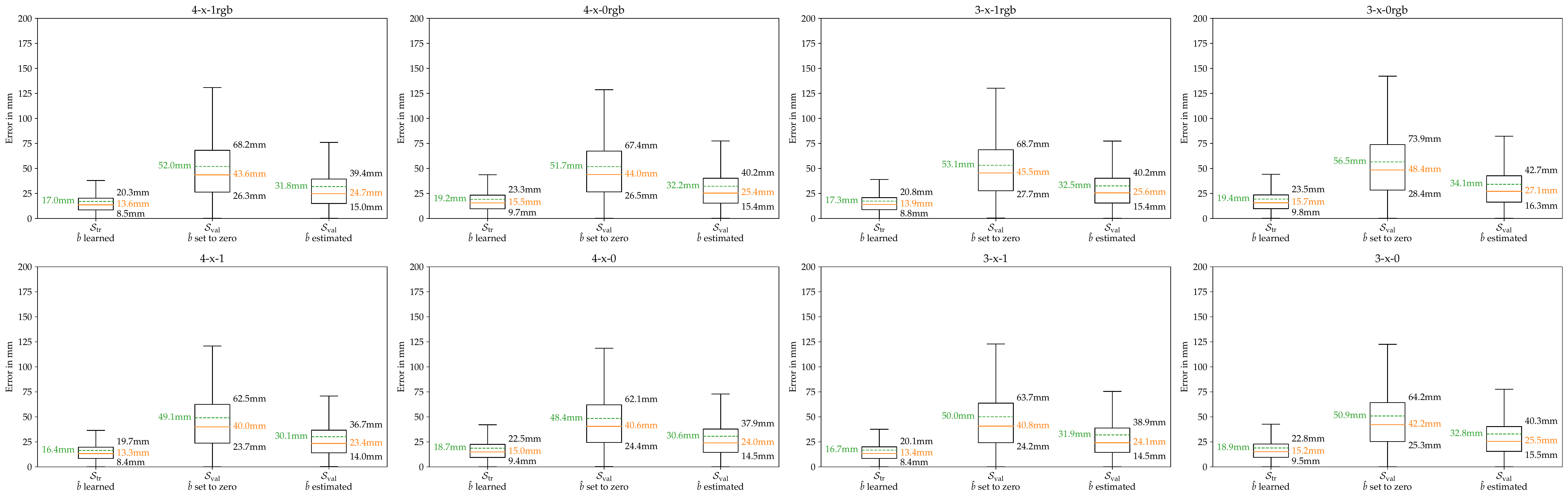}%
    \caption{Box-plots of the results of our RGBDTr model trained on RGB and RGBD images.}%
    \label{f:evaluation:stg:rgbd:results}%
\end{figure}

The results for both the models trained on RGBD images and the models trained on RGB images are visualized as box-plots in \Cref{f:evaluation:stg:rgbd:results}. The top row shows the results of the models that were trained and evaluated without depth input. In the bottom row, we show the results of the models that were trained and evaluated on RGBD input images. These models are identical to the best performing models of the last section and their results were already shown in the last column of \Cref{f:evaluation:stg:gan:results}. However, \Cref{f:evaluation:stg:rgbd:results} allows for a more direct comparison between similar architectures. We employ the same naming strategy as in the previous subsection and add an optional \texttt{rgb} suffix for the models that were trained without the depth input.

The models trained on RGBD images outperform their counterparts trained on RGB images regardless of the input image size and the existence of a fusion block. However, the difference in performance is considerably smaller compared to what we saw in the previous subsection regarding the \ac{GAN} backbone.

For a better comparison, we list the results of all eight models on the validation set $\mathcal{S}_\mathrm{val}$ of the \stg\ dataset with applied calibration in \Cref{t:evaluation:stg:rgbd:results}. The calibration was done using $n_\mathrm{cal} = 50$ randomly sampled calibration images per subject. The selection was the same for all models to ensure comparability. The table shows that for any given architecture, e.g., 4-x-1, the model trained on RGBD images outperforms its counterpart trained on RGB images. The differences are \qty{1.7}{\milli\meter}, \qty{1.6}{\milli\meter}, \qty{0.6}{\milli\meter}, and \qty{1.3}{\milli\meter} for the architectures 4-x-1, 4-x-0, 3-x-1, and 3-x-0, respectively. The differences are smaller for the models with a three layer encoder architecture. It is important to note that the architectures with a fusion block achieve lower errors than their counterparts without a fusion block. This is true for both input image sizes and for regardless of the depth input. This indicates that the fusion block improves the performance even when using only RGB images as input.

\begin{table}[htbp]
    \vspace*{-2mm}%
    \renewcommand{\arraystretch}{1.2}%
    \centering%
    \begin{filecontents*}{stg-experiments-rgb-mean-error-by-architecture.csv}
,\textbf{4-x-1},\textbf{4-x-0},\textbf{3-x-1},\textbf{3-x-0}
\textbf{RGB},31.8,32.2,32.5,34.1
\textbf{RGBD},30.1,30.6,31.9,32.8
\end{filecontents*}
\csvreader[
        tabular=l|cccc,
        no head,
        table head=,
        late after first line={\\\hline},
    ]{stg-experiments-rgb-mean-error-by-architecture.csv}{}{\csvlinetotablerow}
    \caption{Mean distance errors $\bar{d}$ in \qty{}{\milli\meter} of our RGBDTr model on the validation set $\mathcal{S}_\mathrm{val}$ using calibration.}
    \label{t:evaluation:stg:rgbd:results}
\end{table}

Although our RGB-only models achieve worse results than our RGBD models, they all outperform the baseline. Our best performing RGB-only model is 4-x-1rgb, which achieves a median distance error of \qty{24.7}{\milli\meter} and a mean distance error of $\bar{d} = \qty{31.8}{\milli\meter}$. This is still \qty{6.9}{\milli\meter} better than the results reported by \citeauthor*{Lian2019}~\cite{Lian2019} (\qty{38.7}{\milli\meter}) and \qty{0.5}{\milli\meter} better than the results reported by \citeauthor*{Zhang2020a}~\cite{Zhang2020a} (\qty{32.3}{\milli\meter}) on RGBD input.

\subsection{Conclusion on the \stg\ Dataset}
\label{s:evaluation:stg:conclusion}

In this section, we have conducted multiple experiments on the \stg\ dataset to determine basic architecture choices for our RGBDTr model. We discussed the \ac{GAN} and especially the discriminator pre-training process in \Cref{s:evaluation:stg:discriminator} and found well-working four and three layer PatchGAN configurations. We then compared the performance of our RGBDTr model with and without a fusion block in \Cref{s:evaluation:stg:fusion} and found that the fusion block hurts the performance of the model. Furthermore, we found that the model performs better on smaller input images.

However, in \Cref{s:evaluation:stg:gan}, we investigated the impact of the \ac{GAN} backbone on the model performance and found that a pre-trained \ac{GAN} backbone hurts the performance of the model. Contrary to our earlier findings, we also found that the fusion block improves the performance of the model without \ac{GAN} backbone.

When comparing the performance of our RGBDTr model on RGBD images to the performance of our model on RGB images in \Cref{s:evaluation:stg:rgbd}, we found that the RGBD model outperforms the RGB model. However, the difference in performance is smaller than the difference in performance between the models with and without \ac{GAN} backbone. Nevertheless, this is a strong indicator that the depth map contains additional information that is useful for the task of Gaze Point Estimation and, thus, being a valuable input to our RGBDTr model.

\clearpage
\section{Experiments on our \oge\ Dataset}
\label{s:evaluation:oge}

In this section, we conduct experiments on our \oge\ dataset to evaluate the performance of our RGBDTr model in various architecture configurations. We conduct similar experiments as in the previous section on the \stg\ dataset to determine whether the specific task of Gaze Angle Estimation benefits from the same architecture choices as the task of Gaze Point Estimation. As there is no baseline model available on our own dataset, we compare the performance of our RGBDTr models among themselves.

\subsection{Dataset Preprocessing, Training and Evaluation Procedure}
\label{s:evaluation:oge:preprocessing}

Our \oge\ dataset provides both a dedicated test set $\mathcal{D}_\mathrm{te}$ and a 5-fold cross-validation split with training and validation subsets $\left(\mathcal{D}_\mathrm{tr}^{(i)}, \mathcal{D}_\mathrm{val}^{(i)}\right), i \in \left\{1, ..., 5\right\}$. Since the data collection and preprocessing steps described in \Cref{s:processing:dataset} have already been performed, there is no need for further preprocessing.

For \ac{GAN} training, we use the discriminator target set $\tau$ as described in \Cref{s:processing:dataset:preprocessing}. We follow the training procedure described in \Cref{s:processing:models:training} and train for $n_\mathrm{gan} = 100$, $n_\mathrm{rgbdtr} = 25$, and $n_\mathrm{ft} = 10$ epochs using no weight decay.

Similar to our experiments on the \xgaze\ and the \stg\ dataset, we perform no fine-tuning for the models without \ac{GAN} backbone. Furthermore, we use the same $n_\mathrm{cal} = 200$ calibration samples per subject for all models to ensure comparability. Since there are $k=5$ cross-validation splits, we aggregate the results of the five models in the following subsections. Similar to the \xgaze\ dataset, we denote the aggregated results on the training subsets $\mathcal{D}_\mathrm{tr}^{(1)}, \dots, \mathcal{D}_\mathrm{tr}^{(5)}$ as $\mathcal{D}_\mathrm{tr}^{*}$ and on the validation subsets $\mathcal{D}_\mathrm{val}^{(1)}, \dots, \mathcal{D}_\mathrm{val}^{(5)}$ as $\mathcal{D}_\mathrm{val}^{*}$.

To calculate the angular error $e_i$ of the $i$-th sample, we use the same metric as \citeauthor*{Zhang2020}~\cite{Zhang2020}. $e_i$ is calculated using the cosine similarity between the (normalized) ground truth gaze vector $g'_3$ and the (normalized) predicted gaze vector $\hat{g}'_3$ of that sample:
\[e_i = \cos^{-1}\left(\frac{g'_3 \cdot \hat{g}'_3}{\left\Vert g'_3 \right\Vert \cdot \left\Vert \hat{g}'_3 \right\Vert}\right), \qquad \bar{e} = \frac{1}{M} \sum_{i=1}^{M} e_i \]
where $M$ is the number of samples in the evaluation set.

Based on the predicted gaze angles, we can un-normalize them and calculate the predicted gaze point $\hat{p}$ in screen coordinates as described in \Cref{s:processing:sequence:unnorm}. We can then calculate the Euclidean distance between the predicted gaze point $\hat{p}$ and the ground truth gaze point $p$ as $d_i = \left\Vert \hat{p}_i - p_i \right\Vert _2^2$ and the mean distance error as $\bar{d} = \frac{1}{M} \sum_{i=1}^{M} d_i$.

It is important to note that the numbers are not directly comparable to the results on the \stg\ dataset. Although both we and \citeauthor*{Lian2019} used a \SI{27}{\inch} monitor, the physical dimensions of the monitor used by \citeauthor*{Lian2019} are not specified. Furthermore, the distance between the monitor and the subjects is different and the data collection procedure differs significantly.

\subsection{\acs{GAN} Backbone vs. no Depth Reconstruction}
\label{s:evaluation:oge:gan}

Our experiments on the \stg\ dataset in \Cref{s:evaluation:stg:gan} showed a strong favor for a model without \ac{GAN} backbone. Furthermore, we saw little difference between models that were based on the same \ac{GAN} backbone in different training stages. In this section, we investigate the impact of the \ac{GAN} backbone on the performance of our RGBDTr model on our \oge\ dataset and compare the performance of the models with and without \ac{GAN} backbone. We train three model architectures\,\footnote{Because of $k=5$, a total of 15 models were trained.} on the \oge\ dataset and list their configurations in \Cref{t:evaluation:oge:gan:models}. Instead of using the same \ac{GAN} backbone in different training stages, we use two differently trained \ac{GAN} backbones: one was trained on the discriminator target set $\tau$ and one was trained on the full input set $\upsilon$. During the fine-tuning process, we also use the discriminator target set accordingly. We employ the same naming scheme as in \Cref{s:evaluation:stg:gan} and indicate with a suffix whether the model was trained on $\tau$ or $\upsilon$.

\begin{table}[htbp]
    \centering\captionsetup{width=\textwidth}%
    \vspace*{2mm}
    \begin{tabular}{l|c|c|c|c|c|>{\centering\arraybackslash}b{18mm}|r}
        \textbf{Model} & \tableHeaderR{\textbf{\# Enc. Layers}} & \tableHeaderR{\textbf{\# Dec. Layers}} & \tableHeaderR{\textbf{Depth\\Reconstruction}} & \tableHeaderR{\textbf{Fusion Block}} & \tableHeaderR{\textbf{\ac{GAN} Epoch}} & \textbf{Input Image Size}\par\small{(in Pixels)} & \textbf{Remarks} \\
        \hline\hline

        4-4-1g$\tau$     & 4 & 4  & \cmark & \cmark & 100 & $448 \times 448$ & Discriminator trained on $\tau$ \\
        4-4-1g$\upsilon$ & 4 & 4  & \cmark & \cmark & 100 & $448 \times 448$ & Discriminator trained on $\upsilon$ \\
        4-x-1       & 4 & -- & \xmark & \cmark & -- & $448 \times 448$ & No decoder module \\
    \end{tabular}
    \vspace*{3mm}
    \caption{Overview of the 3 RGBDTr model configurations trained on the \oge\ dataset.}
    \label{t:evaluation:oge:gan:models}
\end{table}

\begin{figure}[htb]
    \centering\captionsetup{width=0.95\textwidth}%
    \includegraphics[width=\textwidth]{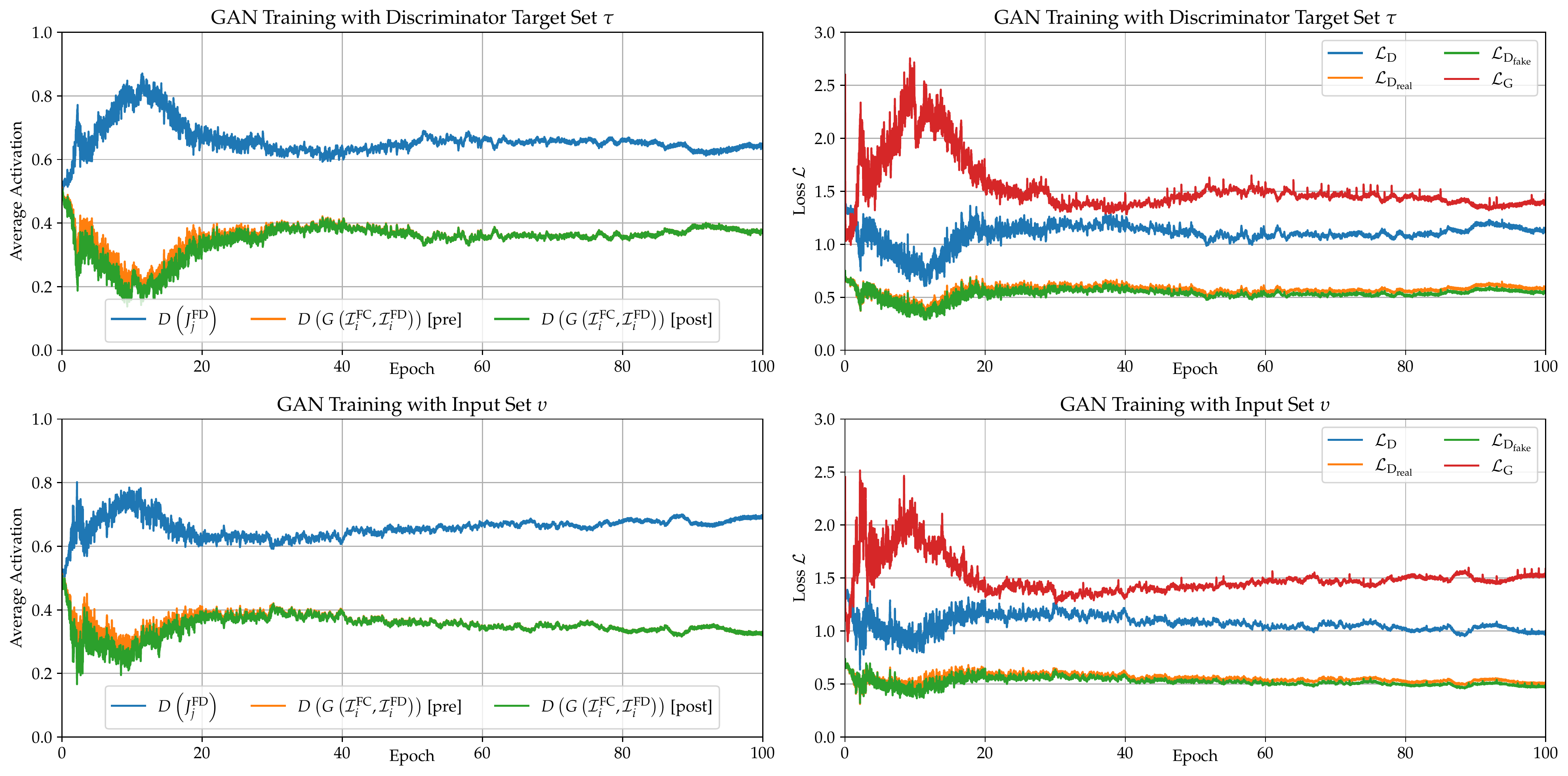}%
    \caption{Progression of the average activation and loss values of G and D during training of the \ac{GAN} backbone on the \oge\ dataset with the discriminator target set $\tau$ (top row) and the full input set $\upsilon$ (bottom row).}%
    \label{f:evaluation:oge:gan:gan-training}%
\end{figure}

\textbf{Training process.} In \Cref{f:evaluation:oge:gan:gan-training}, we show the progression of the average activation and loss values during the \ac{GAN} training. In the top row, the training process using the discriminator target set $\tau$ is depicted, whereas in the bottom row, the training process using the input set $\upsilon$ is shown. We apply a sliding window of 100 batches to smooth the graphs and therefore allow for a more useful visualization.

Although we use a four layer PatchGAN architecture, which is the same as in \Cref{s:evaluation:stg:gan}, the training process graphs look different (cf.~\Cref{f:evaluation:stg:fusion:pgan:ndis4}). However, since the training process graphs of the two models trained on our \oge\ datset (\Cref{f:evaluation:oge:gan:gan-training}) look similar, we assume that the differences to the training process graphs of the models trained on the \stg\ dataset is due to the different datasets. In fact, their depth maps follow dissimilar distributions as described earlier in \Cref{s:processing:models:training,s:processing:dataset:preprocessing}. The training processes depicted in \Cref{f:evaluation:oge:gan:gan-training} are very similar even though the discriminators were trained on different sets. The initial spikes in average activation and loss values within the first 20 epochs are less pronounced on the training process using $\upsilon$ (bottom row). However, this behavior could be due to run-to-run variance and further investigation is needed to confirm a systemic difference.

The RGBDTr training process is visualized in \Cref{f:appendix:oge-rgbdtr-training} as the progression of the loss values of the trained models. \Cref{f:appendix:oge-rgbdtr-training:tau} shows it for models trained with the discriminator target set $\tau$, \Cref{f:appendix:oge-rgbdtr-training:ups} for the input set $\upsilon$, and \Cref{f:appendix:oge-rgbdtr-training:nogan} shows it for the models trained without \ac{GAN} backbone. Similar to our findings on the \stg\ dataset (see~\Cref{s:evaluation:stg:gan}), the losses of the models with \ac{GAN} backbones are significantly higher than of the models without it. This is true for both the training and validation subsets. Although fine-tuning the overall model improves the loss values, the models with \ac{GAN} backbone still perform worse than the models without \ac{GAN} backbone. Furthermore, the models trained with input set~$\upsilon$ achieve slightly lower loss values than the models trained with the discriminator target set~$\tau$. This indicates that our efforts to improve depth reconstruction by using a subset of the input set as discriminator target set lead to worse overall model performance. In fact, this supports our findings on the \stg\ dataset in that the multi-task approach of \citeauthor*{Lian2019} leads to worse model performance due to the conflicting objectives of depth reconstruction and \ges.

Furthermore, it is noticeable that the losses on the first validation subset $\mathcal{D}_\mathrm{val}^{(1)}$ are always significantly less than on the other subsets regardless of the training stage and model configuration. We argue that this is related to the cross-validation splits, but did not conduct further experiments to investigate this phenomenon. During evaluation, we accumulate the results of all five models because of our cross-validation strategy.

\textbf{Evaluation.} As described above, we use $n_\mathrm{cal} = 200$ randomly chosen samples for each subject of the respective validation subsets to estimate the subject-specific bias terms $\hat{b}$. The calibration samples are the same for all models to ensure comparability. Furthermore, we aggregate the results of all $k=5$ subsets.

\begin{figure}[htb]
    \vspace*{-1mm}
    \centering\captionsetup{width=0.95\textwidth}%
    \begin{subfigure}{\textwidth}%
        \includegraphics[width=\textwidth]{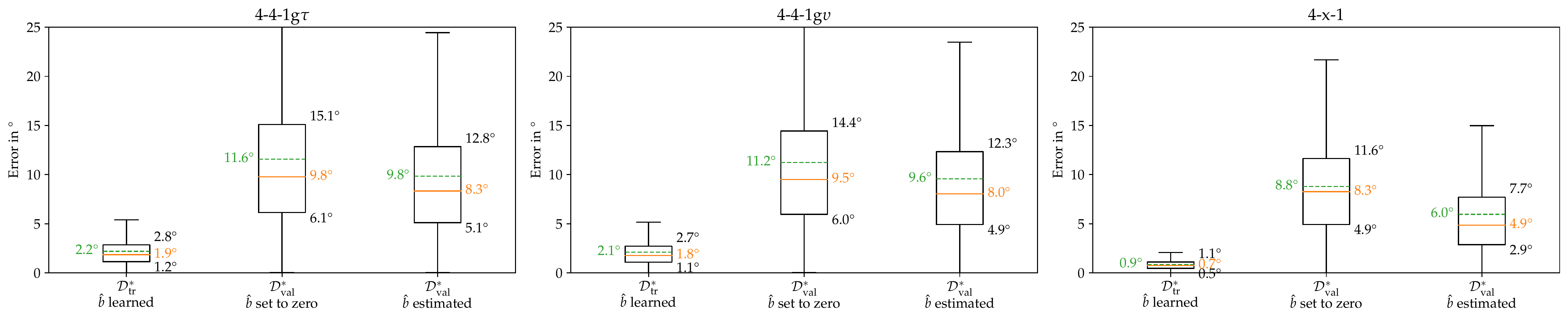}%
        \vspace*{-2mm}%
        \caption{Angular error.}%
        \label{f:evaluation:oge:gan:results:ang}%
        \vspace*{2mm}%
    \end{subfigure}
    \begin{subfigure}{\textwidth}%
        \includegraphics[width=\textwidth]{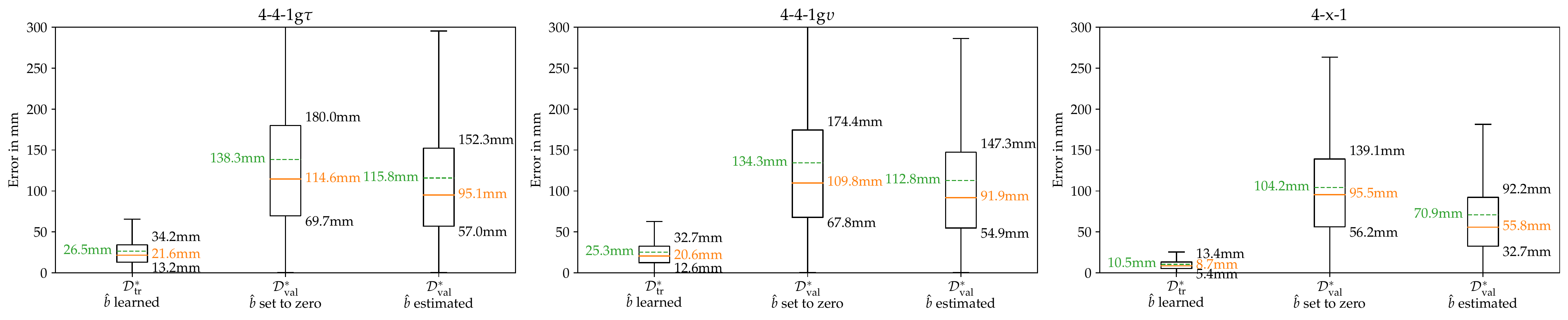}%
        \vspace*{-2mm}%
        \caption{Euclidean error.}%
        \label{f:evaluation:oge:gan:results:distance}%
        \vspace*{2mm}%
    \end{subfigure}
    \caption{Box-plots of the results of our RGBDTr model on our \oge\ dataset in three different configurations.}%
    \label{f:evaluation:oge:gan:results}%
    \vspace*{-5mm}
\end{figure}

In \Cref{f:evaluation:oge:gan:results}, the results of the three model configurations are visualized as box-plots. The angular and Euclidean errors are depicted in \Cref{f:evaluation:oge:gan:results:ang} and \Cref{f:evaluation:oge:gan:results:distance}, respectively. It is clearly visible that the models with \ac{GAN} backbone perform worse than the models without it. The models that were trained on the discriminator target set $\tau$ achieve a mean angular error of $\bar{e} = \ang{9.8}$ and a mean Euclidean error of $\bar{d} = \qty{115.8}{\milli\meter}$ on the validation subsets using calibration. The models that were trained on the input set $\upsilon$ achieve slightly better results with $\bar{e} = \ang{9.6}$ and $\bar{d} = \qty{112.8}{\milli\meter}$. However, these models are significantly worse than the models without \ac{GAN} backbone, which achieve a mean angular error of $\bar{e} = \ang{6.0}$ and a mean Euclidean error of $\bar{d} = \qty{70.9}{\milli\meter}$. This is an improvement of about \qtyrange{37.1}{38.8}{\percent} over the other two model configurations.

It is also noticeable that the models without \ac{GAN} backbone benefit more from the calibration step. The difference in mean angular error between the uncalibrated and calibrated results is \ang{2.8}, whereas the difference for the models with \ac{GAN} backbone is \ang{1.8} and \ang{1.6} for the models trained on $\tau$ and $\upsilon$, respectively.

{\spaceskip=3.4pt plus 1pt minus 1.5pt 
When comparing the results on the training subsets to the results on the validation subsets, we see that there is a larger -- both relative and absolute -- gap in mean and median errors than on the \stg\ and \xgaze\ datasets (see~\Cref{s:evaluation:stg,s:evaluation:xgaze}). We argue that this is due to the fact that there are fewer subjects in our dataset, which leads to more overfitting during training. This behavior of larger gaps between training and validation errors is also visible in the training process graphs in \Cref{f:appendix:oge-rgbdtr-training}.
}

\textbf{Conclusion.} Our experiments on the \oge\ dataset show that the \ac{GAN} backbone hurts the performance of our RGBDTr model, as we saw in previous experiments on the \stg\ dataset (see~\Cref{s:evaluation:stg:gan}). We therefore conclude that the multi-task approach of \citeauthor*{Lian2019} is not suited for the goal of good gaze estimation performance. We support our claims with our findings during the training process described above and the results on both datasets. Furthermore, we propose our 4-x-1 model as baseline for future experiments and comparisons.

\subsection{RGBD Input vs. RGB Input}
\label{s:evaluation:oge:rgbd}

In \Cref{s:evaluation:stg:rgbd}, we saw that adding the depth map as an input to our RGBDTr model improves the mean error on the \stg\ dataset. Within this subsection, we investigate the effect on our \oge\ dataset. Predicting gaze on our dataset is a different task than on the \stg\ dataset, because on our dataset a model takes normalized images as input and outputs gaze angles. Therefore, the results on the \stg\ dataset may not directly translate to our dataset, so it is reasonable to conduct these experiments.

\begin{figure}[htb]
    \centering\captionsetup{width=0.95\textwidth}%
    \begin{subfigure}{0.5\textwidth}%
        \includegraphics[width=\textwidth]{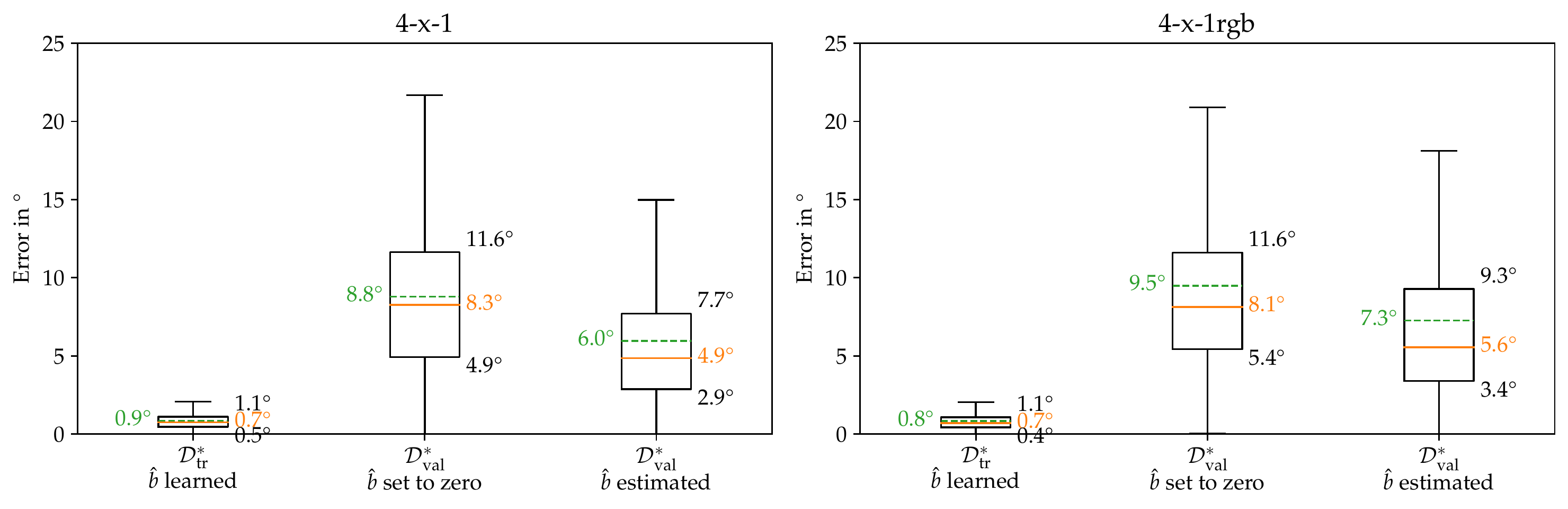}%
        \vspace*{-2mm}%
        \caption{Angular error.}%
        \label{f:evaluation:oge:rgbd:results:ang}%
        \vspace*{2mm}%
    \end{subfigure}%
    \begin{subfigure}{0.5\textwidth}%
        \includegraphics[width=\textwidth]{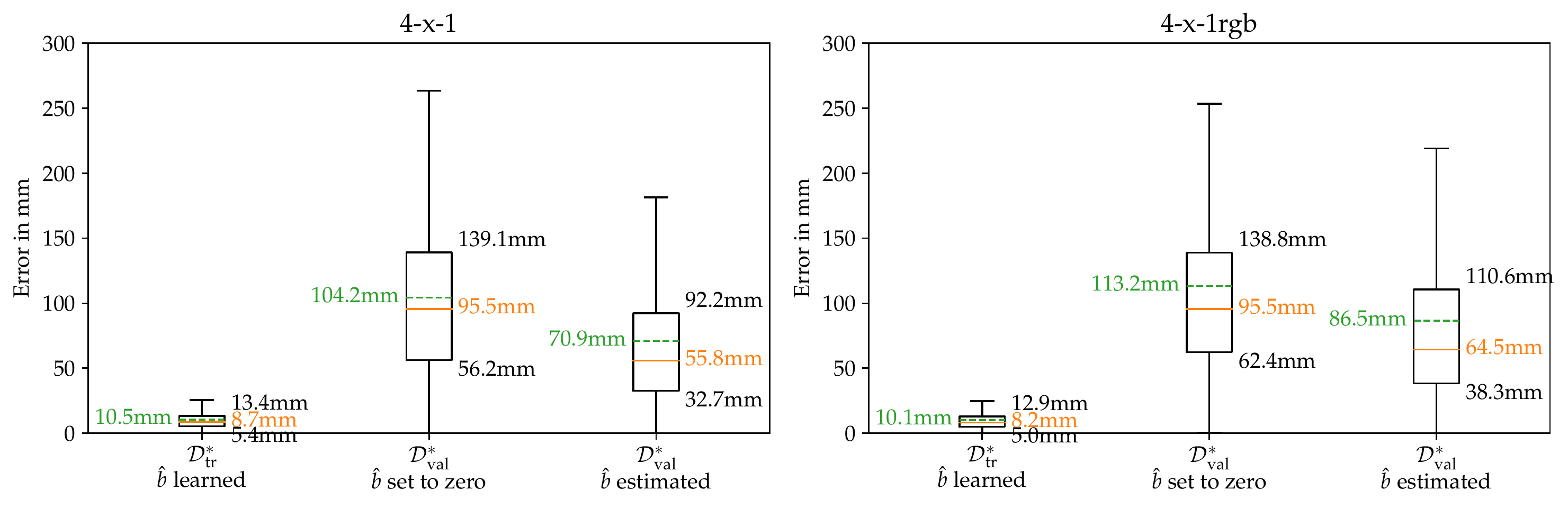}%
        \vspace*{-2mm}%
        \caption{Euclidean error.}%
        \label{f:evaluation:oge:rgbd:results:distance}%
        \vspace*{2mm}%
    \end{subfigure}
    \caption{Box-plots of the results of our RGBDTr model trained on RGB and RGBD images.}%
    \label{f:evaluation:oge:rgbd:results}%
\end{figure}

As described in \Cref{s:evaluation:stg:rgbd}, we remove the depth encoder, the depth decoder, and the depth extraction modules from our model architecture in order to train on RGB images. Furthermore, we employ the same naming strategy. We compare the RGB-only model to its RGBD counterpart in \Cref{f:evaluation:oge:rgbd:results}. Although the angular error on the training subsets is similar or slightly lower for the RGB model, the model trained on RGBD images outperforms its counterpart on both the uncalibrated and calibrated validation subsets. The difference in mean angular error is \ang{0.7} for the uncalibrated validation and \ang{1.3} for the validation subsets. When comparing the mean distance errors on the validation subsets using calibration, the RGBD model outperforms its counterpart by \qty{15.6}{\milli\meter} with a mean Euclidean error of $\bar{d} = \qty{70.9}{\milli\meter}$. Our findings indicate that the depth map contains additional information that is useful for the task of Gaze Angle Estimation on our \oge\ dataset.

We did not conduct more experiments regarding the input image size and the existence of a fusion block due to the long training times of about \qty{45}{\hour} per experiment. However, we expect that adding the depth map as input to our model will improve the performance regardless of the specific architecture, as we saw in \Cref{s:evaluation:stg:rgbd} on the \stg\ dataset.

\subsection{Error Visualization for our Baseline Model}
\label{s:evaluation:oge:error-vis}
\begin{figure}[htb]
    \vspace*{-1.5mm}%
    \centering\captionsetup{width=0.95\textwidth}%
    \includegraphics[width=\textwidth]{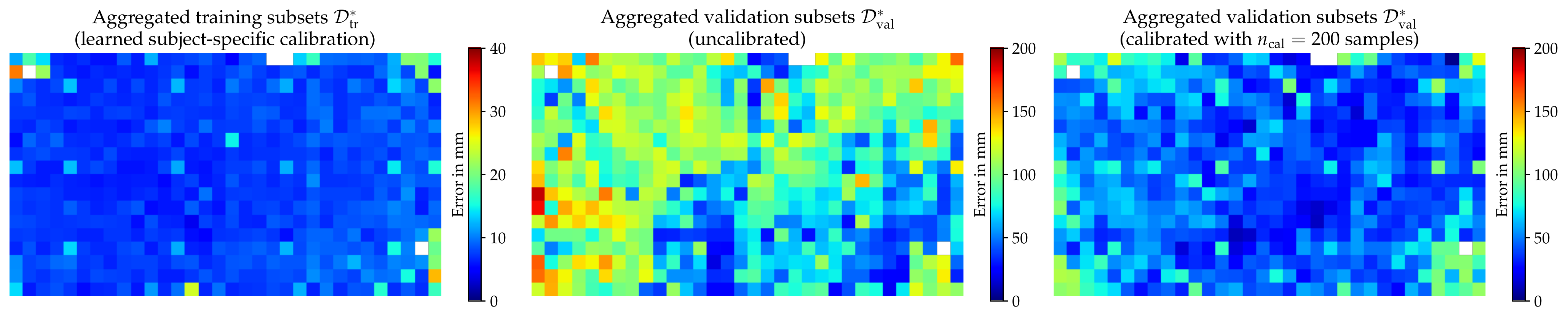}%
    \caption{Distribution of the on-screen Euclidean errors of the baseline model aggregated over all subsets.}%
    \label{f:evaluation:oge:error-vis:hist-agg}%
    \vspace*{-5mm}%
\end{figure}

In this section, we visualize the error of our baseline model 4-x-1 on the training and validation subsets of our \oge\ dataset. In \Cref{f:evaluation:oge:error-vis:hist-agg}, we show a 2D-histogram of the Euclidean errors. We aggregate the $k=5$ training and validation subsets, respectively, and show for each on-screen region the average Euclidean error. The color scales are different for the training and the validation subsets due to the larger errors on the latter ones. In addition to that, we show the distribution for each subset in \Cref{f:appendix:oge-baseline-error-hist} for a more detailed view.

The border areas of the screen are also the regions where the baseline model performs the worst. This is especially true for the corner regions, despite our efforts to generate samples in these regions (see \Cref{s:processing:datacollection:process} and \Cref{f:processing:datacollection:process:grid}). Comparing the uncalibrated and calibrated results, we see that the calibration step improves the performance in all parts of the screen except for the lower right corner region. It is possible that this bias towards the upper and left parts of the screen is due to the underlying data distribution we discussed in \Cref{s:processing:dataset:distribution}.

We can compare the results of the five models trained on the different training and validation subsets visually in \Cref{f:appendix:oge-baseline-error-hist}. While the error distributions on the first and third validation subsets are smaller than on the other subsets, they are not similar. The baseline model evaluated on the first validation subset performs well in the lower right quadrant and the baseline model evaluated on the third validation subset performs well in the lower third of the screen. However, the models evaluated on the second and fourth validation subsets perform bad in the left quarter and the lower left to upper right diagonal, respectively. In contrast, this is where the baseline model evaluated on the fifth validation subset performs well. This phenomenon may be related to the cross-validation splits, as the areas where a model performs good or bad are similar across different model architectures.

However, we can also conclude that the error induced by the models can be reduced significantly using the calibration step. This is true for all five cross-validation splits and is visualized in the third column in \Cref{f:appendix:oge-baseline-error-hist}. As noted in \Cref{f:evaluation:oge:rgbd:results:distance}, the calibration step reduces the mean Euclidean error by \qty{33.3}{\milli\meter} on the validation subsets.

\vspace*{-1.5mm}%
\subsection{Conclusion on the \oge\ Dataset}
\label{s:evaluation:oge:conclusion}
{\spaceskip=3.4pt plus 1pt minus 1.5pt 
We have conducted experiments on our \oge\ dataset to evaluate the performance of our RGBDTr model in different architecture configurations. We compared the performance of our model with and without pre-trained \ac{GAN} backbone and found that it hurts the performance of the model significantly. Furthermore, we found that adding the depth map as input to our model improves the performance on our dataset like on the \stg\ dataset. We therefore conclude that the depth map contains additional information that is best used without a pre-trained \ac{GAN} backbone.

We propose our 4-x-1 model with a mean angular error of $\bar{e} = \ang{6.0}$ on the validation subsets using calibration as baseline model. In the next section, we conduct experiments on various hyperparameters to find the best performing model configuration. We use our 4-x-1 model as a basis for these experiments and compare the results to it.
}

\clearpage
\section{Tuning of Hyperparameters}
\label{s:evaluation:hyperparams}

In this section, we conduct experiments to determine the impact of various hyperparameters on the performance of our RGBDTr model. We use our \oge\ dataset for these experiments, because it features depth maps (vs. \xgaze) and normalized images (vs. \stg). To compare the influence of modified hyperparameters, we compare the newly trained models to our 4-x-1 baseline model (see~\Cref{s:evaluation:oge}) that achieved a mean angular error of $\bar{e} = \ang{6.0}$ and a mean Euclidean error of $\bar{d} = \qty{70.9}{\milli\meter}$ on the validation subsets $\mathcal{D}_\mathrm{val}^{(1)}, \dots, \mathcal{D}_\mathrm{val}^{(5)}$ using $n_\mathrm{cal} = 200$ calibration samples per subject.

In total, we tried to optimize three hyperparameters of our RGBDTr architecture: the number of Transformer layers $n_l$, the Transformer token size $d_\mathrm{model}$, and the number of encoder layers $n_e$ in the two generator encoders for RGB images and depth maps. We did not optimize the number of decoder layers $n_d$, because our experiments are conducted on models without depth reconstruction, i.e., no pre-trained \ac{GAN} backbone.

Our parameter choice is motivated by two reasons: first, the focus of this thesis is on the feature fusion step using the Transformer architecture. Therefore, optimizing the two main hyperparameters -- $n_l$ and $d_\mathrm{model}$ -- of this architecture seems important. Second, we want to validate our earlier choice of using a four layer PatchGAN made in \Cref{s:evaluation:stg:discriminator} and, thus, $n_e = 4$ as our baseline model.

There are many more possibilities to enhance the performance of our model, e.g., by using a different backbone architecture or by using a different loss function. However, we limit our experiments to the three hyperparameters mentioned above due to time constraints and give an outlook on other hyperparameters that can also be optimized in \Cref{s:evaluation:hyperparams:outlook}.

\subsection{Number of Transformer Layers \texorpdfstring{$n_l$}{}}
\label{s:evaluation:hyperparams:layers}

Since we use an encoder-only Transformer architecture, we can vary the number of Transformer layers $n_l$ easily. We train and evaluate four model architectures with $n_l \in \left\{ 4, 6, 8, 10 \right\}$. The rest of the architecture remains unchanged in this experiment. We denote $n_l$ with a suffix \texttt{-t<transformer\,layer>} in the model naming scheme. Therefore, the baseline model to compare against is labeled as 4-x-1-t6.

\begin{figure}[htb]
    \vspace*{2mm}
    \centering\captionsetup{width=0.95\textwidth}%
    \includegraphics[width=\textwidth]{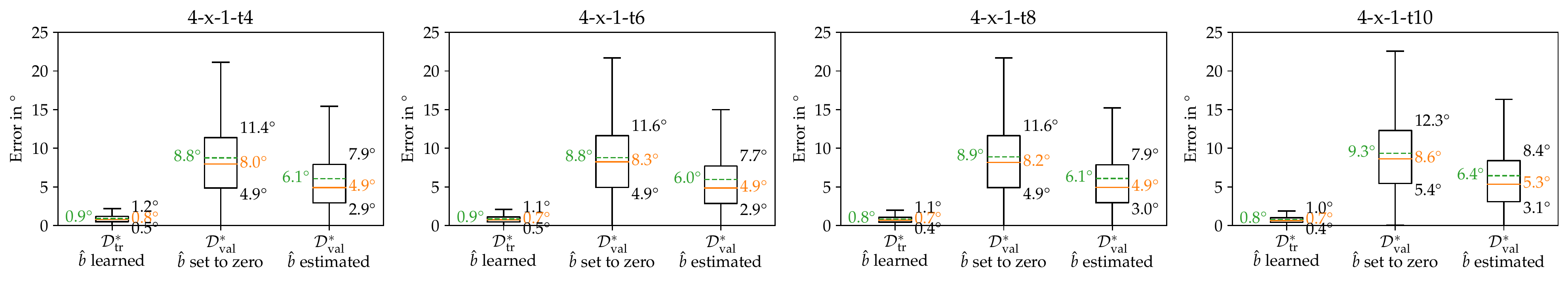}%
    \caption{Box-plots of the results of our RGBDTr model with different numbers of Transformer layers $n_l$.}%
    \label{f:evaluation:hyperparams:layers:results}%
\end{figure}

In \Cref{f:evaluation:hyperparams:layers:results}, we show the results of these four models. When comparing the performance on the validation subsets using calibration, we found our baseline model performing best whereas the model 4-x-1-t10 performs worst with a mean angular error of $\bar{e} = \ang{6.4}$. Both 4-x-1-t4 and 4-x-1-t8 achieved a mean angular error of $\bar{e} = \ang{6.1}$, which is \ang{0.1} worse than the baseline.

However, the differences between the models are small, so that we can conclude that the number of Transformer layers does not have a significant impact on the performance of our RGBDTr model. A smaller $n_l$ leads to a faster training process and a smaller model size, so that we use the best-performing configuration $n_l = 6$ in the following experiments as a balanced middle ground.

\subsection{Transformer Token Size \texorpdfstring{$d_\mathrm{model}$}{}}
\label{s:evaluation:hyperparams:dim}

For all experiments so far the token size of our model has been set to $d_\mathrm{model} = 1024$. In this section, we train another two models with $d_\mathrm{model}$ set to 512 and 2048, respectively. We denote $d_\mathrm{model}$ with a suffix \texttt{-d<token\,size>} in the model naming scheme with 4-x-1-d1024 being the baseline model to compare against.

\begin{figure}[htb]
    \centering\captionsetup{width=0.95\textwidth}%
    \includegraphics[width=\textwidth]{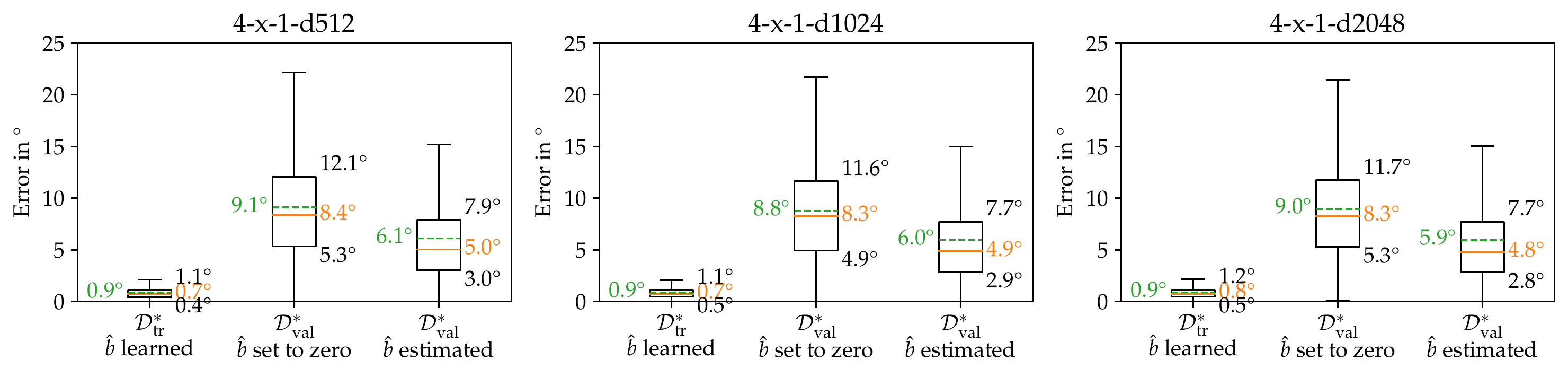}%
    \caption{Box-plots of the results of our RGBDTr model with different Transformer token sizes~$d_\mathrm{model}$.}%
    \label{f:evaluation:hyperparams:dim:results}%
\end{figure}

Although the three models perform similarly on the validation subsets using calibration, there is a slight trend towards a larger token size. We show the results in \Cref{f:evaluation:hyperparams:dim:results}. In fact, the model 4-x-1-d2048 performs best with a mean angular error of $\bar{e} = \ang{5.9}$, which is \ang{0.1} better than the baseline. The model 4-x-1-d512 performs worst with a mean angular error of $\bar{e} = \ang{6.1}$, which is \ang{0.1} worse than the baseline. The same trend is also visible when considering the mean angular errors where the baseline model achieves \ang{4.9} and the other two achieve \ang{0.1} better or worse, respectively.

Since the difference in mean angular error is small, we can conclude that the token size does not have a great impact on model performance. We will still use our default configuration of $d_\mathrm{model} = 1024$ in the following experiments as it allows for a faster training process and smaller model size while still achieving good performance. We did not conduct experiments with a larger token size since both \citeauthor*{Vaswani2017}~\cite{Vaswani2017} and \citeauthor*{Dosovitskiy2020}~\cite{Dosovitskiy2020} used at most $d_\mathrm{model} = 1024$ in their experiments. Therefore setting $d_\mathrm{model} = 2048$ is already a major increase in token size. However, future experiments could investigate the impact of even larger token sizes and could investigate the effect of the $d_\mathrm{ff}$ hyperparameter.

\subsection{Number of Encoder Layers \texorpdfstring{$n_e$}{} and Fusion Block}
\label{s:evaluation:hyperparams:enc}

\begin{figure}[htb]
    \centering\captionsetup{width=0.95\textwidth}%
    \includegraphics[width=\textwidth]{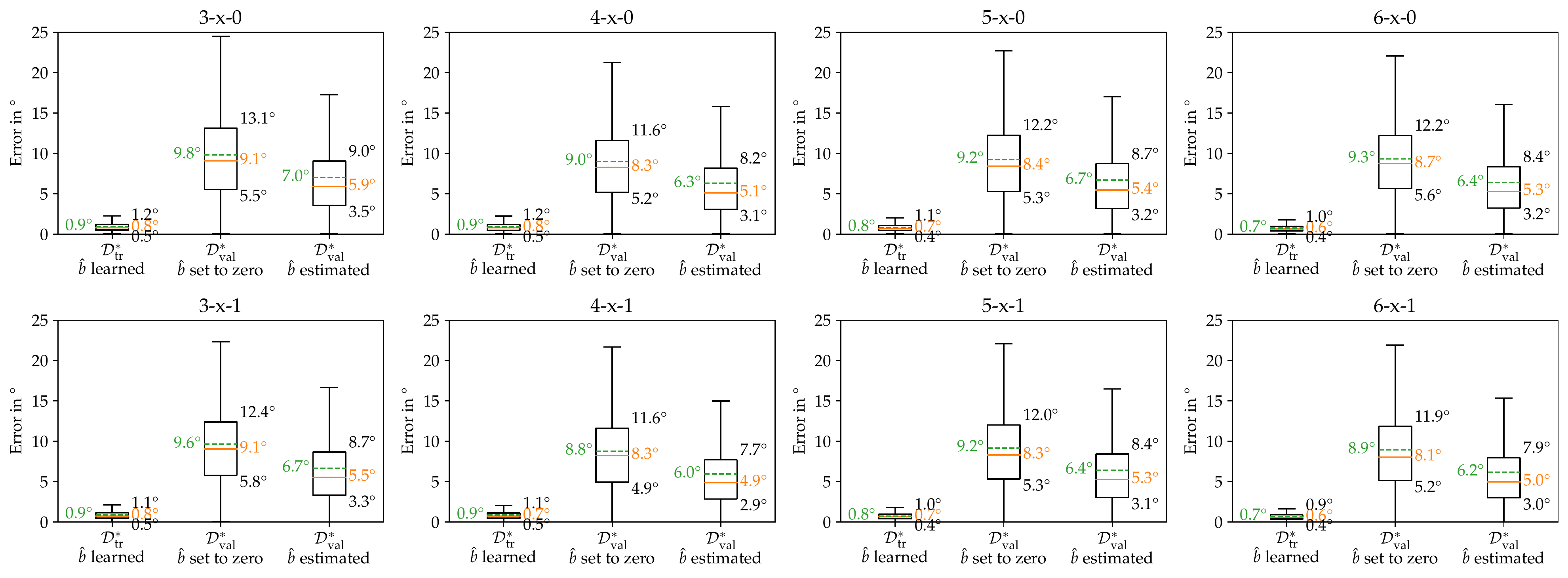}%
    \caption{Box-plots of the results of our RGBDTr model with different numbers of encoder layers~$n_e$ with fusion block (bottom row) and without (top row).}%
    \label{f:evaluation:hyperparams:enc:results}%
\end{figure}

For our next set of experiments, we vary the number of encoder layers $n_e$ in the two generator encoders. We train and evaluate two model architectures per $n_e \in \left\{3, 4, 5, 6 \right\}$, one with and one without a fusion block. In total, we trained eight model architectures and therefore 40 models due to the $k=5$ cross-validation splits. We aggregate the results in \Cref{f:evaluation:hyperparams:enc:results}. The top and bottom rows show the results of the models without and with a fusion block, respectively. Each column corresponds to a different $n_e$.

We found that our baseline model 4-x-1 performs best on the validation subsets using calibration. The model 3-x-0 performs worst in this experiment with a mean angular error of $\bar{e} = \ang{7.0}$, which is \ang{1.0} worse than the baseline. We found that removing the fusion block from our model architecture results in worse performance for all $n_e$. The mean angular error increases by \qtyrange{0.2}{0.3}{\degree} compared to their counterpart models that feature a fusion block.

Interestingly, there is no clear trend towards more or less encoder layers. The model 6-x-1 achieves a mean angular error of $\bar{e} = 6.2$ which is \ang{0.2} better than 5-x-1 and simultaneously \ang{0.2} worse than the baseline. The models with $n_e = 5$ definitely perform worse than their direct siblings regardless of the existence of a fusion block. However, having too few encoder layers hurts the performance as well, making our 3-x-0 and 3-x-1 models the worst performing models in this experiment.

We can conclude that $n_e = 4$ and the existence of a fusion block lead to the best performance in this experiment. These results validate our prior choice of using a four layer PatchGAN backbone architecture. Furthermore, using this configuration this configuration throughout \Cref{s:evaluation:oge,s:evaluation:hyperparams} as a baseline is justified.

\clearpage
\subsection{Summary of Hyperparameter Tuning}
\label{s:evaluation:hyperparams:summary}

To summarize our findings during the hyperparameter optimization, we list the mean angular errors on the validation subsets for each trained model in \Cref{t:evaluation:hyperparams:summary}. We categorized the results of each experiment and colored the respective best model in green, the respective worst model in red, and all other models in yellow.

\begin{table}[htbp]
    \centering
    \renewcommand{\arraystretch}{1.3}%
    \newcommand{\bfour}[2][black]{\Block[fill=#1]{1-3}{#2}}
\newcommand{\bthree}[2][black]{\Block[fill=#1]{1-4}{#2}}
\begin{NiceTabular}{>{\centering\bfseries\arraybackslash}p{5cm}|*{12}{@{}p{0.85cm}@{}}}
    \Block{2-1}{Experiment} & \multicolumn{12}{c}{\textbf{Model Configuration}} \\
                                & \multicolumn{12}{c}{\textbf{Mean Angular Error $\bar{e}$ on $\mathcal{D}_\mathrm{val}^{*}$}} \\
    \hline\hline

    \Block{2-1}{Transformer Layers} & \bfour[okay]{$n_l = 4$} & & & \bfour[good]{$n_l = 6$} & & & \bfour[okay]{$n_l = 8$} & & & \bfour[bad]{$n_l = 10$} & & \\
                                    & \bfour[okay]{\ang{6.1}} & & & \bfour[good]{\ang{6.0}} & & & \bfour[okay]{\ang{6.1}} & & & \bfour[bad]{\ang{6.4}} & & \\
    \hline
    \Block{2-1}{Transformer Token Size} & \bthree[bad]{$d_\mathrm{model} = 512$} & & & & \bthree[okay]{$d_\mathrm{model} = 1024$} & & & & \bthree[good]{$d_\mathrm{model} = 2048$} & & & \\
                                        & \bthree[bad]{\ang{6.1}} & & & & \bthree[okay]{\ang{6.0}} & & & & \bthree[good]{\ang{5.9}} & & & \\
    \hline
    \Block{2-1}{\centering Encoder Layers without Fusion Block $\left(n_f = 0\right)$} & \bfour[bad]{$n_e = 3$} & & & \bfour[good]{$n_e = 4$} & & & \bfour[okay]{$n_e = 5$} & & & \bfour[okay]{$n_e = 6$} & & \\
    & \bfour[bad]{\ang{7.0}} & & & \bfour[good]{\ang{6.3}} & & & \bfour[okay]{\ang{6.7}} & & & \bfour[okay]{\ang{6.4}} & & \\
    \cdottedline{1-13}
    \Block{2-1}{\centering Encoder Layers with Fusion Block $\left(n_f = 2\right)$} & \bfour[bad]{$n_e = 3$} & & & \bfour[good]{$n_e = 4$} & & & \bfour[okay]{$n_e = 5$} & & & \bfour[okay]{$n_e = 6$} & & \\
    & \bfour[bad]{\ang{6.7}} & & & \bfour[good]{\ang{6.0}} & & & \bfour[okay]{\ang{6.4}} & & & \bfour[okay]{\ang{6.2}} & & \\\hline
\end{NiceTabular}%
    \caption{Hyperparameter optimization results.}
    \label{t:evaluation:hyperparams:summary}
\end{table}

In the first experiment, we found that the number of Transformer layers does not have a significant impact on model performance. However, using our default configuration of $n_l = 6$ resulted in the best performance. The second experiment showed that increasing token sizes lead to better performance. However, the difference in mean angular error is small between the three tested configurations. Our model with $d_\mathrm{model} = 2048$ achieved the best performance. In the third experiment, we found that the number of encoder layers significantly affects the performance of our RGBDTr model. We found that $n_e = 4$ and the existence of a fusion block lead to the lowest mean angular error. This validated our choice of using a four layer encoder backbone in our baseline model.

\subsection{Evaluation of the Best Model Configuration on the Test Sets}
\label{s:evaluation:hyperparams:test}
\begin{figure}[htb]
    \centering\captionsetup{width=0.95\textwidth}%
    \includegraphics[width=\textwidth]{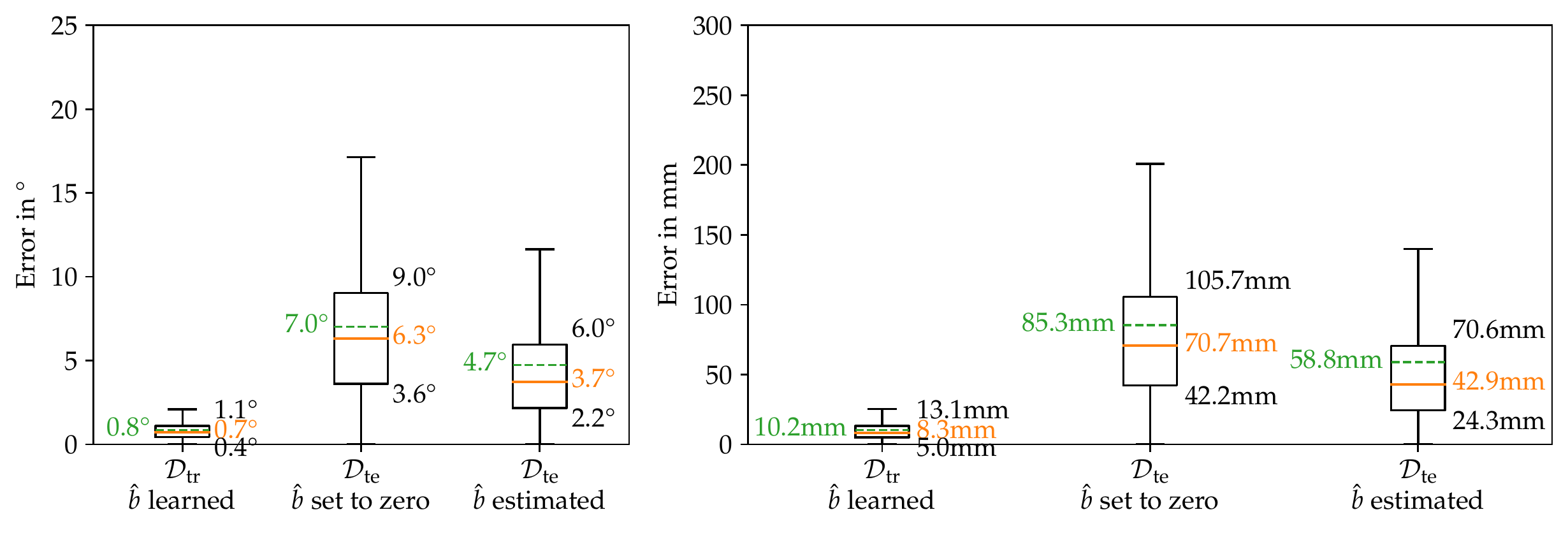}%
    \caption{Box-plots of the results of our best RGBDTr model configuration trained on our \oge\ dataset.}%
    \label{f:evaluation:hyperparams:test:results-oge}%
\end{figure}

Selecting the respective best hyperparameters yields the following model configuration:
\[ n_l = 6 \qquad d_\mathrm{model} = 2048 \qquad n_f = 2 \qquad n_e = 4 \]
All other model and training parameters are set to their default value as listed in \Cref{t:appendix:default-config}. To evaluate our best RGBDTr model, we train it on the train set $\mathcal{D}_\mathrm{tr}$ of our \oge\ dataset and evaluate it on the test set $\mathcal{D}_\mathrm{te}$. We then use $n_\mathrm{cal} = 200$ randomly chosen samples per subject of the test set to estimate the subject-specific bias terms $\hat{b}$. The results are depicted in \Cref{f:evaluation:hyperparams:test:results-oge}. Our RGBDTr model achieves a mean angular error of $\bar{e} = \ang{4.7}$ on the test set and a mean Euclidean error of $\bar{d} = \qty{58.8}{\milli\meter}$.

\clearpage

Although the numbers are not directly comparable to our previous experiments on the validation subsets, we can compare the gaps between the mean errors on the training set and the test set. We found that the difference in mean angular errors is \ang{1.2} lower in \Cref{f:evaluation:hyperparams:test:results-oge} than for our baseline model in \Cref{f:evaluation:oge:gan:results:ang}. Likewise the difference in mean Euclidean errors is \qty{11.8}{\milli\meter} lower (see~\Cref{f:evaluation:oge:gan:results:distance}). This supports our prior findings that more subjects in the training (sub-)set lead to better generalization performance.

\begin{figure}[htb]
    \centering\captionsetup{width=0.95\textwidth}%
    \includegraphics[width=\textwidth]{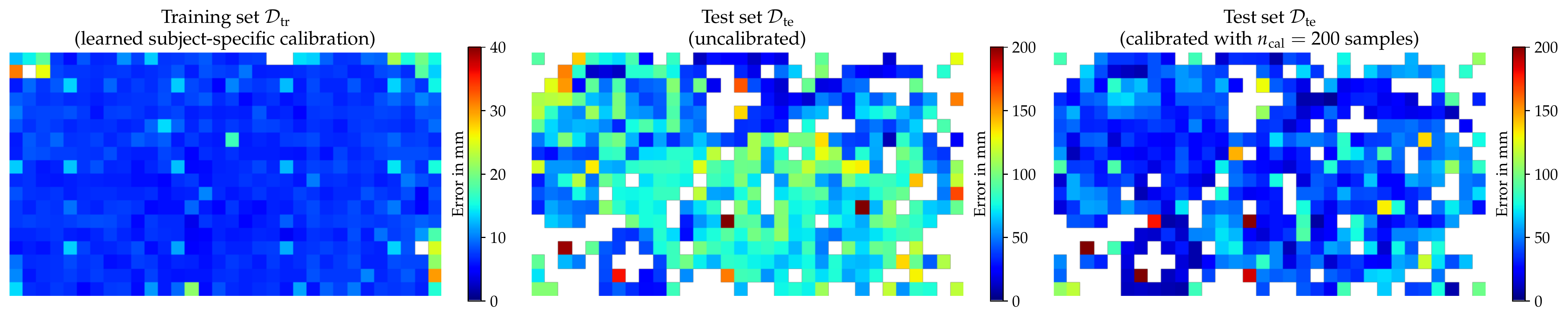}%
    \caption{Distribution of the on-screen Euclidean errors of our best RGBDTr model configuration trained on our \oge\ dataset.}%
    \label{f:evaluation:hyperparams:test:error-hist-oge}%
\end{figure}

We also visualize the error of our best RGBDTr model on the training and test sets in \Cref{f:evaluation:hyperparams:test:error-hist-oge}. This allows for a direct comparison with our baseline model that was trained on the training subsets and evaluated on the validation subsets (see~\Cref{s:evaluation:oge:error-vis} and \Cref{f:evaluation:oge:error-vis:hist-agg}). The test set contains significantly less samples than the training set, which is why the error distributions in \Cref{f:evaluation:hyperparams:test:error-hist-oge} contain more blank spots. However, we can still see general patterns. When applying our model uncalibrated, the worst spots are in the center and in the lower half of the screen. This is different from the error distribution of our baseline model, which has the worst spots in the left third and the upper right corner. After estimating the subject-specific bias terms and applying them to the gaze predictions, the error distribution changes significantly. The worst spots are now in the corner regions, which is similar to the error distribution of our baseline model. However, the error distribution of our best RGBDTr model has more bad spots in the lower half of the screen, but the overall error is lower than for our baseline model as we saw in \Cref{f:evaluation:hyperparams:test:results-oge}.

In order to evaluate our model performance on another dataset, we trained our best RGBDTr model on the train set $\mathcal{X}_\mathrm{tr}$ of the \xgaze\ dataset. We then used the $n_\mathrm{cal} = 200$ calibration samples provided by the authors to estimate the subject-specific bias terms $\hat{b}$ and evaluated our model accordingly. Next, we uploaded our predictions to the \xgaze\ evaluation server\,\footnote{~\url{https://codalab.lisn.upsaclay.fr/competitions/7423} on CodaLab~\cite{Pavao2022}} and retrieved a final mean angular error of $\bar{e} = \ang{3.59}$, which is \ang{0.2} better than the models we trained on the training subsets and evaluated on the validation subsets in \Cref{s:evaluation:xgaze:weight-decay}. However, this is still \ang{1.55} worse than the results reported by \citeauthor*{Zhang2020} on their calibrated model on the same evaluation set~\cite{Zhang2020}. This means that our model causes an additional error of approximately \qty{1.6}{\centi\meter}, given an average eye-to-screen distance of \qty{60}{\centi\meter}.

However, the approach of \citeauthor*{Zhang2020} differs in two main ways: the model architecture itself and the calibration method. The authors of \xgaze\ used a simple pre-trained ResNet-50 model and trained it on the task of \ges~\cite{Zhang2020}. Furthermore, they used a different calibration method by fine-tuning the whole model on the calibration set~\cite{Zhang2020}. Further research is needed to investigate the individual impact of these two differences on the performance of the models.

\begin{figure}[htb]
    \vspace*{-2mm}%
    \centering\captionsetup{width=0.95\textwidth}%
    \includegraphics[width=0.5\textwidth]{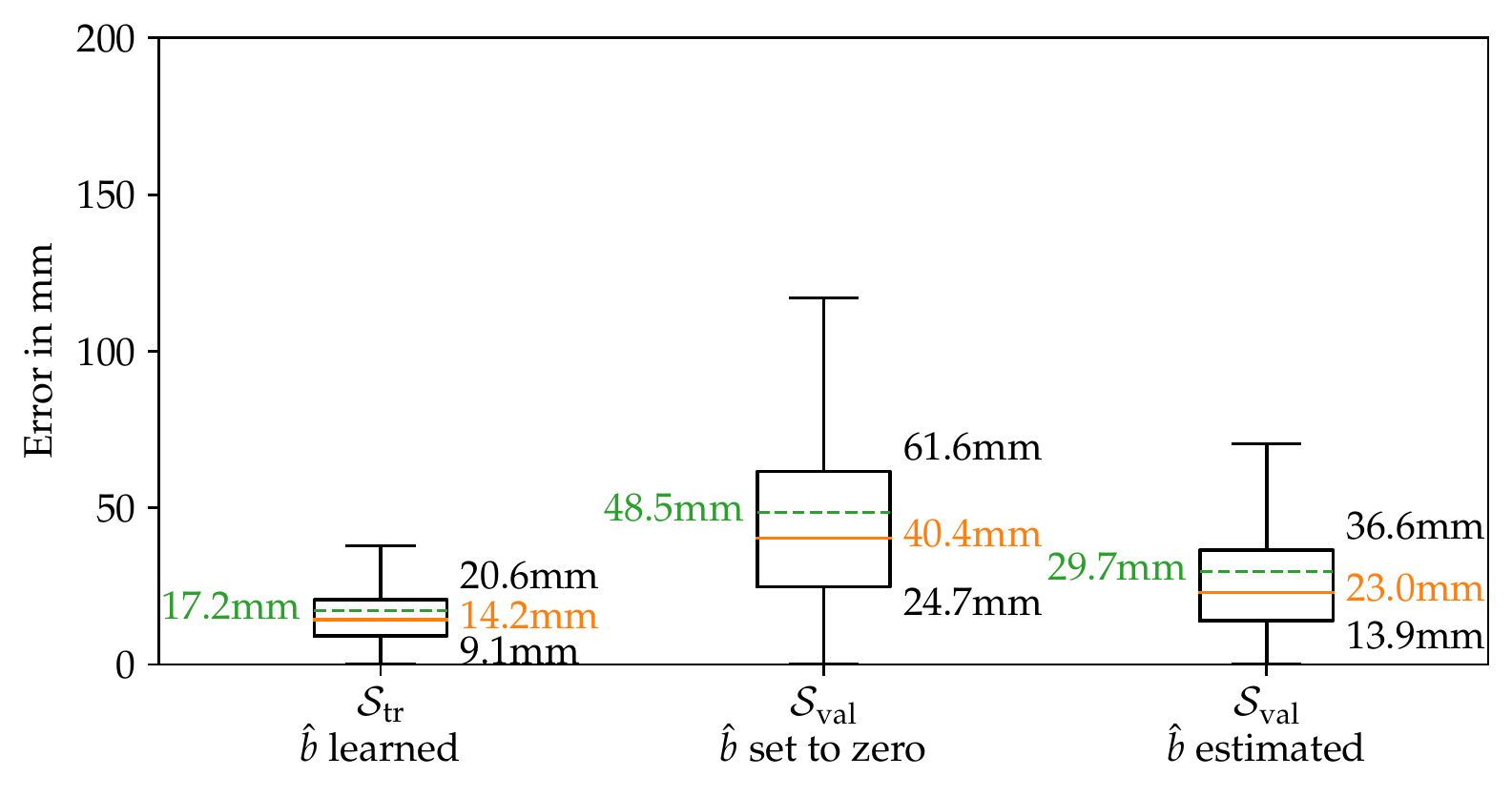}%
    \caption{Box-plots of the results of our best RGBDTr model configuration trained on the \stg\ dataset.}%
    \label{f:evaluation:hyperparams:test:results-stg}%
\end{figure}

We also trained another model on the \stg\ dataset using the model configuration described above. In \Cref{f:evaluation:hyperparams:test:results-stg}, we visualize its results as box-plots. Our model achieves a mean Euclidean error of $\bar{d} = \qty{29.7}{\milli\meter}$, which is \qty{0.4}{\milli\meter} better than our previous best model in \Cref{s:evaluation:stg}. Our results are also \qty{9.0}{\milli\meter} better than the results reported by \citeauthor*{Lian2019}~\cite{Lian2019} and \qty{2.6}{\milli\meter} better than the results reported by \citeauthor*{Zhang2020a}~\cite{Zhang2020a}. To our best knowledge, our model achieves the best performance on the \stg\ dataset to date.

\subsection{Outlook: Other Hyperparameters to Tune}
\label{s:evaluation:hyperparams:outlook}

We already listed all hyperparameters of our RGBDTr model alongside their default values in \Cref{t:appendix:default-config}. In this subsection, we give an outlook on some of these hyperparameters that are worth investigating in future work. Due to time constraints, we could only optimize some of them. Therefore, we focused on some of the hyperparameters of the Fusion Transformer as well as the generator encoders and optimized them in this section. We also give an outlook on possible changes to the training process.
\clearpage 

\textbf{Already optimized hyperparameters.} In this and the previous sections, we tuned some of the hyperparameters. We found in \Cref{s:evaluation:stg:discriminator} that the architecture for the \ac{GAN} should be a four layer PatchGAN. Next, we conducted multiple experiments on the effectiveness of a pre-trained \ac{GAN} backbone and found that it hurts the performance of our RGBDTr model (see~\Cref{s:evaluation:stg:gan,s:evaluation:oge:gan}). Furthermore, we found that adding the depth map as input to our model improves its performance (see~\Cref{s:evaluation:stg:rgbd,s:evaluation:oge:rgbd}). In addition to that, we found that a model with $n_l = 6$ Transformer layers, a token size of $d_\mathrm{model} = 2048$, $n_f = 2$ fusion block layers, and $n_e = 4$ encoder layers performs best on our \oge\ dataset (see~\Cref{s:evaluation:hyperparams:layers,s:evaluation:hyperparams:dim,s:evaluation:hyperparams:enc}).

\textbf{Possibly impactful hyperparameters.} The backbone architecture of the eye pose feature extraction module $t_\mathrm{ea}$ was fixed to a pre-trained ResNet-18 for all of our experiments. Using larger backbones, such as ResNet-34, or using different backbone architectures, such as EfficientNet, open up the possibility to improve model performance. Furthermore, by tweaking the hyperparameters $t_\mathrm{ts}$ and $f$ to enable training the eye pose extraction module from scratch, one can investigate the impact of the backbone architecture on the performance of our RGBDTr model. Although we already found that a fusion block improves model performance, we did not conduct any experiments with $c_f > 2$ or change the fusion block architecture $t_\mathrm{fus}$. Similarly, the Transformer encoder architecture $t_\mathrm{tr}$ was always set to $\mathrm{B2T}$ in our experiments. However, there are two other options possible as described in \Cref{s:processing:models:transformer}. In addition to that, one could switch from the learned positional encoding to a sinusodial one via the $t_\mathrm{tpe}$ hyperparameter and change the number of self-attention heads $n_h$. The depth extraction module could also be optimized using the $r$ hyperparameter. Finally, the number of channels of our \acp{CNN} could be optimized, too.

\textbf{Changes in the training process.} Another set of hyperparameters potentially worth optimizing are the training parameters. We used a batch size dependent on the dataset and training phase, ranging from $b=32$ to $b=64$. We chose the batch sizes in order to be able to fit the training process onto a single NVIDIA~A100~GPU with \SI{40}{\giga\byte} of memory. The number of epochs trained could also be adjusted for each training phase to improve the resulting model. We did not investigate the effect of different learning rates and learning rate schedulers. For example, the learning rate and its decay during the RGBDTr training was adapted from \citeauthor*{Zhang2020}~\cite{Zhang2020}. The amount of dropout within the Transformer and the weighting of the loss terms could also be optimized. Furthermore, one could investigate the effect of different optimizers and their respective hyperparameters. We used the Adam optimizer~\cite{Kingma2014} with $\beta_1 = 0.9$ and $\beta_2 = 0.999$ for all experiments. Finally, to improve the multi-task learning, one could modify the training process in a way that no pre-trained \ac{GAN} backbone is used, but instead only the fine-tuning phase gets applied with a larger number of epochs.

However, one has to balance the time and effort put into optimizing the hyperparameters with the resulting performance gain. In this section, we found some hyperparameters to have more impact than others. In the next section, we conduct experiments on the effect of removing modules from our model architecture to investigate their impact on the performance.

\clearpage
\addtocontents{toc}{\protect\newpage} 
\section{Ablation Study}
\label{s:evaluation:ablation}

Our model architecture consists of multiple modules as described in \Cref{s:processing:models} and visualized in \Cref{f:processing:models:overview}. In order to determine the impact of these modules on the final gaze prediction performance, we remove them -- one at a time -- from our model architecture and train the resulting model on the train set $\mathcal{D}_\mathrm{tr}$ of our \oge\ dataset. We then evaluate the models on the test set $\mathcal{D}_\mathrm{te}$ to compare their performance to our baseline model 4-x-1. We use $n_\mathrm{cal} = 200$ randomly chosen samples per subject of the test set to estimate the subject-specific bias terms $\hat{b}$. These samples are identical for each model to ensure comparability. We denote the resulting models with a suffix \texttt{-no-<module>} in the model naming scheme.

\subsection{Removal of Modules and Architecture Changes}
\label{s:evaluation:ablation:modules}
We refer to the module colors as in \Cref{f:processing:models:overview}. Some modules can be removed easily from our model architecture, e.g., the depth extraction module (\textcolor{magenta!80!black}{magenta}) and the eye pose feature extractors (\textcolor{cyan!80!black}{cyan}). However, removing the head pose feature extractor (\textcolor{orange!80!black}{orange}) leads to the removal of both generator encoders (\textcolor{green!80!black}{green} and \textcolor{blue!80!black}{blue}) and the fusion block (\textcolor{blk80}{gray}) since we do not use depth reconstruction, i.e., no pre-trained \ac{GAN} backbone. The aforementioned modules can be removed completely and only reduce the number of tokens $n_t$ by two (eye pose) or one (all others).

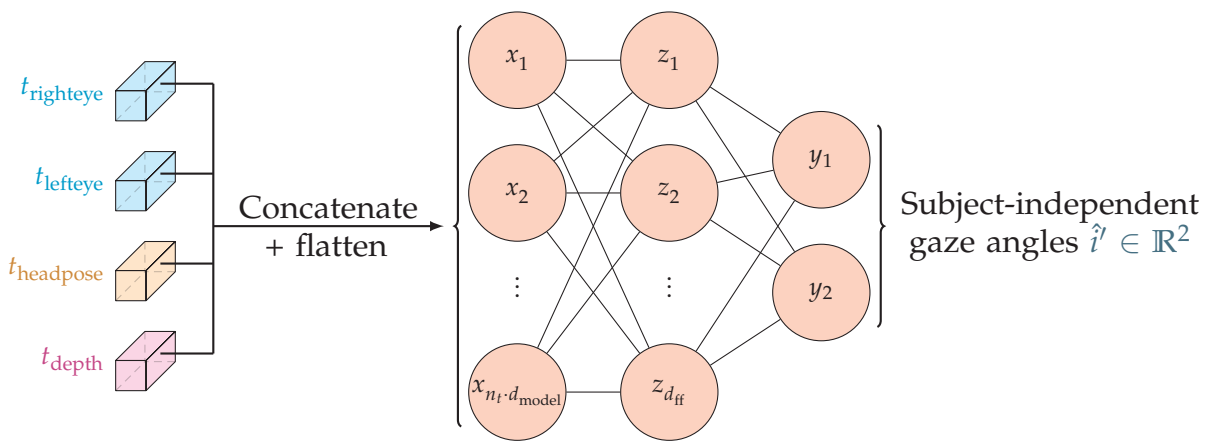
\begin{figure}[htb]
    \centering\captionsetup{width=0.95\textwidth}%
    \begin{tikzpicture}
    \tikzmath{\scale=1.0;}
    \tikzset{every node/.style={scale=\scale}}

    \pic [pic=tre, fill=cya20] at (0, 0) {cuboid={width=4, height=4, depth=10}};
    \pic [pic=tle, fill=cya20] [below=\scale * 8mm of tre] {cuboid={width=4, height=4, depth=10}};
    \pic [pic=thp, fill=ora20] [below=\scale * 8mm of tle] {cuboid={width=4, height=4, depth=10}};
    \pic [pic=td, fill=mag20] [below=\scale * 8mm of thp] {cuboid={width=4, height=4, depth=10}};

    \node [left=0cm of tre, align=right, text=cyan!80!black] {\footnotesize $t_\mathrm{righteye}$};
    \node [left=0cm of tle, align=right, text=cyan!80!black] {\footnotesize $t_\mathrm{lefteye}$};
    \node [left=0cm of thp, align=right, text=orange!80!black] {\footnotesize $t_\mathrm{headpose}$};
    \node [left=0cm of td, align=right, text=magenta!80!black] {\footnotesize $t_\mathrm{depth}$};

    \coordinate (tconcat1) at ($(tre.east) + (\scale * 5mm, \scale * 1mm)$);
    \coordinate (tconcat2) at ($(td.east) + (\scale * 5mm, \scale * 1mm)$);

    \draw [thick] (tconcat1) -- (tconcat2);
    \foreach \t in {tre, tle, thp, td} {
        \coordinate (temp) at ($(\t.center)!0.5!(\t.east) + (0, \scale * 1mm)$);
        \draw [thick] (temp) -- (temp -| tconcat1);
    }

    \pic [pic=mlp, fill=red20, draw=black] (mlp) at ($(tre.east)!0.5!(td.east)$) [xshift=\scale * 4.5cm] {fc={scale=0.8, circlescale=1.6, vscale=2.2, hscale=2.5, nodedef={{0, "x", "-", "n_t \cdot d_\mathrm{model}"}, {0, "z", "-", "d_\mathrm{ff}"}, {2, "y"}}}};

    \draw [decoration={brace, raise=\scale * 8pt}, decorate, thick] (mlp-node-1-3.south west) -- (mlp-node-1-1.north west) node [midway, left=\scale * 2mm, text width=0cm, align=center] (tconcat3) {};

    \draw [arrowr] (tconcat3) -- (tconcat3 -| tconcat1) node [midway, text width=\scale * 3cm, align=center] {Concatenate + flatten};

    \draw [decoration={brace, raise=\scale * 8pt}, decorate, thick] (mlp-node-3-1.north east) -- (mlp-node-3-2.south east) node [midway, right=\scale * 4mm, text width=\scale * 4cm, align=center] {Subject-independent gaze angles \textcolor{teal!80!black}{$\hat{i}' \in \mathbb{R}^2$}};
\end{tikzpicture}%
    \caption{Architecture of the \ac{MLP} to replace the Fusion Transformer during ablation study.}%
    \label{f:evaluation:ablation:modules:new-fusion}%
\end{figure}

However, we also want to ablate our Fusion Transformer (\textcolor{red!80!black}{red}). To do so, we replace the entire Transformer with an \ac{MLP} depicted in \Cref{f:evaluation:ablation:modules:new-fusion}. Furthermore, we remove the learnable class token as well as the positional encoding when removing the Fusion Transformer module. Since the replacement \ac{MLP} outputs two values, we can also remove the gaze estimator network (\textcolor{teal!80!black}{teal}). The \ac{MLP} consists of an input layer with $n_t \cdot d_\mathrm{model}$ neurons, a single hidden layer with $d_\mathrm{ff}$ neurons, and an output layer with two neurons predicting the subject-independent gaze angels. We also employ dropout before each layer in the \ac{MLP} and apply layer normalization after the first and second layers. We do not change the calibration step.

\subsection{Results on our \oge\ Dataset}
\label{s:evaluation:ablation:oge}
In total, we have trained four models with different modules removed and two other models to compare against: our baseline model and an RGB-only baseline model. We show the results of these six models trained on our \oge\ dataset in \Cref{t:evaluation:ablation:oge:results}.

\begin{table}[htbp]
    \vspace*{-2mm} 
    \centering\captionsetup{width=0.95\textwidth}
    \renewcommand{\arraystretch}{1.3}%
    \begin{NiceTabular}{l|>{\centering\arraybackslash}p{2cm}:>{\centering\arraybackslash}p{2cm}|>{\centering\arraybackslash}p{2cm}:>{\centering\arraybackslash}p{2cm}}
    \Block{2-1}{\textbf{Model}} & \Block{1-2}{\textbf{Mean Angular Error $\bar{e}$}} & & \Block{1-2}{\textbf{Mean Euclidean Error $\bar{d}$}} & \\
    & Abs~\textsuperscript{a} & Rel~\textsuperscript{b} & Abs~\textsuperscript{a} & Rel~\textsuperscript{b} \\\hline

    baseline & \ang{4.84} & -- & \qty{60.2}{\milli\meter} & -- \\
    \rowcolor{cyan!10!white} baseline-no-depth & \ang{4.80} & \ang{-0.04} & \qty{59.6}{\milli\meter} & \qty{-0.6}{\milli\meter} \\
    \rowcolor{cyan!10!white} baseline-no-eye & \ang{4.88} & \ang{0.04} & \qty{60.8}{\milli\meter} & \qty{0.6}{\milli\meter} \\
    baseline-no-headpose & \ang{9.46} & \ang{4.62} & \qty{113.3}{\milli\meter} & \qty{53.1}{\milli\meter} \\
    \rowcolor{blue!10!white} baseline-no-transformer & \ang{4.71} & \ang{-0.13} & \qty{58.6}{\milli\meter} & \qty{-1.6}{\milli\meter} \\
    baseline-rgb & \ang{6.29} & \ang{1.45} & \qty{77.5}{\milli\meter} & \qty{17.3}{\milli\meter} \\\hline
    \Block[l]{1-5}{
        \small\textsuperscript{a} Absolute error (lower is better) \\
        \small\textsuperscript{b} Relative error to baseline model (lower is better)
    } & & & & \\
\end{NiceTabular}%
    \caption{Results of the ablation study on our \oge\ dataset.}%
    \label{t:evaluation:ablation:oge:results}
\end{table}

The results are both interesting and somewhat unexpected. Removing the depth extraction or the eye pose feature extraction modules (\textcolor{cyan!80!black}{cyan} rows) does not have a significant impact on the performance of our RGBDTr model. The mean angular error is \ang{0.04} lower or higher than the baseline, respectively, which could be due to initialization differences. This is also true for the mean Euclidean error.

Removing the headpose feature extraction module leads to a massive drop in performance and almost doubles the mean angular and Euclidean errors. This indicates that this module, which takes both RGB and depth face patches as input, is crucial for our RGBDTr model. Furthermore, the RGB-only baseline model performs significantly worse than our baseline model. The mean angular error increases by \ang{1.45}. This is expected, because the RGB-only baseline model does not have access to the depth map and therefore cannot use the additional information contained in it. We have already seen in \Cref{s:evaluation:stg:rgbd,s:evaluation:oge:rgbd} that using the depth map as input improves the performance of our RGBDTr model.

However, removing the Fusion Transformer (\textcolor{blue!80!black}{blue} row) and replacing it with an \ac{MLP} as described in \Cref{s:evaluation:ablation:modules} improves the performance of our model not insignificantly. The mean angular error decreases by \ang{0.13} and the mean Euclidean error decreases by \qty{1.6}{\milli\meter}. This is surprising, because the \ac{MLP} has a much smaller number of parameters and no attention or self-attention mechanism. However, the \ac{MLP} substitute was able to learn the desired mapping function better than the original Transformer. This indicates that using the Transformer architecture in our way is not beneficial for the task of \ges. This might be different for other hyperparameters, e.g., when using no pre-trained eye pose backbones. However, additional research is needed to substantiate or disprove this conjecture.

\subsection{Results on the \stg\ Dataset}
\label{s:evaluation:ablation:stg}
We also conducted an ablation study on the \stg\ dataset using the same model configurations and module removals as in \Cref{s:evaluation:ablation:oge}. We show the results of these six models in \Cref{t:evaluation:ablation:stg:results}.

\begin{table}[htbp]
    \centering\captionsetup{width=0.95\textwidth}
    \renewcommand{\arraystretch}{1.3}%
    \begin{NiceTabular}{l|>{\centering\arraybackslash}p{2cm}:>{\centering\arraybackslash}p{2cm}}
    \Block{2-1}{\textbf{Model}} & \Block{1-2}{\textbf{Mean Euclidean Error $\bar{d}$}} & \\
    & Abs~\textsuperscript{a} & Rel~\textsuperscript{b} \\\hline

    baseline                & \qty{30.1}{\milli\meter}  & -- \\
    baseline-no-depth       & \qty{31.5}{\milli\meter}  & \qty{1.4}{\milli\meter} \\
    baseline-no-eye         & \qty{30.2}{\milli\meter}  & \qty{0.1}{\milli\meter} \\
    baseline-no-headpose    & \qty{119.2}{\milli\meter} & \qty{89.1}{\milli\meter} \\
    baseline-no-transformer & \qty{26.9}{\milli\meter}  & \qty{-3.2}{\milli\meter} \\
    baseline-rgb            & \qty{31.8}{\milli\meter}  & \qty{1.7}{\milli\meter} \\\hline
    \Block[l]{1-3}{
        \small\textsuperscript{a} Absolute error (lower is better) \\
        \small\textsuperscript{b} Relative error to baseline model (lower is better)
    } & & \\
\end{NiceTabular}%
    \caption{Results of the ablation study on the \stg\ dataset.}%
    \label{t:evaluation:ablation:stg:results}
\end{table}

The results are similar to the previous section. However, the differences in performance are more pronounced, especially when considering percentage differences. Removing the depth extraction module leads now to a significantly worse result. This can be explained by the task our model has to perform on the \stg\ dataset as it predicts gaze points rather than gaze angles. In this scenario, the depth data from the depth extraction module provides more valuable information than when predicting gaze angles in the normalized camera space. Removing the eye pose feature extraction module increases the error less on the \stg\ dataset.

As we saw in the previous section, removing the headpose feature extraction module degrades model performance massively. On the \stg\ dataset, it quadruples the mean Euclidean error, which is probably a combination of the limited information available through the other modules and the task of Gaze Point Estimation. However, future research is needed to investigate in the reasons behind this extraordinary performance drop. The RGB-only baseline model performs only slightly worse than our baseline model with a substantially lower mean Euclidean error gap than on our \oge\ dataset. This indicates that the depth maps in our \oge\ dataset are more valuable for the task of \ges\ than the depth maps in the \stg\ dataset.

Replacing the Fusion Transformer with an \ac{MLP} improves the performance of our model on this dataset as well. The mean Euclidean error decreases by \qty{3.2}{\milli\meter}, which is both absolute and relative to the baseline model a larger improvement than on our \oge\ dataset. This indicates that the Transformer architecture used in our model is also not beneficial for the task of Gaze Point Estimation. As stated above, this might be different for other hyperparameters.

\subsection{Effect of the Feature Fusion Transformer on \xgaze}
\label{s:evaluation:ablation:xgaze}
{\spaceskip=3.4pt plus 1pt minus 1.5pt 
Since the \xgaze\ dataset contains only RGB images, we cannot evaluate the effect of the depth extraction module on this dataset. Furthermore, we concluded in experiments on the other two datasets that the headpose feature extraction module is key to the performance of our RGBDTr model. Similarly, we saw few to no effect when ablating the eye pose feature extraction module. Therefore, we focus on the effect of the Fusion Transformer on the performance of our RGBDTr model on the \xgaze\ dataset. We trained two models in total, a baseline model with the optimized hyperparameter configuration and a model where the Fusion Transformer was replaced by an \ac{MLP} as described in \Cref{s:evaluation:ablation:modules}. We show the results of these two models in \Cref{t:evaluation:ablation:xgaze:results}.
}

\begin{table}[htbp]
    \centering\captionsetup{width=0.95\textwidth}
    \renewcommand{\arraystretch}{1.1}%
    \begin{NiceTabular}{l|>{\centering\arraybackslash}p{2cm}:>{\centering\arraybackslash}p{2cm}}
    \Block{2-1}{\textbf{Model}} & \Block{1-2}{\textbf{Mean Angular Error $\bar{e}$}} & \\
    & Abs~\textsuperscript{a} & Rel~\textsuperscript{b} \\\hline
    baseline                & \ang{3.59} & -- \\
    baseline-no-transformer & \ang{3.26} & \ang{-0.33} \\\hline
    \Block[l]{1-3}{
        \small\textsuperscript{a} Absolute error (lower is better) \\
        \small\textsuperscript{b} Relative error to baseline model (lower is better)
    } & & \\
\end{NiceTabular}%
    \caption{Results of the ablation study on the \xgaze\ dataset.}%
    \label{t:evaluation:ablation:xgaze:results}%
    \vspace*{-4mm}%
\end{table}

Our experiment on the \xgaze\ dataset confirm our findings on the other two datasets. Replacing the Fusion Transformer improves the performance of our model. In the case of this dataset, our model achieves a mean angular error that is \ang{0.33} lower than our baseline model. This is a larger improvement than on our \oge\ dataset (see~\Cref{s:evaluation:ablation:oge}) and therefore supports our findings that the Transformer architecture used in our model architecture is not beneficial for the task of Gaze Angle and Gaze Point Estimation.

\subsection{Summary of the Ablation Study}
\label{s:evaluation:ablation:summary}
We saw a similar pattern throughout all three datasets on which we conducted our ablation study. When removing or replacing modules of our model architecture, the effects on our \oge\ dataset were smaller compared to the ones on the \stg\ dataset. We found that the headpose feature extraction module is a crucial component of our model and its removal increases the error of our model significantly. Future work might investigate in alternative architectures for this module to improve the overall model performance.

However, we also found that the replacement of the Fusion Transformer module with an \ac{MLP} leads to significantly improved model performance, which indicates that the Transformer architecture used in our model is not suited for the task of \ges. Since the \xgaze\ dataset contains high-quality RGB images, we can conclude that the observed performance drop is not affected by the existence of depth maps. Furthermore, the errors on the training set of our \oge\ dataset and the \stg\ dataset are also higher for the model with the Fusion Transformer. This indicates that the root cause is not overfitting to the training set but rather the complexity of the Transformer architecture.

\clearpage
\section{Real-Time Gaze Point Estimation Pipeline}
\label{s:evaluation:pipeline}

The third objective of this thesis is to implement a pipeline that uses a previously trained model to estimate the gaze point of a person in real-time. In this chapter, we present our pipeline and its functional architecture. We first describe the prerequisites for our pipeline and discuss its limitations. Next, we explain the processing and describe the pipeline implementation. Finally, we show the real-time visualization capabilities of our pipeline as well as its extensibility.

\subsection{Prerequisites and Limitations}
\label{s:evaluation:pipeline:prerequisites}
Since our pipeline is based on our RGBDTr model and is written in Python, it is compatible with all operating systems that support Python. We tested it on Windows~10 and Linux~Fedora~38. The required Python packages for our pipeline are OpenCV (\package{cv2})~\cite{Bradski2000}, \package{NumPy}~\cite{Harris2020}, \package{screeninfo}~\cite{Kurczewski2022}, \package{pyrealsense2}~\cite{IntelCorporation2018}, and PyTorch (\package{torch})~\cite{Paszke2019}. To estimate the subject-specific bias terms $\hat{b}$, we use scikit-learn (\package{sklearn})~\cite{Pedregosa2011}. For the 3D visualization described in \Cref{s:evaluation:pipeline:visualization}, we use \package{matplotlib}~\cite{Hunter2007} and \package{plotly}~\cite{PTI2023}.

In order to use our gaze estimation pipeline, we need to calibrate the camera using the method described in \Cref{s:processing:datacollection:calibration}. However, the amount of tiles in the checkerboard pattern and the size of each tile can be adjusted to the specific monitor used. Since we use a \SI{27}{\inch} monitor, we use a $10 \times 5$ checkerboard pattern with a tile size of \SI{50}{\milli\meter} for our calibration. Due to this calibration step, our gaze estimation pipeline is inherently independent of the screen resolution, the screen size, and the aspect ratio of the screen used. However, our implementation is limited to a flat monitor setup and assumes square pixels. We do not support curved monitors. Furthermore, in a multi-monitor setup, we assume that both monitors lie in the same plane. Since our gaze estimation dataset contains only data from a single monitor setup, we cannot guarantee that our pipeline works in a multi-monitor setup as we did not test it.

For our pipeline to work, one needs an Intel RealSense camera placed beneath the monitor. Although some models may not use the depth map, our implementation still captures them and feeds them into the gaze angle estimation models. We tested only the Intel RealSense D435 camera system, which we used to capture our \oge\ dataset.

The third prerequisite is the trained model itself. Our pipeline implementation is compatible with all model configurations used in this thesis as it shares a common code basis with the training and evaluation scripts. They can be easily extended to support other model configurations in the future. Since our pipeline expects the model to use the depth map and to predict gaze angles, we use models trained on our \oge\ dataset as it provides both RGBD images and gaze angles. However, the processing pipeline can be easily extended to support models that predict gaze points instead of gaze angles. In this case, the model is more dependent on the monitor used during data collection and, thus, less versatile than a model predicting gaze angles. This is why we implemented our pipeline to support only gaze angle prediction models for now.

\subsection{Processing Pipeline}
\label{s:evaluation:pipeline:processing}
The processing steps of our gaze estimation pipeline was depicted in \Cref{f:processing:sequence}. In essence, our real-time pipeline performs four steps. First, face and facial landmark detection is performed on the RGB input image as described in \Cref{s:processing:sequence:landmark}. Next, the pipeline normalizes both the RGB image and the aligned depth map using the same parameters as in \Cref{s:processing:sequence:normalization}. It is important to use the same normalization parameters as during data collection for the model to work properly.

After that, the normalized RGB face patch $I^\mathrm{FC}$, the corresponding eye patches $I^\mathrm{REC}$ and $I^\mathrm{LEC}$, the normalized face depth map $I^\mathrm{FD}$, and the warped facial landmarks $L'$ are fed into the model. The model then predicts the gaze angles $\hat{g}'$, which are then used to calculate the gaze point $\hat{p}$ in pixels as described in \Cref{s:processing:sequence:unnorm}. Since we strive for a real-time pipeline and, thus, want to reduce the latency, we have to apply the model on each sample individually and cannot use batch processing. However, we can still use the GPU to speed up the processing.

As mentioned in \Cref{s:processing:sequence:unnorm}, we use up to three filters in our gaze estimation pipeline to reduce the noise of the estimated gaze point. Per default, we use Kalman filters but there is also the option to switch to a simple three-samples averaging filter. The averaging filters have a higher latency compared to their Kalman counterparts. The first filter helps to smooth the landmarks $L$ detected by the yolov7-face landmark detection model and, thus, also smooths the face normalization process. The second filter is applied to the predicted (normalized) gaze angles $\hat{g}'$ and the third filter is applied to the final gaze point $\hat{p}$. All three filters are optional and can be disabled individually during runtime. We implemented the Kalman filters to model 2D objects with a position and a velocity. The velocity vector has independent axes, i.e., the velocity in $x$-direction does not influence the velocity in $y$-direction and vice versa.

The subject-specific bias terms $\hat{b}$ are set to zero when using our pipeline. In order to estimate them using a Least Squares Linear Regression, our pipeline allows for the collection of calibration samples. During the calibration sample collection process, it shows a red dot which must be focused on and clicked by the user. This is similar to the single-point-single-sample phase of our data collection process described in \Cref{s:processing:datacollection:process}. A minimum of $n_\mathrm{cal} \geq 3$ samples must be collected to estimate the subject-specific bias terms $\hat{b}$. The user can decide on how many samples to collect. Once the user finished the calibration sample collection process, $\hat{b}$ are estimated and used for future gaze point predictions. The calibration process can be repeated at any time to update the subject-specific bias terms $\hat{b}$.

Although the implementation of our pipeline uses Python as its programming language and is a single-threaded application, we achieve real-time performance on a modern platform. We use an AMD~Ryzen~7900X processor and a NVIDIA~RTX~4080 GPU. The image preprocessing steps, i.e., landmark detection and normalization, take between \SIrange{5}{10}{\milli\second} on our platform. The estimation of gaze angles and the gaze point calculation take another \SIrange{8}{12}{\milli\second}. Since the Intel~RealSense~D435 camera system is limited to 30\,fps and our pipeline is dependent on the monitor frequency to display its result, we achieve an average framerate of 25\,fps, which we consider real-time.

\subsection{Real-Time 3D Visualization}
\label{s:evaluation:pipeline:visualization}
We implemented two 3D visualizations to show the gaze point estimation process of our pipeline. The first uses \package{matplotlib}~\cite{Hunter2007} and the second uses \package{plotly}~\cite{PTI2023}. They both show the five facial landmarks $L$, the face center $c$, the screen plane, and the estimated gaze vector $\hat{g}$ as well as the estimated gaze point $\hat{p}$ in 3D space. Additionally, the position and orientation of the camera is shown in the visualization. For debugging purposes, we also show the current cursor position as well as angle between it and $c$.

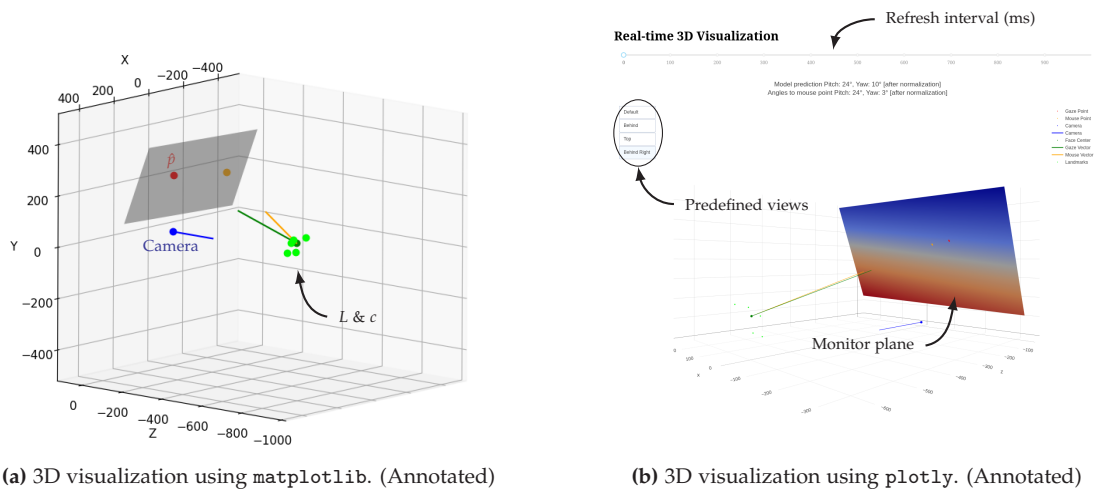
\begin{figure}[htb]
    \vspace*{-2mm}%
    \centering\captionsetup{width=0.95\textwidth}%
    \begin{subfigure}{0.5\textwidth}%
        \centering%
        \begin{tikzpicture}
            \node (img) at (0, 0) [inner sep=0pt] {\includegraphics[width=0.8\textwidth]{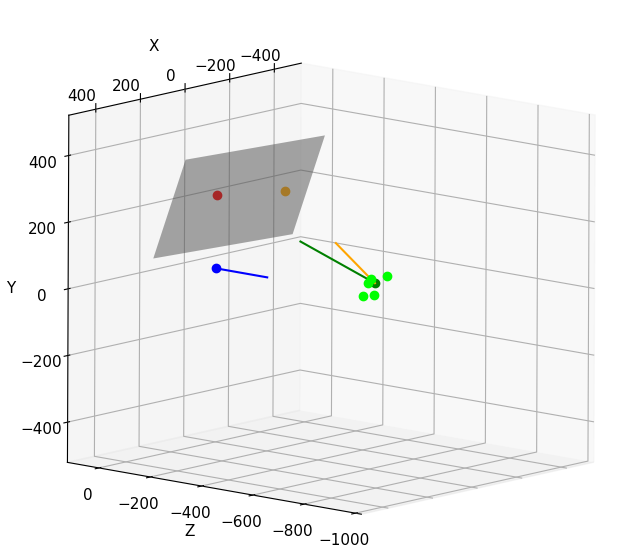}};

            \path let \p1=($(img.west)!0.60!(img.east)$), \p2=($(img.north)!0.55!(img.south)$) in node (landmarks) at (\x1, \y2) {};
            \node (landmarks-label) [below right=0.05\textwidth and 0.03\textwidth of landmarks] {\tiny $L$ \& $c$};
            \draw [arrow] (landmarks-label.west) to [out=170, in=280] (landmarks.south);

            \path let \p1=($(img.west)!0.34!(img.east)$), \p2=($(img.north)!0.32!(img.south)$) in node (p) at (\x1, \y2) [color=red!70!black!85!blue] {\tiny $\hat{p}$};

            \path let \p1=($(img.west)!0.34!(img.east)$), \p2=($(img.north)!0.52!(img.south)$) in node (cam) at (\x1, \y2) [color=blue!90!white] {\tiny Camera};

        \end{tikzpicture}
        \caption{3D visualization using \package{matplotlib}. (Annotated)}%
        \label{f:evaluation:pipeline:visualization:matplotlib}%
    \end{subfigure}%
    \begin{subfigure}{0.5\textwidth}%
        \centering%
        \begin{tikzpicture}
            \node (img) at (0, 0) [inner sep=0pt] {\includegraphics[width=0.8\textwidth]{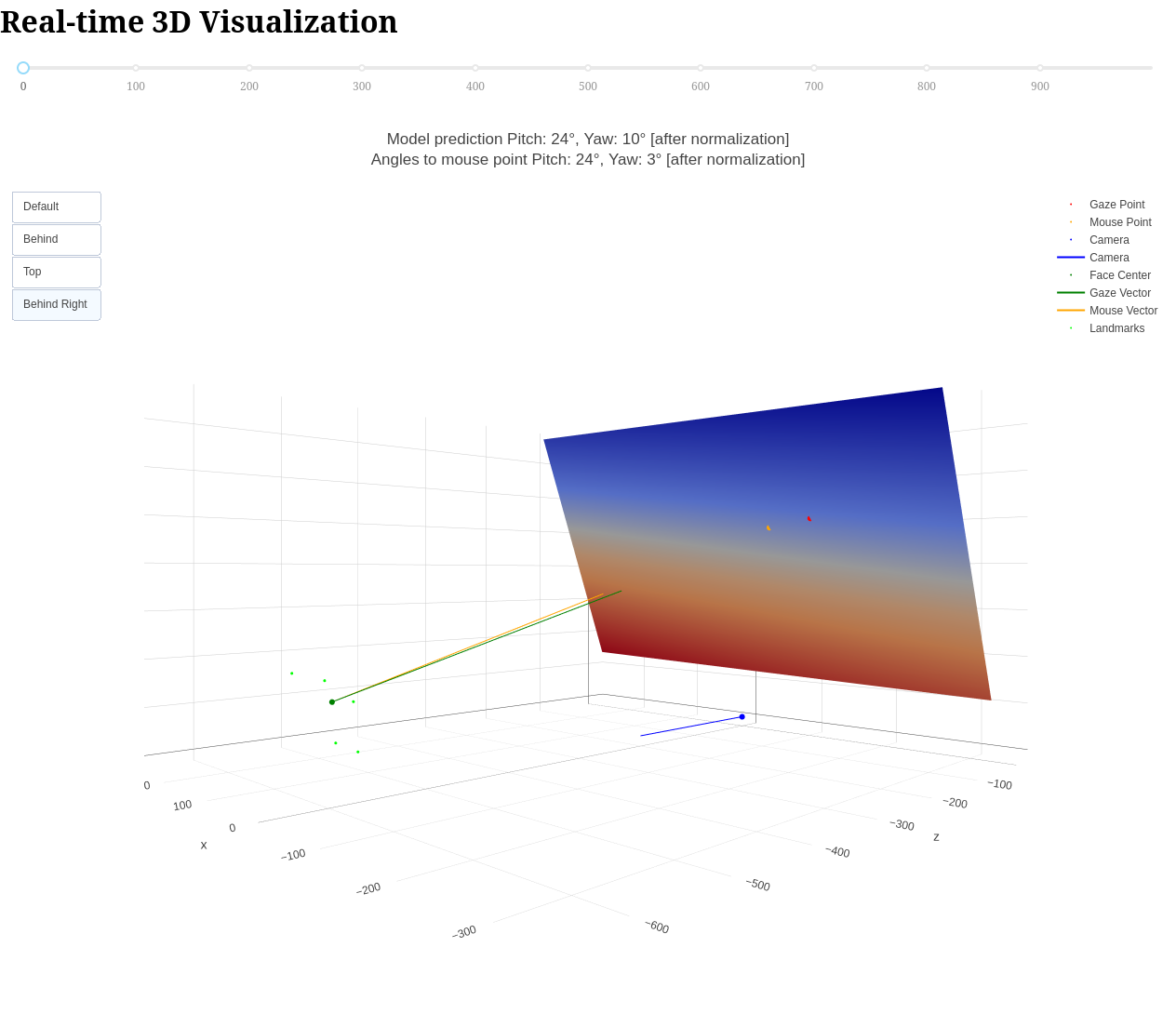}};

            \path let \p1=($(img.west)!0.05!(img.east)$), \p2=($(img.north)!0.25!(img.south)$) in node (predef) at (\x1, \y2) [draw=black, ellipse, inner sep=0pt, minimum width=0.08\textwidth, minimum height=0.11\textwidth] {};
            \node (predef-label) [below right=0.05\textwidth and 0.03\textwidth of predef] {\tiny Predefined views};
            \draw [arrow] (predef-label.west) to [out=180, in=280] (predef.south);

            \path let \p1=($(img.west)!0.45!(img.east)$), \p2=($(img.north)!0.08!(img.south)$) in node (slider) at (\x1, \y2) {};
            \node (slider-label) [above right=0.02\textwidth and 0.05\textwidth of slider] {\tiny Refresh interval (\unit{\milli\second})};
            \draw [arrow] (slider-label.west) to [out=180, in=80] (slider.north);

            \path let \p1=($(img.west)!0.70!(img.east)$), \p2=($(img.north)!0.61!(img.south)$) in node (monitor) at (\x1, \y2) {};
            \node (monitor-label) [below left=0.05\textwidth and 0.03\textwidth of monitor] {\tiny Monitor plane};
            \draw [arrow] (monitor-label.east) to [out=10, in=260] (monitor.south);
        \end{tikzpicture}
        \caption{3D visualization using \package{plotly}. (Annotated)}%
        \label{f:evaluation:pipeline:visualization:plotly}%
    \end{subfigure}%
    \caption{3D visualization of our gaze estimation pipeline.}%
    \label{f:evaluation:pipeline:visualization}%
\end{figure}

The main difference between the two visualization is in the way they are rendered. The \package{matplotlib} visualization is rendered onto the same screen as the outputs (gaze point) of our pipeline. However, this slows down the process to such an extent that we are only able to achieve 10\,fps to 15\,fps on our platform. The \package{plotly} visualization uses a web browser to render the 3D points, vectors, and planes. This allows us to offload the expensive rendering to another process and achieve the aforementioned 25\,fps. Since this visualization is based on web technology, it is also possible to render it on a different device. Furthermore, it allows for interactive exploration of the 3D scene, e.g., by rotating the virtual camera or zooming in and out. We show both visualizations in \Cref{f:evaluation:pipeline:visualization}. The Intel~RealSense~D435 camera system is located in the center of the coordinate system in both the \package{matplotlib} visualization depicted in \Cref{f:evaluation:pipeline:visualization:matplotlib} and the \package{plotly} visualization depicted in \Cref{f:evaluation:pipeline:visualization:plotly}.

\vspace*{-3mm}
\subsection{Extensibility}
\label{s:evaluation:pipeline:extensibility}
Our pipeline already supports different model configurations and can be used with models trained on our dataset as well as on the \xgaze\ dataset. Furthermore, models that predict gaze points instead of gaze angles could be used as well with minor modifications as described above. To incorporate our real-time pipeline into other workflows, it is possible to provide an \ac{API} that allows external software to retrieve the estimated gaze point $\hat{p}$. For example, the pipeline could be extended to publish the gaze points to a MQTT server allowing listener programs to react to them. However, such an \ac{API} is not part of our implementation yet.

  \makeatletter
\let\@savedchapterhead\@makechapterhead
\def\@makechapterhead{\vspace*{-4mm}\@savedchapterhead}
\chapter[Conclusion and Future Work]{\vspace*{-2mm}Conclusion and Future Work}
\label{c:conclusion}
\let\@makechapterhead\@savedchapterhead
\makeatother
\vspace*{-3mm}
In this final chapter, we will discuss the execution of the three tasks of this thesis that were introduced in \Cref{s:intro:problem}. Furthermore, we summarize the results of our experiments and our findings. Finally, we give an outlook on future work.

\section{Discussion and Conclusion}
\label{c:conclusion:discussion}

\textbf{Main objectives.} The task of this thesis consisted of three components. First, we created our own dataset -- \oge\ -- featuring more than 130,000 samples of 12 different subjects. Each sample consists of a normalized RGB face patch, its corresponding normalized depth map, two colored eye patches, facial landmarks and the gaze angles as labels. For future work, we also included information from the normalization process as well as the specific task of the subject in the sample. Furthermore, we include scene-relevant information such as the camera extrinsic parameters and the screen parameters. We described the data collection process in \Cref{s:processing:datacollection} and analyzed our dataset in \Cref{s:processing:dataset}.

Second, we designed, trained and evaluated \dl\ models for the task of \ges. Our models were based on the works of \citeauthor*{Lian2019}~\cite{Lian2019}, which we extended by the combination of \acp{CNN} and Transformers for the task of \ges\ on RGBD images. The model architecture, its modules, and the training process were described in \Cref{s:processing:models}. We then trained and evaluated various model configurations on the \xgaze, the \stg, and our \oge\ dataset in \Cref{s:evaluation:xgaze,s:evaluation:stg,s:evaluation:oge}, establishing a baseline for future experiments. Furthermore, we performed hyperparameter optimization on our dataset and evaluated our best model configuration on the test sets of all three datasets in \Cref{s:evaluation:hyperparams}. Next, we conducted an ablation study on our baseline model configuration in \Cref{s:evaluation:ablation} to analyze the impact of the different modules of our model architecture.

Third, we implemented a real-time pipeline for the task of \ges\ on RGBD images using our models. We described the processing sequence in \Cref{s:processing:sequence} since there is a common basis for both the data collection process and the real-time inference. We then described the implementation in \Cref{s:evaluation:pipeline} and demonstrated its visualization capabilities. Our pipeline can be extended to provide an \ac{API} for other applications.

\textbf{Results and discussion.} We trained and evaluated various model configurations on the datasets \xgaze, \stg, and \oge. We found that applying weight decay during training increases the error in \Cref{s:evaluation:xgaze:weight-decay}, which is why we conducted all other experiments without it. In \Cref{s:evaluation:stg}, our experiments showed that using a pre-trained \ac{GAN} backbone increases the error significantly compared to a backbone which is trained from scratch. However, using depth maps improves our model performance, indicating that RGBD images contain additional information useful for the task of \ges\ compared to RGB-only input. These findings were confirmed by our experiments on our \oge\ dataset in \Cref{s:evaluation:oge}. Furthermore, we found the calibration step to be crucial for the performance of our models on all datasets. On our dataset, calibration reduced the on-screen Euclidean errors predominantly in the center, upper and left parts of the screen.

{\spaceskip=3.4pt plus 1pt minus 1.5pt 
The hyperparameter tuning was conducted on our dataset and yielded a model configuration that achieved a mean angular error of $\bar{e} = \ang{4.7}$ on the test set using $n_\mathrm{cal} = 200$ calibration samples for the estimation of the subject-specific bias terms $\hat{b}$. This model configuration achieved a mean Euclidean error of $\bar{d} = \qty{29.7}{\milli\meter}$ on the \stg\ dataset, making it the best-performing model on this dataset that we know of. However, our best model configuration achieved a mean angular error of $\bar{e} = \ang{3.59}$ on the \xgaze\ test set, which is worse than the baseline model of the authors \citeauthor*{Zhang2020}.

In our final ablation study in \Cref{s:evaluation:ablation}, we found that the head pose feature extraction module is crucial for our model architecture. Its removal increased the mean error depending on the dataset tested by a factor of \qtyrange{2}{4}{}. On the other hand, we found little to no difference when removing the depth extraction module or the eye pose feature extraction module. However, we found that our model performs better on RGBD datasets when using the depth map compared to an RGB-only input. This supports our earlier findings that the depth maps contain useful information for the task of \ges. To our surprise, we found that the Transformer module for feature fusion leads to worse performance compared to a simple substitute \ac{MLP}. We could confirm this on all three datasets, which indicates that the used Fusion Transformer module is not suitable for the tasks of both Gaze Angle Estimation and Gaze Point Estimation. However, we do not conclude that Transformers are not suitable for these tasks in general, since we only tested our specific implementation. In particular, previous work indicates that models incorporating the Transformer architecture are beneficial~\cites{Cai2021}{Cheng2021}{Ding2021}{Li2023}{Nagpure2023}.
}

\textbf{Conclusion.} In this thesis, we created our own dataset -- \oge\ -- for the task of \ges\ on RGBD images. We then designed, trained and evaluated \dl\ models for this task. Our experiments show that the depth map of our dataset and the \stg\ dataset contain additional information that is useful for the task. Furthermore, we found that the calibration step and the head pose feature extraction module are crucial for our model performance. We also found that our implementation of the Transformer module to fuse features generated by the \ac{CNN}-based modules hurts the performance, whereas the increase in error is considerably less than the decrease in error achieved by using RGBD images instead of RGB-only ones on our dataset. Finally, we implemented a real-time pipeline for the task of \ges\ on RGB and RGBD images using our models.

\section{Future Work}
\label{c:conclusion:future-work}
{\spaceskip=3.4pt plus 1pt minus 1.5pt 
Our work can be extended in multiple ways. In this section, we take a look at some potential enhancements for our dataset, our model architecture, and our real-time pipeline.
}

\textbf{Improvements to our dataset.} Since our dataset contains samples from only 12 subjects, its variety and size are limited. Therefore, it would be beneficial to extend our dataset with more samples from more subjects. \Citeauthor*{Krafka2016} showed that the number of subjects is an important factor when training \dl\ models~\cite{Krafka2016}. Furthermore, we only recorded samples in a semi-controlled environment. It would be interesting to see how our models perform on samples recorded in various locations featuring different environmental influence. In contrast to other datasets, such as \xgaze, we allowed the subjects to move their head freely and encouraged them to do so. Future work could investigate the impact of head movement on the model's performance.

\textbf{Improvements to our model architecture.} We derived our model architecture from the work of \citeauthor*{Lian2019}~\cite{Lian2019} and adjusted in particular the feature fusion step. However, there are many more modifications that could be made to our model architecture. For example, \citeauthor*{Chen2019}~\cite{Chen2019} showed that dilated convolutions can be used to increase the receptive field of a \ac{CNN} without increasing the number of parameters. Furthermore, future work could use different pre-trained models for the feature extraction modules. In particular, a different pre-trained model for the eye pose feature extraction module could improve model performance. Since we found the head pose feature extraction module to be crucial, it might be worth investigating different architectures not based on \acp{GAN}. Apart from that, there is always the option to further improve hyperparameters as discussed in \Cref{s:evaluation:hyperparams:outlook}. Additionally, one could use \ac{NAS} as shown by \citeauthor*{Nagpure2023}~\cite{Nagpure2023} to reduce the number of parameters and, thus, improve the performance of our models on weaker platforms. We used the calibration method of \citeauthor*{Chen2020}~\cite{Chen2020} in this thesis. However, it could be replaced by or combined with other calibration methods, such as the differential calibration method of \citeauthor*{Liu2019}~\cite{Liu2019} or the subject-specific fine-tuning method of \citeauthor*{Zhang2020}~\cite{Zhang2020}.

{\spaceskip=3.4pt plus 1pt minus 1.5pt 
\textbf{Improvements to our real-time pipeline.} Although we achieve real-time performance with our pipeline implementation, there is still room for improvement. For example, both the pipeline implementation and the model size could be optimized to achieve better performance on weaker platforms. Future work could also investigate in the combination of different face detection and facial landmark detection methods. As outlined in \Cref{s:evaluation:xgaze:conclusion}, it might be beneficial to combine the yolov7-face detection method with the dlib face landmark detection method. Future work could investigate into the performance characteristic of such a combination and decide whether the landmark detection is better and if so, whether the additional computational cost is worth it. Furthermore, we use the temporal information of the continuous video stream only to reduce noise in the gaze point estimation pipeline using Kalman filters or moving averages. However, \citeauthor*{Li2023}~\cite{Li2023} showed a novel architecture based on \ac{LSTM} networks to incorporate temporal information into the \ges\ process. Future work could investigate this approach and combine it with a Transformer-based fusion architecture described in~\cite{Sharir2021}. However, the datasets used in this thesis are not suited for this task.
}

  \phantomsection
  \addcontentsline{toc}{chapter}{Bibliography}
  \setnowidow[10]\setnoclub[10]
  \printbibliography
  \newpage

  \phantomsection
  \addcontentsline{toc}{chapter}{List of Figures}
  \listoffigures
  \newpage

  \phantomsection
  \addcontentsline{toc}{chapter}{List of Tables}
  \listoftables
  \newpage

  \appendix
  \renewcommand{\chaptermark}[1]{\markboth{Appendix \thechapter. #1}{}}
  \chapter{Additional Plots}

\begin{figure}[htb]
    \centering\captionsetup{width=0.95\textwidth}%
    \begin{subfigure}{\textwidth}%
        \includegraphics[width=\textwidth]{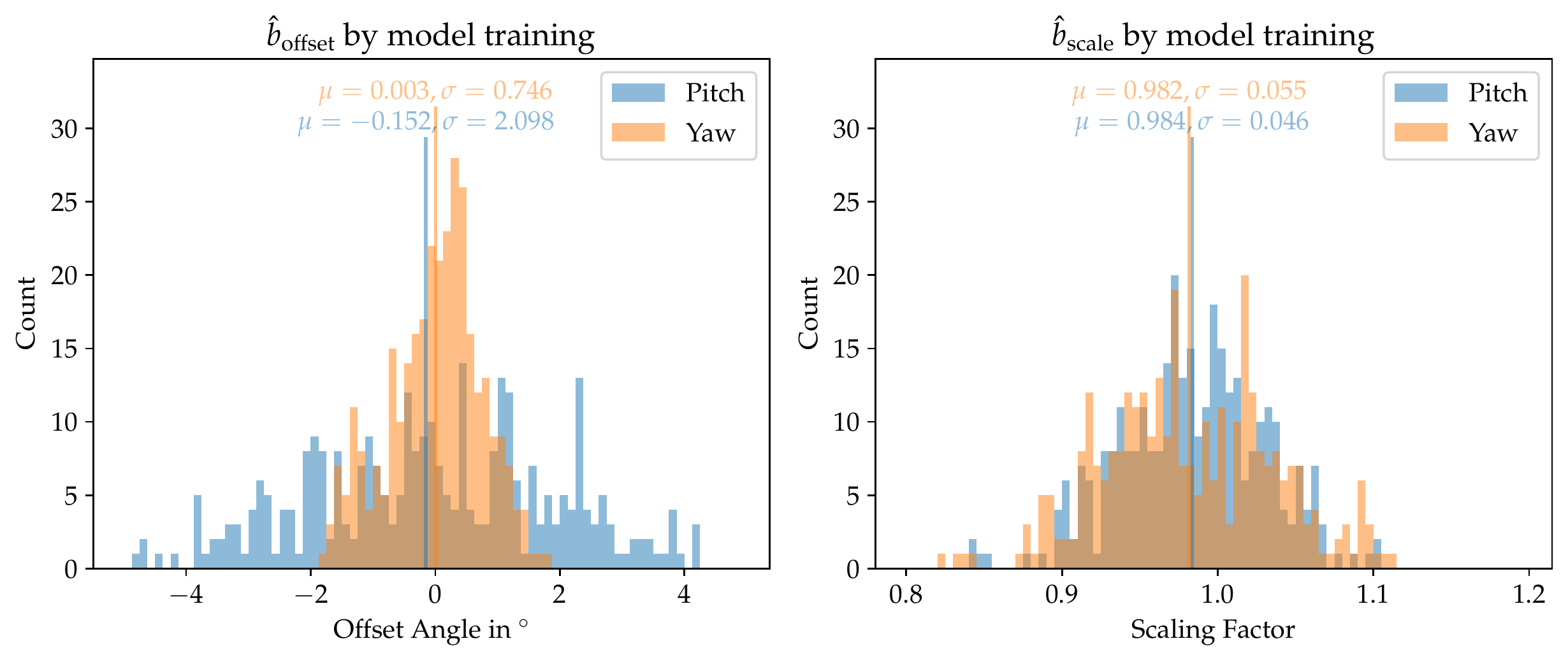}%
        \vspace*{-2mm}%
        \caption{Distribution of the learned subject-specific bias terms $\hat{b}$ after training.}%
        \label{f:appendix:xgaze-parameter-distribution:learned}%
        \vspace*{2mm}%
    \end{subfigure}
    \begin{subfigure}{\textwidth}%
        \includegraphics[width=\textwidth]{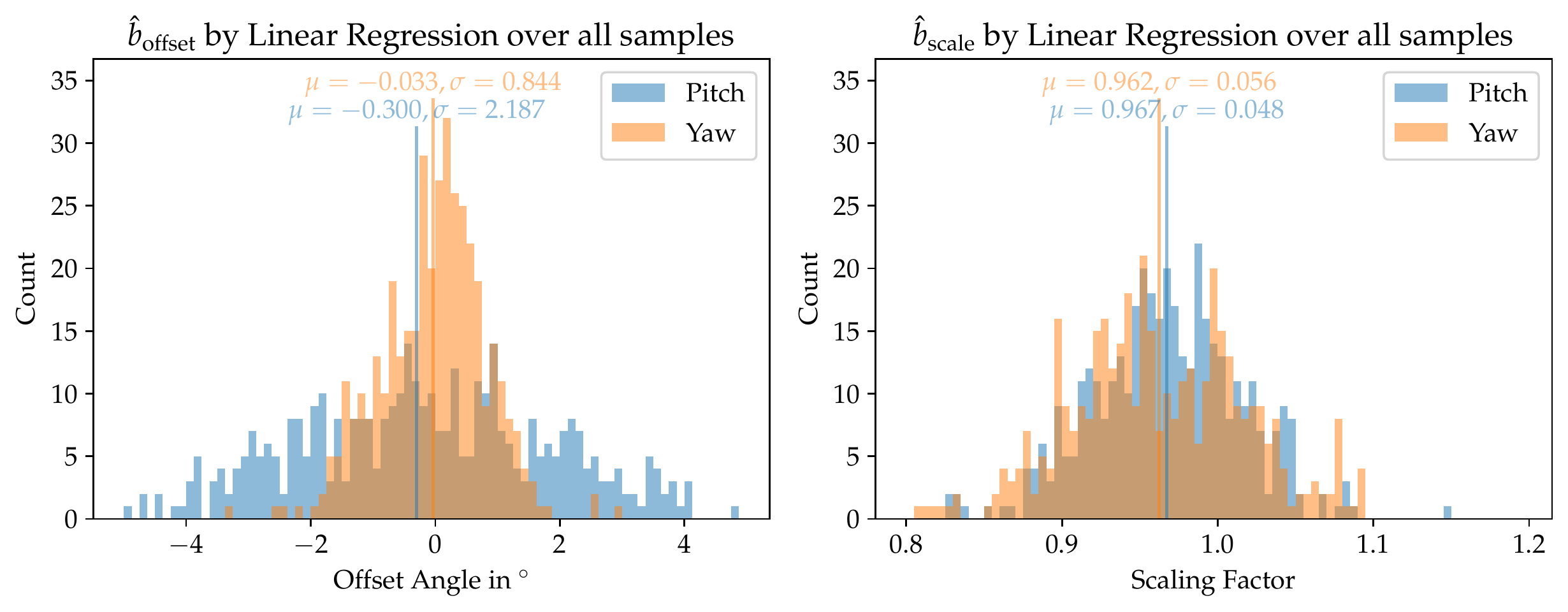}%
        \vspace*{-2mm}%
        \caption{Distribution of the subject-specific bias terms $\hat{b}$ estimated using a Linear Regression model.}%
        \label{f:appendix:xgaze-parameter-distribution:linreg}%
    \end{subfigure}%
    \caption{Distributions of subject-specific bias terms $\hat{b}$ for the \xgaze\ dataset aggregated over all 5-fold cross-validation splits.}%
    \label{f:appendix:xgaze-parameter-distribution}%
\end{figure}

\begin{figure}[htb]
    \centering%
    \includegraphics[width=\textwidth]{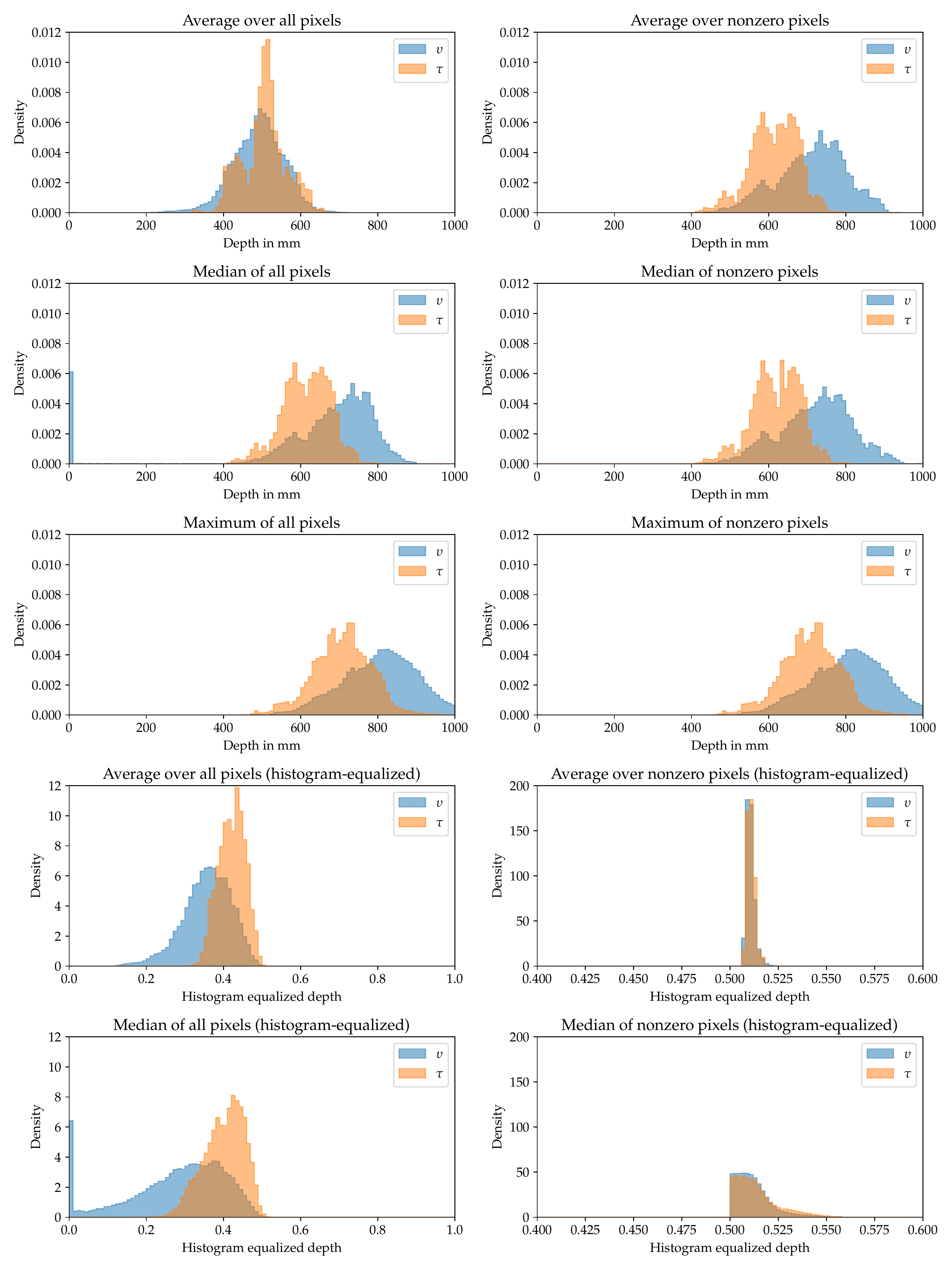}%
    \caption{Distribution of the depth values in the \stg\ input set $\upsilon$ and the target subset $\tau$.}%
    \label{f:appendix:stg-depth-distribution}%
\end{figure}

\begin{figure}[htb]
    \centering%
    \includegraphics[width=0.986\textwidth]{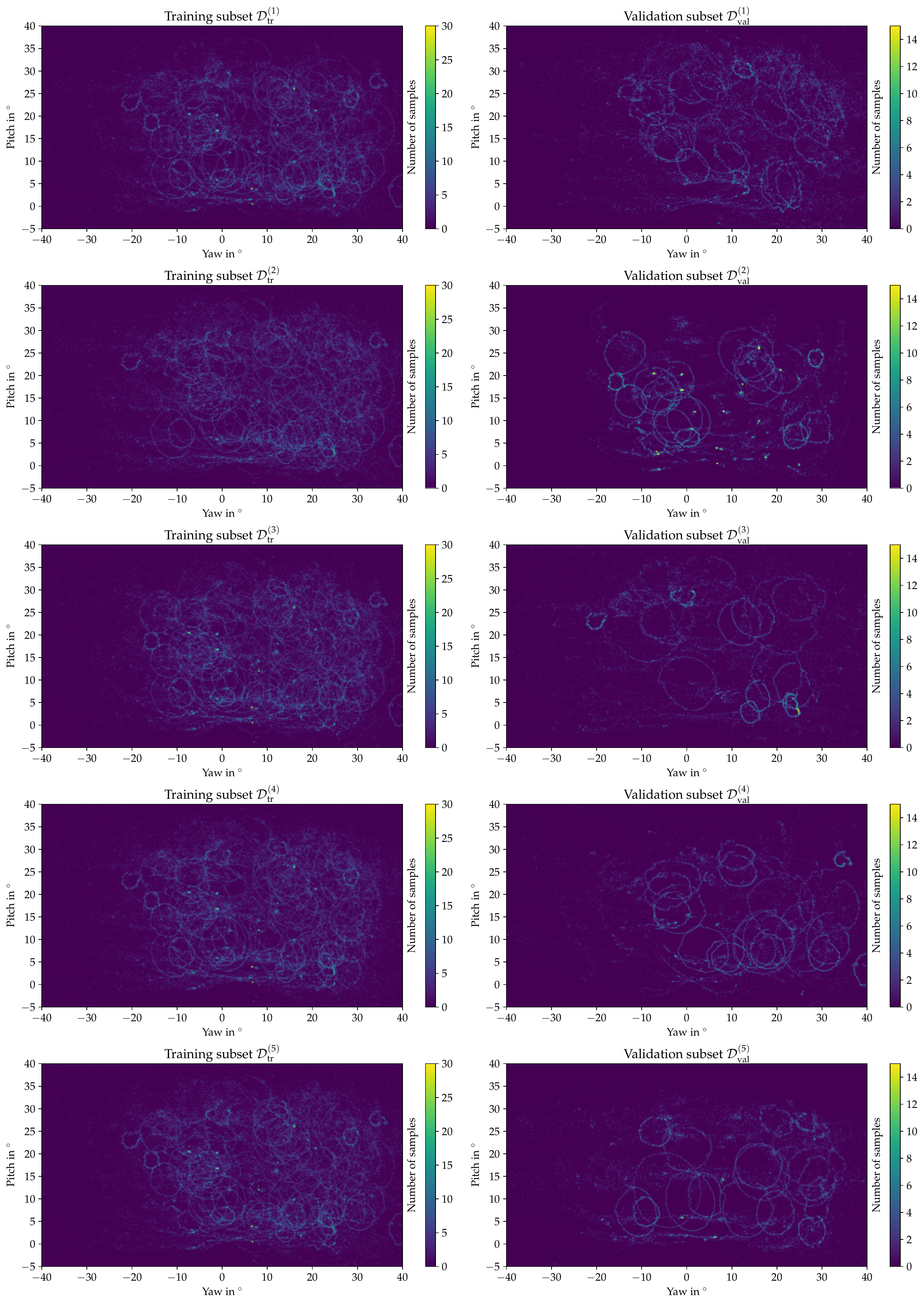}%
    \caption{Distribution of the normalized gaze angles in the training and validation subsets of our \oge\ dataset.}%
    \label{f:appendix:oge-gaze-angles-train-val}%
\end{figure}
\begin{figure}[htb]
    \centering%
    \includegraphics[width=0.986\textwidth]{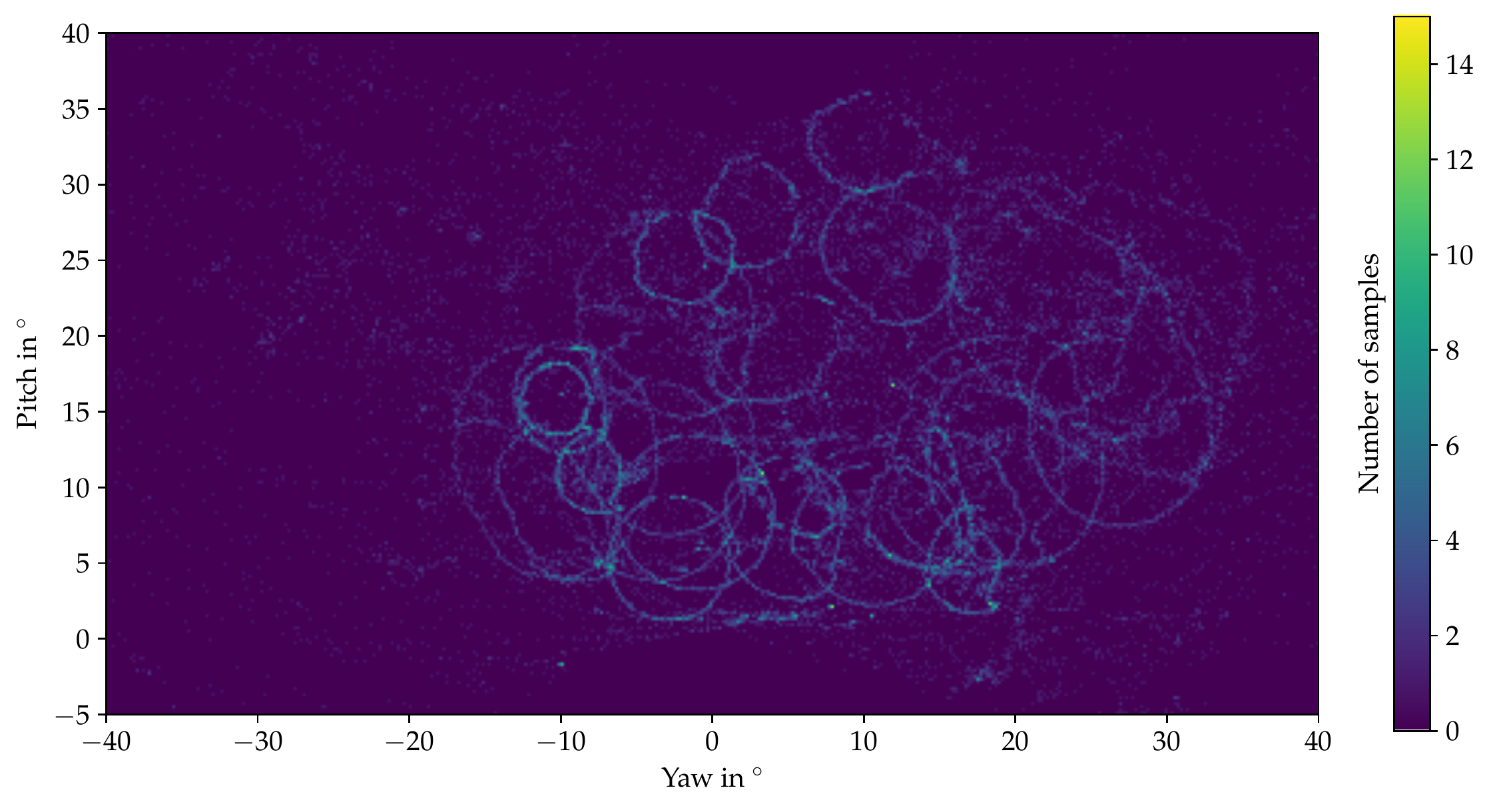}%
    \caption{Distribution of the normalized gaze angles in the test set of our \oge\ dataset.}%
    \label{f:appendix:oge-gaze-angles-test}%
\end{figure}

\begin{figure}[htb]
    \centering%
    \includegraphics[width=0.986\textwidth]{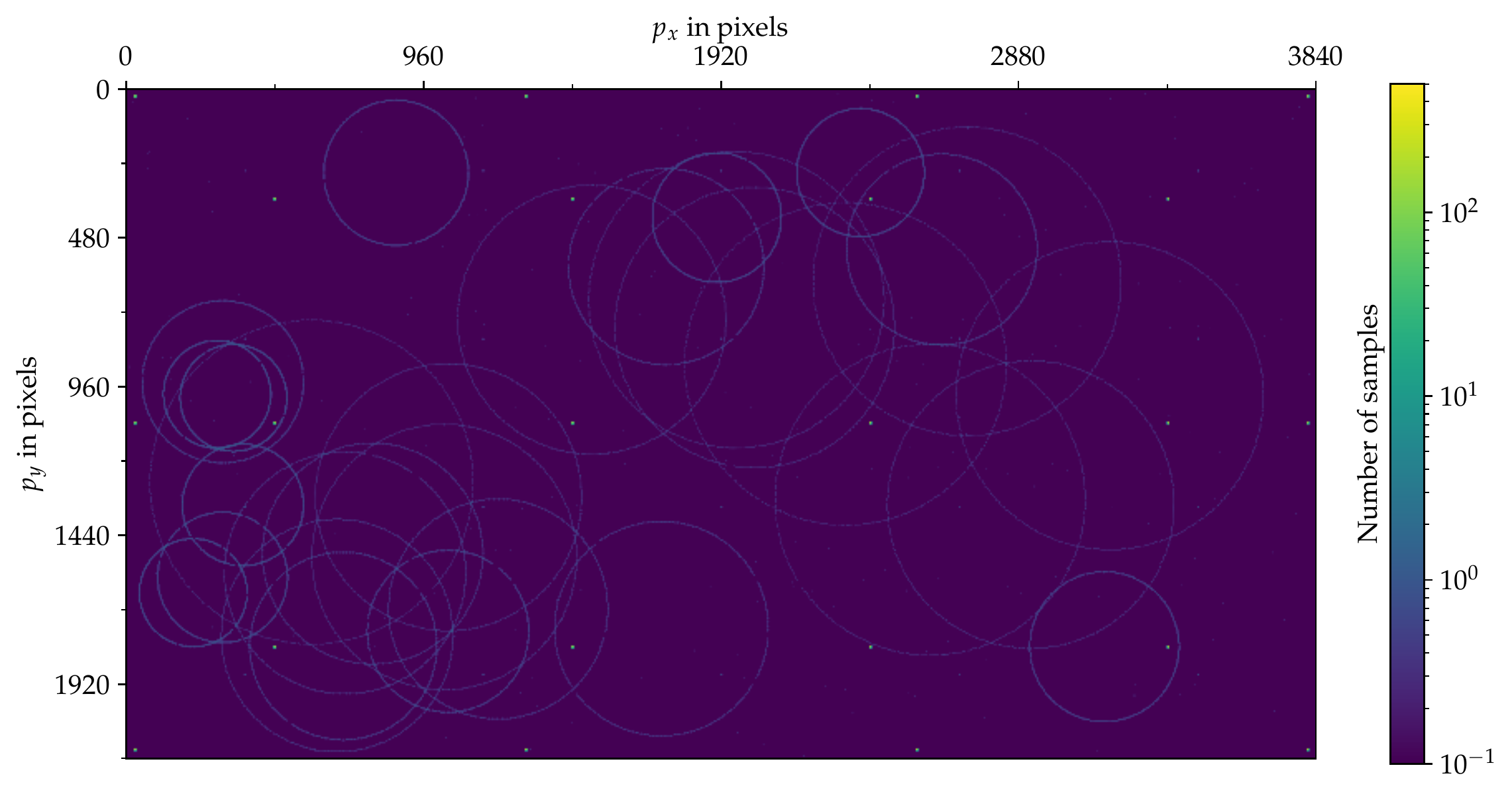}%
    \caption{Distribution of the on-screen gaze points in the test set of our \oge\ dataset (logarithmic scale).}%
    \label{f:appendix:oge-gaze-points-test}%
\end{figure}
\begin{figure}[htb]
    \centering%
    \includegraphics[width=0.986\textwidth]{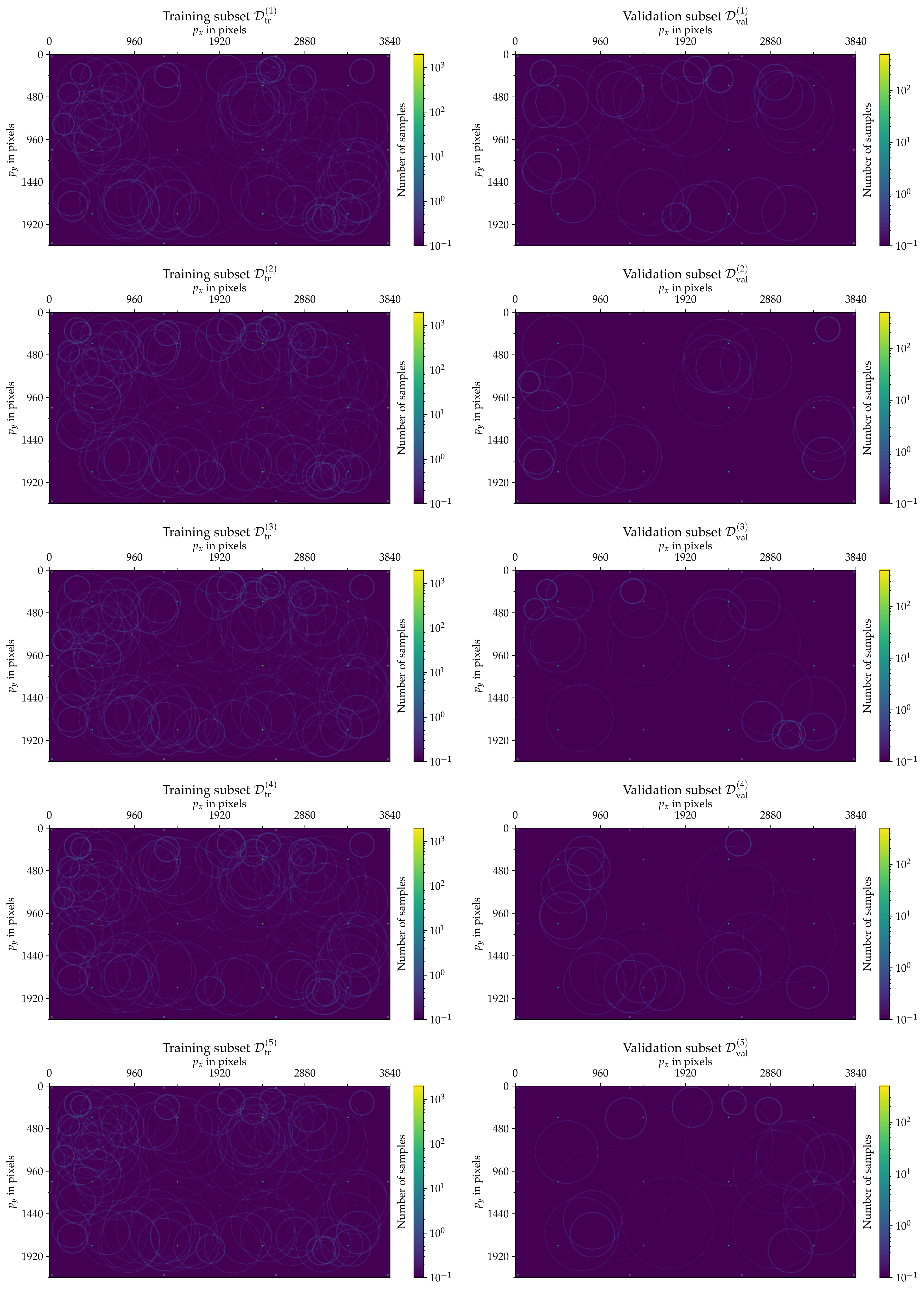}%
    \caption{Distribution of the on-screen gaze points in the training and validation subsets of our \oge\ dataset (logarithmic scale).}%
    \label{f:appendix:oge-gaze-points-train-val}%
\end{figure}

\begin{figure}[htb]
    \centering%
    \includegraphics[width=\textwidth]{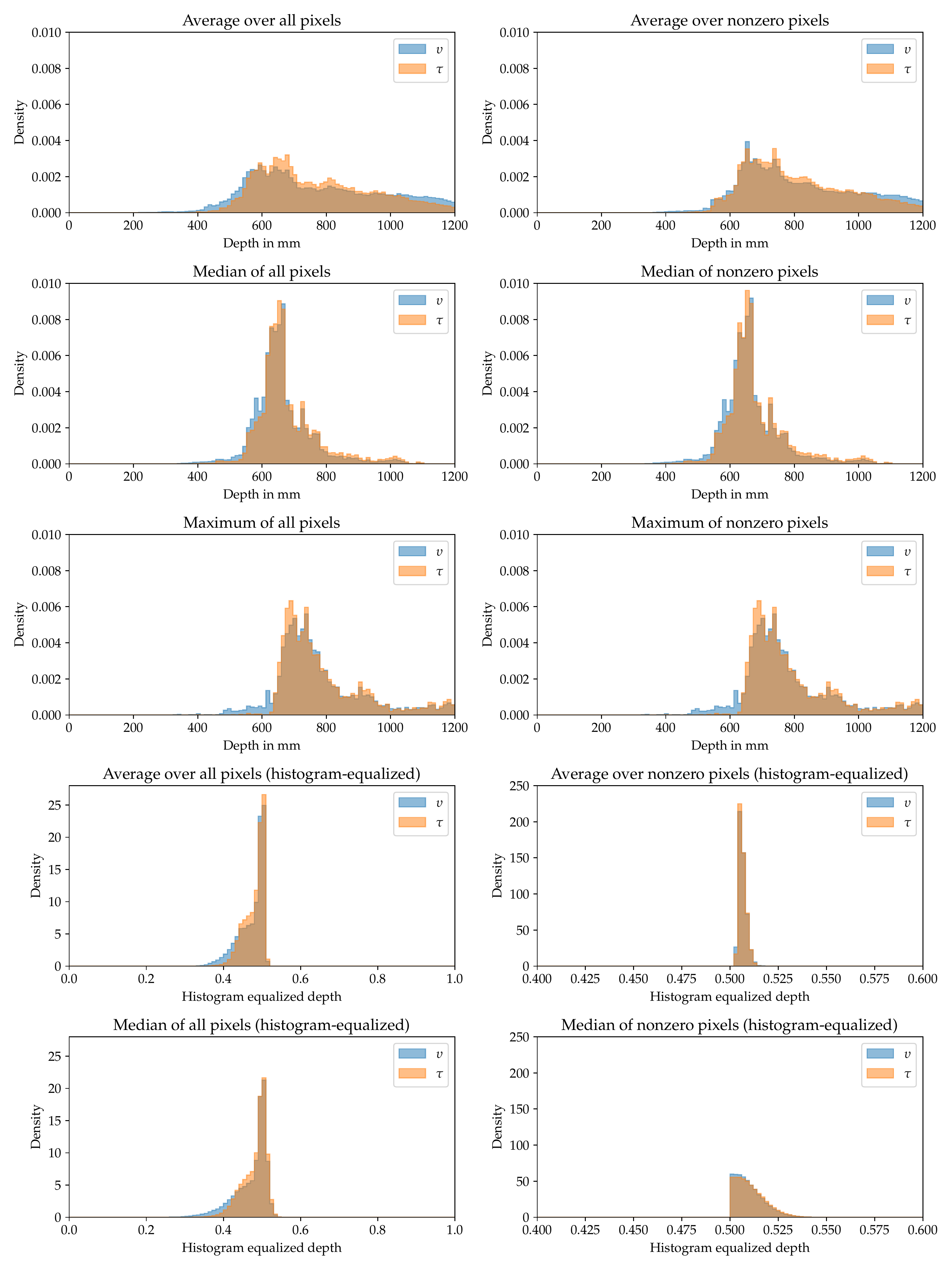}%
    \caption{Distribution of the depth values in the \oge\ input set $\upsilon$ and the target subset $\tau$.}%
    \label{f:appendix:oge-depth-distribution}%
\end{figure}

\begin{figure}[htb]
    \centering\captionsetup{width=0.9\textwidth}%
    \begin{tikzpicture}
        \node (redacted) {redacted due to unclear copyright situation};
        \node [below=of redacted, text width=0.8\textwidth] {Affected samples: \begin{itemize}
            \item Subject 4 (train), Sample 90
            \item Subject 38 (train), Sample 182
            \item Subject 38 (train), Sample 576
            \item Subject 54 (test\_specific), Sample 287
            \item Subject 54 (test\_specific), Sample 288
            \item Subject 59 (train), Sample 124
            \item Subject 66 (train), Sample 389
            \item Subject 75 (train), Sample 92
            \item Subject 113 (test\_specific), Sample 235
            \item Subject 118 (test), Sample 78
            \item Subject 119 (test\_specific), Sample 205
        \end{itemize}};
    \end{tikzpicture}
    \caption{Images from the \xgaze\ dataset where the conversion of the dlib landmarks failed (marked red) and neighboring samples for comparison.}%
    \label{f:appendix:xgaze-landmarks:dlib-failed}%
\end{figure}

\begin{figure}[htb]
    \centering%
    \begin{subfigure}{\textwidth}%
        \includegraphics[width=\textwidth]{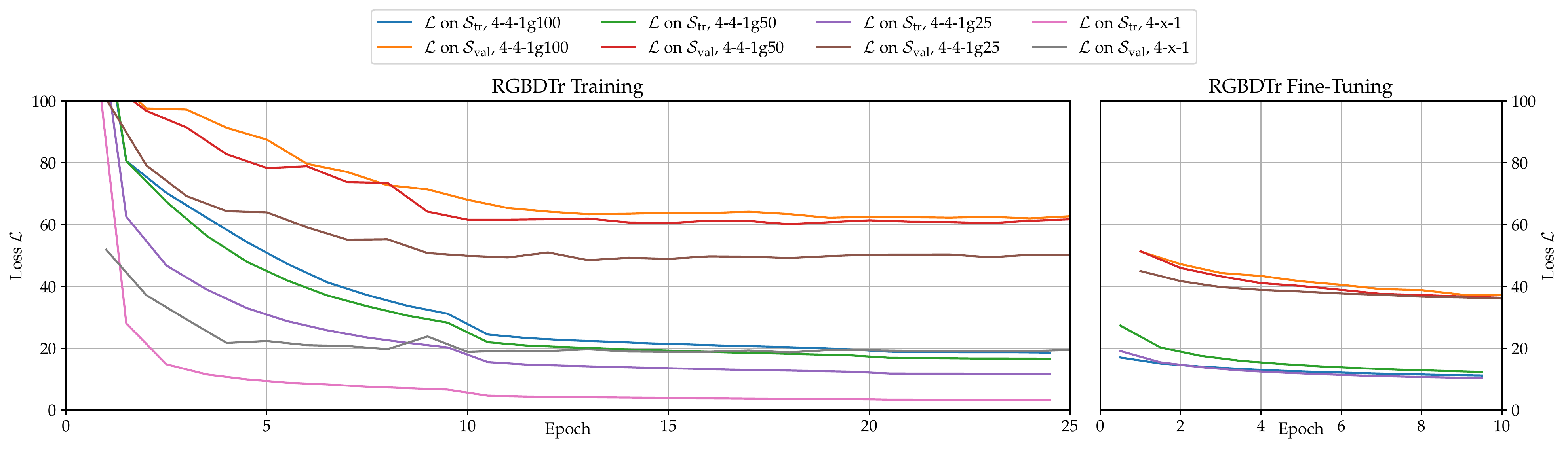}%
        \vspace*{-2mm}%
        \caption{Four layer PatchGAN/encoder on images of size $448 \times 448$ pixels with fusion block.}%
        \label{f:appendix:stg-losses:fus-448}%
        \vspace*{2mm}%
    \end{subfigure}
    \begin{subfigure}{\textwidth}%
        \includegraphics[width=\textwidth]{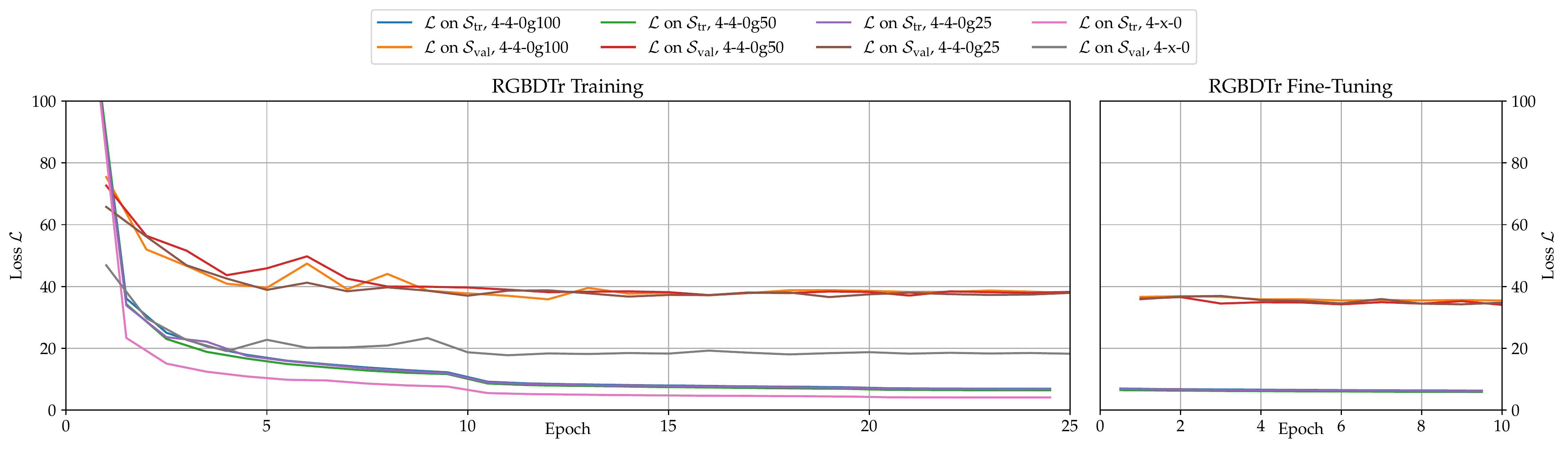}%
        \vspace*{-2mm}%
        \caption{Four layer PatchGAN/encoder on images of size $448 \times 448$ pixels without fusion block.}%
        \label{f:appendix:stg-losses:nofus-448}%
        \vspace*{2mm}%
    \end{subfigure}
    \begin{subfigure}{\textwidth}%
        \includegraphics[width=\textwidth]{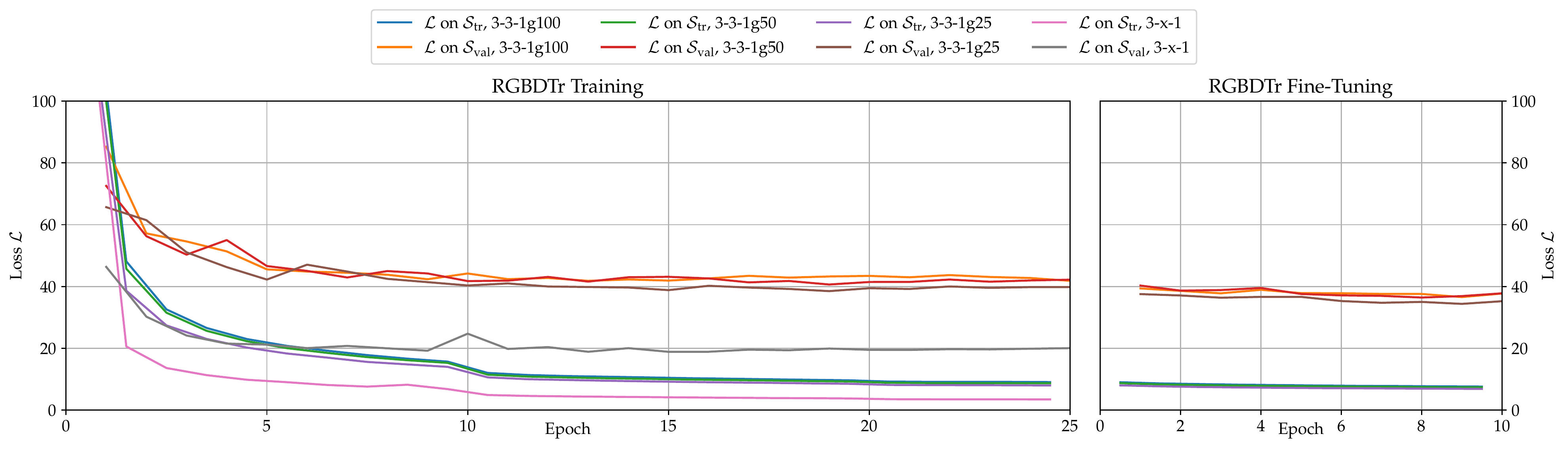}%
        \vspace*{-2mm}%
        \caption{Three layer PatchGAN/encoder on images of size $224 \times 224$ pixels with fusion block.}%
        \label{f:appendix:stg-losses:fus-224}%
        \vspace*{2mm}%
    \end{subfigure}
    \begin{subfigure}{\textwidth}%
        \includegraphics[width=\textwidth]{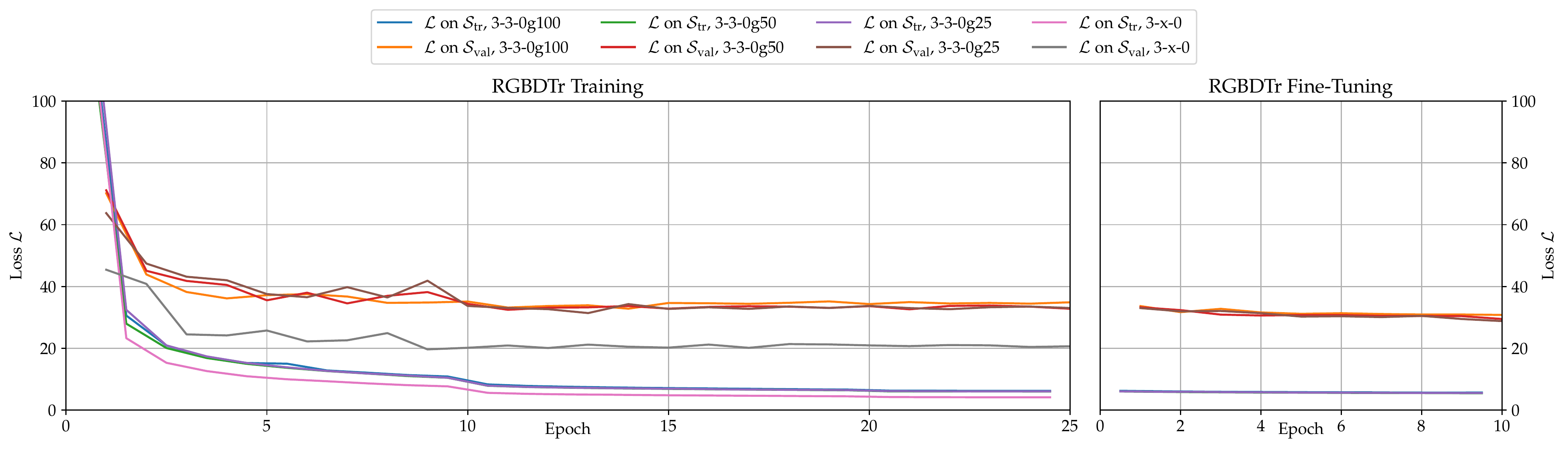}%
        \vspace*{-2mm}%
        \caption{Three layer PatchGAN/encoder on images of size $224 \times 224$ pixels without fusion block.}%
        \label{f:appendix:stg-losses:nofus-224}%
        \vspace*{2mm}%
    \end{subfigure}
    \caption{Progression of the loss values during training of the 16 RGBDTr models on the \stg\ dataset.}
    \label{f:appendix:stg-losses}%
\end{figure}

\begin{figure}[htb]
    \centering\captionsetup{width=0.97\textwidth}%
    \begin{subfigure}{\textwidth}%
        \includegraphics[width=\textwidth]{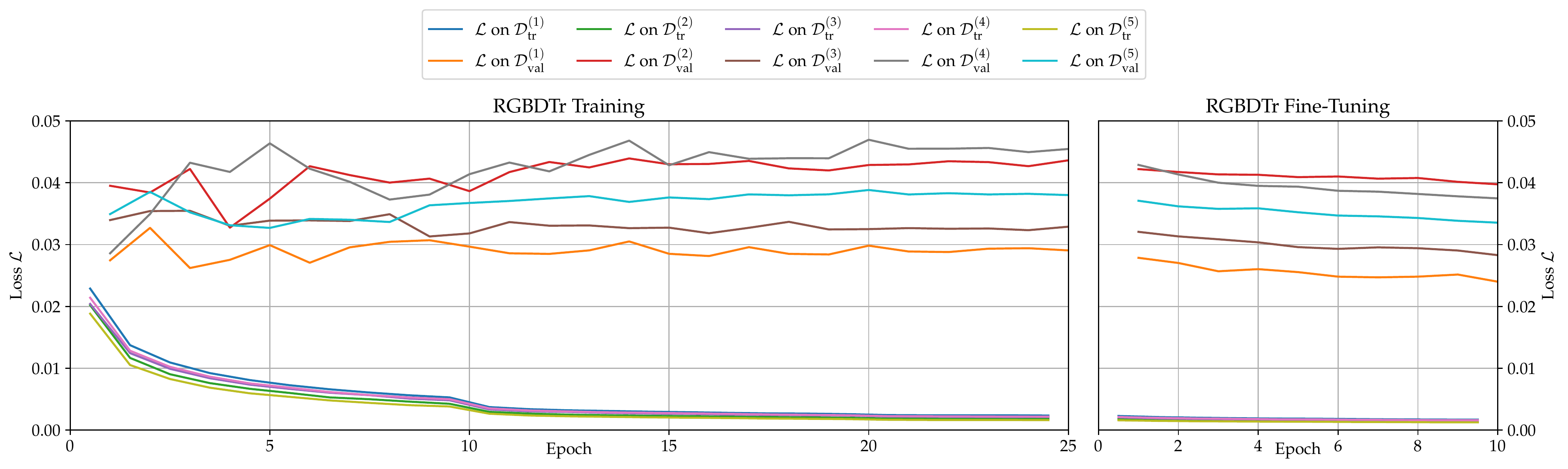}%
        \vspace*{-2mm}%
        \caption{Models trained with discriminator target set $\tau$.}%
        \label{f:appendix:oge-rgbdtr-training:tau}%
        \vspace*{6mm}%
    \end{subfigure}
    \begin{subfigure}{\textwidth}%
        \includegraphics[width=\textwidth]{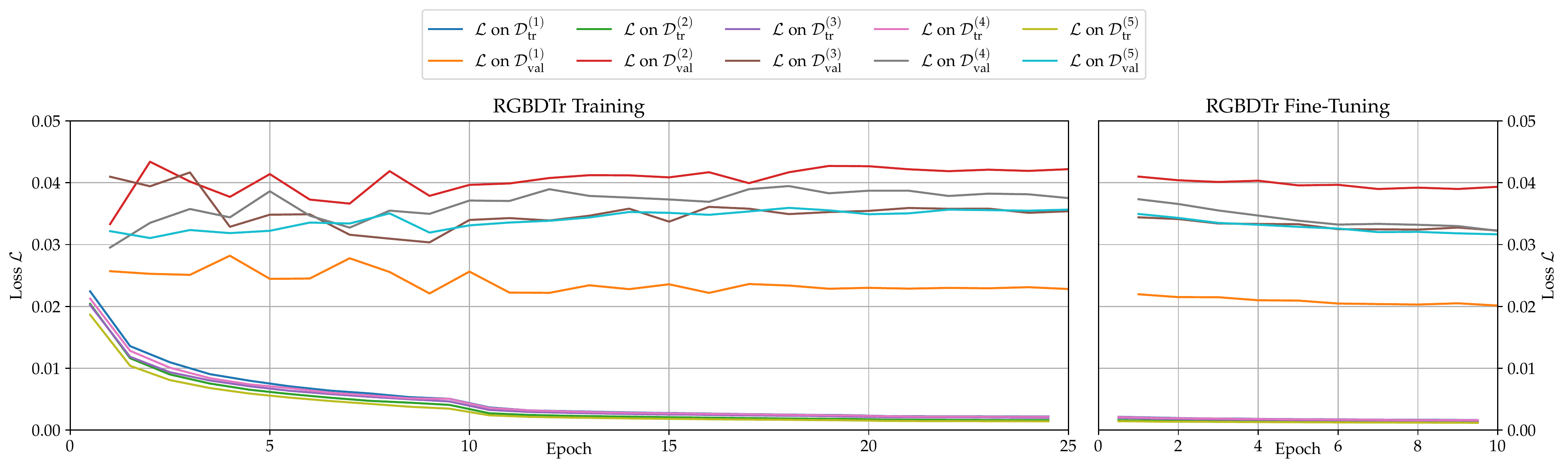}%
        \vspace*{-2mm}%
        \caption{Models trained with input set $\upsilon$.}%
        \label{f:appendix:oge-rgbdtr-training:ups}%
        \vspace*{6mm}%
    \end{subfigure}
    \begin{subfigure}{\textwidth}%
        \includegraphics[width=\textwidth]{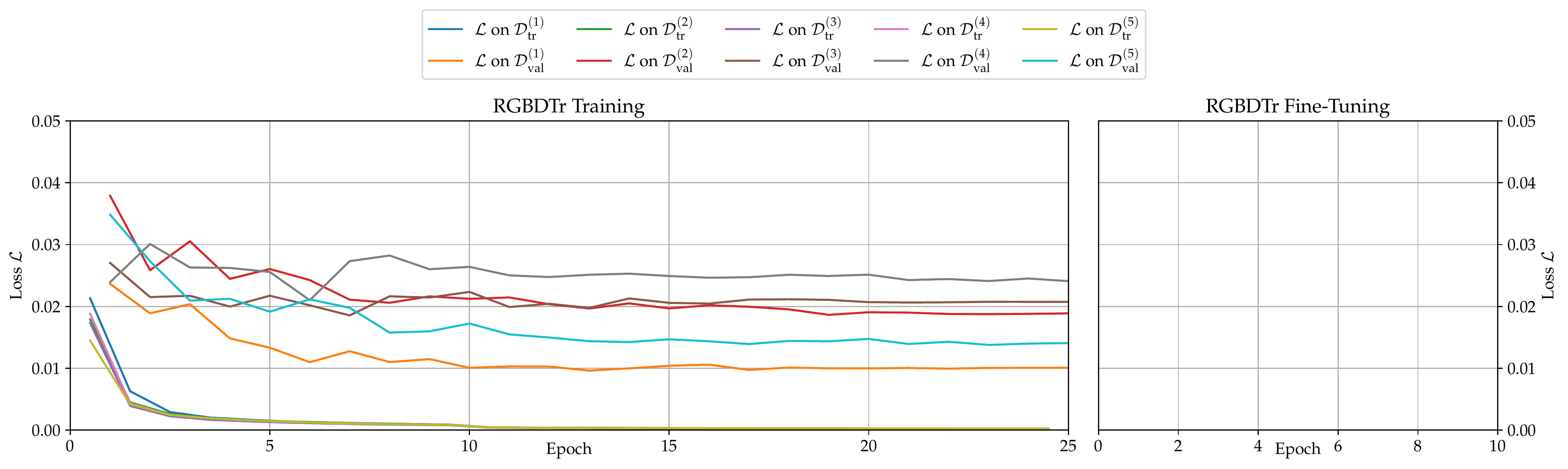}%
        \vspace*{-2mm}%
        \caption{Models trained without \ac{GAN} backbone (no fine-tuning).}%
        \label{f:appendix:oge-rgbdtr-training:nogan}%
        \vspace*{6mm}%
    \end{subfigure}

    \caption{Progression of the loss values during training and fine-tuning of our RGBDTr model on the \oge\ dataset in three different configurations.}%
    \label{f:appendix:oge-rgbdtr-training}%
\end{figure}

\begin{figure}[htb]
    \centering\captionsetup{width=0.95\textwidth}%
    \includegraphics[width=\textwidth]{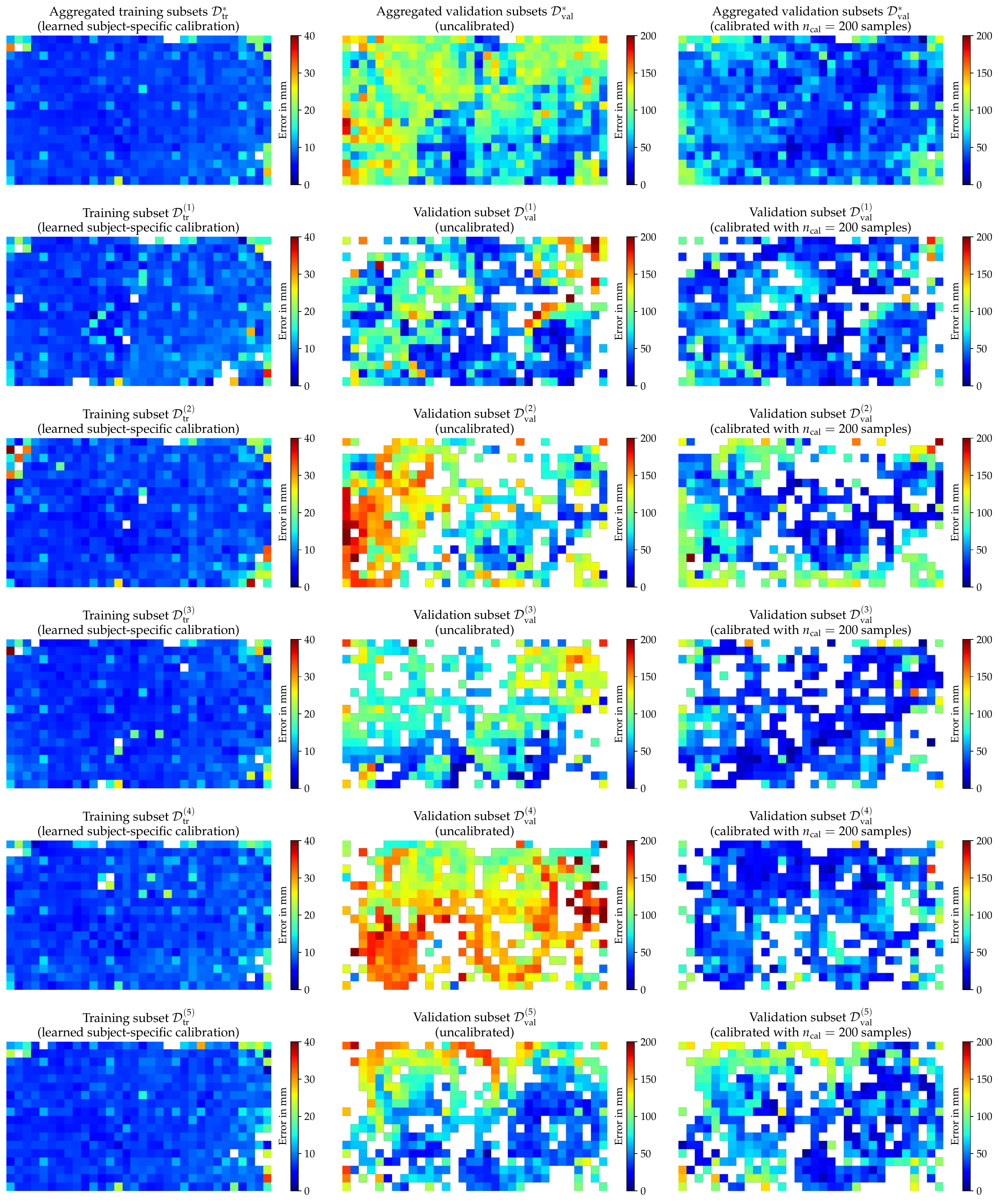}%
    \caption{Distribution of the on-screen Euclidean errors of the baseline model trained on our \oge\ dataset.}%
    \label{f:appendix:oge-baseline-error-hist}%
\end{figure}

\chapter{Additional Tables}

\begin{table}[htb]
    \centering
    \renewcommand{\arraystretch}{1.2}
    \begin{tabular}{>{\ttfamily}c||r|r|r|r}
        \multirow{2}{*}{Subject} & \multicolumn{4}{c}{Samples} \\
        \cline{2-5}
        & \multicolumn{1}{c|}{\# Total} & \multicolumn{1}{c|}{\# Usable} & \multicolumn{1}{c|}{\# Pruned} & \multicolumn{1}{c}{\% Usable} \\
        \hline\hline
        p000 & 9830 & 9487 & 343 & 96.5 \\\hline
        p001 & 9785 & 9464 & 321 & 96.7 \\\hline
        \multirow{2}{*}{p002} & 9559 & 9263 & 296 & 96.9 \\\cline{2-5}
        & 9731 & 9570 & 161 & 98.3 \\\hline
        p003 & 9839 & 9812 & 27 & 99.7 \\\hline
        p004 & 10,480 & 10,477 & 3 & 100.0 \\\hline
        p005 & 9685 & 8666 & 1019 & 89.5 \\\hline
        p006 & 9833 & 9603 & 230 & 97.7 \\\hline
        p007 & 9809 & 9688 & 121 & 89.8 \\\hline
        p008 & 8665 & 8652 & 13 & 99.9 \\\hline
        p009 & 9812 & 9469 & 343 & 96.5\\\hline
        \multirow{2}{*}{p010} & 9771 & 9771 & 0 & 100.0 \\\cline{2-5}
        & 9822 & 9822 & 0 & 100.0 \\\hline
        p011 & 9501 & 9230 & 271 & 97.1 \\\hline\hline
        \rmfamily\textbf{Total} & \textbf{136,122} & \textbf{132,974} & \textbf{3148} & \textbf{97.7} \\
    \end{tabular}
    \caption{Number of samples per recording session of our \oge\ dataset. (Data visualized in \Cref{f:processing:dataset:setup:sessions})}
    \label{t:appendix:oge-samples-per-session}
\end{table}

\begin{table}
    \vspace*{-9mm}%
    \centering\captionsetup{width=0.95\textwidth}%
    \renewcommand{\arraystretch}{1.15}%
    \begin{tabular}{@{}>{\raggedleft$}m{1.25cm}<{$}@{\hspace{1mm}}>{$}c<{$}@{\hspace{1mm}}>{\raggedright$}m{1.25cm}<{$}@{\hspace{1mm}}|m{0.79\textwidth}@{}}%
        \multicolumn{3}{c|}{\textbf{Parameter}} & \textbf{Description} \\\hline\hline
        c_d &= & 64 & \# Channels in the First Discriminator Layer \\\hline
        c_g &= & 64 & \# Channels in the First Generator Encoder Layer \\\hline
        c_h &= & 256 & \# Channels in the First Head Pose Feature Extractor Layer \\\hline
        d_\mathrm{model} &= & 1024 & Transformer Token Size \\\hline
        d_\mathrm{ff} &= & 2048 & Transformer Feed-Forward Layer Size \\\hline
        f &\in & \left\{\frac{1}{4}, \frac{1}{2}\right\} & Eye Region Resize Factor \newline (Default $f = 0.5$ for Pre-Trained Eye Pose Backbones) \\\hline
        h_\mathrm{in} &= & 448 & Input Image Height in Pixels \\\hline
        n_d &= & n_e & \# Decoder Layers \\\hline
        n_\mathrm{dis} &= & n_e & \# Discriminator Layers \\\hline
        n_e &= & 4 & \# Encoder Layers \\\hline
        n_f &= & 2 & \# Fusion Block Layers \newline ($n_f = 0$: No Fusion Block, $n_f = 2$: One Fusion Block) \\\hline
        n_g &\in &\left\{1, 2\right\} & \# Encoder Networks \newline ($n_g = 1$ for RGB Input, $n_g = 2$ for RGBD Input) \\\hline
        n_h &= & 8 & \# Transformer Self-Attention Heads \\\hline
        n_l &= & 6 & \# Transformer Layers \\\hline
        n_t &\in & \left\{4, 5\right\} & \# Transformer Tokens \newline ($n_t = 4$ for RGB Input, $n_t = 5$ for RGBD Input) \\\hline
        \mathcal{O} &= & 0 & Missing Depth Value \\\hline
        r &= & 5 & Depth Extraction Patch Size $\left(r \times r\right)$ Pixels \\\hline
        t_\mathrm{ea} &= & \mathrm{R18} & Eye Pose Backbone Architecture \newline (\textbf{R}esNet-\textbf{18}|\textbf{34}, \textbf{E}fficientNet-\textbf{B2}|\textbf{B3}|\textbf{B4}) \\\hline
        t_\mathrm{es} &= & \mathrm{P} & Eye Pose Backbone State (\textbf{P}re-Trained, From \textbf{S}cratch) \\\hline
        t_\mathrm{dis} &= & \mathrm{P} & Discriminator Architecture (\textbf{F}ull-Image, \textbf{P}atchGAN, \textbf{B}igPatchGAN) \\\hline
        t_\mathrm{fus} &= & \mathrm{D} & Fusion Block Architecture (\textbf{D}efault, \textbf{B}ottleneck) \\\hline
        t_\mathrm{tpe} &= & \mathrm{L} & Transformer Positional Encoding (\textbf{L}earnable, \textbf{S}inusoidal) \\\hline
        t_\mathrm{tr} &= & \mathrm{B2T} & Transformer Architecture (B2T, Pre-LN, Post-LN) \\\hline

        w_\mathrm{in} &= & 448 & Input Image Width in Pixels \\\hline\hline

        \beta_1 &= & 0.9 & Adam Optimizer $\beta_1$ Parameter \\\hline
        \beta_2 &= & 0.999 & Adam Optimizer $\beta_2$ Parameter \\\hline
        &b & & Batch Size (Dependent on Dataset and Training Phase) \\\hline
        d_\mathrm{ta} &= & 0.1 & Transformer Attention Dropout Rate \\\hline
        d_\mathrm{tf} &= & 0.1 & Transformer (Feed-Forward) Dropout Rate \\\hline
        \lambda_\mathrm{L1} &= & 10 & L1-Reconstruction-Loss Weight \\\hline
        &l & & Learning Rate (Dependent on Training Phase and Epoch) \\\hline
        n_\mathrm{ft} &= & 10 & \# Epochs for Fine-Tuning \\\hline
        n_\mathrm{gan} &= & 100 & \# Epochs for \ac{GAN} Training \\\hline
        n_\mathrm{rgbdtr} &= & 25 & \# Epochs for RGBDTr Training \\\hline
        \mathrm{wd} &= & 0 & Weight Decay \\\hline
    \end{tabular}
    \caption{Default model and training parameters of our proposed RGBDTr model.}
    \label{t:appendix:default-config}
\end{table}
\end{document}